\documentclass{article}

\usepackage[preprint,nonatbib]{Styles/neurips_2024}
\usepackage{graphicx}
\usepackage{multirow}
\usepackage{array} 
\usepackage{float}
\usepackage[export]{adjustbox}

\usepackage[utf8]{inputenc} % allow utf-8 input
\usepackage[T1]{fontenc}    % use 8-bit T1 fonts
\usepackage{hyperref}       % hyperlinks
\usepackage{url}            % simple URL typesetting
\usepackage{booktabs}       % professional-quality tables
\usepackage{amsfonts}       % blackboard math symbols
\usepackage{nicefrac}       % compact symbols for 1/2, etc.
\usepackage{microtype}      % microtypography
\usepackage{xcolor}         % colors
\usepackage{amsmath}
\usepackage{lipsum}

\title{VP-LLM: Text-Driven 3D Volume Completion with Large Language Models through Patchification}

\renewcommand*{\thefootnote}{\fnsymbol{footnote}}

\author{%
  Jianmeng Liu$^1$\footnotemark{}~, 
  Yichen Liu$^1$\footnotemark[\value{footnote}]~, 
  Yuyao Zhang$^1$\footnotemark[\value{footnote}]~, 
  Zeyuan Meng$^1$\footnotemark[\value{footnote}]~, 
  Yu-Wing Tai$^{2}$, 
  Chi-Keung Tang$^{1}$ \\
  \AND
  $^1$The Hong Kong University of Science and Technology\\
  $^2${\bf Dartmouth College}\\
}

\begin{document}

\maketitle

\footnotetext{Co-first authors, ranked by alphabetical order of first names}
\renewcommand*{\thefootnote}{\arabic{footnote}}

\newcommand{\opl}{output projection layer~}
\newcommand{\ourwork}{VP-LLM}
\newcommand{\leftfig}{\includegraphics[width=0.49\linewidth, trim = 6cm 0.5cm 0.5cm 0.5cm]}
\newcommand{\lrightfig}{\includegraphics[width=0.49\linewidth, trim = 0.5cm 0.5cm 6cm 0.5cm]}
\begin{figure*}[h]
    \centering
    \includegraphics[width=\linewidth]{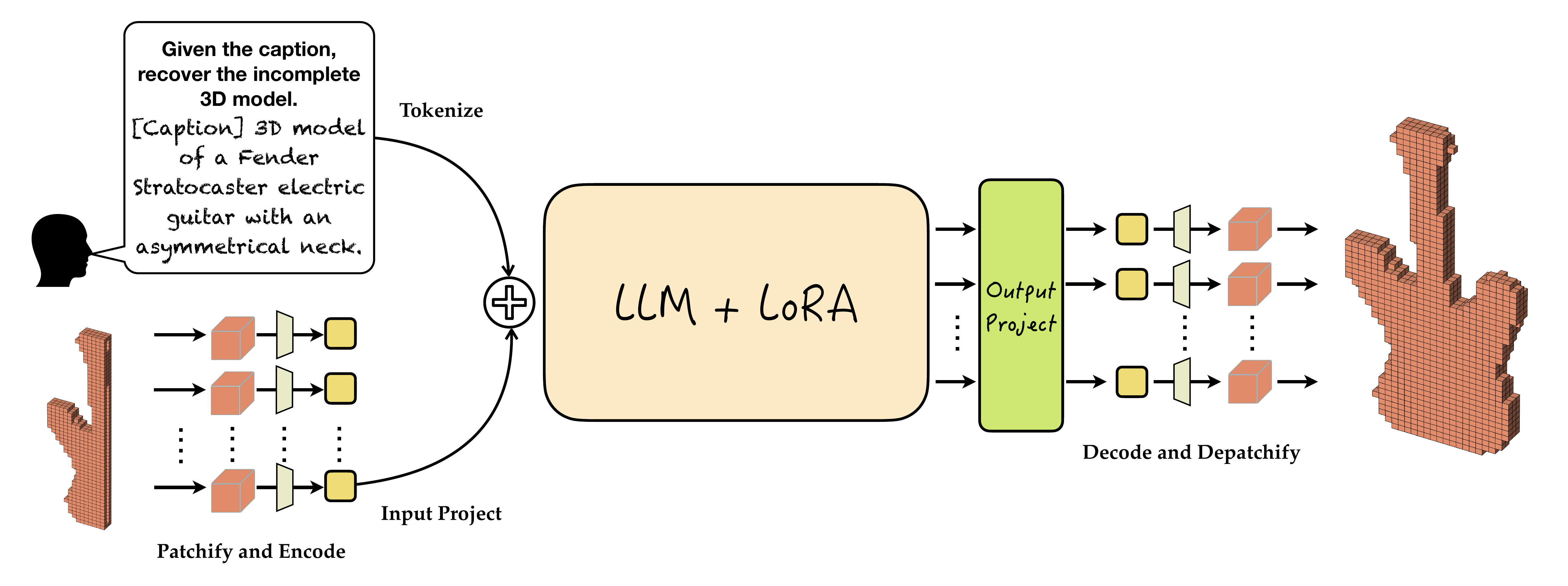}
    \caption{{\bf Overview}. \ourwork~leverages the long-context comprehension capability of Large Language Models (LLMs) to process 3D models. It takes either incomplete or noisy 3D models along with textual instructions as input, and generate a complete model. 
    This is achieved by segmenting the 3D object into patches and processing each independently.}
    \label{fig:teaser}
\end{figure*}
\begin{abstract}
    % This paper explores a new framework to perform 3D completion tasks.

% {\color{red}{LJM:}} Previous conditional 3D completion works mainly focused on diffusion-based methods, which take a significant amount of time to complete each object. Meanwhile, large language models (LLMs) have shown great potential in multi-modal understanding and generation tasks. 
% In this work, we leverage LLMs to perform conditional 3D completion in a single-forward pass by first dividing the incomplete 3D object into independent patches and then feeding the patches into an LLM along with the text prompt to let the LLM capture the relations between those patches as well as injecting semantic meanings to the 3D object. Our results show that LLMs demonstrate a strong ability to interpret complicated text instructions and understand 3D objects, surpassing state-of-the-art diffusion-based 3D completion models in both generation quality and speed.

% {\color{purple}{LYC:}}
Recent conditional 3D completion works have mainly relied on CLIP or BERT to encode textual information, which cannot support complex instruction. Meanwhile, large language models (LLMs) have shown great potential in multi-modal understanding and generation tasks. Inspired by the recent advancements of LLM, we present Volume Patch LLM (\ourwork), which leverages LLMs to perform conditional 3D completion in a single-forward pass. To integrate a 3D model into the LLM tokenization configuration, the incomplete 3D object is first divided into small patches that can be encoded independently. These encoded patches are then fed into an LLM along with the text prompt, instructing the LLM to capture the relations between these patches as well as injecting semantic meanings into the 3D object. Our results demonstrate a strong ability of LLMs to interpret complex text instructions and understand 3D objects, surpassing state-of-the-art diffusion-based 3D completion models in generation quality.

% {\color{blue}MZY:}
% Recently, There have been many Diffusion-based 3D model completion works, and using large language models to achieve multimodal tasks has also been a recent hot topic. Inspired by these two trends, we present (), which is a promptable 3D object completion model (e.g., text input or incomplete 3D model with text guidance)  Unlike previous diffusion-CLIP-based pipelines that involve tedious per-prompt optimizations, () is capable of completing a variety of 3D objects with a single forward pass, leveraging a patchify technique and LLM. The patchify technique,  integrated with the tokenization mechanism of Large Language Models, substantially improves the model's ability to generalize and scale across various contexts and facilitates the generation of 3D models with intricate details.

% Our methodology marks a considerable advancement in narrowing the divide between textual descriptions and 3D visual representations, providing a dynamic and efficient means for the creation and interpretation of sophisticated 3D models derived from natural language inputs. This breakthrough sets the stage for further innovations in the domain of machine interaction with the 3D environment, representing a critical evolution in the discipline.

\end{abstract}

% \vspace{2in}
% {\LARGE \color{red} Please Compile and Read {\tt Styles/neurips\_2024.tex} Before Writing Anything!!}

\section{Introduction}
\label{sec-intro}
3D modeling serves as a pivotal component in a multitude of 3D vision applications including robotics and virtual reality, where the quality of 3D data critically influences model performance. Despite the advancement in 3D scanning technology, the raw data acquired are often noisy, clustered, and may contain large portions of missing data due to occlusion, complex real-world scenes and restricted camera angles, resulting in incomplete 3D acquisition. 
%Moreover, real-world scene complexities and restricted camera angles during data capturing can result in incomplete 3D reconstructions. 
This necessitates robust pre-processing to recover or complete the 3D objects, which can enhance the efficiency of subsequent 3D vision tasks.

Current approaches for 3D shape completion typically operate on a depth map or partial point cloud, converting it into a voxel representation or sampling points to restore the original 3D objects. While~\cite{wu2020multimodal, Zhang_2021_CVPR} showcase advancements, they are confined to specific categories and lack the ability of cross-object generation. Although efforts have been made~\cite{yan2022shapeformer, yu2021pointr, Wu_2018_ECCV, wen2021pmp} to create a unified model that handles multi-category 3D completion, these models often overlook the inclusion of textual input in guiding the completion process, leading to uncertainty when the input is ambiguous, as well as degradation of feasibility when given captions deviate from the training set.
Consequently, methods are needed to generate a completed shape aligning precisely with the provided text description. Some attempts mimic the 2D diffusion method~\cite{cheng2023sdfusion} or score distillation sampling (SDS)~\cite{NEURIPS2023_284afdc2} to incorporate text guidance in the 3D completion tasks, but they cannot be precisely controlled when the description is complicated, and are very time-consuming to generate the results.

To this end, we propose Volume Patch Large Language Model (\ourwork), which achieves 3D completion with precise textual control. Inspired by the recent progress in 3D multi-modality models~\cite{yin2023shapegpt, chen2023octavius, wang2023chat}, we believe that Large Language Models (LLMs) can underpin our approach by decoding the complex associations between 3D structures and textual descriptions.
LLMs, pretrained on large-scale text datasets, have the capability to process long sequences and comprehend complex human languages, while 3D models represented by voxel grids can be straightforwardly converted into a one-dimensional format through flattening. Therefore, we investigate how to enable LLMs to understand a 3D model by decoding complex correlations between 3D structures and textual descriptions, or ``translating'' it into a ``sentence''.

For seamless incorporation of 3D data into the LLM tokenization framework, 3D models are initially segmented into smaller patches, facilitating independent encoding and decoding. Different from most previous methods that manage the 3D object as a unified, this idea of patchification is more scalable and extendable.
% (consider removing this sentence) The patch variational autoencoder eliminates the need for recurrent training across various datasets, as the elementary patches remain consistent across all 3D objects. 
The patchified 3D voxel volume can be processed as a sequence and fed into the LLM with the textual description after alignment. The LLM can fuse the 3D and textural features into its hidden latents, which are finally decoded into complete 3D models.
% The LLM, pre-trained on various textual tasks, has the capability to merge the 3D and textual features (why pretrained on textual tasks indicates it has ability to merge 3d and textual features? need to align first?), and the fused latents (what are fused latents?) are finally decoded into the complete 3D models.
% YAOO ADDED TECHNICAL DETAILS

Our whole pipeline is presented in Fig.~\ref{fig:teaser}. The 3D volumes are first patchified into individual patches and processed by a patch-wise Variational Autoencoder (VAE) to encode individually. The encoded patches are then projected and concatenated with user-specified text conditions to the LLM. Finally, the output projection layer extracts the features generated by the LLM and lets the VAE decode back each patch individually.

In summary, the major contributions of our papers are
\begin{enumerate}
    \item VP-LLM is the first work  leveraging {\em LLM} to achieve text-guided 3D completion, outperforming existing state-of-the-art text-conditioned 3D completion works.
    \item The {\em patchification} method we proposed enables a scalable integration of 3D volumes into the LLM, solving the difficulty in handling high-resolution voxel grids faced by existing works.
    % \item the first significant attempt for integrating 3D models into LLM through {\em patchification} akin to LLM's discrete tokenization.
    \item Our work can perform 3D completion and denoising for multiple categories using detailed text control within one unified model.
\end{enumerate}

Thus, this experimental paper offers insights and lessons learned, providing the first LLM solution to text-guided 3D object completion. Codes and data will be made public upon the paper's acceptance. 
\section{Related Works}
\label{sec-related-works}

\subsection{Multimodality Large Language Models}
The advent of Large Language Models (LLMs) has significantly accelerated advancements in natural language processing. Several studies \cite{jiang2023mistral, touvron2023llama, gemma} have demonstrated the capabilities of LLM in comprehending long contexts, ensuring scalability and adaptability, and facilitating the understanding and generation of natural language. Benefiting from these advantages of LLM, many works have already employed LLM across different modalities, including images~\cite{wang2023visionllm, openai2024gpt4, alayrac2022flamingo}, motion~\cite{jiang2023motiongpt}, and video~\cite{zhang2023videollama, li2024videochat}. Recently, several works combined LLM with 3D data. For example, \cite{yin2023shapegpt} leverages a 3D-aware VQ-VAE~\cite{oord2018neural} and integrates its codebook into the LLM's vocabulary, enabling the LLM to generate and understand 3D objects. %(Add other 3D llm work here) 
But the codebook size may bottleneck the capability of LLM to tackle 3D objects with more complex and various structures. 
LLM-Grounder~\cite{yang2023llm} carefully designs LLM prompt to translate the instructions into regular sub-tasks and instructs some pre-trained 3D grounders~\cite{kerr2023lerf, peng2023openscene} for 3D reasoning, where 3D models are not integrated into the LLM. 
Octavius~\cite{chen2023octavius} adopts the object detector to first discover candidate regions, followed by the application of pre-trained point cloud encoders for extracting features at the instance level. These features are then aggregated and mapped into an LLM for diverse 3D understanding tasks. However, this process reduces the entire 3D model to a single feature, thereby omitting crucial detailed information.
%(I don't think we need to mention this para here) 
In contrast, \ourwork employs a VAE~\cite{kingma2022autoencoding} combined with projection layers, capable of effectively aligning the 3D latent space with the LLM's text space, thus enhancing the model's generalizability, especially for out-of-distribution data.

\subsection{Text-to-3D Generation}
Prior to the era of machine learning, primitive works attempted to retrieve 3D assets from large databases, such as~\cite{chang2014learning, chang2015text}. With the rise of GANs~\cite{goodfellow2014generative}, attempts such as Text2Shape~\cite{chen2019text2shape} started to dominate the 3D generation field. 
Recently, due to promising advancements in text-to-image generation, research focus on text-control 3D generation has shifted to diffusion model~\cite{ho2020denoising}. 
Some works~\cite{liu2023prompt2nerf, sanghi2022clip, jain2022zero} adopt CLIP~\cite{radford2021learning} to align the rendered images with the input text, thus ensuring the semantic meaning of the 3D model, while others like~\cite{poole2022dreamfusion, wang2024prolificdreamer,wang2023score,chen2023fantasia3d,lorraine2023att3d,babu2023hyperfields} leverage pre-trained 2D diffusion models to provide text control and score distillation sampling to improve the 3D consistency. Although text-to-3D generation serves as an inspiration for text-guided 3D completion, none of the existing methods employ large language models for 3D-text interaction to guide the completion results.
% However, this adversarial training fashion reaches its bottleneck due to the insufficiency of diverse paired text and 3D assets. Fortunately, leveraging the massive 2D data, works like~\cite{liu2023prompt2nerf, sanghi2022clip, jain2022zero} adopt CLIP for text-to-3D guidance, while others~\cite{poole2022dreamfusion} ignites the passion that breeds a series of works~\cite{wang2024prolificdreamer,wang2023score,chen2023fantasia3d,lorraine2023att3d,babu2023hyperfields, metzer2023latent,li2023instant} that based on the score distillation sampling methods.

\subsection{3D Completion}
3D completion is a crucial process in various industries, enabling accurate and efficient design and production, and enhancing the overall quality of products and projects. Early works such as~\cite{choy20163d,dai2017shape,girdhar2016learning,han2017high,stutz2018learning,stutz2020learning,wu20153d} that use 3D convolutions with structured representation require high memory usage and compute. Followed by~\cite{yuan2018pcn}, many works~\cite{an2024pointtr,tchapmi2019topnet,yu2022point, wu2020multimodal} that adopt point clouds as 3D representation for shape completion were proposed. For example,~\cite{an2024pointtr,tchapmi2019topnet,yu2022point} generate the final shape in an auto-regressive manner, while~\cite{wu2020multimodal} uses GANs~\cite{goodfellow2014generative} to complete the model.  But none of these offer satisfactory user control. With the introduction of DDPM~\cite{ho2020denoising}, works such as ~\cite{li2023generalized, liu2023meshdiffusion, luo2021diffusion, vahdat2022lion, wu2024blockfusion, zhou20213d,cheng2023sdfusion,li20233dqd} have advanced the 3D shape completion pipelines. Notably,~\cite{cheng2023sdfusion} adopts a 3D diffusion model with VAE to achieve 3D completion with multi-type control, and~\cite{kasten2024point} uses score distillation to perform test-time optimization. However, neither of them can perform completion tasks at a larger scale %(we cannot do large scale as well?)(yaoo:maybe change to "with finer details"? I am not sure what to write, TBD)(cannot emphasize speed here ) 
where model details are required, nor even generate satisfactory results without tedious denoising steps. Thus, scalability, speed and level of control are still issues to be addressed, providing strong motivation for our work.
\section{Methodology}
\label{sec-method}
Given an incomplete 3D model and user-supplied textual description of the target 3D model, our model aims to recover the underlying 3D model aligned with the input text. %Figure~\ref{fig:main} illustrates the pipeline of our model. 
First, the incomplete 3D model undergoes {\em patchification}, where it is split into small patches, and each patch is independently encoded by our Variational Autoencoder (VAE). Next, a shared-weight linear layer maps the patch features to the embedding space of the LLM, which are then combined with the textual description and input into the LLM. The LLM, concatenated with our specially-designed output projection layer, %(should we still use output projection layer?) 
generates the features %(should we say patch tokens instead of patch features?) 
of patches at all positions, allowing for the separate decoding and subsequent assembly, or de-patchification, to the underlying complete 3D model.
\begin{figure*}
    \centering
    \includegraphics[width=\linewidth]{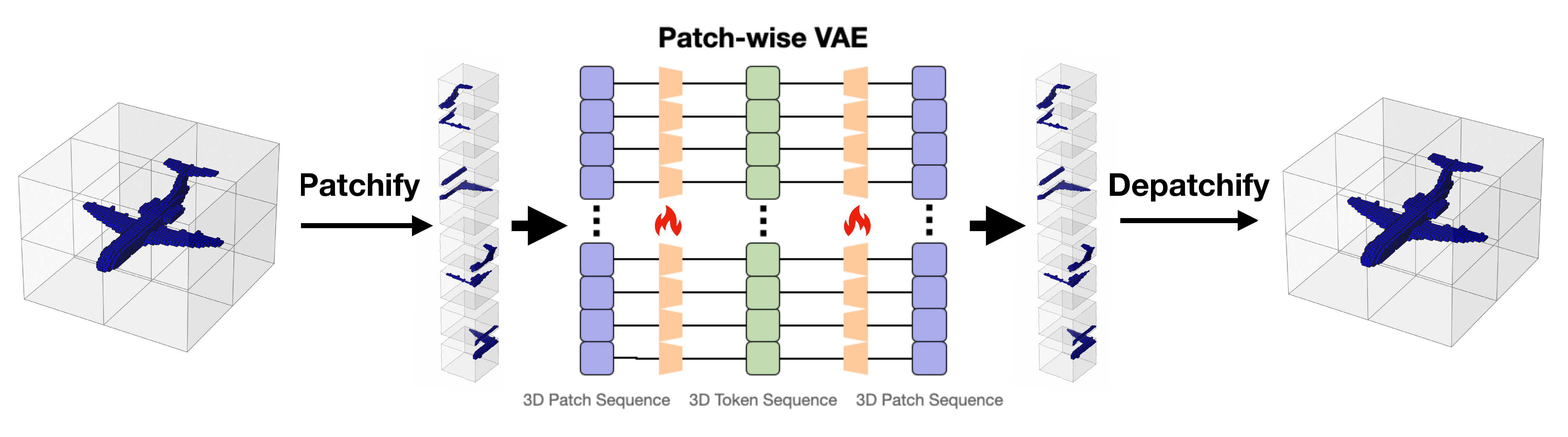}
    \caption{{\bf Patchification}: given a 3D object, we first fit it into a voxel grid and then divide it into a sequence of small patches. Next, we utilize a patch-wise Variational Autoencoder (VAE) to extract the features of each patch individually and then reconstruct it back. It is important to note that only one VAE is trained for all the patches throughout the entire dataset, making our method a scalable approach.}
    \label{fig:patchification}
\end{figure*}

\subsection{Patchification}
\label{sec-method-patch}
The first step of our method is to divide the 3D models into small patches, dubbed patchification. Figure~\ref{fig:patchification} demonstrates the process of patchification. For a 3D object represented in voxel $V \in \{0, 1\}^{H\times W \times D}$, $V(x,y,z) = 1$ if the position $x,y,z$ is occupied and 0 otherwise. Patchification uniformly partitions the 3D voxel volume into $p$ small patches of the same size, each containing a local region of the entire object. For each patch $P_{i,j,k} \in \{0, 1\}^{h\times w \times d}$, the coordinate for position $(x, y, z), 0\le x\le h, 0\le y\le w, 0\le z\le d$ is: 
\begin{align}
    P_{i,j,k}(x,y,z) = V(i \cdot h + x, \; j \cdot w + y, \; k \cdot d + z). 
\end{align}
%CK: P_ijk description is OK
%(I'm confused by this equation)
%(I think a figure is better so as I commented, feel free to delete it)
% lyc: not sure which part is important. Feel free to delete sth.
% , where $h,w,d$ represent the resolution of the patch and $x \in [0, h), y \in [0,w), z \in [0,d)$.
Thus, $p = \lfloor H/h \rfloor \cdot \lfloor W/w \rfloor \cdot \lfloor D/d \rfloor$. In our experiments, we set $H=W=D=64$ and $h=w=d=8$. %(Add something like we will make furthur ablation on the HWD?)
%(Then we can direct delete this part
%(I think we should elaborate more on what is HWD and what is hwd in this part. And since people usually use hwd to describe 2D image data, using hwd in here seems a little bit weird.)

After patchification, a patch Variational Autoencoder (VAE) is adopted to extract the feature for each patch independently. Our patch VAE consists of an encoder $\mathbf{E}$ and a decoder $\mathbf{D}$, where $\mathbf{E}$ encodes a patch into a Gaussian distribution $\mathcal{N}({\mu}, {\sigma})$ where $\mu$ and $\sigma$ are mean and variance, respectively, and $\mathbf{D}$  recovers the original patch from this distribution. The VAE training loss for a single patch $P$ is defined as 
\begin{align}
    \mathcal{L}_{\mathit{VAE}}(P) = \mathcal{L}_{\mathit{BCE}}\left(P, \;\mathbf{D}(\mathbf{E}(P))\right) + \beta \mathcal{L}_{kld} (\mathbf{E}(P)),
\end{align}
where $\mathcal{L}_{\mathit{BCE}}$ is the binary cross entropy loss, $\mathcal{L}_{kld}$ is the KL-divergence and $\beta$ is a hyperparameter. 

Benefiting from this patchification structure that allows for independent encoding and decoding of each patch, \ourwork~ensures that when the completions of certain patches are undesired, they do not affect the performance of other well-performed patches, solving the problems faced by previous works that encode or decode the entire scene collectively.
% Due to the large proportion of empty patches (i.e. $P(x,y,z)=0$ at all places) in 3D voxels, we introduce a gaussian weight $w$ for each patch to accelerate the training, where 
% \begin{align}
%     w(P) = \exp\left( \frac{(-0.5)}{}\right)\frac{\sum_{x,y,z} P\left(x,y,z\right)}{h\cdot w \cdot d}
% \end{align}
\begin{figure*}
    \centering
    \includegraphics[width=\linewidth]{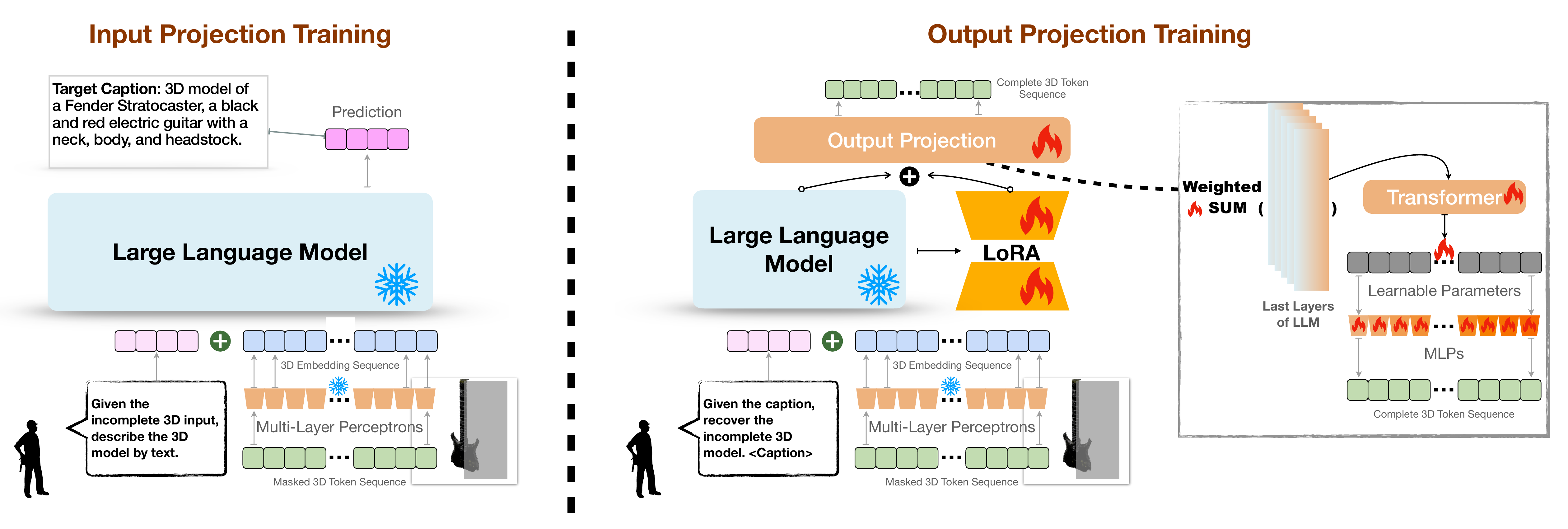}
    \caption{The training process of the {\bf input projection} (left) and {\bf output projection} (right). During the input projection training, a single share-weighted MLP maps the masked or noisy 3D tokens encoded by our patch-wise VAE to the embedding space of the LLM. After wrapping the prompt with the 3D tokens as input and feeding them to the LLM, we back-propagate the loss calculated between the ground-truth caption and the LLM's prediction, enabling the LLM to learn to generate captions that accurately describe the 3D object from the input patches. For the output projection, we freeze the input projection layer and train the output projection layer, while also fine-tuning the LLM with LoRA. The output projection layer comprises a Transformer and a cluster of MLPs, such that after passing the Transformer, every 3D token is processed independently with an MLP.}
    % This process allows the LLM to generate more accurate and coherent captions by projecting the learned features back to the 3D space.}
    \label{fig:input_output}
\end{figure*}

\subsection{Mask Strategy}
To enhance the understanding of incomplete 3D models, we designed three different strategies to mask out different parts of the original 3D input, aiming to mimic the possible user inputs during the inference stage. Specifically, the following three strategies will be applied randomly with the same possibility:%(Where should we put Data augmentation?)(don't mention or in experiment? not sure)
\begin{enumerate}
    \item {\it Random Mask}: Given the input 3D model in $p$ patches, we randomly set $m_r\cdot p$ patches to $0$ (unoccupied), where $m_r$ is a  mask ratio sampled within a pre-defined range;
    \item {\it Plane Mask}: Given the input 3D model represented in voxel $V(x, y, z)$, we first project the model onto $x$-axis and find the first and last occupied voxels, denoted as $x_1$ and $x_2$, along $x$-axis. Next, a plane parallel to $yOz$ is sampled with $x$-coordinate between $[x_1, x_2]$. Intuitively, such a plane cuts the 3D model into two parts, and we then discard one of them by setting all voxels to be $0$ (unoccupied) to simulate large portions of missing data in real capture;
    \item {\it Random Noise}: Since real-world 3D models usually contain noises and artifacts, we mimic this situation by randomly inverting voxel occupancy (setting occupied voxels to be unoccupied, and vice versa). The noise level is also sampled in a pre-defined range indicating how many voxels to invert.
\end{enumerate}

\subsection{Input Projection Layer Training}
\label{sec-method-input-proj}
The input project layer, which can map the VAE latent space into the LLM input embedding space, is a single linear model operated on each VAE latent patch. 
After patchifying and encoding the incomplete 3D model, each patch, represented by its respective $\mu$ and $\sigma$, is first reparameterized into a single feature $f$ by sampling from the Gaussian distribution $\mathcal{N}({\mu}, {\sigma})$. %(Consider removing the re-parameterize process, just VAE latent)(You said to write in this way...)
The feature (or each encoded patch) going through the input projection layer becomes a 3D token, which then can be understood by the LLM. To train the input projection layer, the LLM is instructed to predict the caption of the incomplete 3D model, given the prompt ``Given the incomplete 3D input, describe the 3D model by text.'' %The sequence of 3D tokens, following the instruction prompt, is fed into the LLM and the LLM outputs the corresponding caption of the 3D model (this sentence is duplicated, consider removing). 
We use the training loss in~\cite{radford2019language} %(I don't think this is mentioned in GPT? should just be NLL loss or simply describe it as the loss between generated and gt captions), 
which is the negative log-likelihood on the caption. We use the training loss from the LLM to supervise this stage of training.

\subsection{Output Projection Layer Training}
\label{sec-method-output-proj}
The LLM, with the \opl appended, takes as input the tokenized sequence of the underlying incomplete 3D model, the user-supplied caption of the complete model as well as the instructions for completion, and outputs the complete 3D model aligned with the textual description.  To be specific, we formulate the prompt for LLM as {\tt ``Given the caption, recover the incomplete 3D model, <tokenized incomplete 3D sequence>, <Caption>''}. The \opl architecture employs a transformer consisting of a 2-layer encoder and a 2-layer decoder, translating the LLM hidden states into a sequence of separable latent codes. We observe that to sufficiently explore the highly fused information in the LLM, more layers of hidden states are necessary. Thus, we select $5$ layers in our experiments, which balances the amount of information with computational complexity. To ensure the length of the generated sequence, we set an additional learnable token sequence as the target. We then utilize Multi-layer Perceptrons (MLPs) to individually map each token to the desired 3D token. The final result is obtained through the concatenation of the mapped tokens.

During training, the input projection layer is frozen, while the LLM is finetuned with LoRA~\cite{hu2021lora}, while the \opl is trained from scratch. We use mean-squared error (MSE) loss between the \opl output and the VAE latent of ground-truth 3D model patches to update our model.

\section{Experiments}
\label{sec-exp}

\subsection{Dataset}
We train our model on a subset of ShapeNet~\cite{chang2015shapenet} dataset, comprising over 3000 objects. To obtain the detailed textual description of 3D models in human languages, we adopt Cap3D~\cite{luo2024scalable}, which leverages BLIP to predict and GPT-4 to refine captions for 3D models. In our experiments, the resolutions of the 3D voxels and patches are respectively set as $64 \times 64 \times 64$ and $8 \times 8 \times 8$, while we explore the capability of the model to handle higher resolution formats in Sec.~\ref{sec-ablation-scalability}. 

To improve the robustness of our model, we apply data augmentation during training. For the 3D models, we rotate them along one random axis with an arbitrary angle making the order of sequence different, while for the captions of the 3D model, we adjust the GPT configurations in Cap3D. Only 3D data augmentation is used in input projection training, while both 3D data and caption augmentation are used in output projection training. More details of the data augmentation can be found in Appendix~\ref{sec-supp-data-aug}.

\subsection{Comparison}

\begin{table*}
\centering
\caption{\textbf{Quantitative results compared with SDFusion and 3DQD}. We can observe that our method consistently hits the lowest (best) Chamfer Distance (CD) and the highest (best) CLIP-s score compared with SDFusion (text-conditioned completion) and 3DQD (label-conditioned completion). Moreover, our method is capable of denoising extremely noisy 3D inputs, while the baselines cannot accomplish the task.}
\resizebox{0.99\linewidth}{!}{%
\begin{tabular}{ccccccccccc} \toprule  \multirow{2}{*}{\textbf{Methods}} & \multicolumn{2}{c}{{Seg 20\%}} & \multicolumn{2}{c}{{Seg 50\%}} & \multicolumn{2}{c}{{Seg 80\%}} & \multicolumn{2}{c}{{Noise 1\%}} & \multicolumn{2}{c}{{Noise 2\%}} \\ \cmidrule(lr){2-3} \cmidrule(lr){4-5} \cmidrule(lr){6-7} \cmidrule(lr){8-9} \cmidrule(lr){10-11} & \small{CD.}$\downarrow$ & \small{CLIP-s.}$\uparrow$ & \small{CD.}$\downarrow$ & \small{CLIP-s.}$\uparrow$ & \small{CD.}$\downarrow$ & \small{CLIP-s.}$\uparrow$ & \small{CD.}$\downarrow$ & \small{CLIP-s.}$\uparrow$ & \small{CD.}$\downarrow$ & \small{CLIP-s.}$\uparrow$\\
\midrule  {Ours} & \textbf{10.96}& \textbf{27.80\%} & \textbf{11.37} & \textbf{27.71}\% &\textbf{ 17.42}& \textbf{25.17\%} & 16.03& 23.92\% & 34.44& 23.07\% \\
{SDFusion} & 95.44 & 26.66\% & 137.31 & 26.20\% & 235.98 & 22.22\% & \multicolumn{2}{c}{--} & \multicolumn{2}{c}{--} \\
{3DQD} & 172.89& 22.63\% & 170.20& 22.62\% &  196.12& 22.62\% & \multicolumn{2}{c}{--} & \multicolumn{2}{c}{--} \\
\bottomrule \end{tabular}
}
\label{tab:comparison_main} 
\end{table*}

\begin{table*}
    \newcommand{\dummyfigure}{\fbox{\rule{0pt}{0.5in} \rule{0.25\linewidth}{0pt}} }
    \centering
    \caption{{\bf Comparison of our method with SDFusion and 3DQD, on Airplane dataset.} We can clearly see our method outperforms their methods when the input is segmented by a plane and performs reasonably well when the input is added with noises. Since the two baselines cannot work on noisy inputs, ``N/A'' is placed instead.}
    \resizebox{0.99\linewidth}{!}{%
    \begin{tabular}{>{\centering\arraybackslash}m{0.1\linewidth}>{\centering\arraybackslash}m{0.2\linewidth}>
    {\centering\arraybackslash}m{0.2\linewidth}>
    {\centering\arraybackslash}m{0.2\linewidth}>
    {\centering\arraybackslash}m{0.2\linewidth}>
    {\centering\arraybackslash}m{0.2\linewidth}}
       \toprule
          & Ground Truth & Masked/Noisy & Ours & SDFusion & 3DQD \\
        \midrule
        \multirow{4}{*}{Seg 20\%} & 
        \adjincludegraphics[trim={{0.28\width} {0.35\width} {0.18\width} {0.3\width}},clip,width=0.2\textwidth]{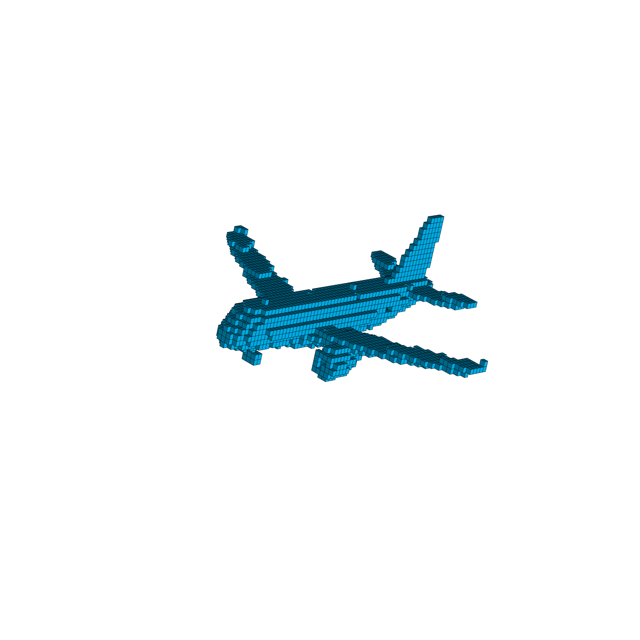} & 
        \adjincludegraphics[trim={{0.28\width} {0.35\width} {0.18\width} {0.3\width}},clip,width=0.2\textwidth]{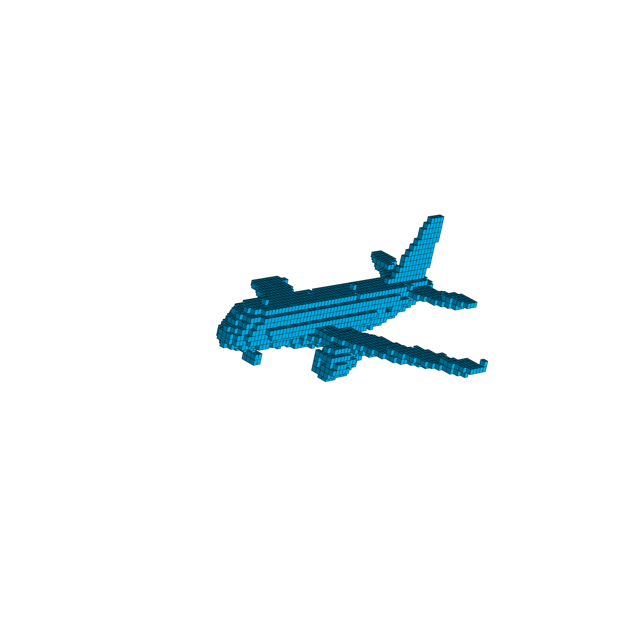} & 
        \adjincludegraphics[trim={{0.28\width} {0.35\width} {0.18\width} {0.3\width}},clip,width=0.2\textwidth]{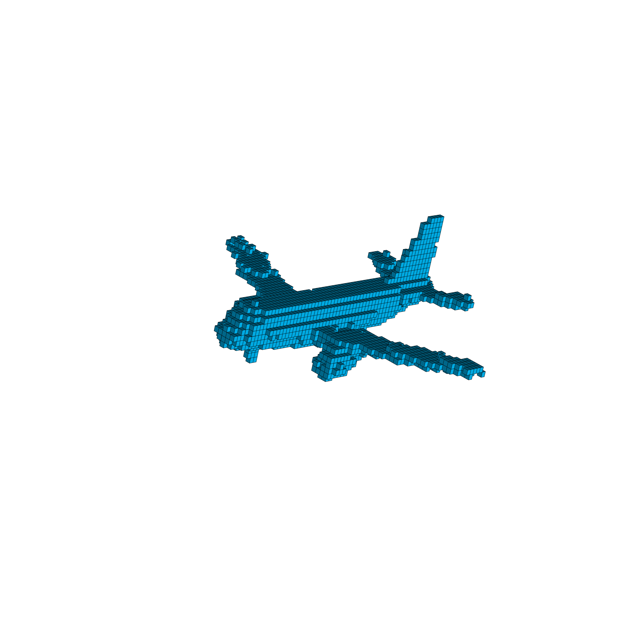} & 
        \adjincludegraphics[trim={{0.28\width} {0.35\width} {0.18\width} {0.3\width}},clip,width=0.2\textwidth]{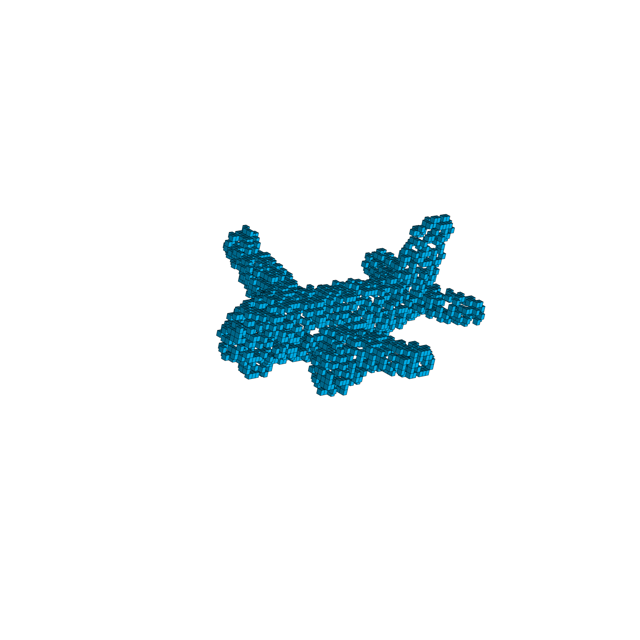} & 
        \adjincludegraphics[trim={{0.28\width} {0.35\width} {0.18\width} {0.3\width}},clip,width=0.2\textwidth]{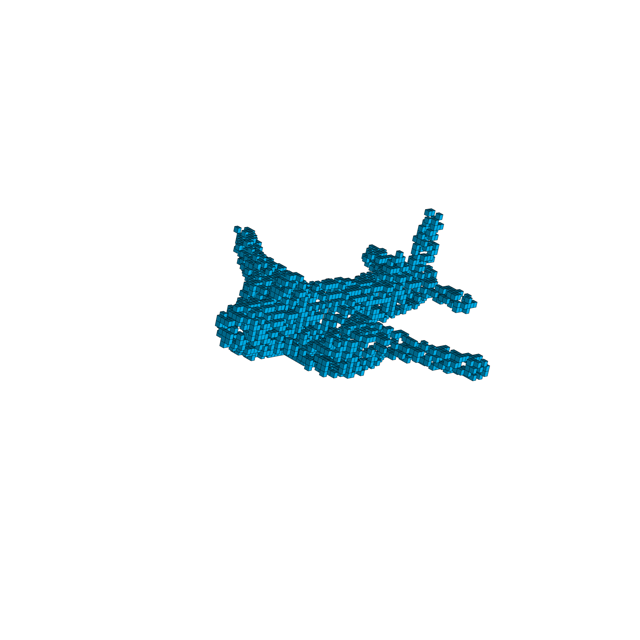} \\
        & \multicolumn{5}{c}{\it ``3D model of a Boeing 747-400 featuring a spherical fuselage shell, truncated oblate wings,}\\
        & \multicolumn{5}{c}{\it and made of aluminum and steel.''}\\
        \multicolumn{6}{c}{}\\
        \multirow{4}{*}{Seg 50\%} & 
        \adjincludegraphics[trim={{0.28\width} {0.35\width} {0.18\width} {0.3\width}},clip,width=0.2\textwidth]{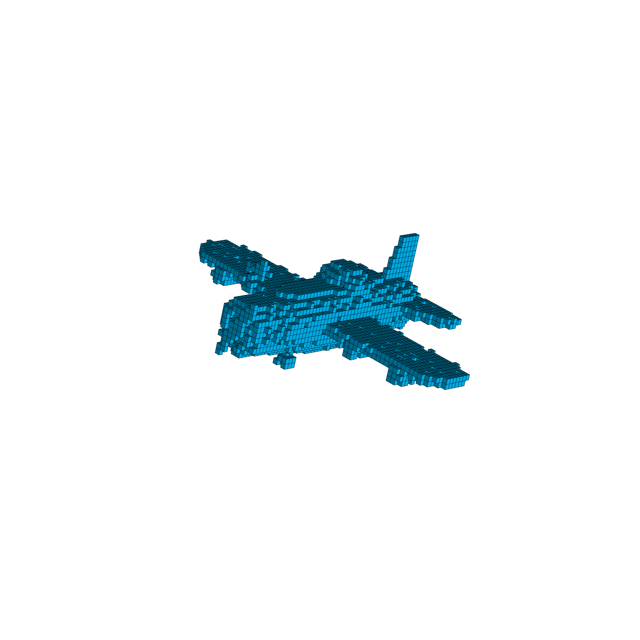} & 
        \adjincludegraphics[trim={{0.28\width} {0.35\width} {0.18\width} {0.3\width}},clip,width=0.2\textwidth]{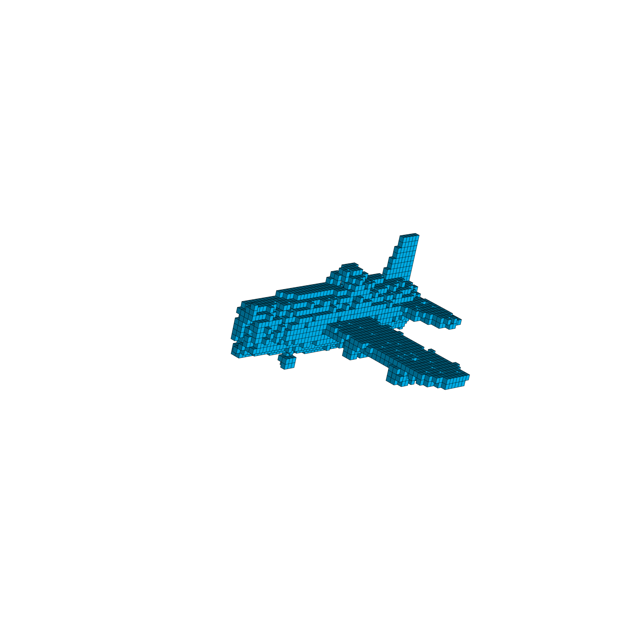} & 
        \adjincludegraphics[trim={{0.28\width} {0.35\width} {0.18\width} {0.3\width}},clip,width=0.2\textwidth]{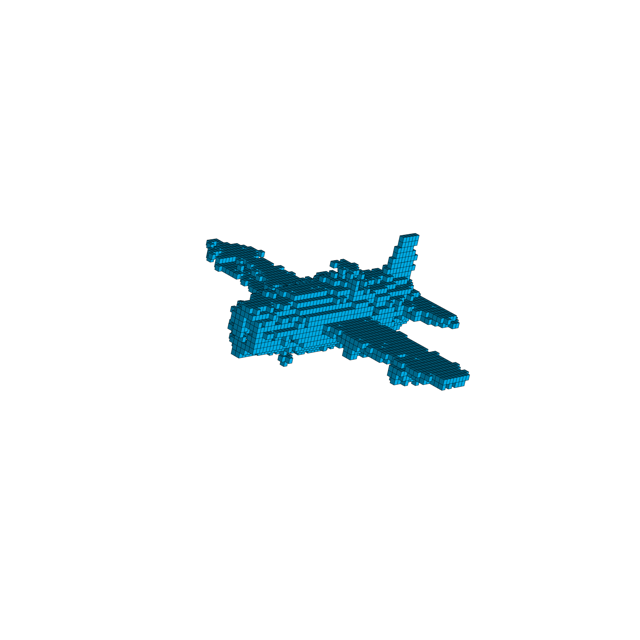} & 
        \adjincludegraphics[trim={{0.28\width} {0.35\width} {0.18\width} {0.3\width}},clip,width=0.2\textwidth]{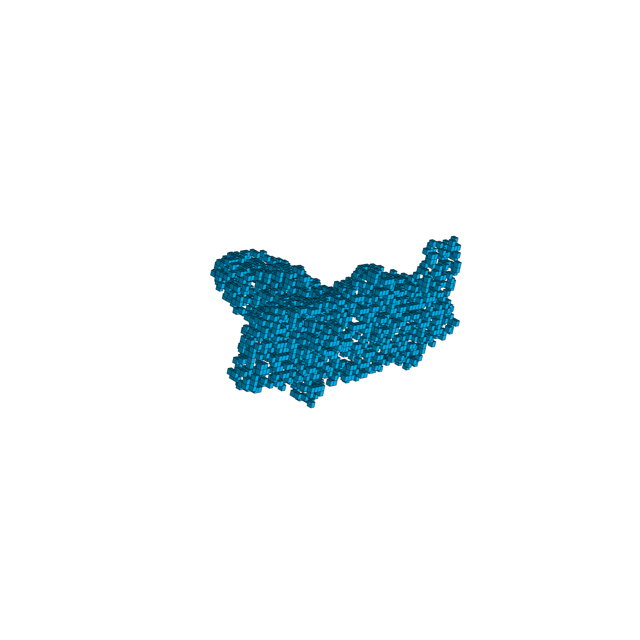} & 
        \adjincludegraphics[trim={{0.28\width} {0.35\width} {0.18\width} {0.3\width}},clip,width=0.2\textwidth]{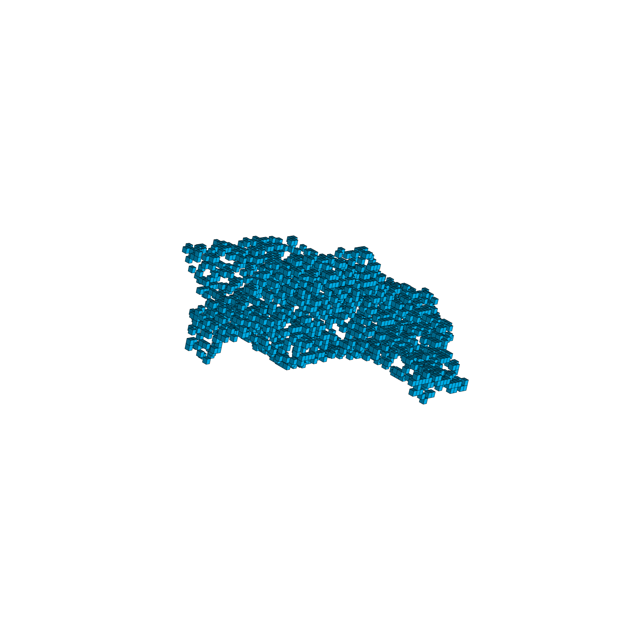} \\
        & \multicolumn{5}{c}{\it ``3D model of a toy airplane featuring a wing, fuselage, tail, rudder, and propeller,}\\
        & \multicolumn{5}{c}{\it available in 3ds Max, OBJ, FBX, and C4D formats.''}\\
        \multicolumn{6}{c}{}\\
        \multirow{4}{*}{Seg 80\%} & 
        \adjincludegraphics[trim={{0.28\width} {0.35\width} {0.18\width} {0.3\width}},clip,width=0.2\textwidth]{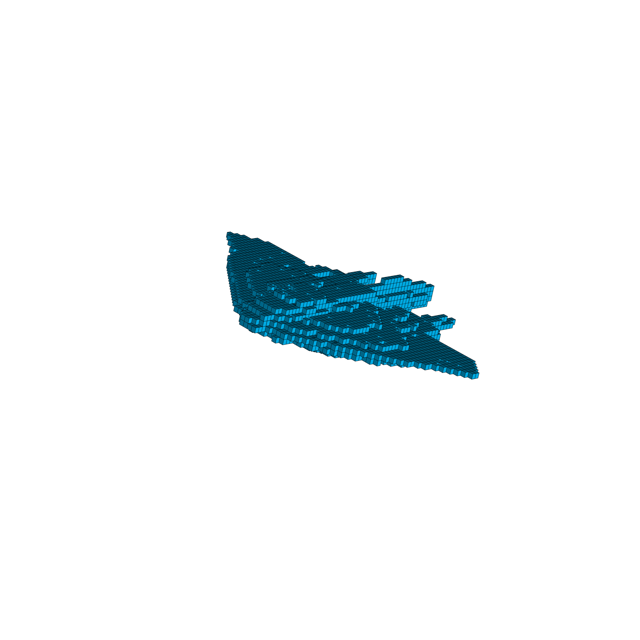} & 
        \adjincludegraphics[trim={{0.28\width} {0.35\width} {0.18\width} {0.3\width}},clip,width=0.2\textwidth]{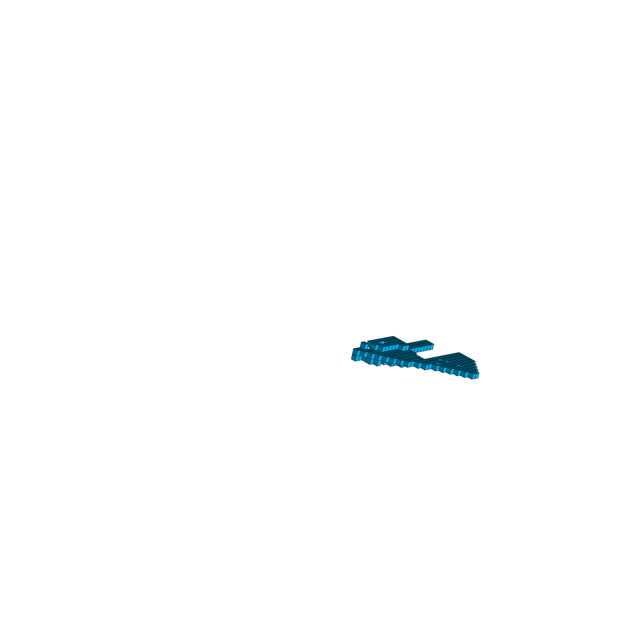} & 
        \adjincludegraphics[trim={{0.28\width} {0.35\width} {0.18\width} {0.3\width}},clip,width=0.2\textwidth]{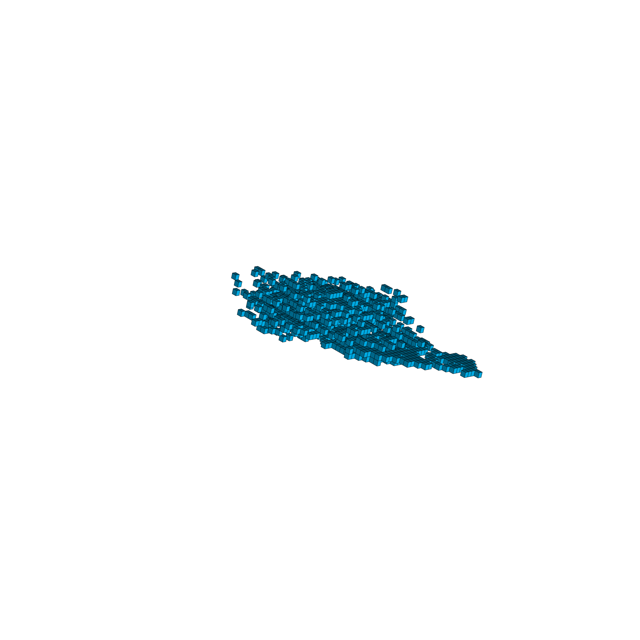} & 
        \adjincludegraphics[trim={{0.28\width} {0.35\width} {0.18\width} {0.3\width}},clip,width=0.2\textwidth]{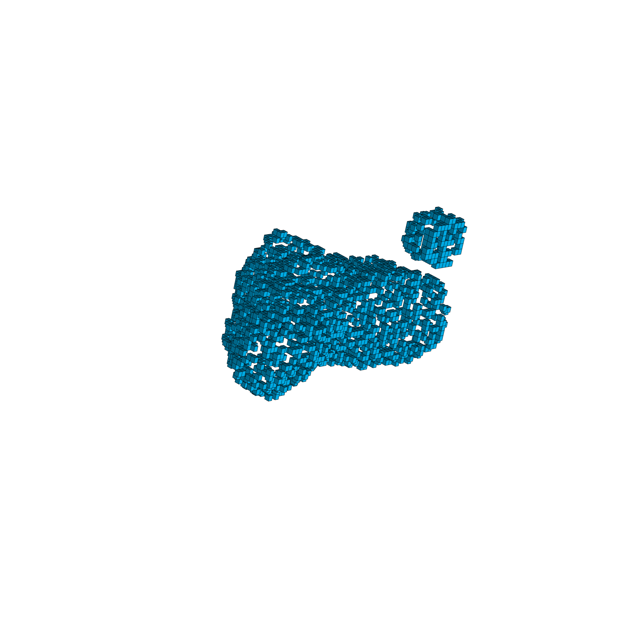}  & 
        \adjincludegraphics[trim={{0.28\width} {0.35\width} {0.18\width} {0.3\width}},clip,width=0.2\textwidth]{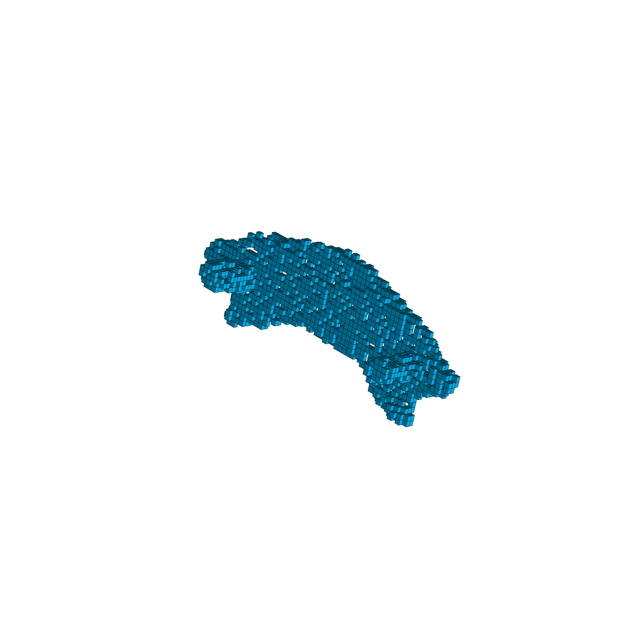} \\
        & \multicolumn{5}{c}{\it ``Royalty-free 3D model of a stealth fighter jet, featuring a delta wing}\\
        & \multicolumn{5}{c}{\it with horizontal and vertical stabilizers''}\\
        \multicolumn{6}{c}{}\\
        \multirow{4}{*}{Noise 1\%} & 
        \adjincludegraphics[trim={{0.28\width} {0.35\width} {0.18\width} {0.3\width}},clip,width=0.2\textwidth]{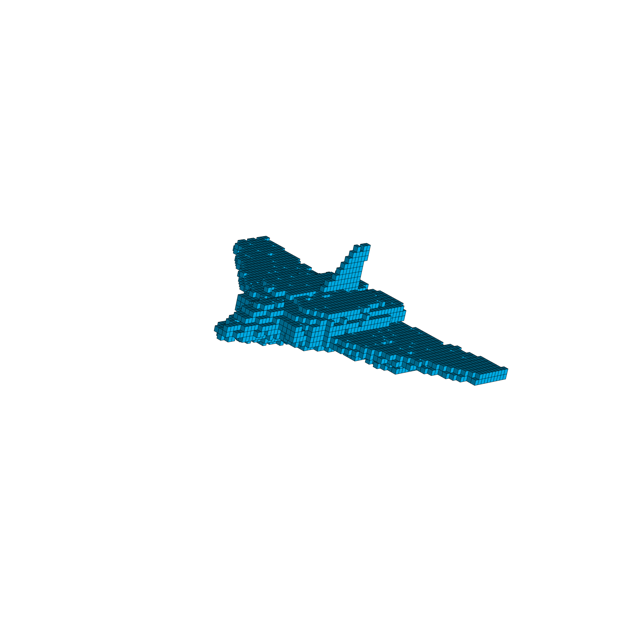} & 
        \adjincludegraphics[trim={{0.28\width} {0.35\width} {0.18\width} {0.3\width}},clip,width=0.2\textwidth]{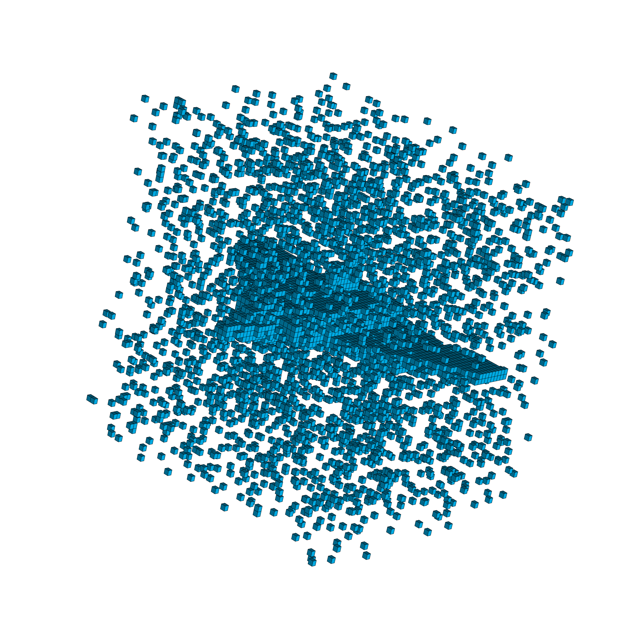} & 
        \adjincludegraphics[trim={{0.28\width} {0.35\width} {0.18\width} {0.3\width}},clip,width=0.2\textwidth]{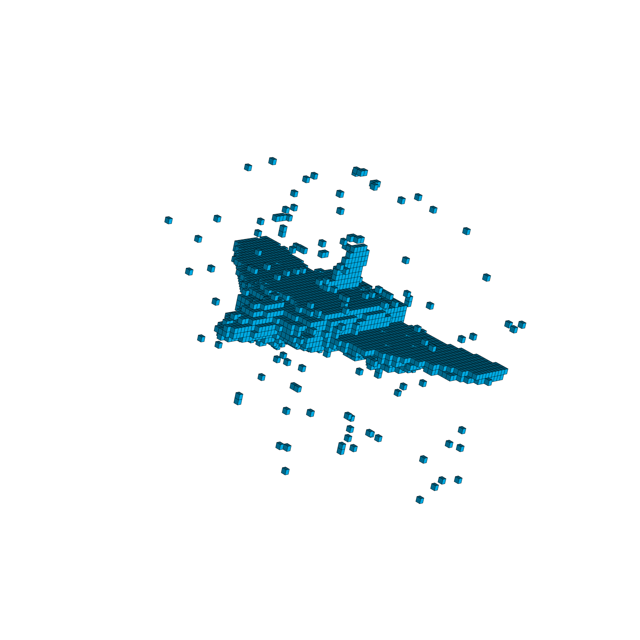} & 
        N/A & 
        N/A \\
        & \multicolumn{5}{c}{\it ``3D model of a Saber fighter jet featuring detailed wings, fuselage, and multiple landing gears,}\\
        & \multicolumn{5}{c}{\it compatible with 3ds Max, Maya, Blender, and other 3D software.''}\\
        \multicolumn{6}{c}{}\\
        \multirow{3}{*}{Noise 2\%} & 
        \adjincludegraphics[trim={{0.28\width} {0.35\width} {0.18\width} {0.3\width}},clip,width=0.2\textwidth]{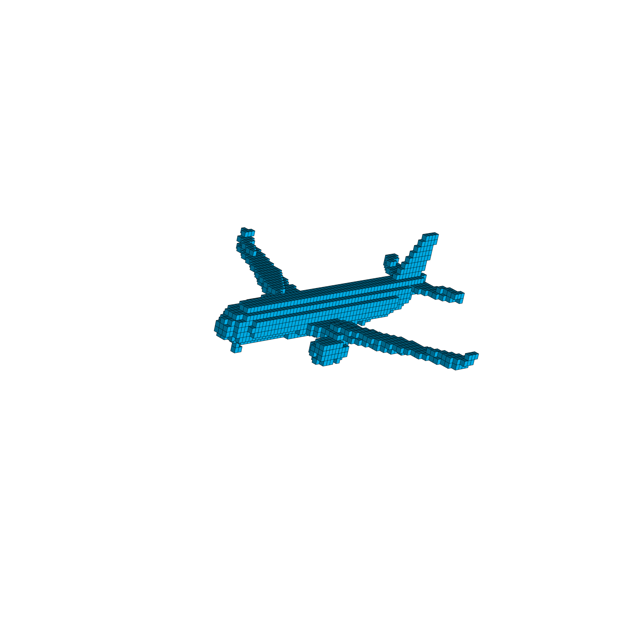} & 
        \adjincludegraphics[trim={{0.28\width} {0.35\width} {0.18\width} {0.3\width}},clip,width=0.2\textwidth]{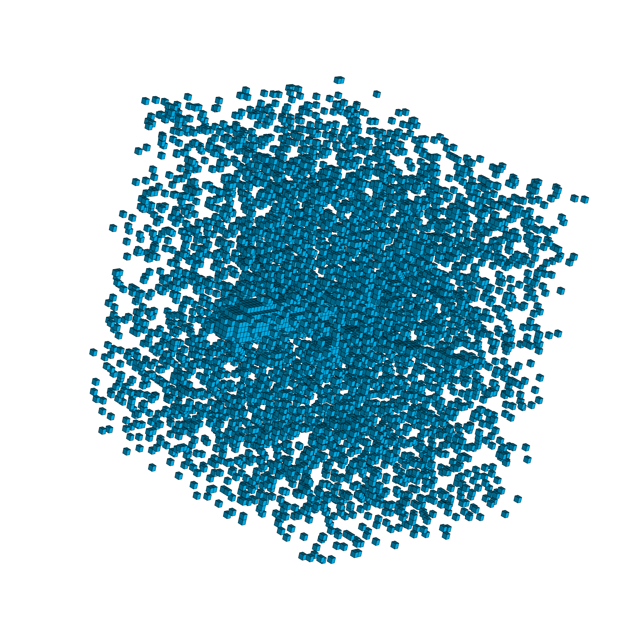} & 
        \adjincludegraphics[trim={{0.28\width} {0.35\width} {0.18\width} {0.3\width}},clip,width=0.2\textwidth]{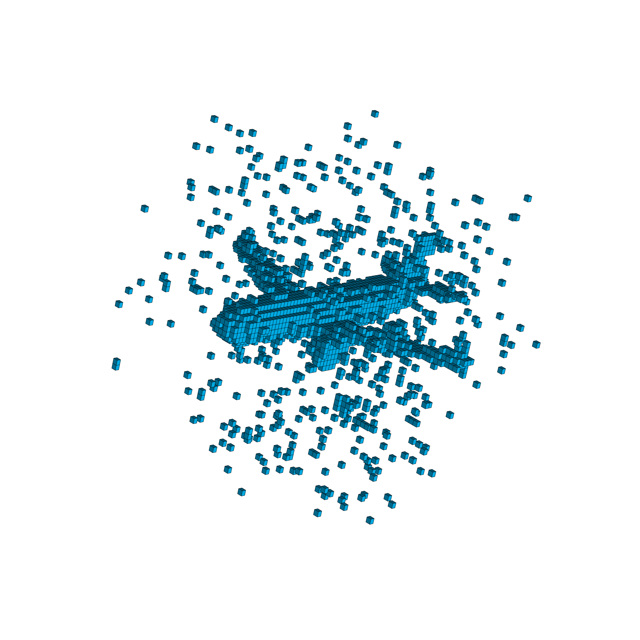} &  
        N/A & 
        N/A \\
        & \multicolumn{5}{c}{\it ``Royalty-free 3D model of a Boeing 747-400 jumbo jet.''}\\
        \multicolumn{6}{c}{}\\
        \bottomrule
    \end{tabular}%
    }
    \label{tbl:table_of_figures_plane}
\end{table*}

\begin{table*}
    \newcommand{\dummyfigure}{\fbox{\rule{0pt}{0.5in} \rule{0.25\linewidth}{0pt}}}
    \centering
    \caption{{\bf Comparison of our method with SDFusion and 3DQD, on Car dataset.} We can clearly see our method outperforms their methods when the input is segmented by a plane and performs reasonably well when the input is added with noises. Since the two baselines cannot work on noisy inputs, ``N/A'' is placed instead.}
    \resizebox{0.99\linewidth}{!}{%
    \begin{tabular}{>{\centering\arraybackslash}m{0.1\linewidth}>{\centering\arraybackslash}m{0.2\linewidth}>
    {\centering\arraybackslash}m{0.2\linewidth}>
    {\centering\arraybackslash}m{0.2\linewidth}>
    {\centering\arraybackslash}m{0.2\linewidth}>
    {\centering\arraybackslash}m{0.2\linewidth}}
       \toprule
          & Ground Truth & Masked/Noisy & Ours & SDFusion & 3DQD \\
        \midrule
        \multirow{3}{*}{Seg 20\%} & 
        \adjincludegraphics[trim={{0.28\width} {0.35\width} {0.18\width} {0.3\width}},clip,width=0.2\textwidth]{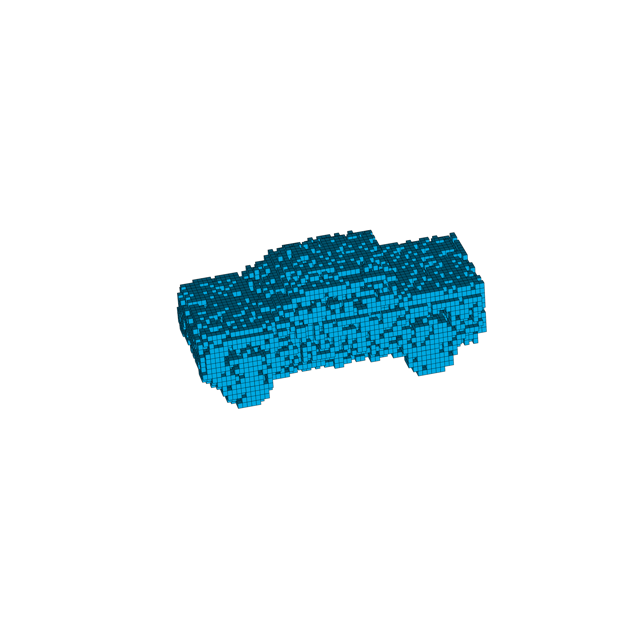} & 
        \adjincludegraphics[trim={{0.28\width} {0.35\width} {0.18\width} {0.3\width}},clip,width=0.2\textwidth]{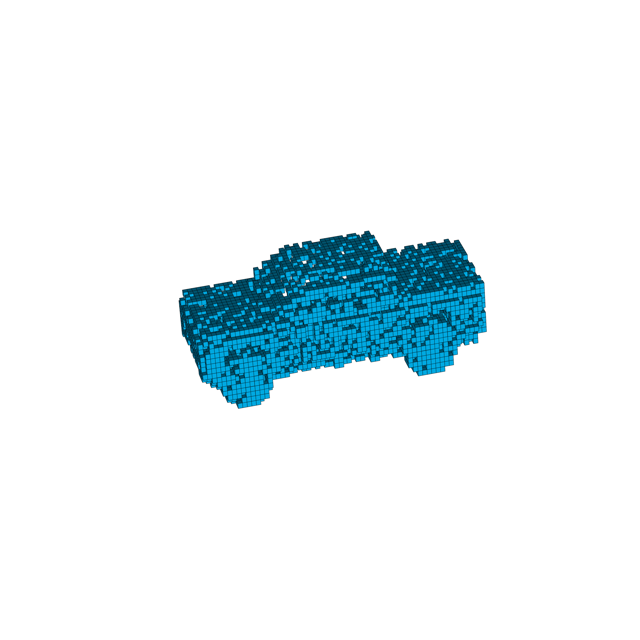} & 
        \adjincludegraphics[trim={{0.28\width} {0.35\width} {0.18\width} {0.3\width}},clip,width=0.2\textwidth]{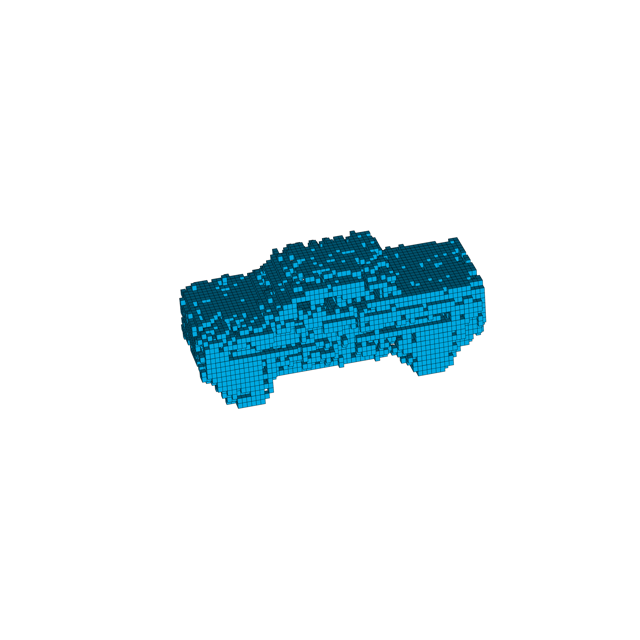} & 
        \adjincludegraphics[trim={{0.28\width} {0.35\width} {0.18\width} {0.3\width}},clip,width=0.2\textwidth]{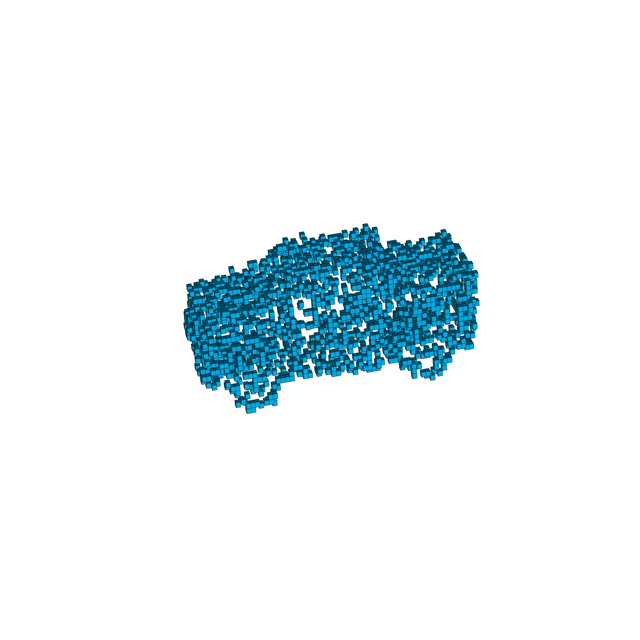}  & 
        \adjincludegraphics[trim={{0.28\width} {0.35\width} {0.18\width} {0.3\width}},clip,width=0.2\textwidth]{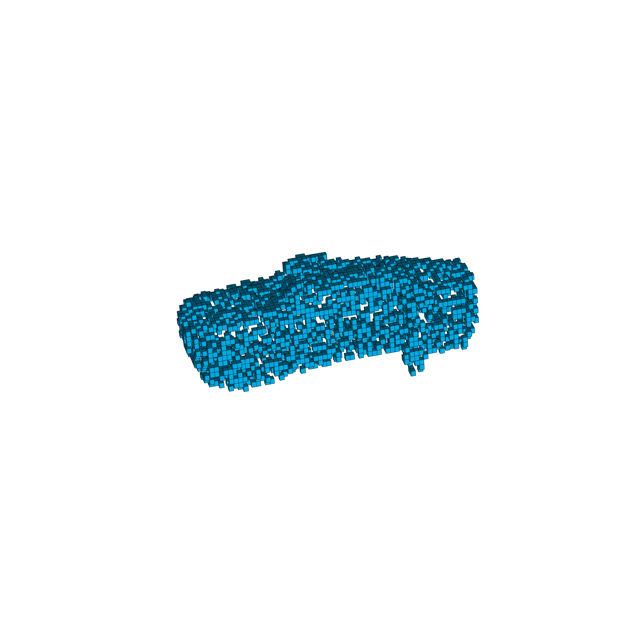}  \\
        & \multicolumn{5}{c}{\it ``3D model of a Chevrolet Tahoe pickup truck in black and red, available in multiple formats including OBJ and FBX.''}\\
        \multicolumn{6}{c}{}\\
        \multirow{3}{*}{Seg 50\%} & 
        \adjincludegraphics[trim={{0.28\width} {0.35\width} {0.18\width} {0.3\width}},clip,width=0.2\textwidth]{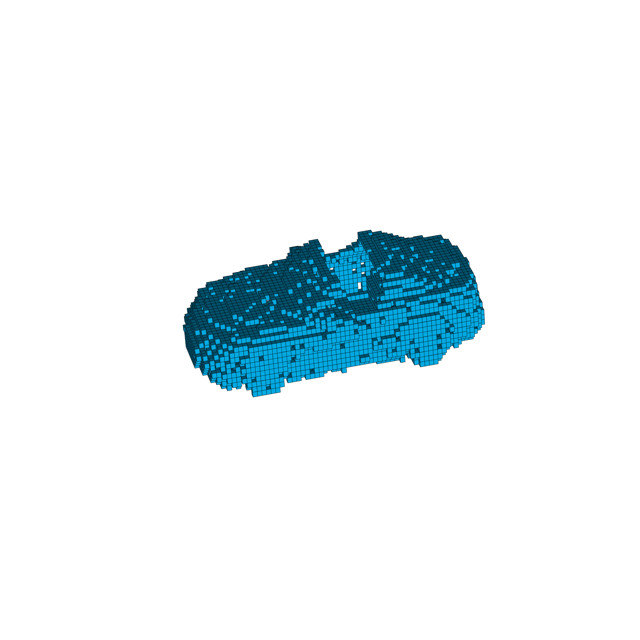} & 
        \adjincludegraphics[trim={{0.28\width} {0.35\width} {0.18\width} {0.3\width}},clip,width=0.2\textwidth]{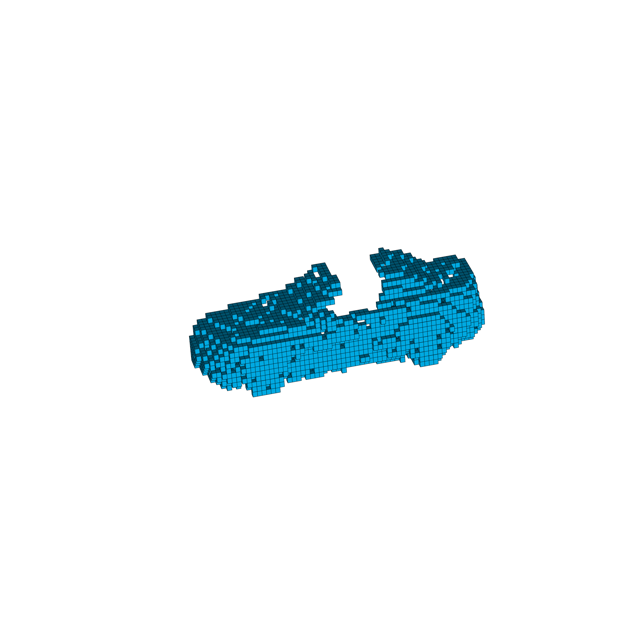} & 
        \adjincludegraphics[trim={{0.28\width} {0.35\width} {0.18\width} {0.3\width}},clip,width=0.2\textwidth]{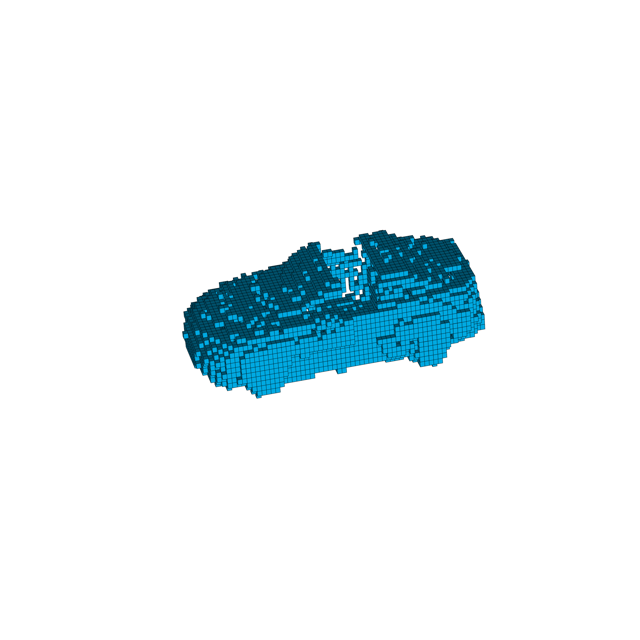} & 
        \adjincludegraphics[trim={{0.28\width} {0.35\width} {0.18\width} {0.3\width}},clip,width=0.2\textwidth]{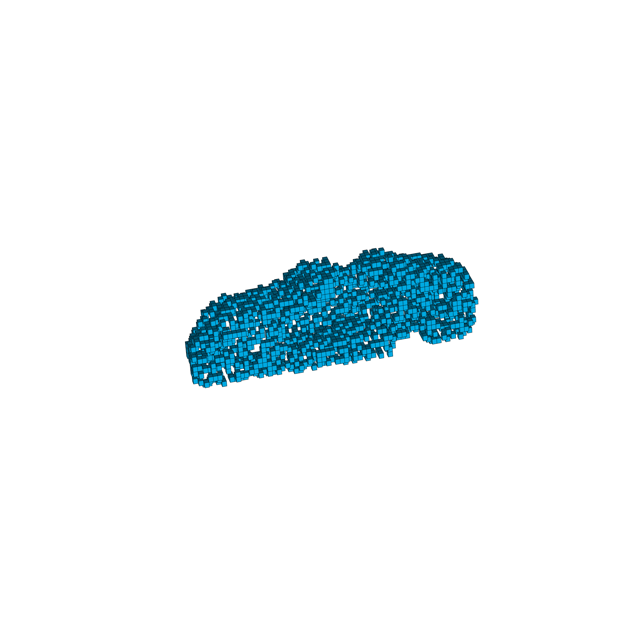}  & 
        \adjincludegraphics[trim={{0.28\width} {0.35\width} {0.18\width} {0.3\width}},clip,width=0.2\textwidth]{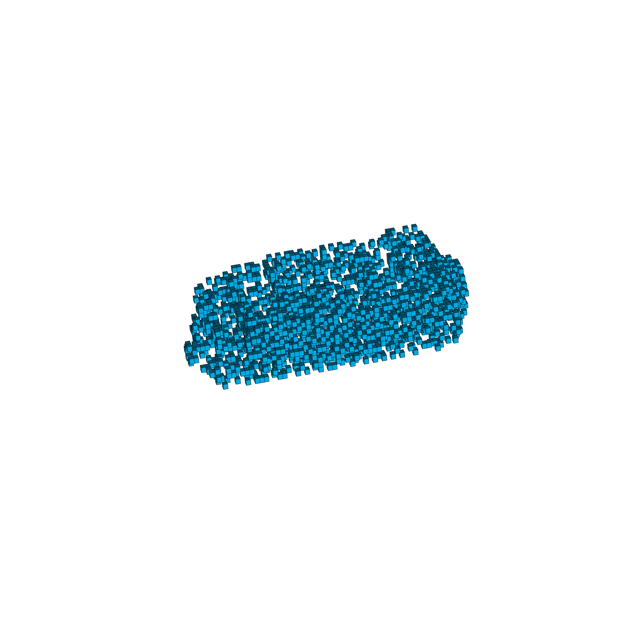} \\
        & \multicolumn{5}{c}{\it ``Royalty-free 3D model of a Mercedes-Benz SLK sports car''}\\
        \multicolumn{6}{c}{}\\
        \multirow{3}{*}{Seg 80\%} & 
        \adjincludegraphics[trim={{0.28\width} {0.35\width} {0.18\width} {0.3\width}},clip,width=0.2\textwidth]{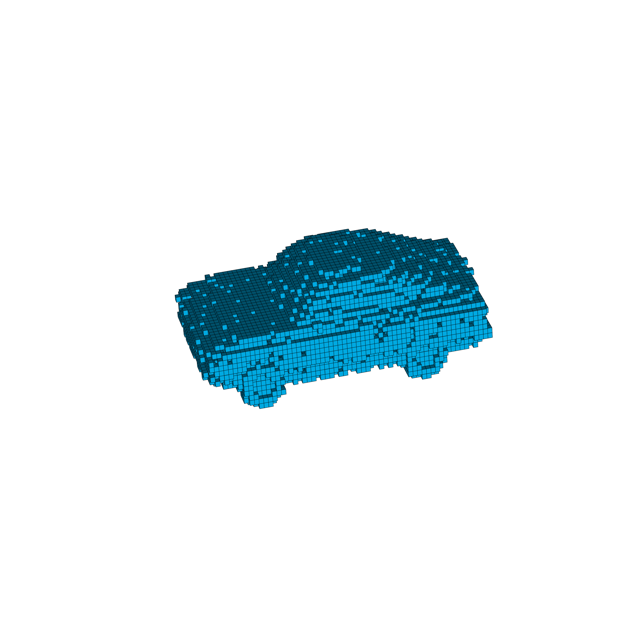} & 
        \adjincludegraphics[trim={{0.28\width} {0.35\width} {0.18\width} {0.3\width}},clip,width=0.2\textwidth]{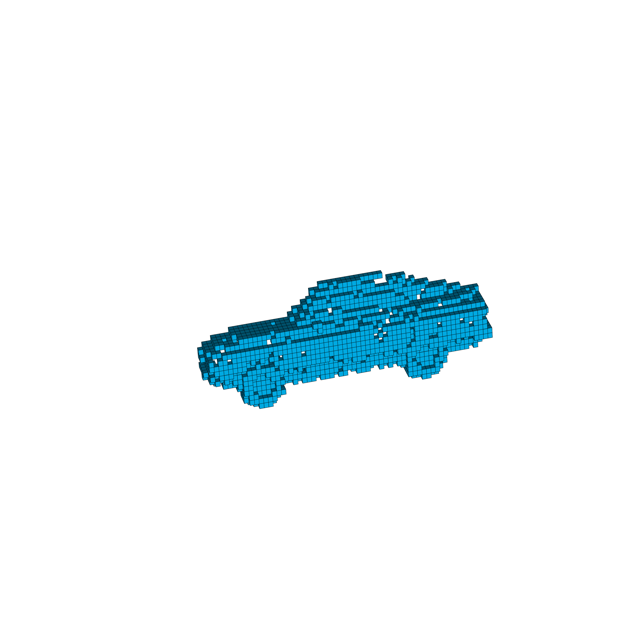} & 
        \adjincludegraphics[trim={{0.28\width} {0.35\width} {0.18\width} {0.3\width}},clip,width=0.2\textwidth]{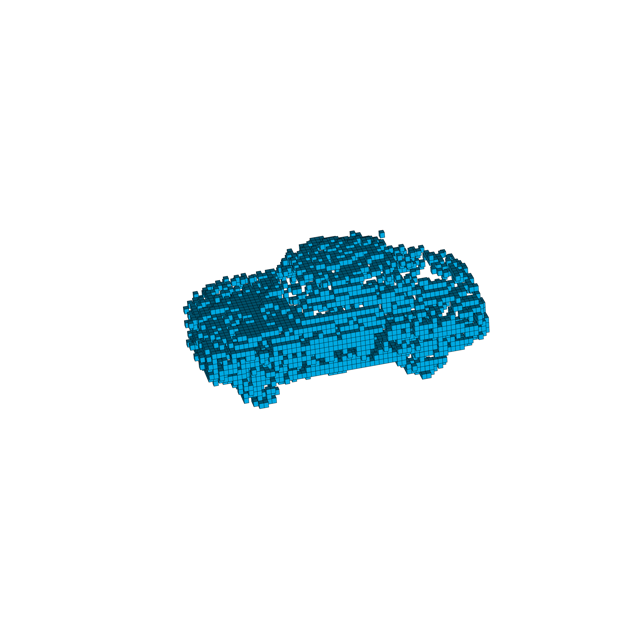} & 
        \adjincludegraphics[trim={{0.28\width} {0.35\width} {0.18\width} {0.3\width}},clip,width=0.2\textwidth]{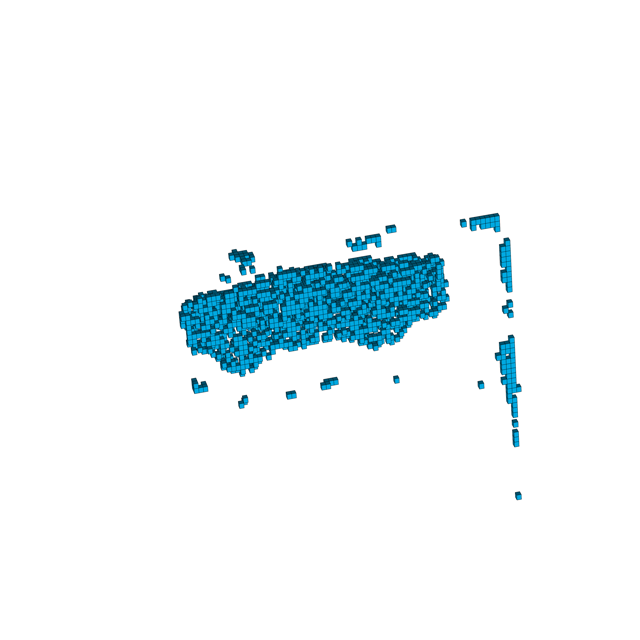} & 
        \adjincludegraphics[trim={{0.28\width} {0.35\width} {0.18\width} {0.3\width}},clip,width=0.2\textwidth]{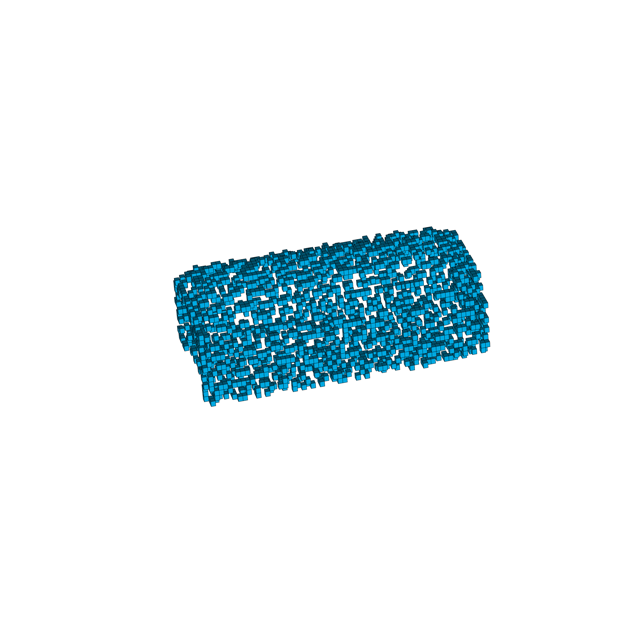} \\
        & \multicolumn{5}{c}{\it ``3D model of a muscle car with a hood, fenders, and a hood scoop.''}\\
        \multicolumn{6}{c}{}\\
        \multirow{3}{*}{Noise 1\%} & \adjincludegraphics[trim={{0.28\width} {0.35\width} {0.18\width} {0.3\width}},clip,width=0.2\textwidth]{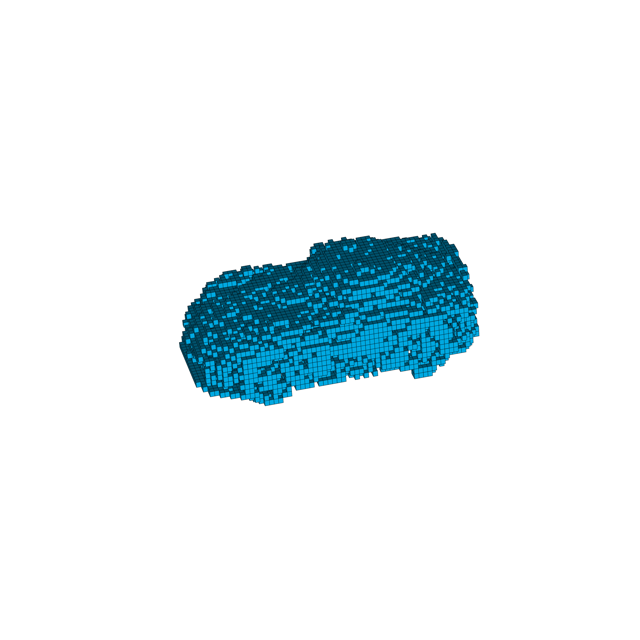} & 
        \adjincludegraphics[trim={{0.28\width} {0.35\width} {0.18\width} {0.3\width}},clip,width=0.2\textwidth]{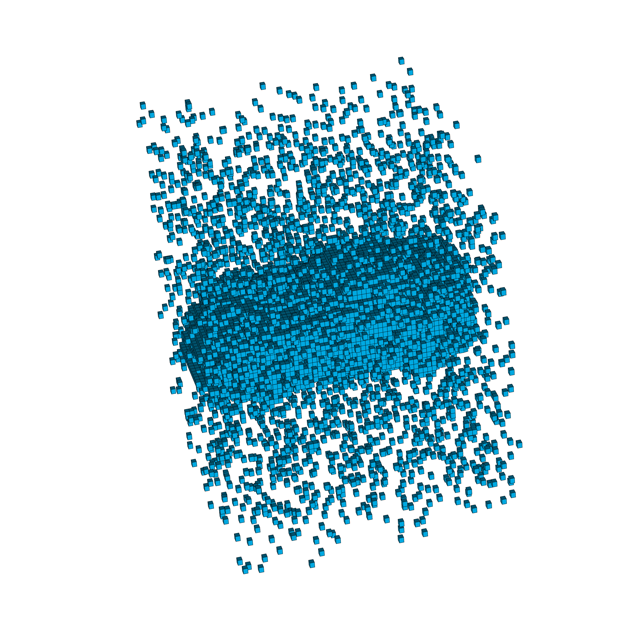} & 
        \adjincludegraphics[trim={{0.28\width} {0.35\width} {0.18\width} {0.3\width}},clip,width=0.2\textwidth]{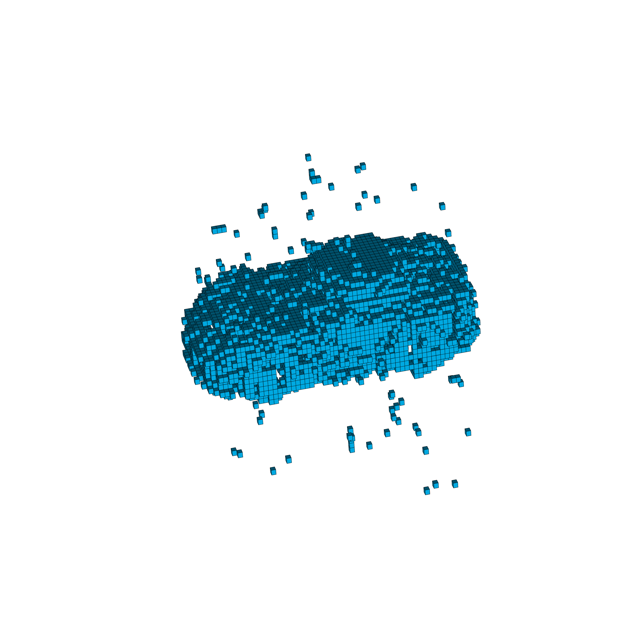} &
        N/A & 
        N/A \\
        & \multicolumn{5}{c}{\it ``3D model of a yellow Dodge Viper SRT10 sports car.''}\\
        \multicolumn{6}{c}{}\\
        \multirow{3}{*}{Noise 2\%} & 
        \adjincludegraphics[trim={{0.28\width} {0.35\width} {0.18\width} {0.3\width}},clip,width=0.2\textwidth]{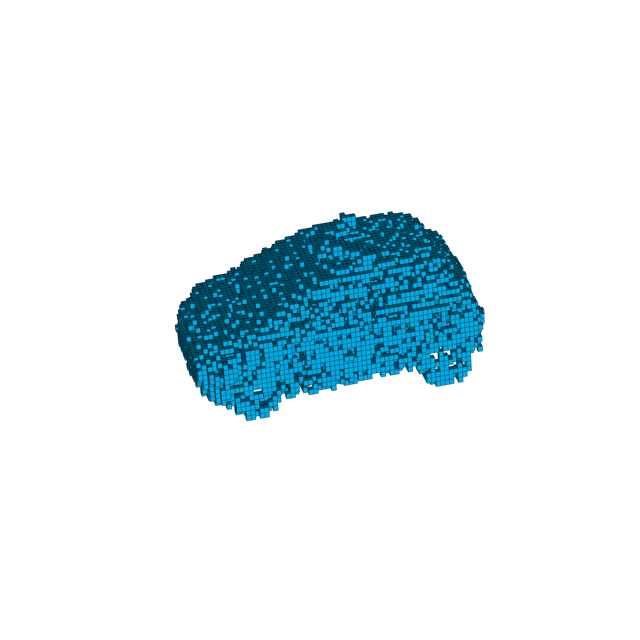} & 
        \adjincludegraphics[trim={{0.28\width} {0.35\width} {0.18\width} {0.3\width}},clip,width=0.2\textwidth]{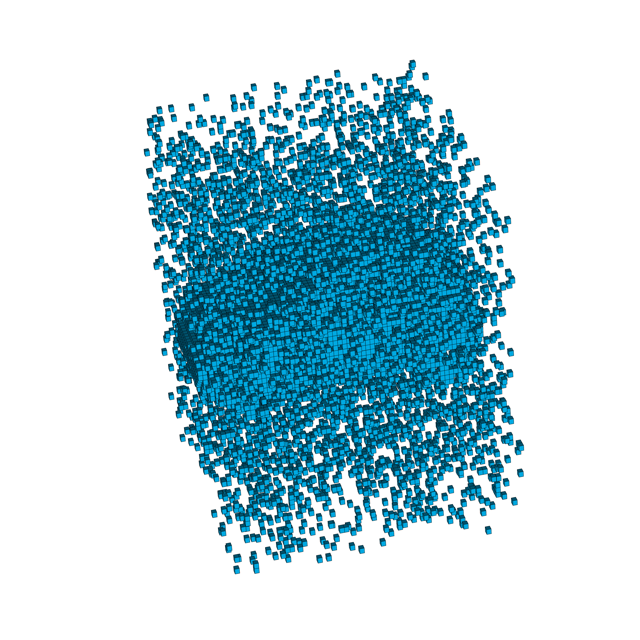} & 
        \adjincludegraphics[trim={{0.28\width} {0.35\width} {0.18\width} {0.3\width}},clip,width=0.2\textwidth]{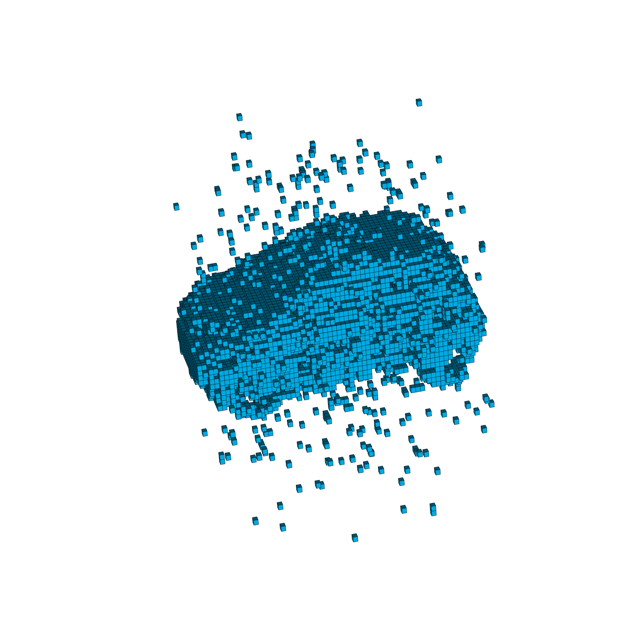} &
        N/A & 
        N/A \\
        & \multicolumn{5}{c}{\it ``A 3D model of a police car, available royalty-free, with previews from different angles.''}\\
        \multicolumn{6}{c}{}\\
        \bottomrule
    \end{tabular}%
    }
    \label{tbl:table_of_figures_car}
\end{table*}
% c778e4d590417da82205cec171865d70_gt.png

We compare our model with SDFusion~\cite{cheng2023sdfusion} and 3DQD~\cite{li20233dqd}, which are two state-of-the-art diffusion-based methods on conditional 3D completion tasks. SDFusion accepts multi-modality input for shape completion, so in our case, we only enable text-conditioned completion and enforce no image condition. 

The quantitative comparison results are shown in Tab.~\ref{tab:comparison_main}. Following~\cite{cui2024neusdfusion, li20233dqd}, we use Chamfer Distance (CD) and CLIP-s score for evaluation. For Chamfer Distance, we transform our volume representation into point clouds and use the coordinates in the voxel grids. For the CLIP-s score, we render 20 different-view 2D images around each volume and take the maximums among the CLIP feature scores between the images and the textual description. We test the performance of the models under various circumstances, by segmenting the different fractions of the object and adding random noise to the objects.
We also provide visualization results in Tab.~\ref{tbl:table_of_figures_plane} and Tab.~\ref{tbl:table_of_figures_car}. 

It is worth noticing that 3DQD is a label-conditional completion model that does not include detailed text control during inference, leading to large variance in the prediction results.
Though SDFusion contains text control in completion, it adopts BERT~\cite{devlin2018bert} for text encoding, thereby suffering from complex text understanding. 

More qualitative results 
are presented in Appendix~\ref{supp-more-results}.
% will be presented in our Gallery.
% More qualitative results are presented in Appendix~\ref{supp-more-result}.
\section{Ablation Study}
\label{sec-ablation}

\begin{table*}
  \centering
  \begin{minipage}[t]{0.49\linewidth}
    \centering
    \caption{{\bf Comparison using different LLM structures.} We utilize two LLMs, namely Mistral-7B and Gemma-2B, to train our whole pipeline with the same dataset as before. For comparison purposes, we demonstrate the Chamfer Distance (CD) and CLIP-s score on data containing 1\% noise and 20\% mask.}
    \resizebox{\linewidth}{!}{
        \begin{tabular}{ccccc} \toprule  \multirow{2}{*}{\textbf{Methods}} & \multicolumn{2}{c}{{Seg 20\%}} &  \multicolumn{2}{c}{{Noise 1\%}} \\ \cmidrule(lr){2-3} \cmidrule(lr){4-5} & \small{CD.}$\downarrow$ & \small{CLIP-s.}$\uparrow$ & \small{CD.}$\downarrow$ & \small{CLIP-s.}$\uparrow$\\
        \midrule
        {Mistral-7B} & {\bf 10.96} & 27.80\% & 16.03& 23.92\% \\
        {Gemma-2B}  & 11.19& {\bf 28.08\%} & {\bf 10.64} & {\bf 26.34\%} \\
        \bottomrule
        \end{tabular}
    }
    \label{tab:comparison_ablation_llm}
  \end{minipage}
  \hfill % This adds an optional space between the two minipages, if needed
  \begin{minipage}[t]{0.49\linewidth}
    \centering
    \caption{{\bf Comparison on different voxel volume resolutions.} We selected two different voxel volume resolutions, namely $H=W=D=64$ and $H=W=D=72$. We can see the two resolutions have comparable results. Thanks to our patchification method that enables each patch to be processed and generated independently, our method perfectly scales when the resolution is larger.}
    \resizebox{\linewidth}{!}{
        \begin{tabular}{ccccc} \toprule  \multirow{2}{*}{\textbf{Methods}} & \multicolumn{2}{c}{{Seg 20\%}} &  \multicolumn{2}{c}{{Noise 1\%}} \\ \cmidrule(lr){2-3} \cmidrule(lr){4-5} & \small{CD.}$\downarrow$ & \small{CLIP-s.}$\uparrow$ & \small{CD.}$\downarrow$ & \small{CLIP-s.}$\uparrow$\\
        \midrule
        {Resolution $64^3$} & 10.96 & 27.80\% & 16.03 & 23.92\% \\
        {Resolution $72^3$}  & 12.33& 27.38\% & 12.11& 27.05\% \\
        \bottomrule
        \end{tabular}
    }
    \label{tab:comparison_ablation_resolution}
  \end{minipage}
\end{table*}

\subsection{Comparison of different LLM architectures}
\label{sec-ablation-llm}
LLM is one of the most important components in our model, whose ability to capture multi-modal semantic information and token relations may greatly affect the performance of the entire pipeline. To investigate the effect of different LLMs on the performance of our model, we re-trained our method on Mistral-7B~\cite{mistral7b} and Gemma-2B~\cite{gemma}. To ensure a fair comparison, we keep the LoRA ranks for the two models to be the same. From Tab.~\ref{sec-ablation-llm}, we find that under this setting, the two LLMs demonstrate similar results on mask completion, while the performance of Gemma-2B is better than Mistral-7B on denoising tasks while slightly worse in completion from masked inputs. Higher performance can be expected if larger LLM models or higher LoRA ranks are deployed.

% From the results in~\ref{sec-ablation-llm}, we can observe that the ability of LLM indeed severely affects the performance of our model, with the 7B model outperforming the 2B model. This result also aligns with common LLM benchmarks such as~\cite{open-llm-leaderboard}. 
% {\it (Remove the last sentence if we don't find some benchmarks that support ``Mistral-7B is better than Gemma-7B'')}

\subsection{Scalability on Higher Resolution Voxel Volumes}
\label{sec-ablation-scalability}
Our VAE encodes and decodes each patch of the 3D model individually, thus enabling our model to scale to higher voxel resolutions.
To demonstrate the scalability, we have expanded upon our previous experiments by setting $H=W=D=64$, and further increasing the voxel resolution to $H=W=D=72$, while maintaining the patch size at $8$. By fixing the patch size, we can ensure that the fine details of the data remain consistent as we increase the input scale. After this operation, the sequence length of each 3D object will increase from $512$ to $729$. We can see from Tab.~\ref{tab:comparison_ablation_resolution} that the two resolutions have comparable results, indicating our method can successfully generalize to higher voxel volume resolutions.

% (Add results)
% (I think it is better to be a subsection of experiment)

\section{Limitation}

Our method has been proven effective on small LLMs such as Mistral-7B. However, due to the limitation of computational resources, LLMs with stronger capability to understand long sequences, such as 70B or larger scale models, are not employed in our model. Thus, huge potential of our method is still left to probe. Additionally, though we verify the effectiveness of our method by patchification on voxel volumes, which can be intuitively extended to point clouds and SDFs, it is still very hard to employ it on those nascent 3D representations like NeRF~\cite{mildenhall2020nerf} and 3D Gaussian Splatting (3DGS~\cite{kerbl3Dgaussians}) that encoded 3D in an implicit (MLP weights for NeRF and Gaussians for 3DGS). More investigations on how to patchify such representations are left in future works.

\section{Conclusion}
In this paper, we present \ourwork, which combines text and 3D through LLMs to achieve text-guided 3D completion. We introduce a novel approach called patchification to incorporate 3D models into LLMs, and adopt a two-stage training process that allows LLMs to understand input incomplete 3D models and generate entire 3D models. Such an approach also allows an independent encoding and decoding process, thereby ensuring the scalability of our method.  Experiments on the ShapeNet dataset validate that our method surpasses the state-of-the-art methods in the 3D completion task. Moreover, our model can achieve satisfactory results from noisy 3D inputs, a practical issue in real 3D data capture.

{
\small
\bibliographystyle{plain}
\bibliography{egbib}
}

\newpage
\appendix

% \section{Appendix / supplemental material}

% Optionally include supplemental material (complete proofs, additional experiments and plots) in appendix.
% All such materials \textbf{SHOULD be included in the main submission.}

%%%%%%%%%%%%%%%%%%%%%%%%%%%%%%%%%%%%%%%%%%%%%%%%%%%%%%%%%%%%

\section{Implementation Details}

\paragraph{Dataset preparation}
We randomly selected 3130 data from ShapeNetCore~\cite{chang2015shapenet}, and randomly split data of each category into 90\% training data and 10\% testing data. 

\paragraph{Model structure}

We present the hyperparameters in Tab.~\ref{tab:hyperparameter}. Those values can determine the detailed structures of each component in our pipeline.

\begin{table}[H]
\renewcommand{\arraystretch}{1.5}
\centering
\begin{tabular}{cc}
\toprule
\textbf{\textsc{Hyperparameter}}                                        & \textbf{\textsc{Value}} \\ 
\midrule
VAE Encoder Convolution Layer Num                                          & 2            \\ \hline
VAE Decoder Convolution Layer Num                                          & 2            \\ \hline
VAE Hidden Dimension
& 64            \\ \hline
VAE Latent Size                                                & 128            \\ \hline
LoRA Rank                                                      & 32             \\ \hline
LoRA Alpha                                                     & 32             \\ \hline
LoRA Dropout                                                   & 0.05           \\ \hline
LLM Type                                                      & Mistral-7B       \\ \hline
Output Projection Transformer Encoder Layer Num                & 2              \\ \hline
Output Projection Transformer Decoder Layer Num                & 2              \\ \hline
{Output Projection Transformer Feedforward Dimension} & 2048           \\ \hline
{Output Projection Transformer Num Heads}      & 4              \\
\bottomrule\\
\end{tabular}
\caption{Hyperparameters used to configure model structure.}
\label{tab:hyperparameter}
\end{table}

\paragraph{Training configuration}
We train our patch VAE on 2 RTX 4090 cards for 100 epochs and batch size using ShapeNet voxels volumes with resolution $64^3$ and adopt the same VAE for voxels with different resolutions. This training takes approximately 2 hours. The model is trained with AdamW~\cite{adamw} optimizer with a learning rate $3e^{-4}$.

Our input projection and output projection models are trained on 8 RTX 6000 Ada GPU cards until converge, for around 100 and 500 epochs respectively. For Mistral-7B, these processes take around 1 hour and 18 hours, respectively, while for Gemma-2B, the numbers go down to 20 minutes and 8 hours. Both stages are trained with AdamW optimizer as well, with input projection training using a learning rate of $3e^{-4}$ and output projection training using $5e^{-4}$ or $5e^{-5}$, depending on the LLM size.

\section{Examples of Ground-Truth and Predicted Captions}

Here we present some examples of ground-truth captions and the captions predicted by our LLM model during input projection layer training. We notice that the captions are not perfect, while they provide adequate semantic meanings.

{\sc Predicted Caption 1}: 3AD model of a ", featuring a exterior such as wings, fuselage, and, andinglets, and, and, andilerons, and flaps" with for  3 and7-400 and 747-800 variants.",

{\sc Ground-Truth Caption 1}: "3D model of Boeing aircraft, featuring detailed components such as wings, fuselage, tail, winglets, rudder, elevators, ailerons, and flaps, available in both 747-400 and 737-800 variants."

{\sc Predicted Caption 2}: 3D model of a rectangular 737- featuring a fuselage body with a wings, and tail tail tail, and, and, andilerons, and a gear.,

{\sc Ground-Truth Caption 2}: 3D model of a Boeing 747, featuring a cylindrical fuselage, elliptical wings, a truncated cone tail, rudder, elevators, ailerons, and landing gear.

{\sc Predicted Caption 3}: 3D model of of a guitars  Ghostcar aircraft jets, including a-A-18E F-14., with for download3ds Max. OBJ.,

{\sc Ground-Truth Caption 3}: 3D model collection of various Phantom and Super Hornet fighter jets, including F/A-18 and F-16 variants, available for 3ds Max and Maya.

{\sc Predicted Caption 4}: 3D model of a rectangular 737-800 aircraft a fuselage dome, with a conelate spher, and a- a. steel.,

{\sc Ground-Truth Caption 4}: 3D model of a Boeing 747-400 featuring a spherical fuselage shell, truncated oblate wings, and made of aluminum and steel.

{\sc Predicted Caption 5}: 3  with a cylindrical, a, and, and, and,, and landing landing gear.,

{\sc Ground-Truth Caption 5}: A spaceship featuring a wing, fuselage, tail, propeller, rotor blade, and retractable landing gear.

{\sc Predicted Caption 6}: 3D model of a electric with a, a, and, and, and, andilerons, and gear, and a.,

{\sc Ground-Truth Caption 6}: 3D model of an aircraft featuring wings, fuselage, tail, rudder, elevators, ailerons, landing gear, and propeller.

{\sc Predicted Caption 7}: 3Aalty-free 3D model of a female 737-400 aircraft featuring a exterior such as a fuselage, fuselage, and tail.",

{\sc Ground-Truth Caption 7}: "Royalty-free 3D model of a Boeing 747-400, featuring detailed components such as a wing, fuselage, and tail."

{\sc Predicted Caption 8}: 3D model of a rectangularliner with a fuselage wing, a fuselage, and, and, andders, and, and gear, and landing.,

{\sc Ground-Truth Caption 8}: 3D model of a jet plane featuring a delta wing, triangular fuselage, tail, fin, rudders, propeller, landing gear, and hull.

% \section{Comparison of Mask Strategies}
% As mentioned in Sec.~\ref{sec-method-input-proj}, our model can cope with different types of masks including random masks, plane masks, and random noise. 
% % We demonstrate the performance of our models under various circumstances. 
% Here we present the result of those mask strategies with different mask/noise ratios to provide readers with examples of how they look like.

\section{Illustrations of Data Augmentation}
\label{sec-supp-data-aug}

\paragraph{3D Data Augmentation}

\begin{figure}[H]
    \centering
    \includegraphics[width=0.9\linewidth]{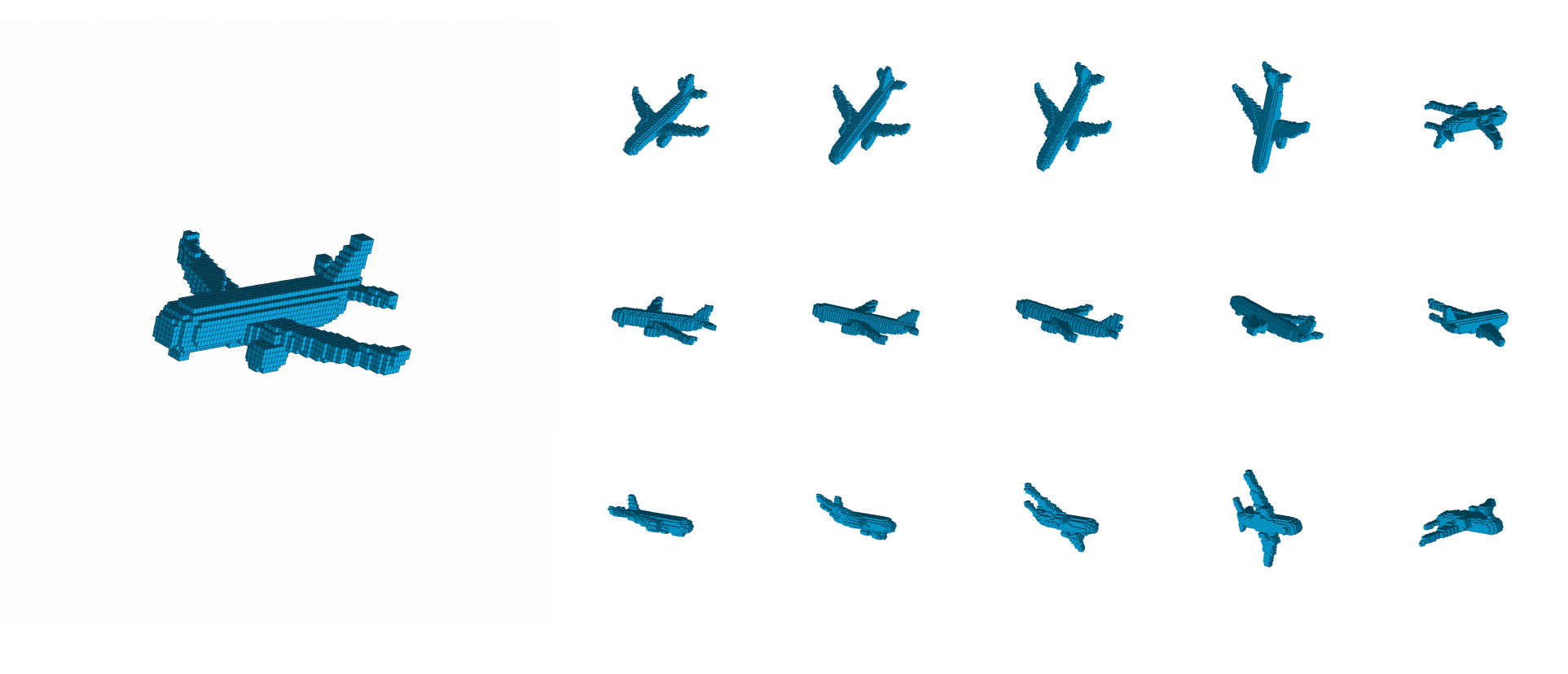}
    \caption{3D data augmentation example result of an airplane.}
    \label{fig:supp-3d-aug}
\end{figure}

During training, we performed 3D augmentation by randomly rotating each 3D object with a different angle with respect to either $x$, $y$, or $z$ axis. Figure~\ref{fig:supp-3d-aug} visualizes this result.

\paragraph{Caption Augmentation}

During training, we leveraged Cap3D~\cite{luo2024scalable} to generate ground-truth captions for every 3D model. To perform caption augmentation, we run Cap3D for three times, using GPT-4-Turbo, GPT-4-Turbo with another seed, and ChatGPT (GPT-3.5). Here we present the captions generated by them.

\begin{figure}[H]
    \centering
    \includegraphics[width=0.24\linewidth]{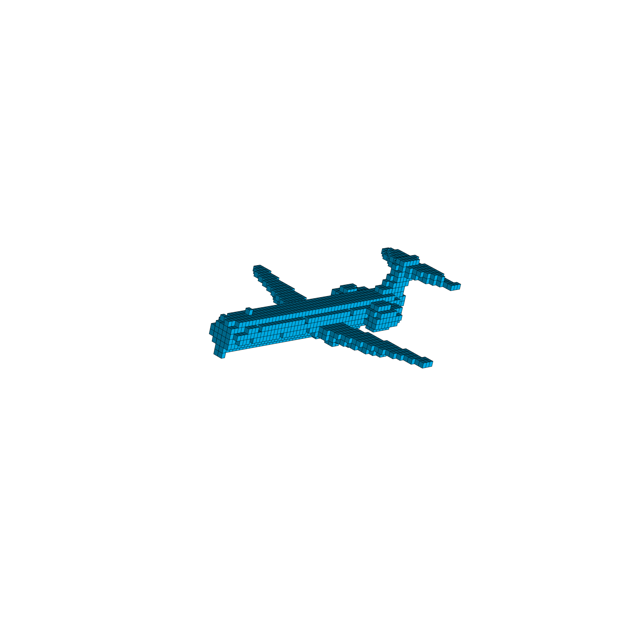}
    \includegraphics[width=0.24\linewidth]{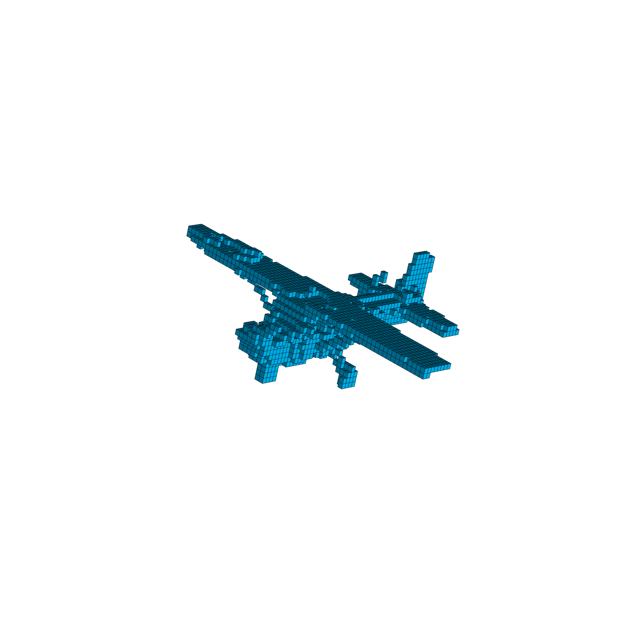}
    \includegraphics[width=0.24\linewidth]{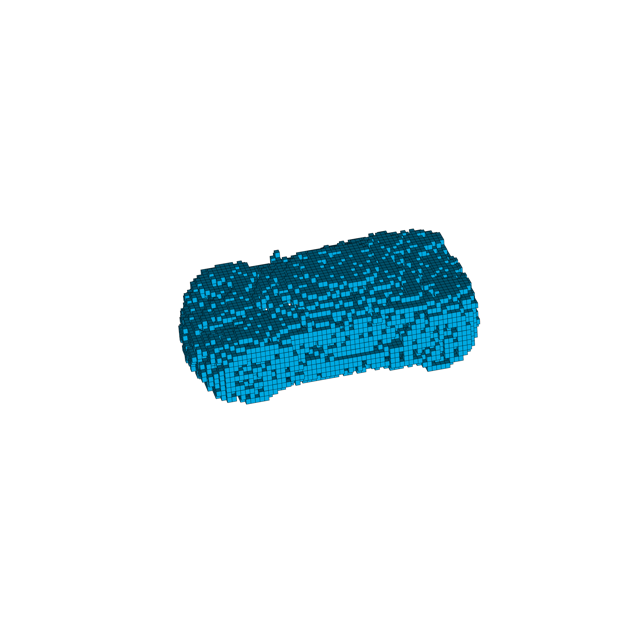}
    \includegraphics[width=0.24\linewidth]{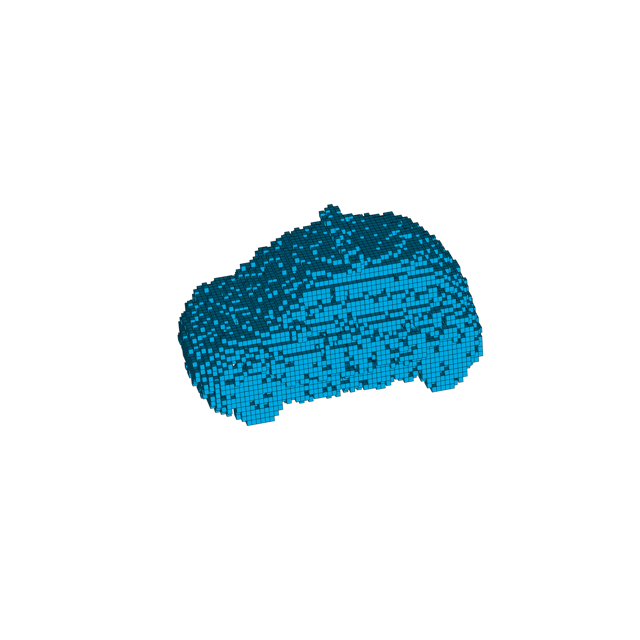}
    \caption{From left to right: Object 1, 2, 3, 4.}
    \label{fig:supp-cap-aug}
\end{figure}

Captions for Object 1:
\begin{itemize}
    \item [] {\sc GPT-4 Seed 1:} Royalty-free 3D model of a Boeing 747-400 featuring detailed components including a cylindrical fuselage, delta wings, tail, rudder, elevators, and ailerons.
    \item [] {\sc GPT-4 Seed 2:} Royalty-free 3D model of a Boeing 747-400 featuring a cylindrical fuselage, delta wings, a tail, rudder, elevators, and ailerons, representing a four-engine jet airliner.
    \item [] {\sc ChatGPT:} Boeing 747-400 3D model featuring a fuselage, wings, and tail, a jumbo jet with a delta wing design and four engines.
\end{itemize}

Captions for Object 2:
\begin{itemize}
    \item [] {\sc GPT-4 Seed 1:} A 3D model of a two-seater, single-engine RC airplane featuring a four-bladed, fixed-pitch propeller, retractable tricycle landing gear, and control surfaces including ailerons, rudder, and elevator.
    \item [] {\sc GPT-4 Seed 2:} A 3D model of a small, two-seater, single-engine RC airplane featuring a retractable tricycle landing gear, a fixed-pitch, four-bladed propeller, and control surfaces including wings, fuselage, tail, rudder, elevator, and ailerons.
    \item [] {\sc ChatGPT:} 3D model of a two-seater single-engine airplane with retractable landing gear and a four-bladed propeller.
\end{itemize}

Captions for Object 3:
\begin{itemize}
    \item [] {\sc GPT-4 Seed 1:} Royalty-free 3D model of a blue McLaren MP4-12C sports car with polygonal geometry.
    \item [] {\sc GPT-4 Seed 2:} Royalty-free 3D model of a McLaren MP4-12C sports car with polygonal geometry.
    \item [] {\sc ChatGPT:} 3D model of a McLaren MP4-12C sports car.
\end{itemize}

Captions for Object 4:
\begin{itemize}
    \item [] {\sc GPT-4 Seed 1:} 3D model of a police car, available royalty-free, featuring detailed polygonal geometry.
    \item [] {\sc GPT-4 Seed 2:} 3D model of a police car, featuring detailed polygonal vertices and edges, available royalty-free.
    \item [] {\sc ChatGPT:} A detailed 3D model of a police car.
\end{itemize}

% \section{More Qualitative Results}
% \label{supp-more-result}

% \section{Unlock Better Quality via Iterative Completion}
% We also found that our model is able to refine the results without further training, by directly passing the output from the last step and the caption into our model. We show this using the denoising example as the changes are more obvious. After applying the denoising twice, we observe that the quality in Fig.~\ref{fig:iterative_fig} has significantly increased.

\section{Unlock Better Quality via Iterative Completion}
\begin{figure}[H]
    \centering
    \includegraphics[width=0.99\linewidth]{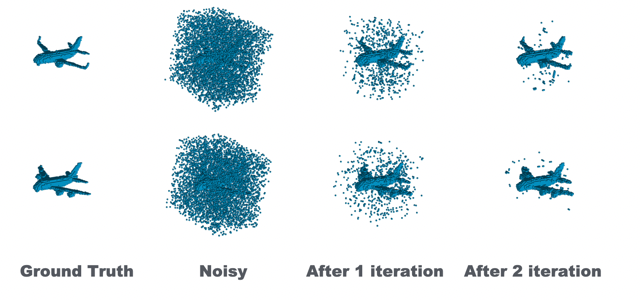}
    \caption{Result on Iterative Denoising}
    \label{fig:iterative_fig}
\end{figure}

We also found that our model is able to refine the results without further training, by directly passing the output from the last step and the caption into our model. We show this using the denoising example as the changes are more obvious. After applying the denoising twice, we observe that the quality in Fig.~\ref{fig:iterative_fig} has significantly increased.

\section{More Results}
\label{supp-more-results}
\begin{figure}[H] 
 \centering 
\leftfig{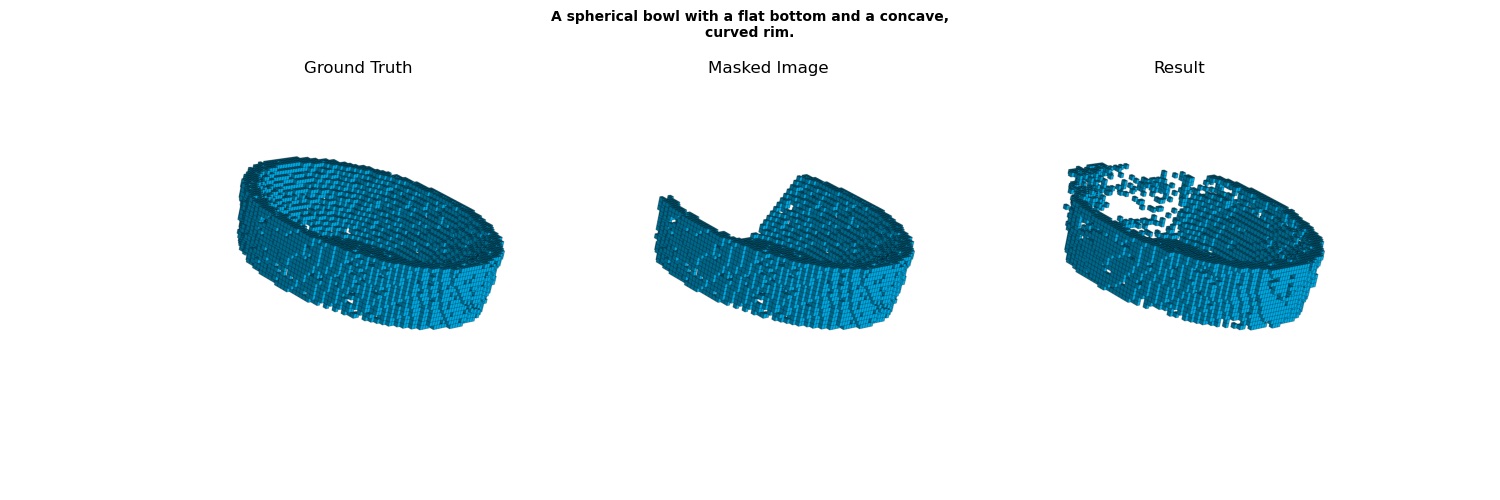}
\lrightfig{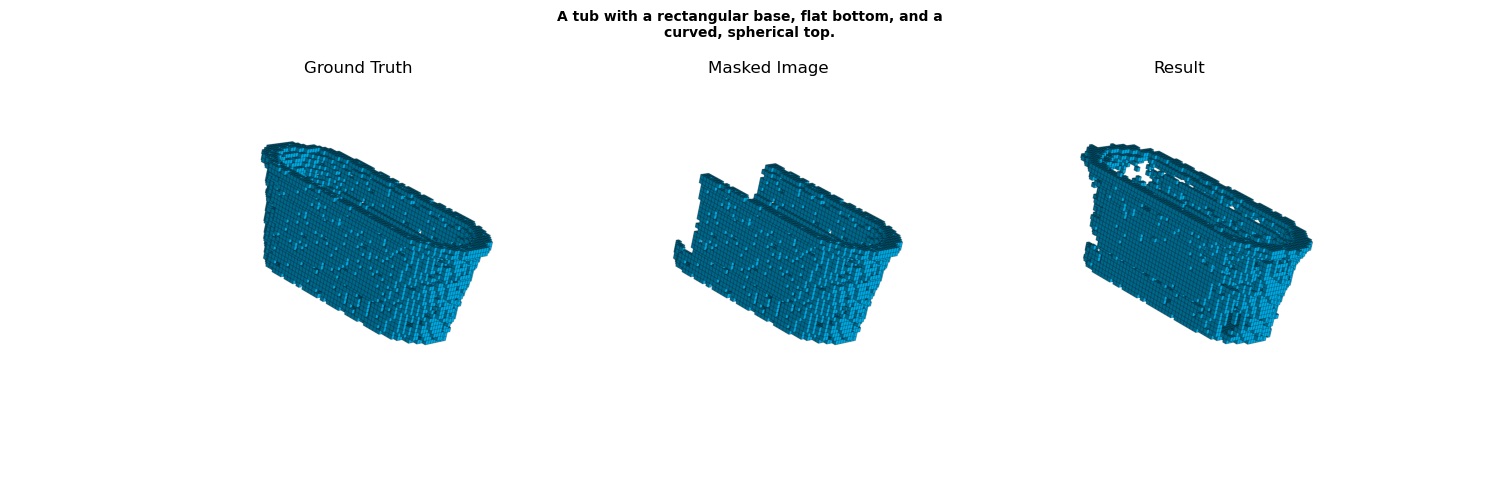}
\leftfig{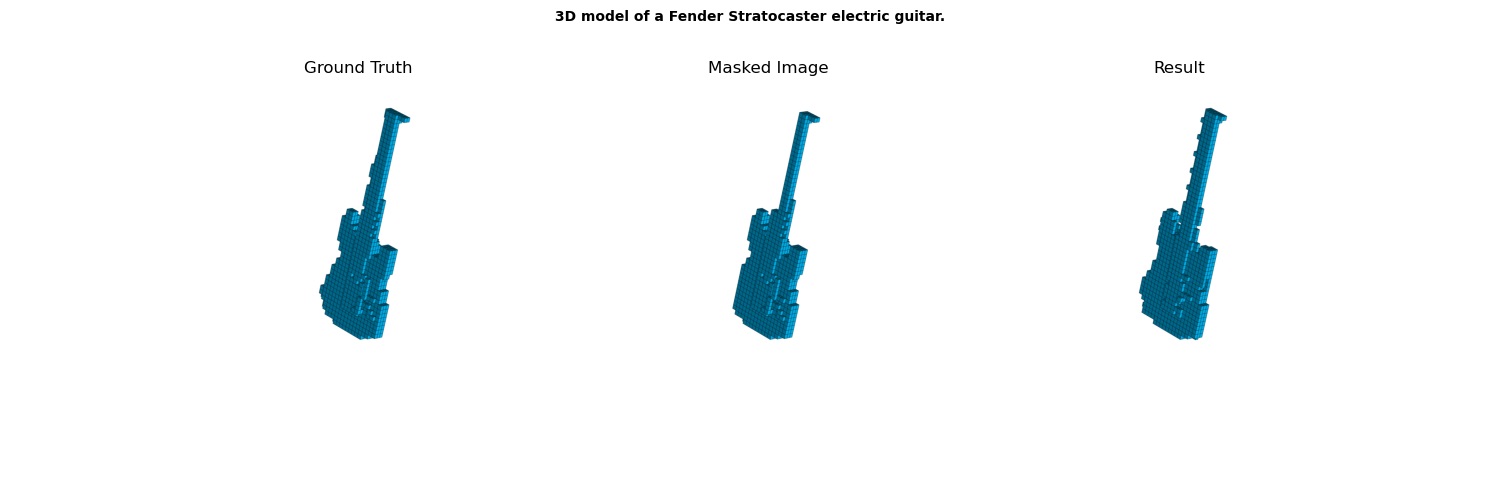}
\lrightfig{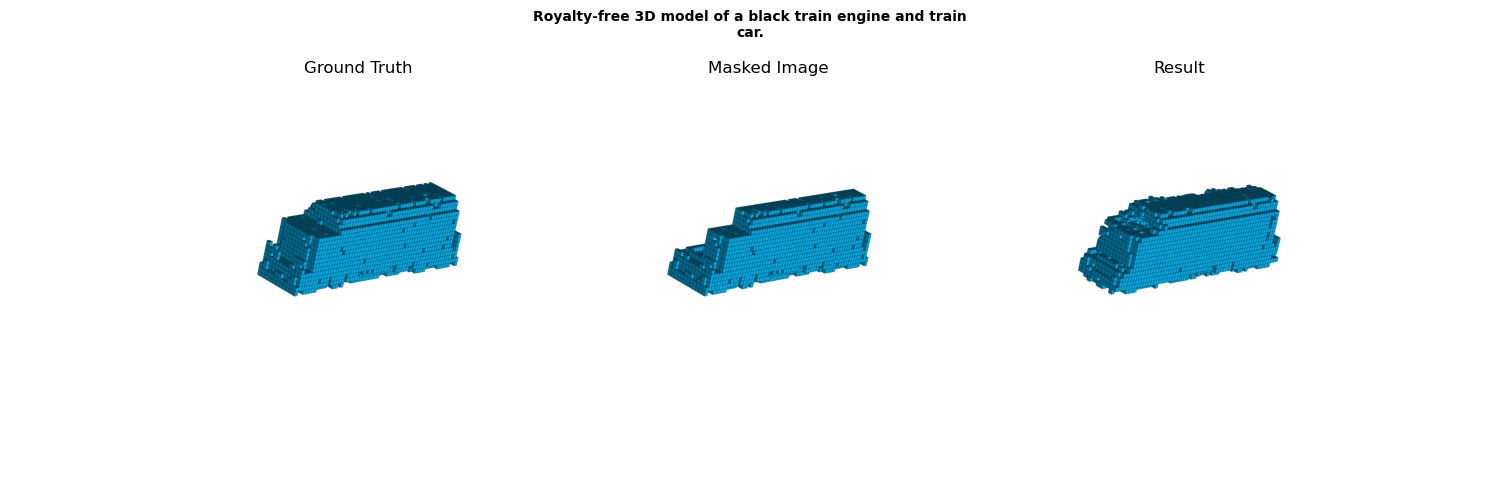}
\leftfig{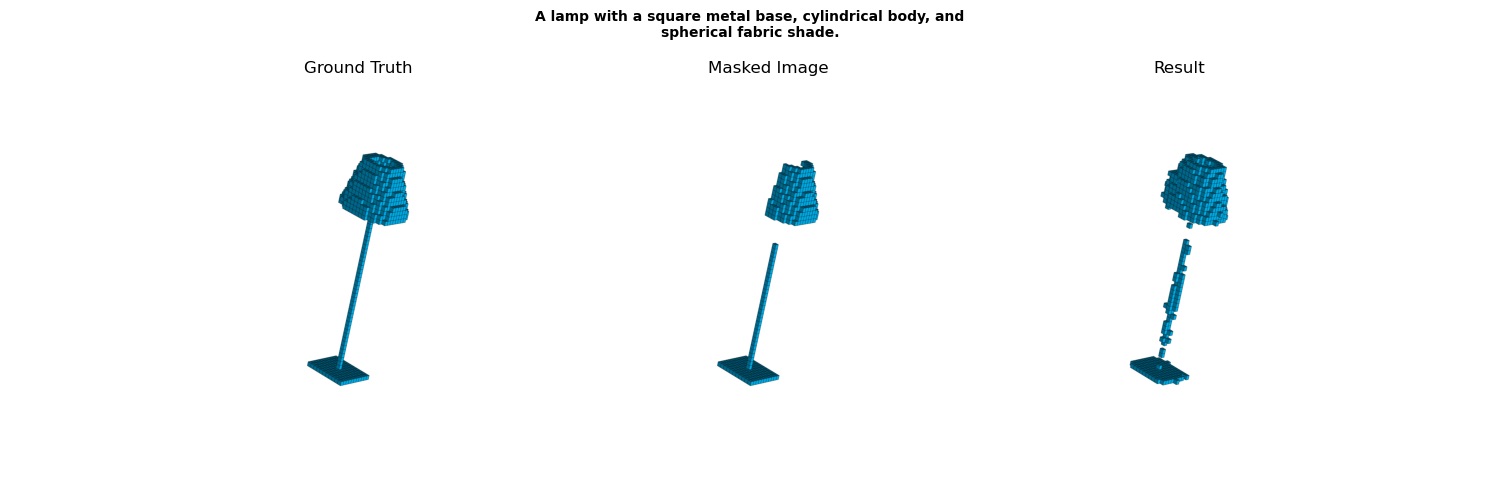}
\lrightfig{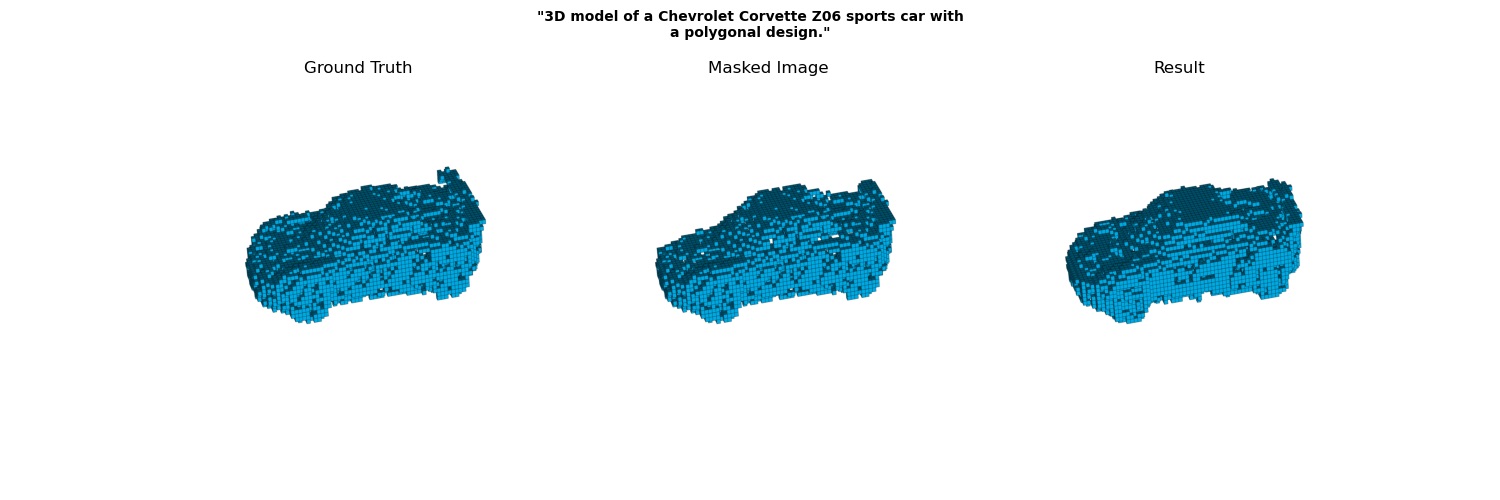}
\leftfig{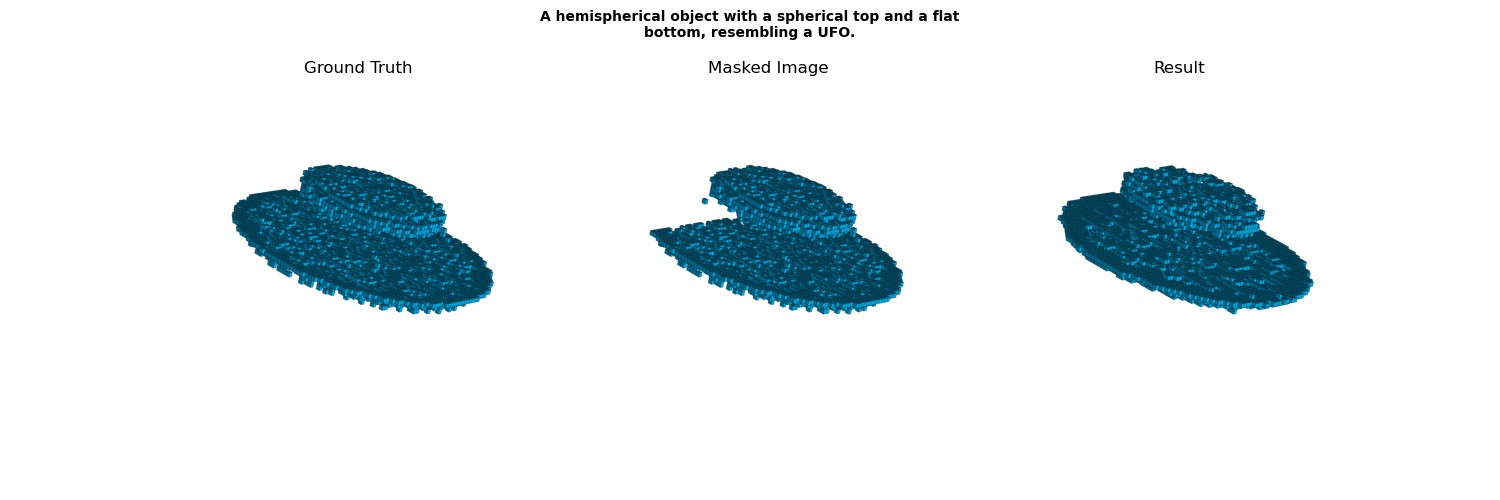}
\lrightfig{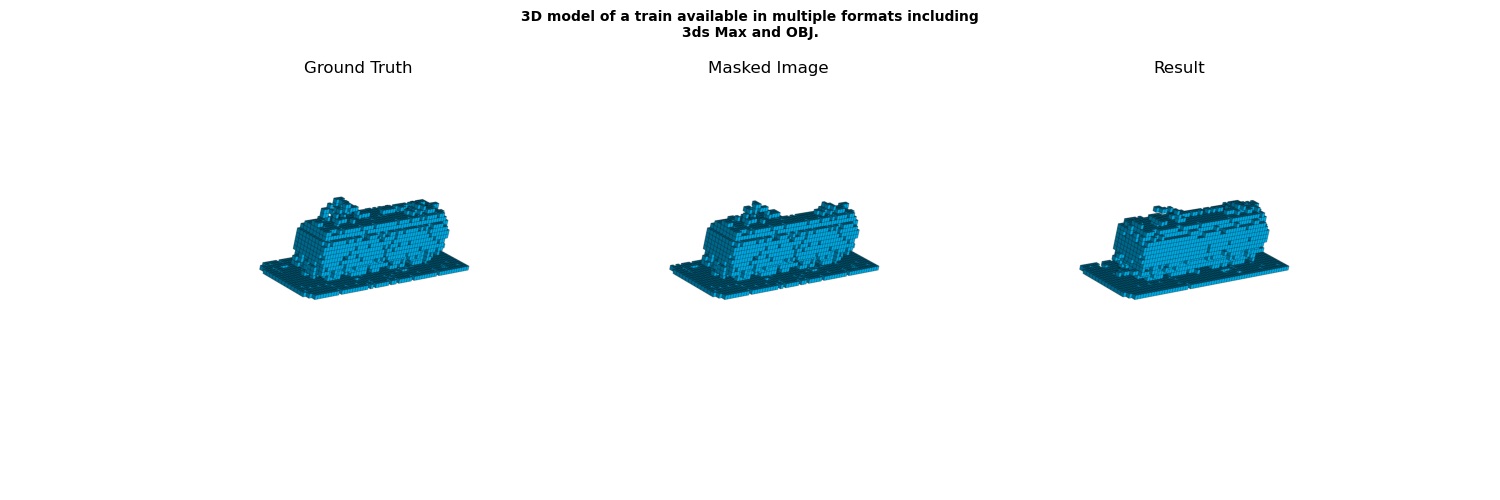}
\leftfig{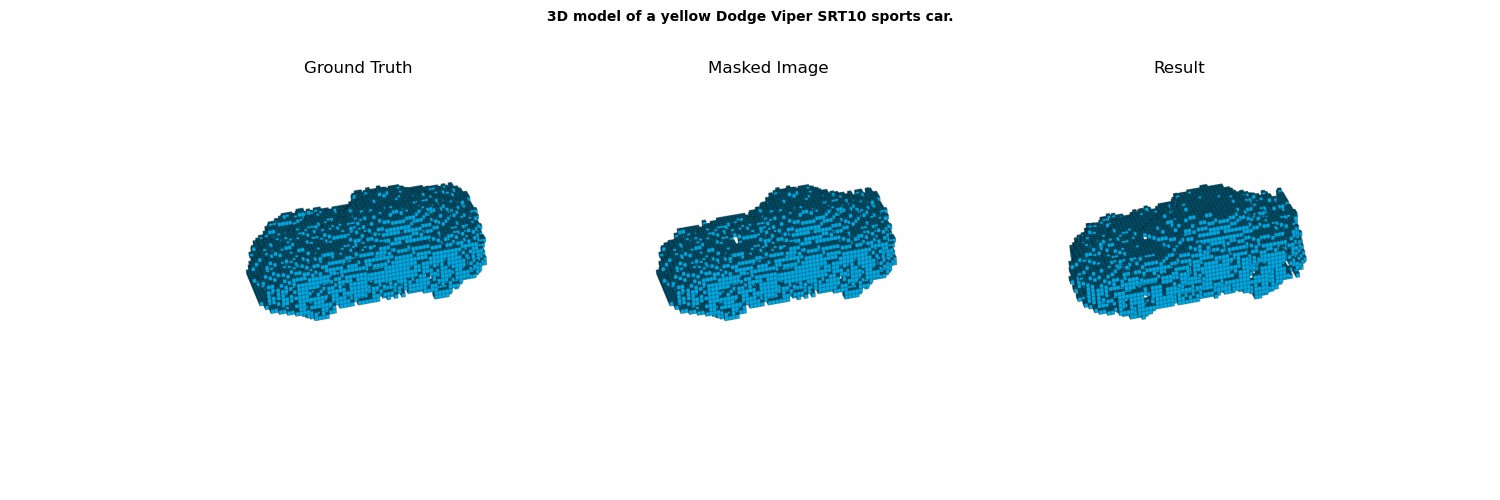}
\lrightfig{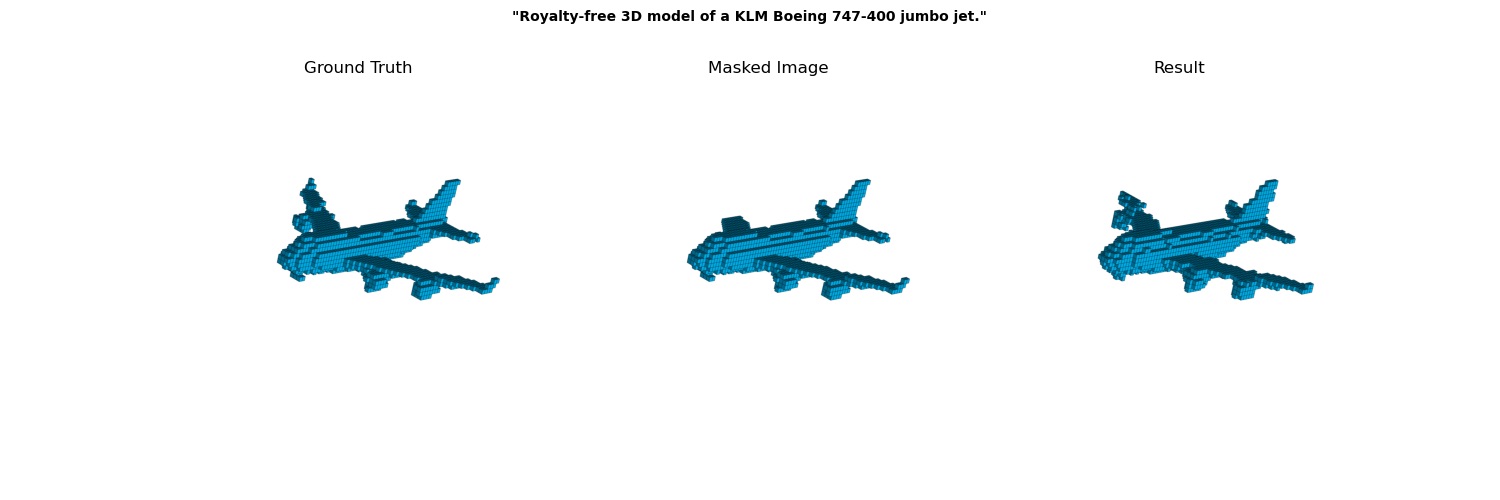}
\leftfig{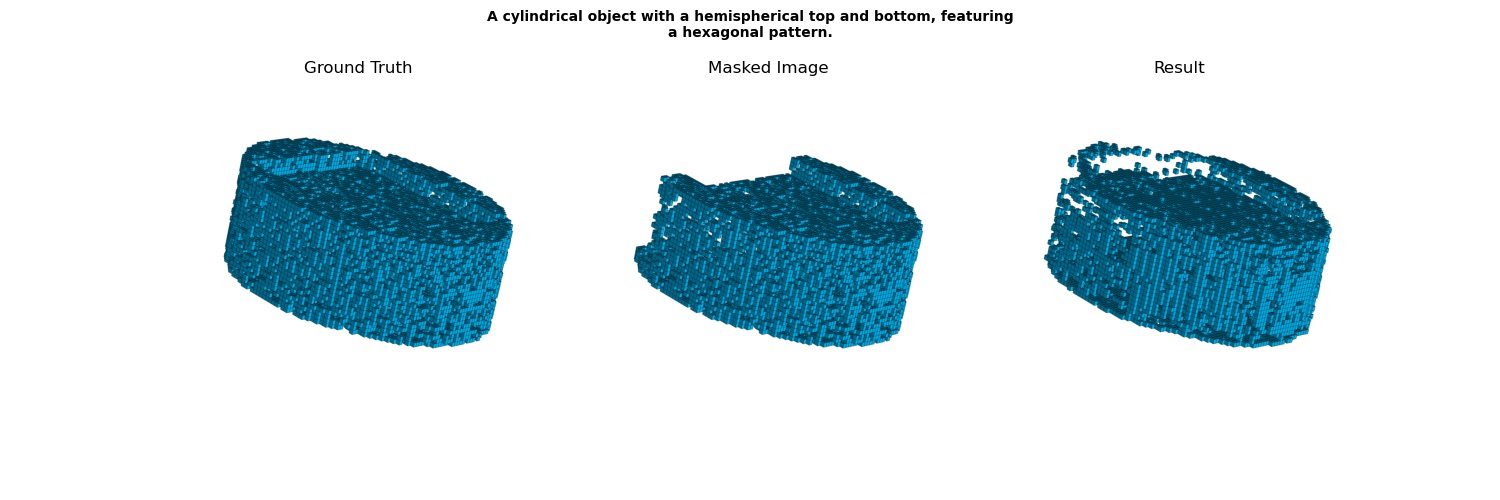}
\lrightfig{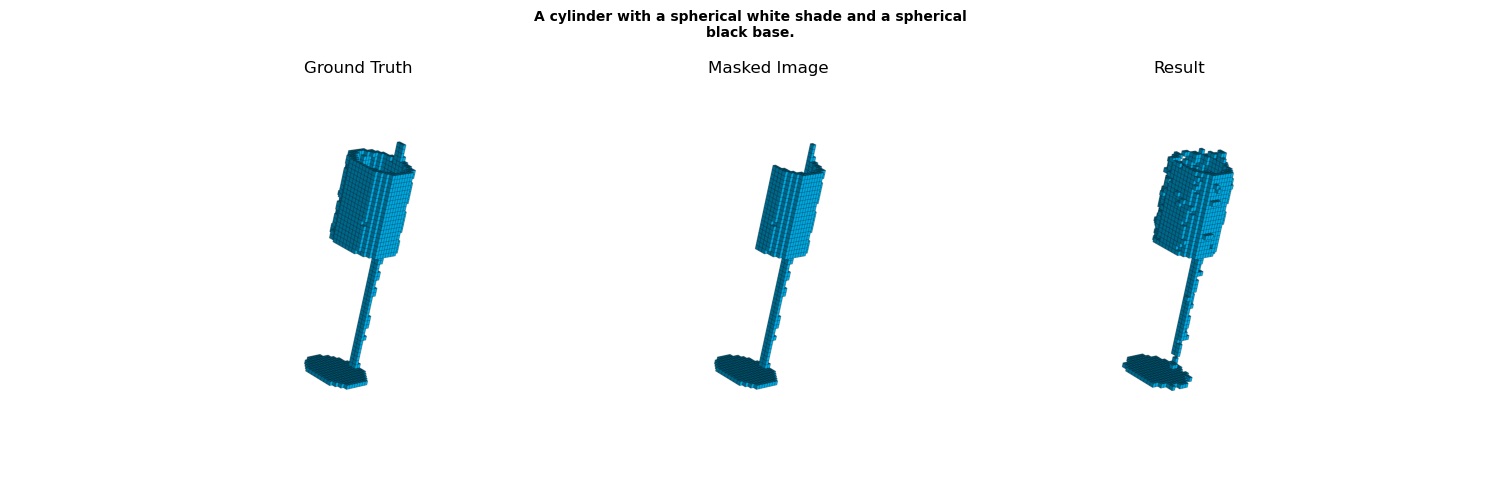}
\leftfig{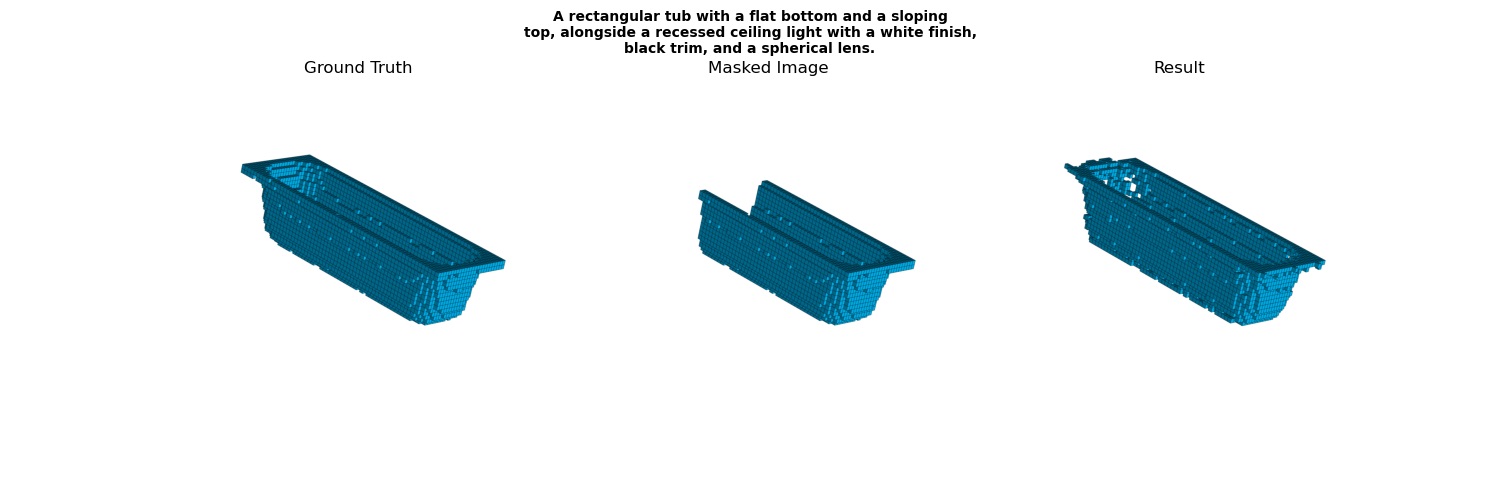}
\lrightfig{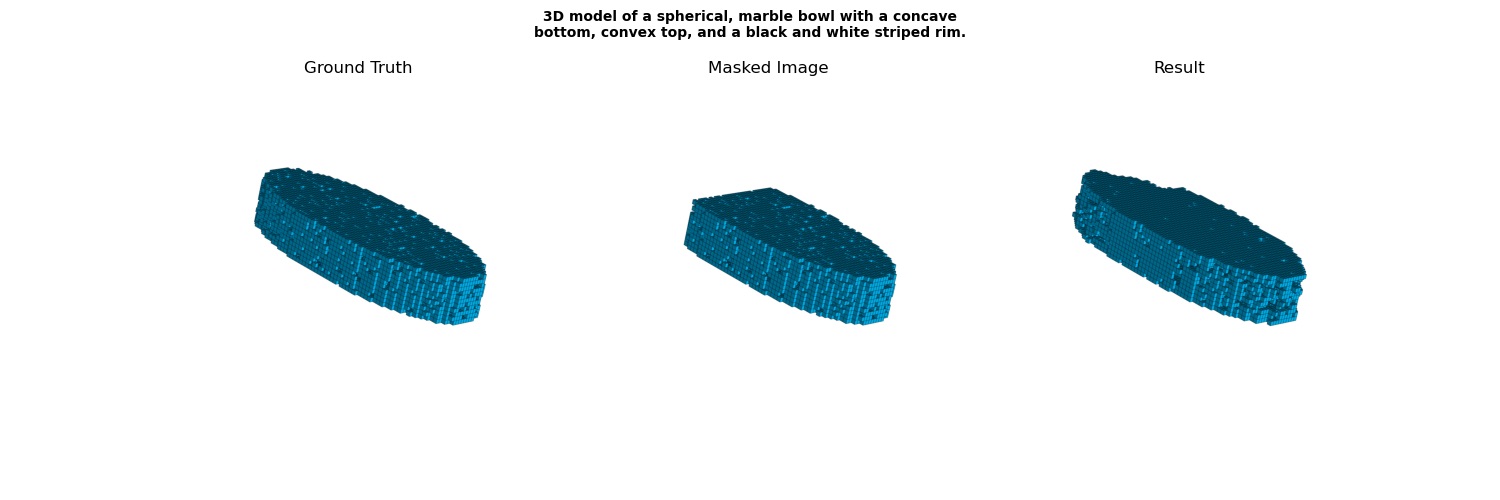}
\leftfig{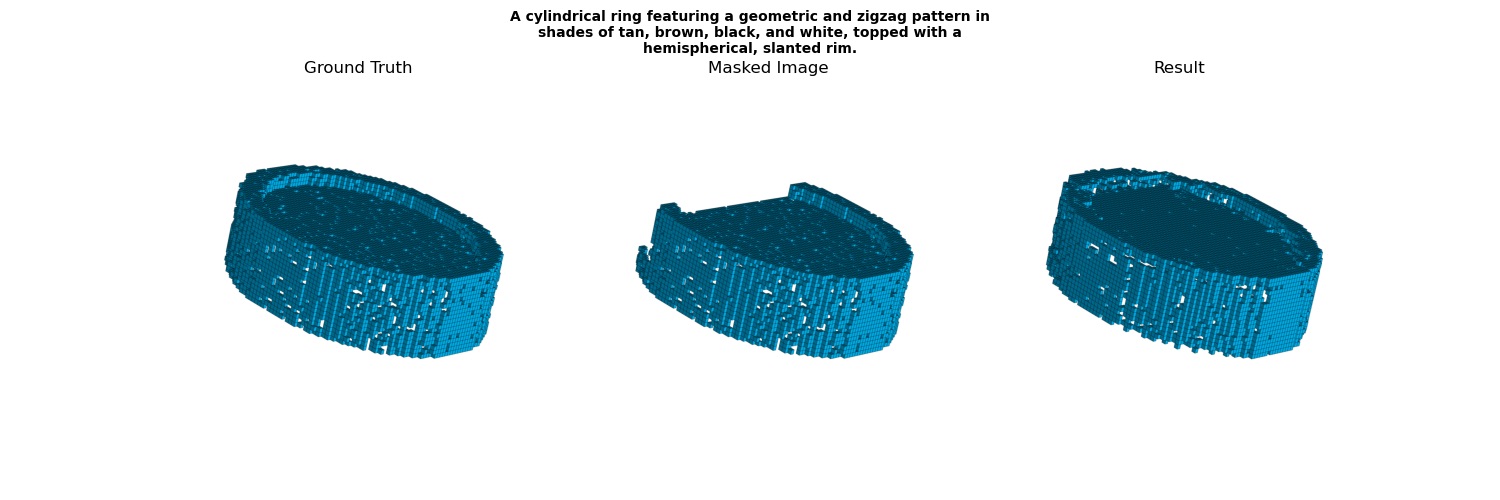}
\lrightfig{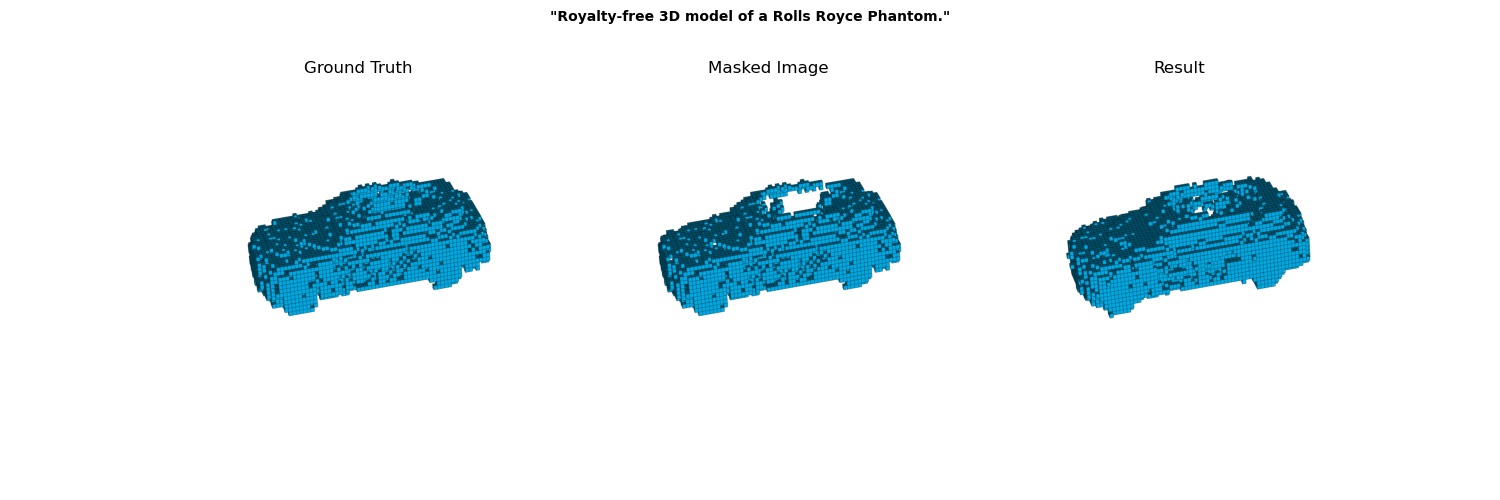}
\caption{Results Seg$20\%$}
\label{fig:plane02-1}
\end{figure}

\begin{figure}[H] 
 \centering 
\leftfig{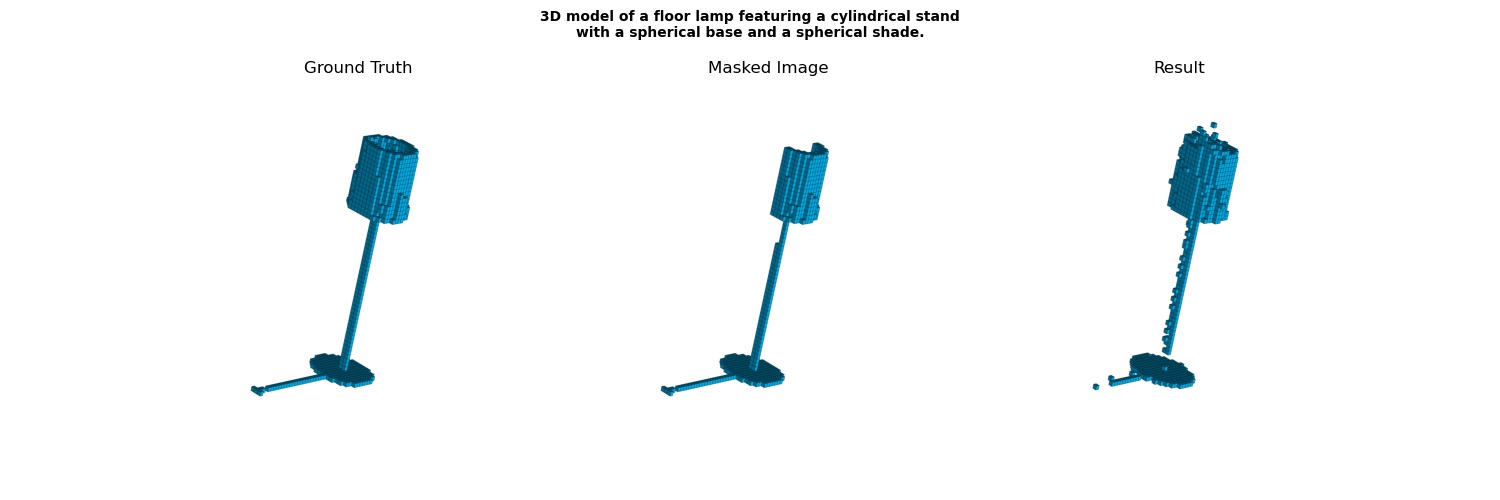}
\lrightfig{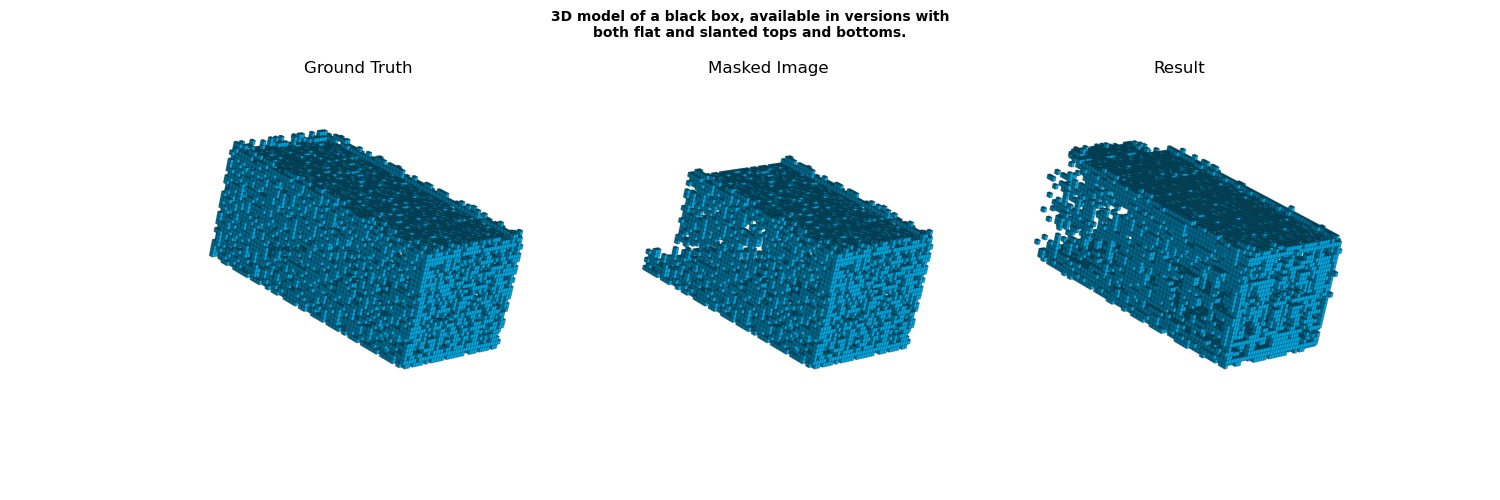}
\leftfig{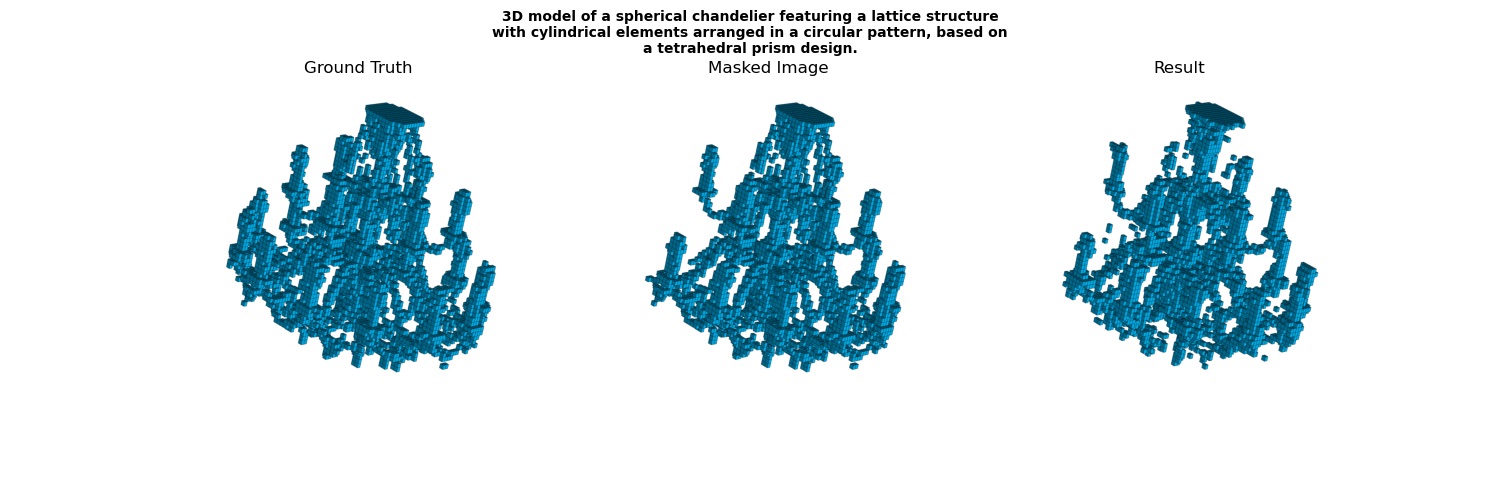}
\lrightfig{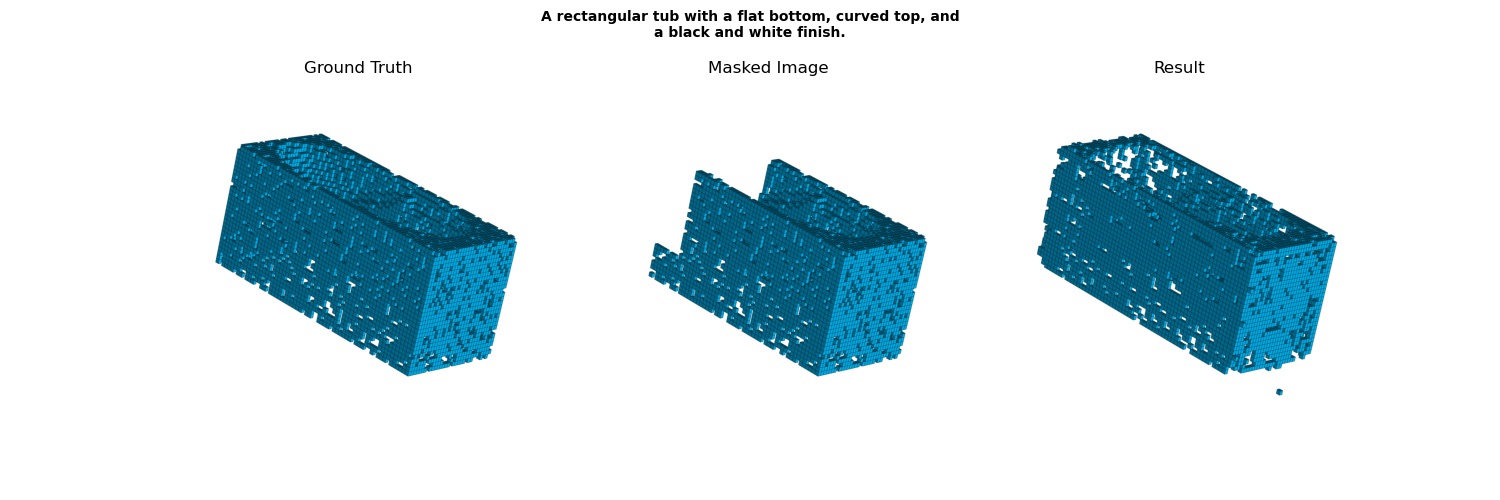}
\leftfig{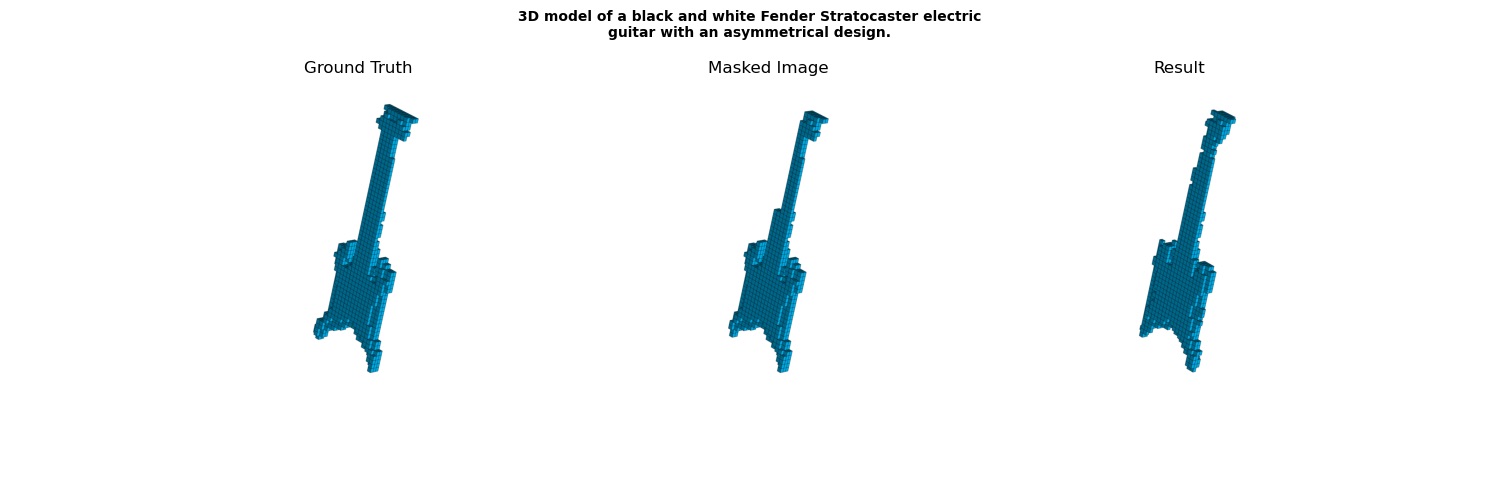}
\lrightfig{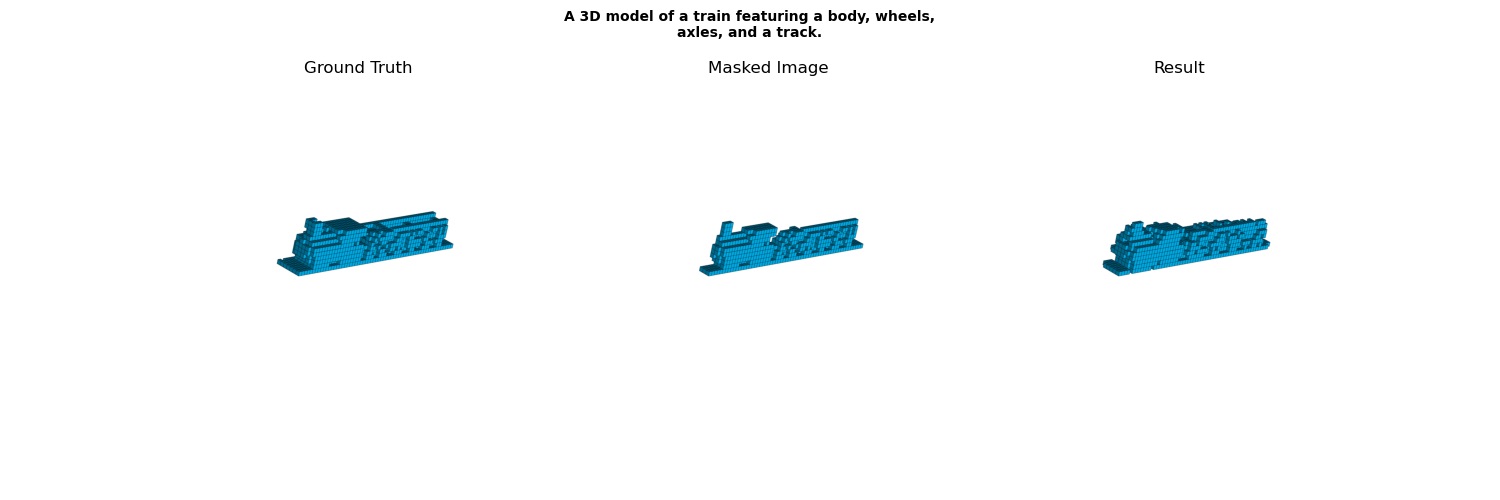}
\leftfig{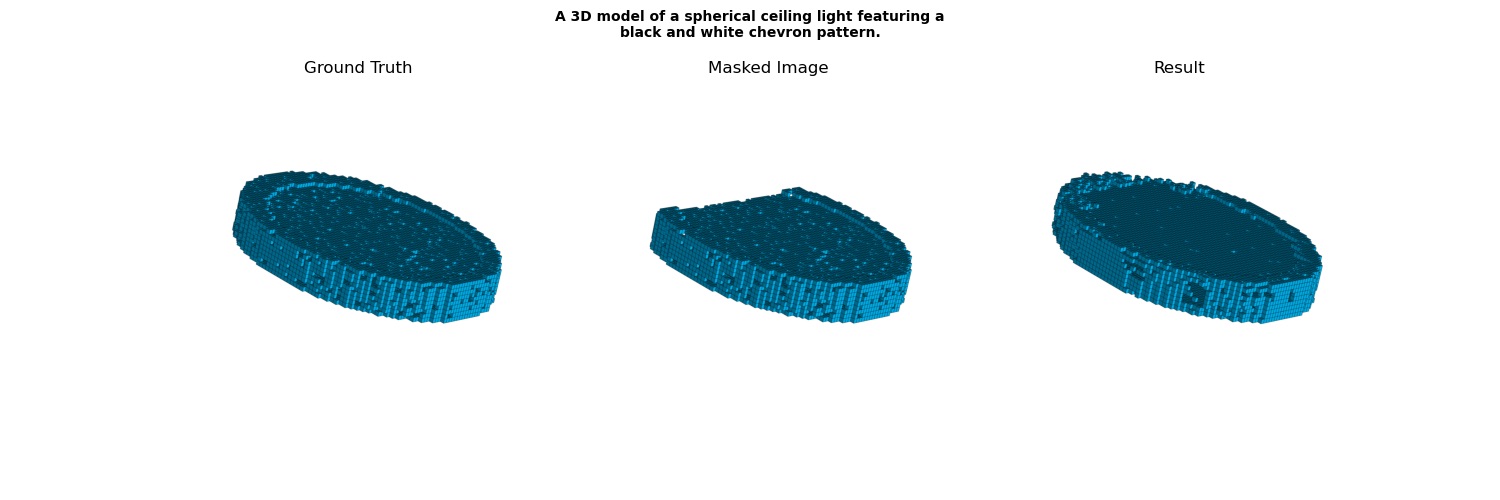}
\lrightfig{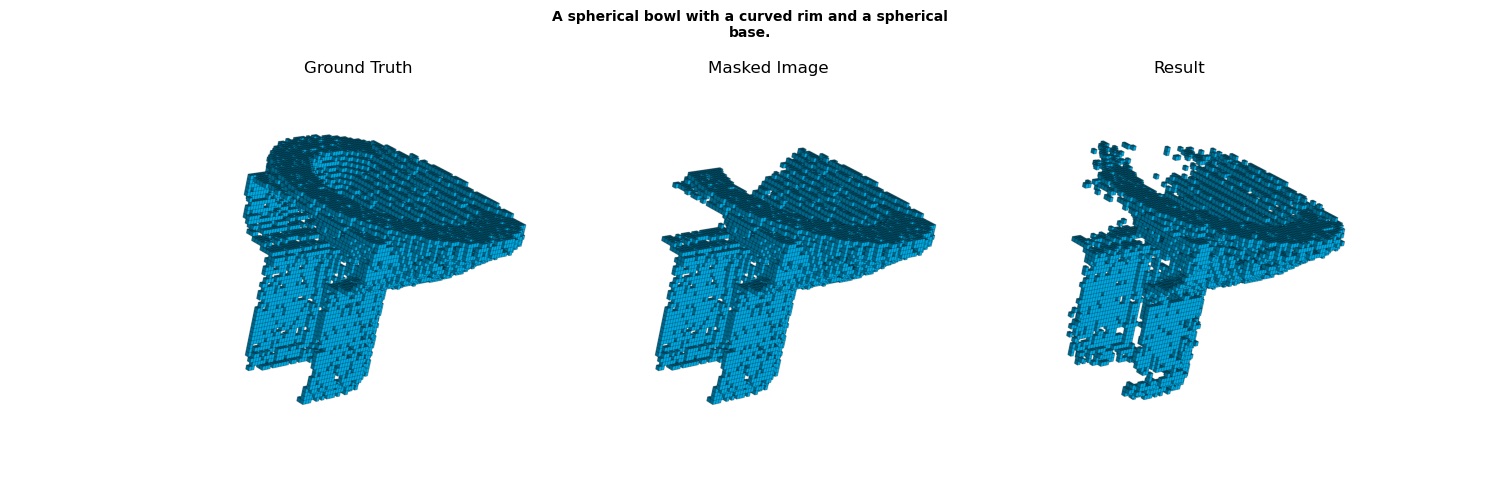}
\leftfig{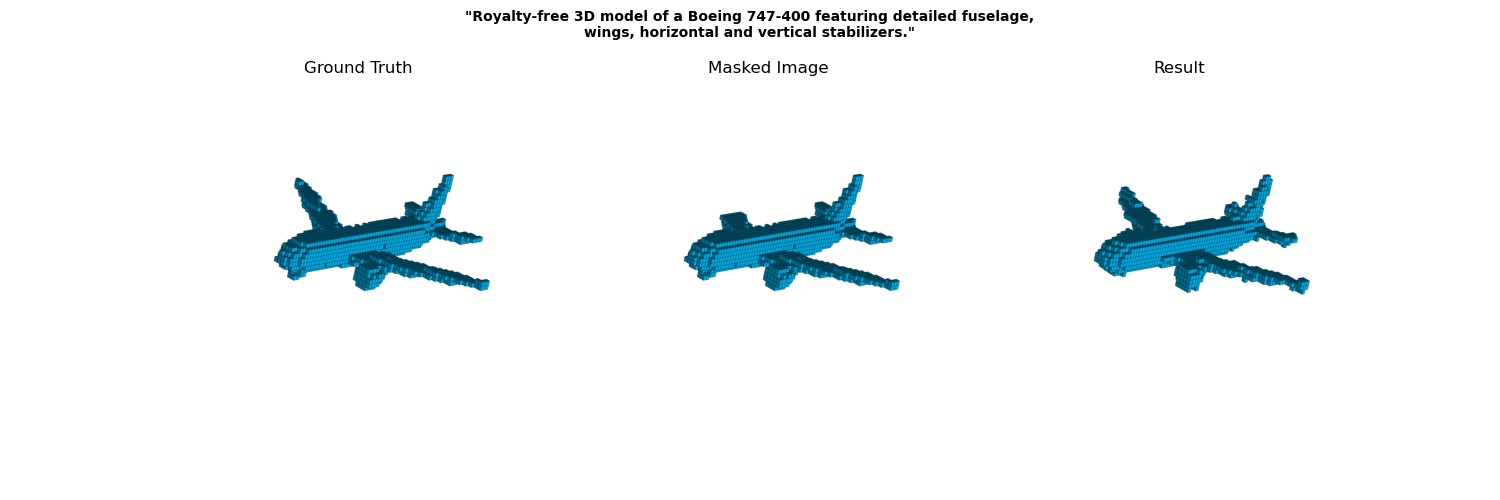}
\lrightfig{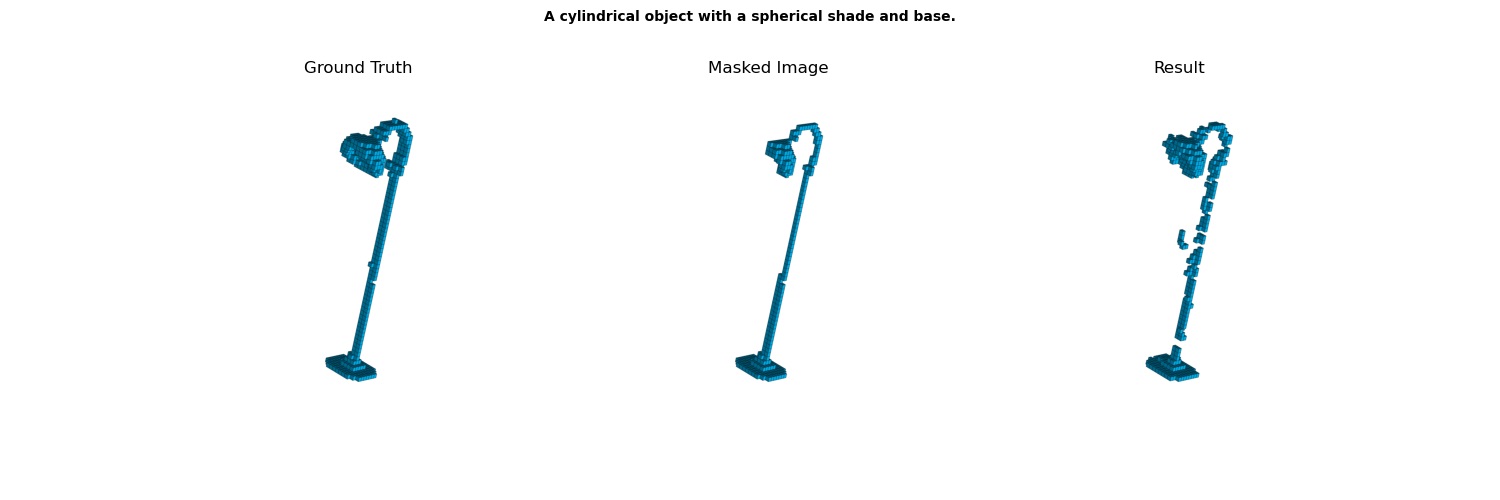}
\leftfig{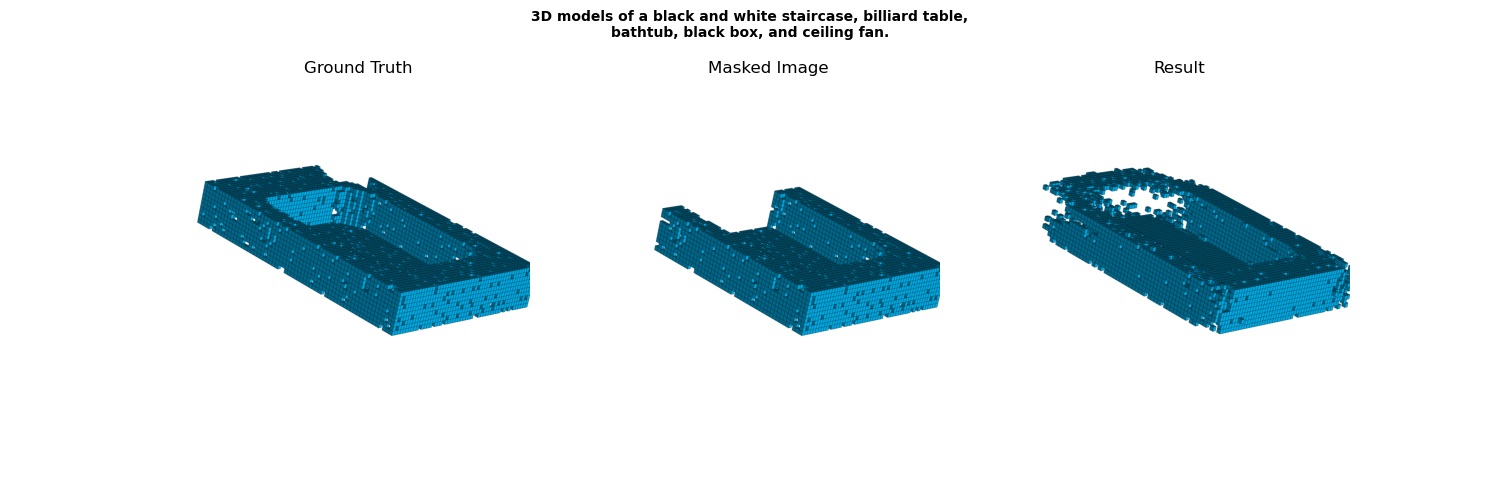}
\lrightfig{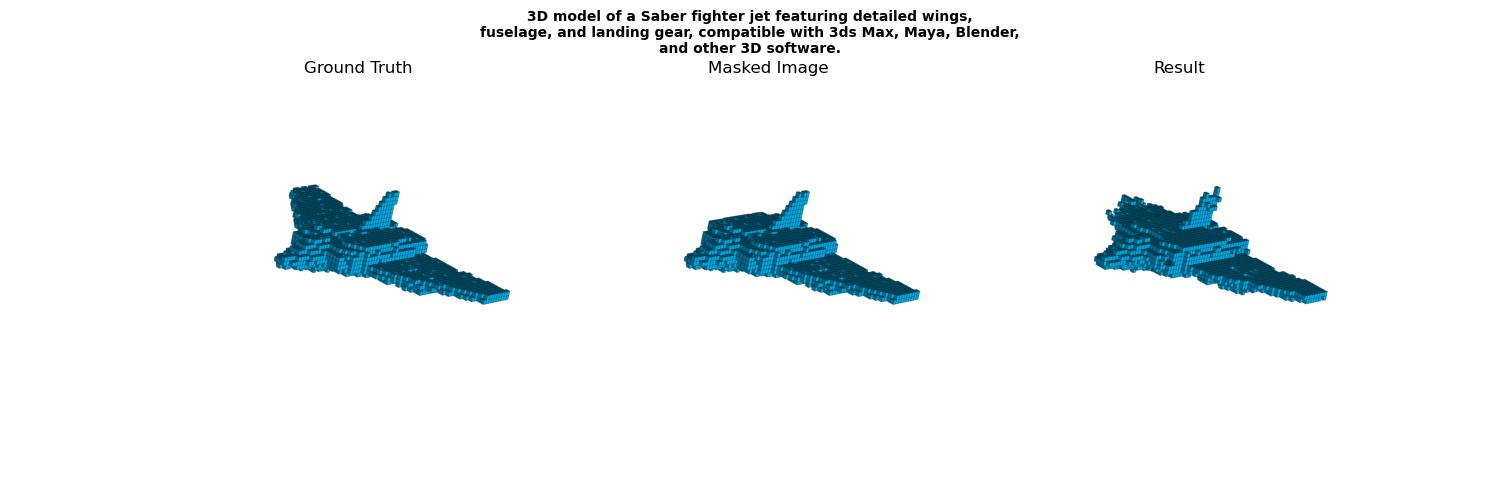}
\leftfig{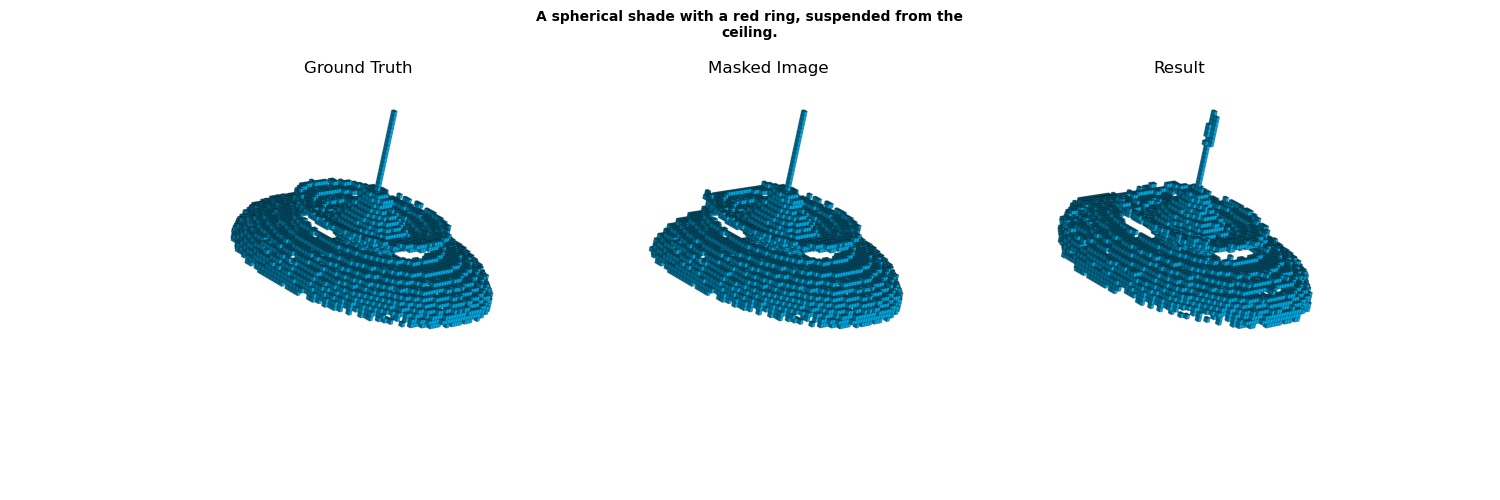}
\lrightfig{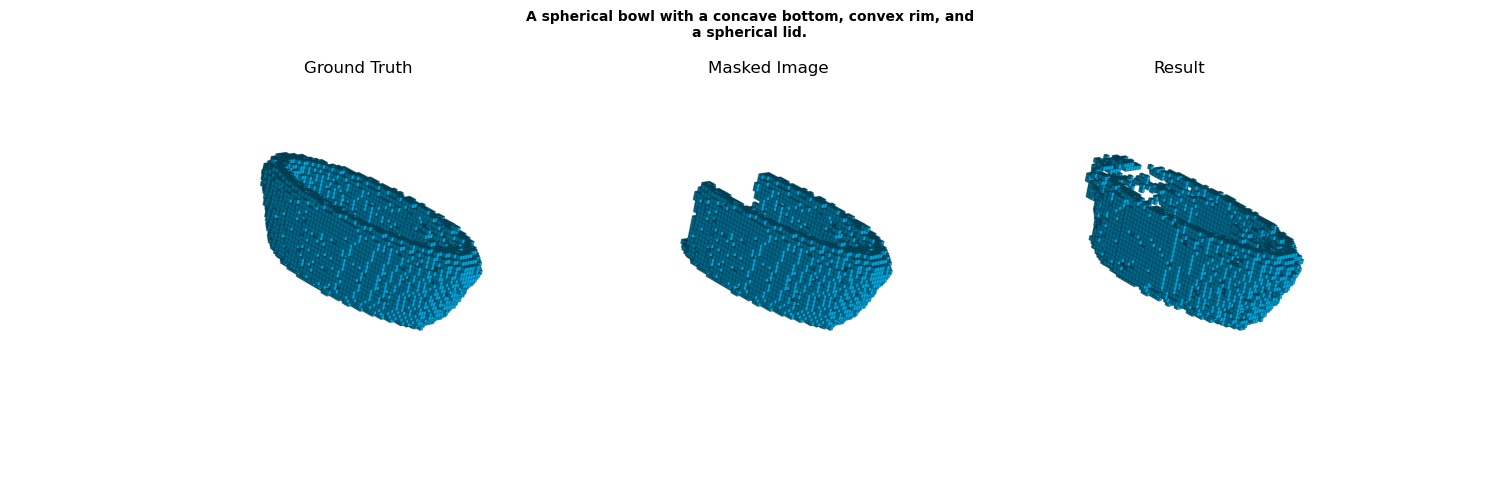}
\leftfig{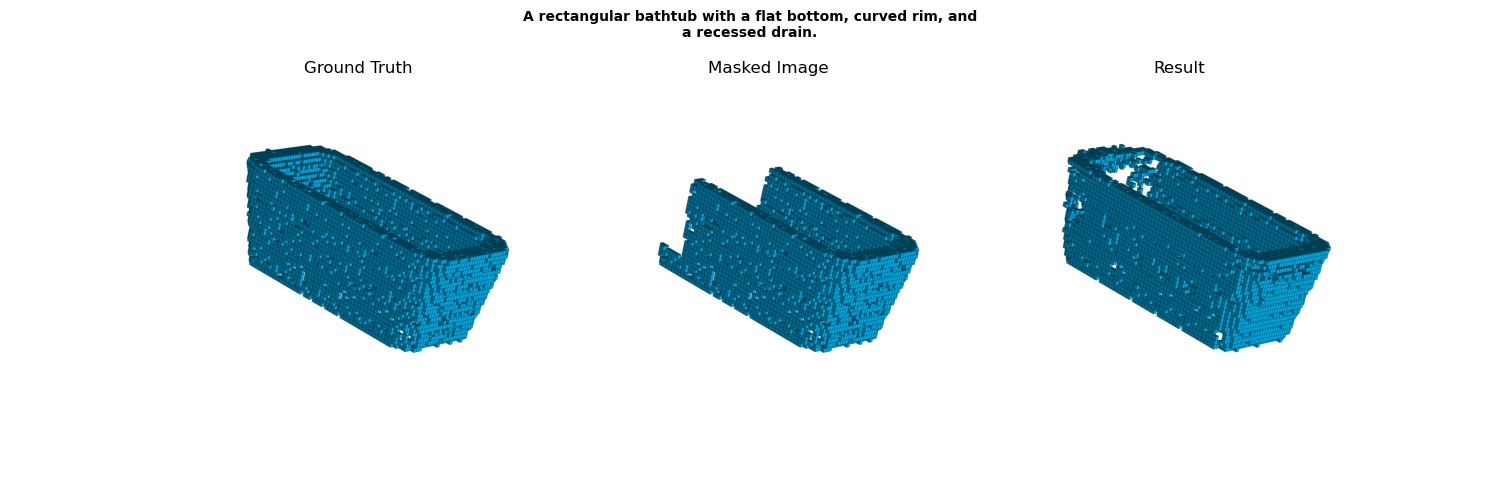}
\lrightfig{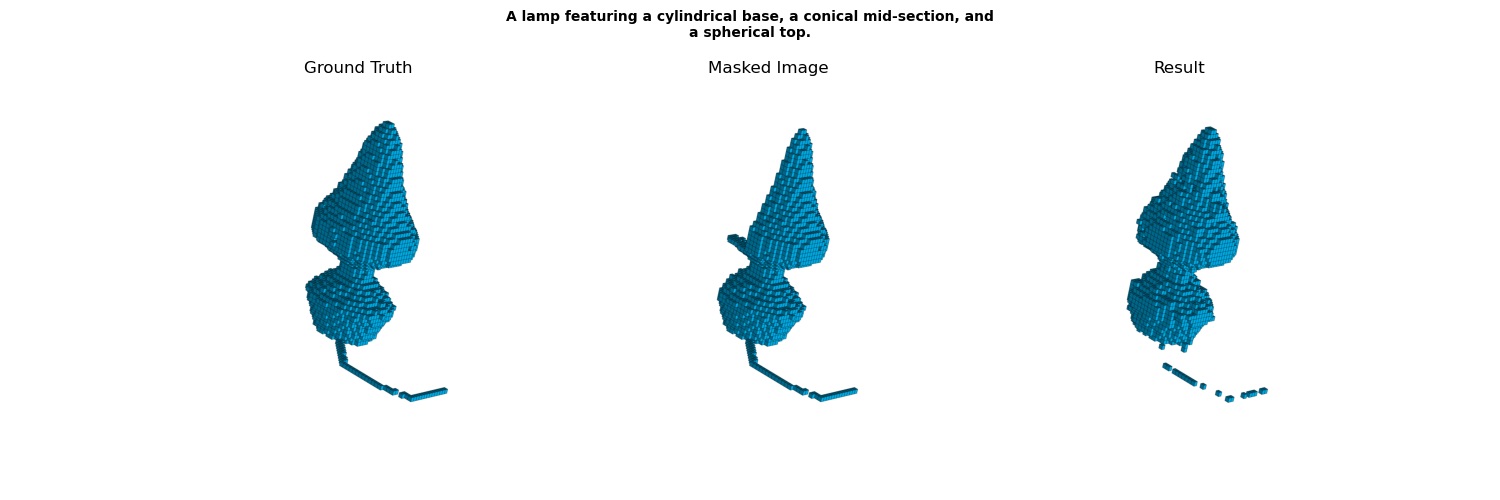}
\caption{Results Seg$20\%$}
\label{fig:Plane02-2}
\end{figure}

\begin{figure}[H] 
 \centering 
\leftfig{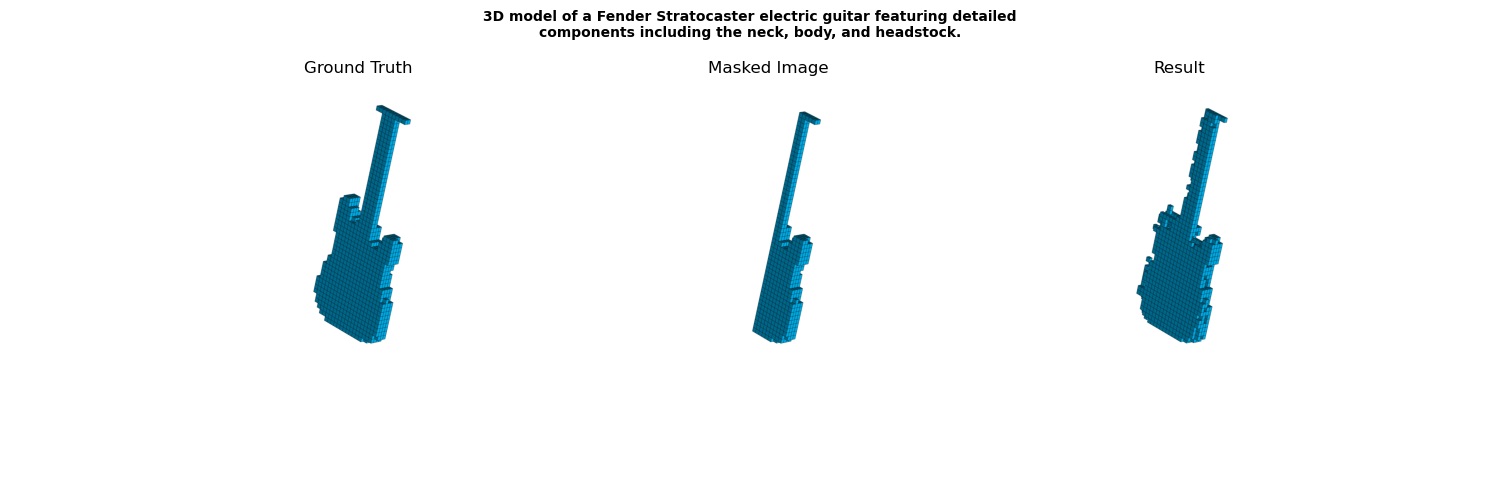}
\lrightfig{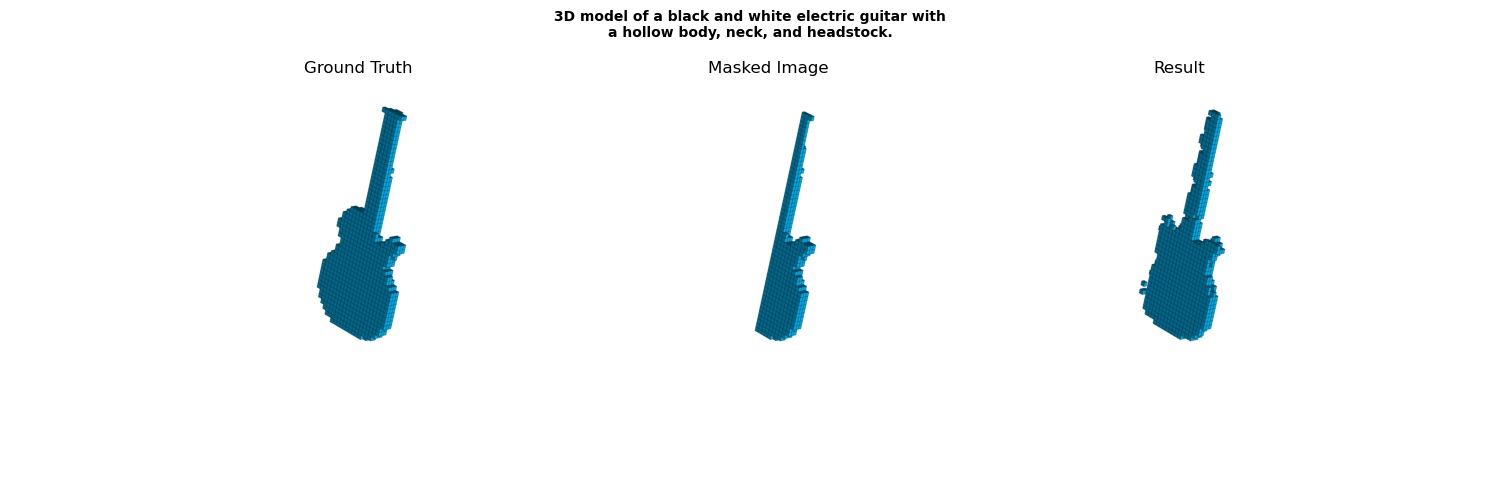}
\leftfig{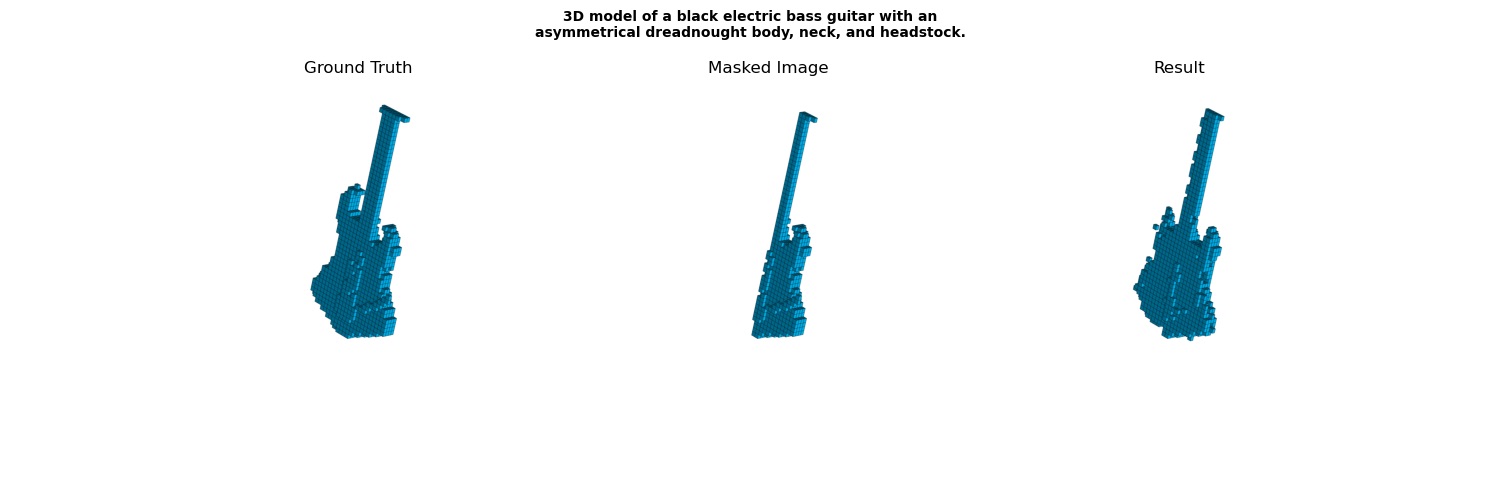}
\lrightfig{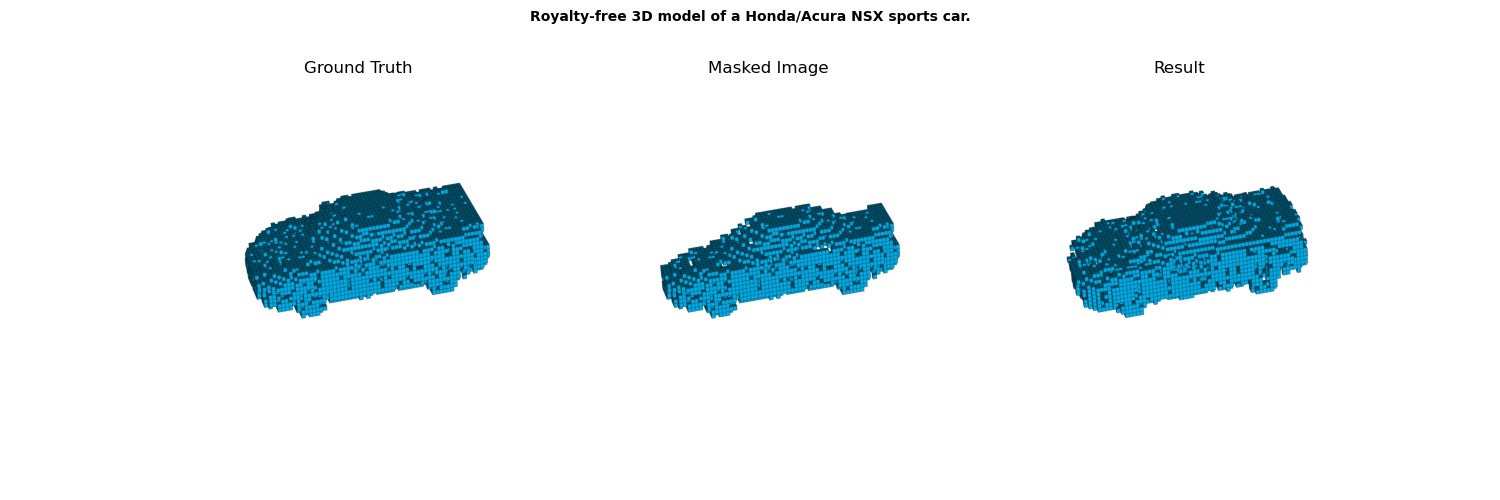}
\leftfig{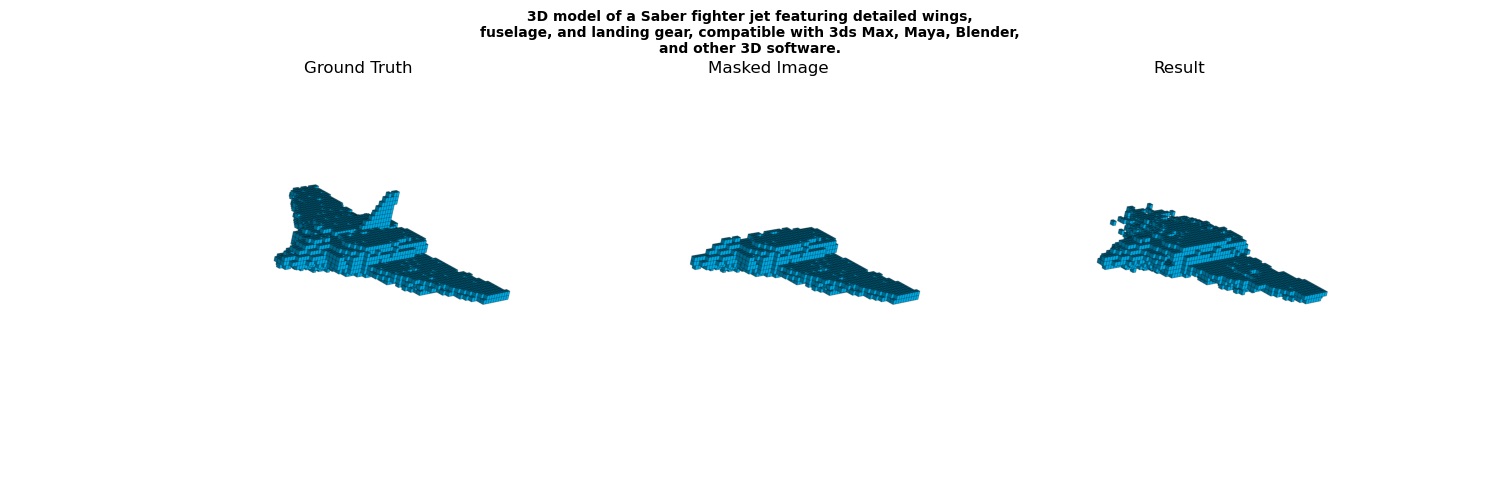}
\lrightfig{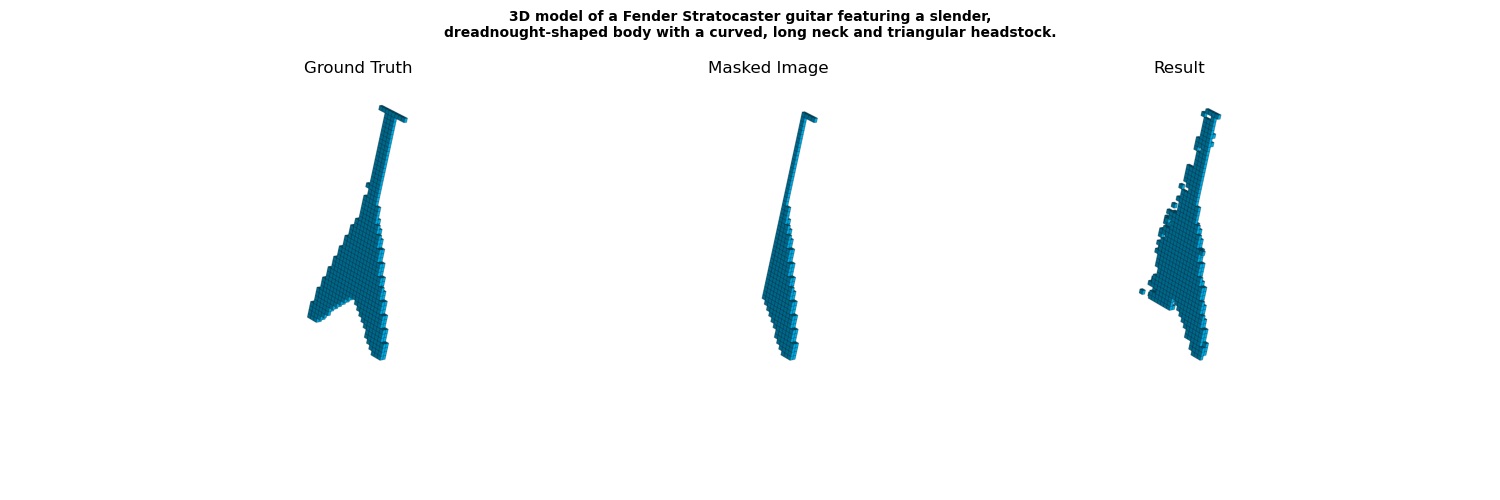}
\leftfig{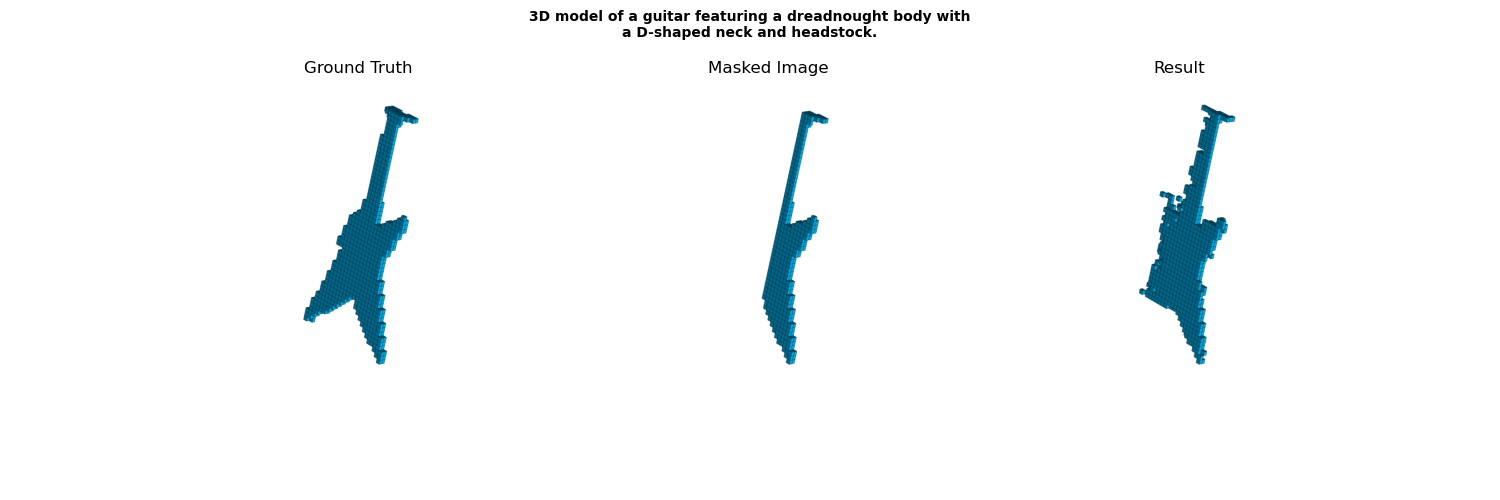}
\lrightfig{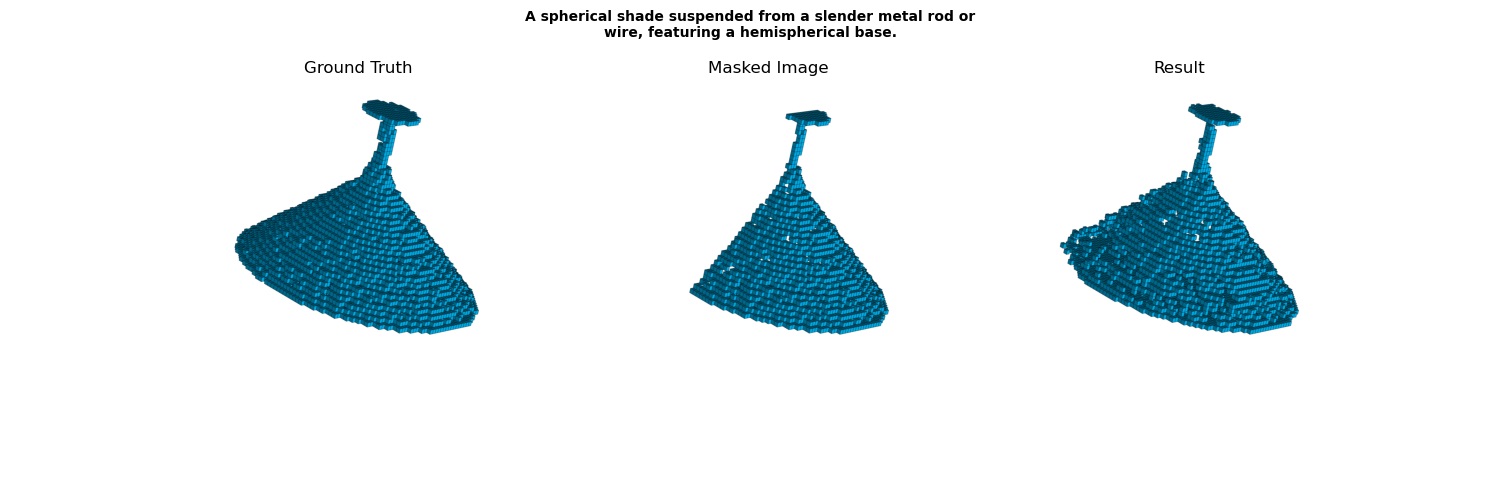}
\leftfig{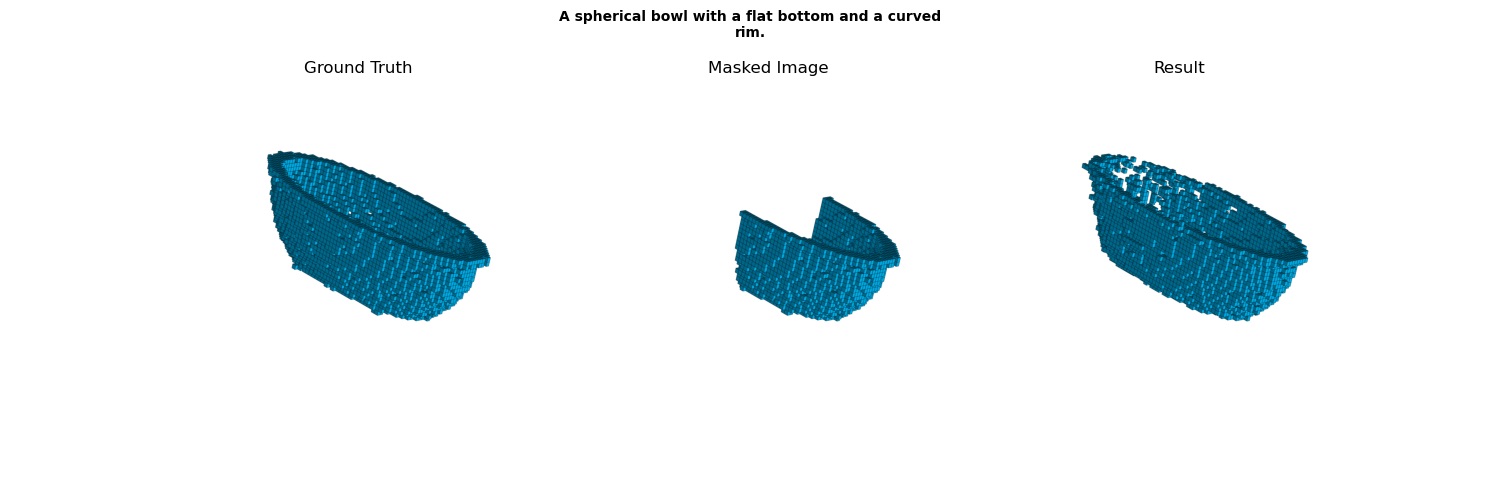}
\lrightfig{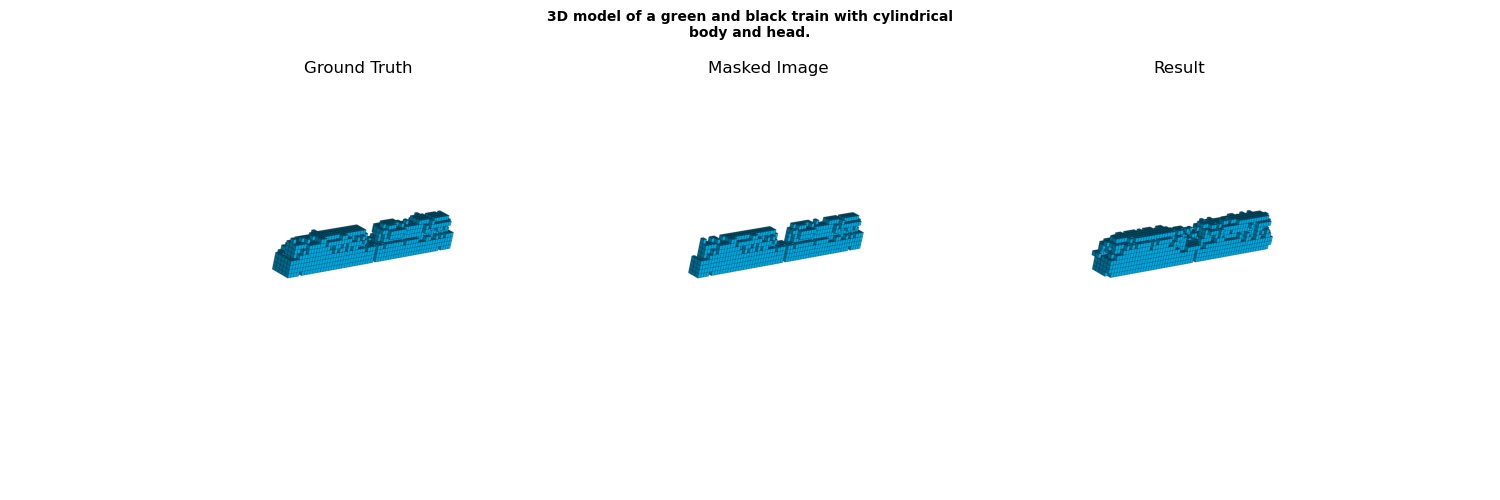}
\leftfig{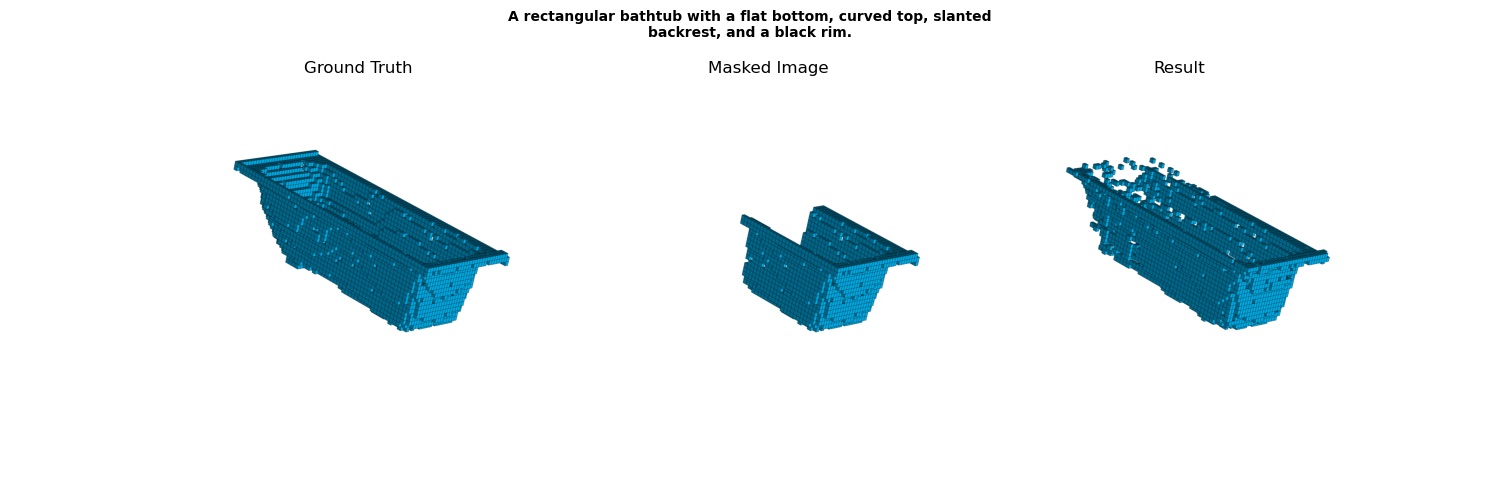}
\lrightfig{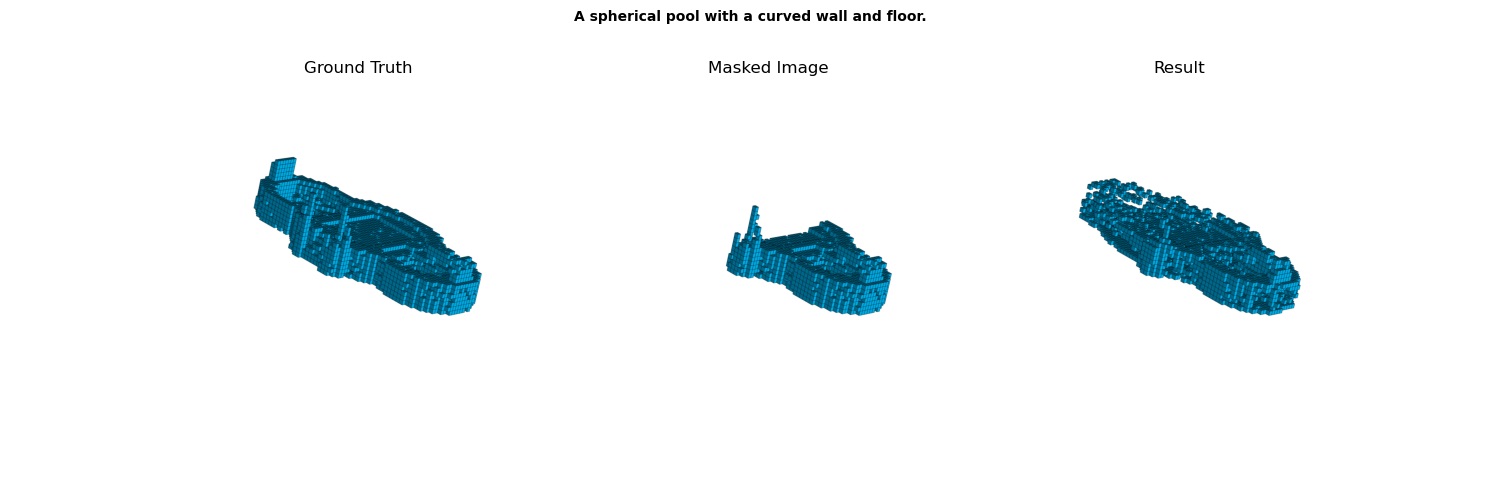}
\leftfig{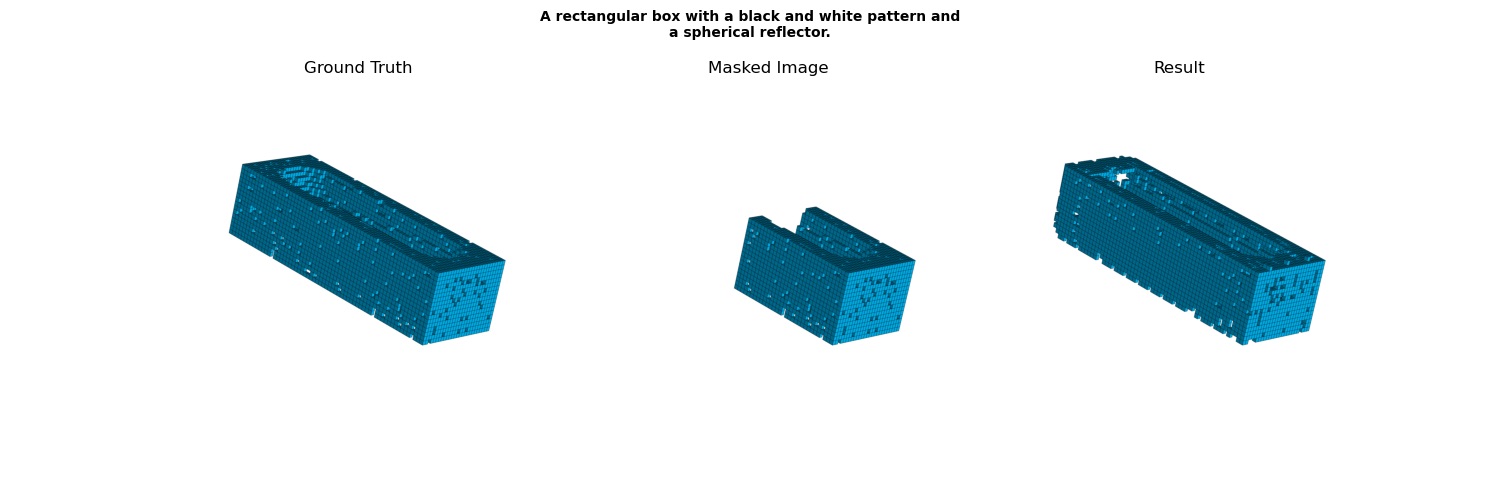}
\lrightfig{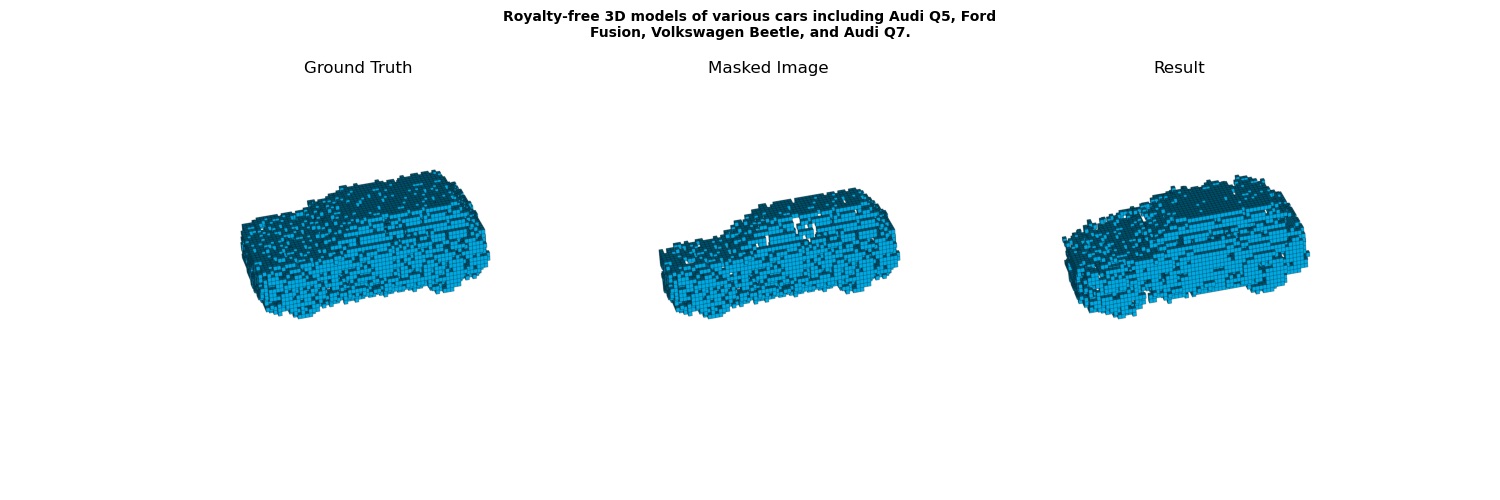}
\leftfig{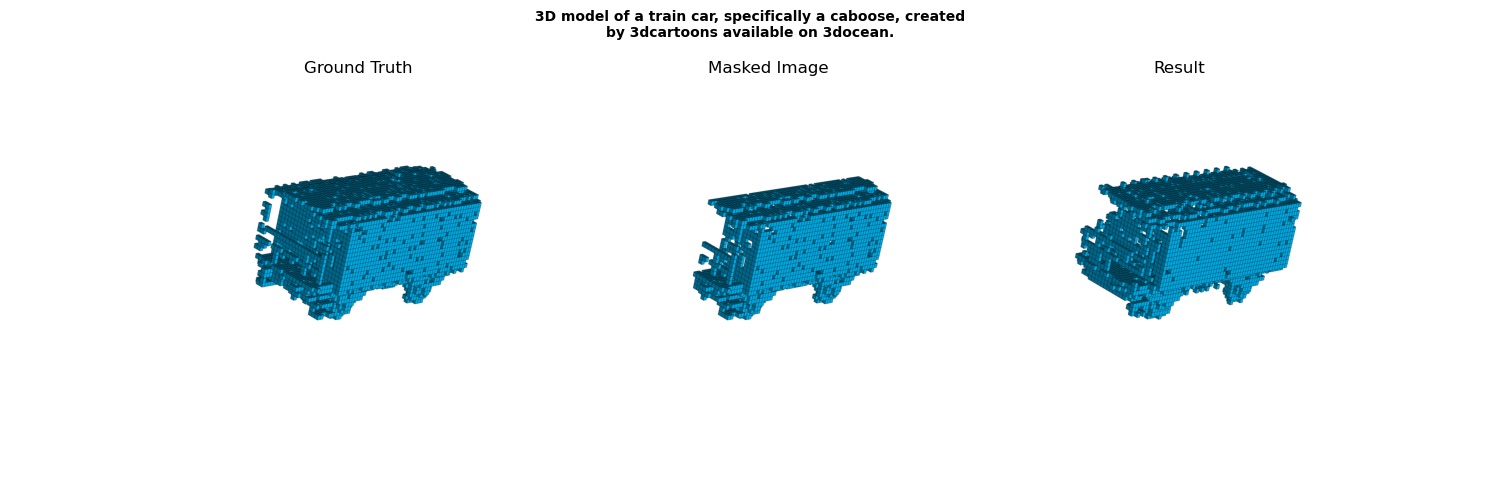}
\lrightfig{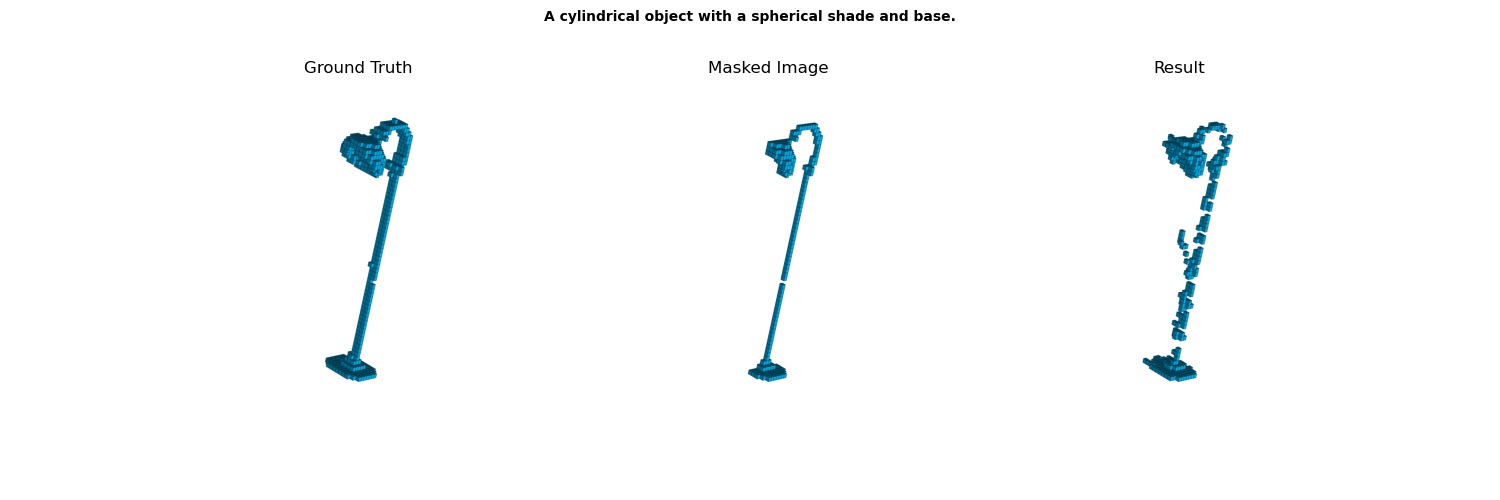}
\caption{Results Seg$50\%$}
\label{fig:plane05-1}
\end{figure}

\begin{figure}[H] 
 \centering 
\leftfig{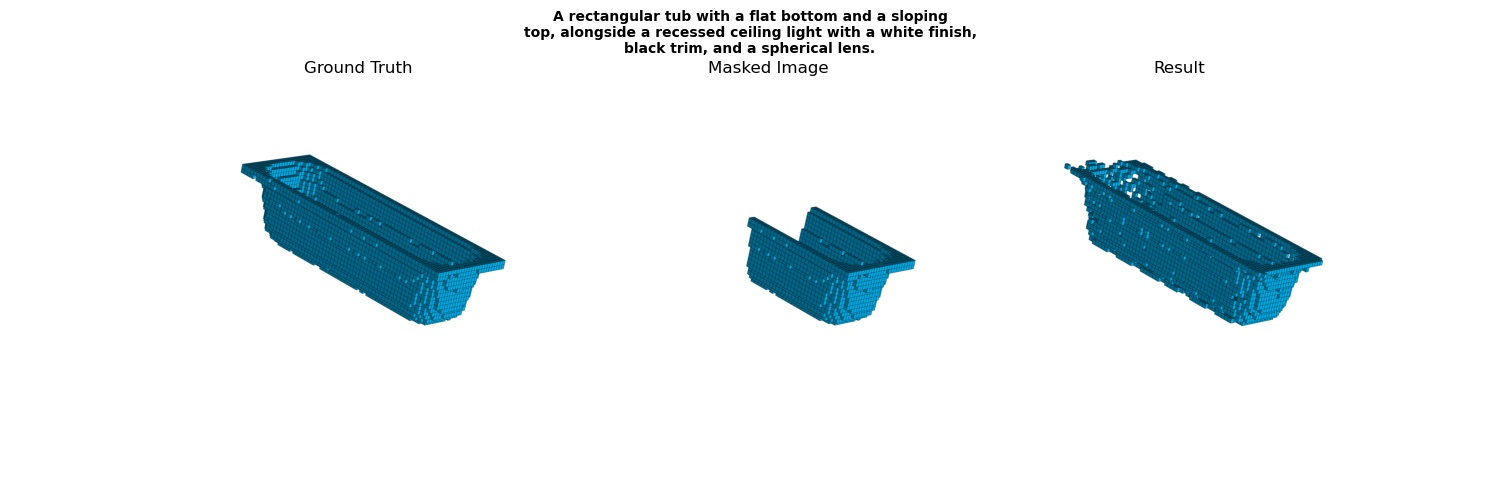}
\lrightfig{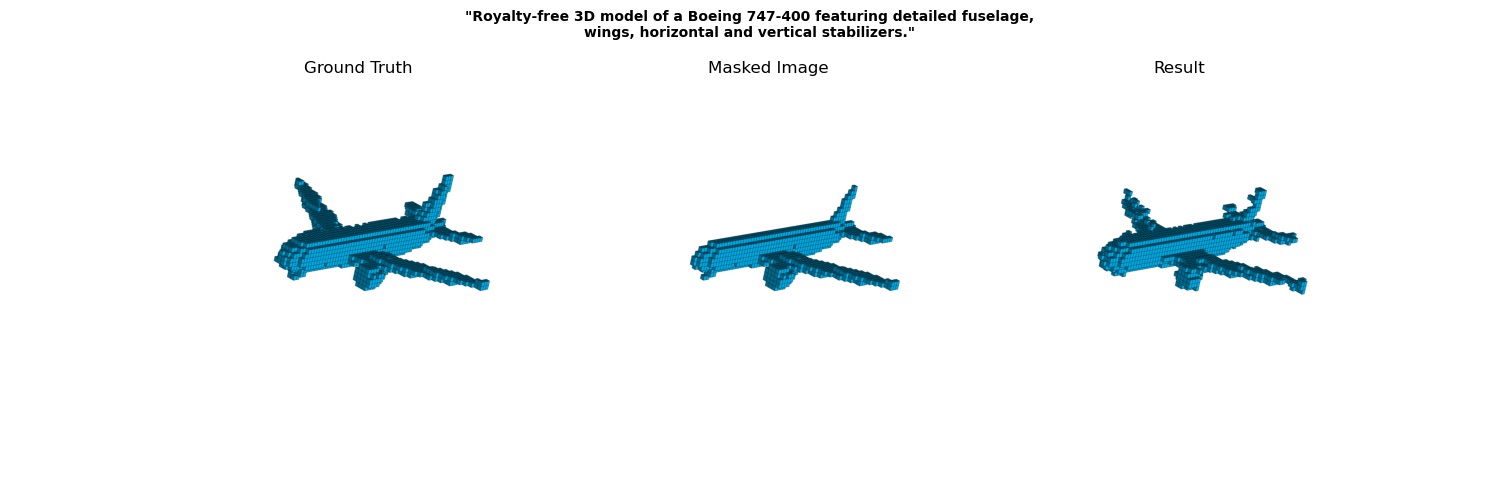}
\leftfig{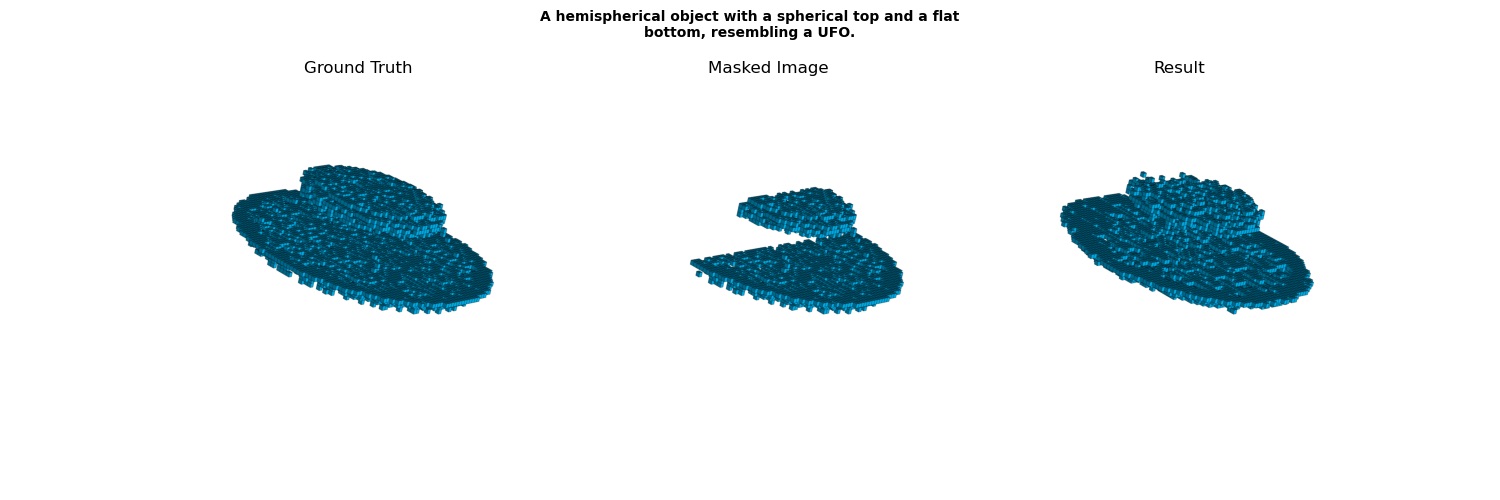}
\lrightfig{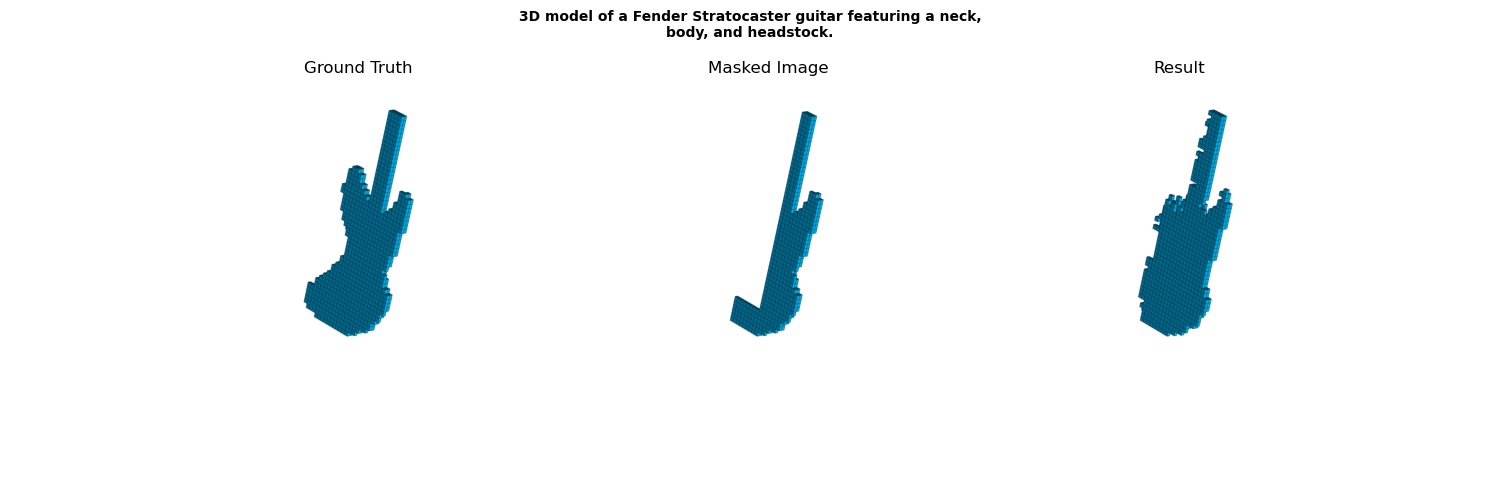}
\leftfig{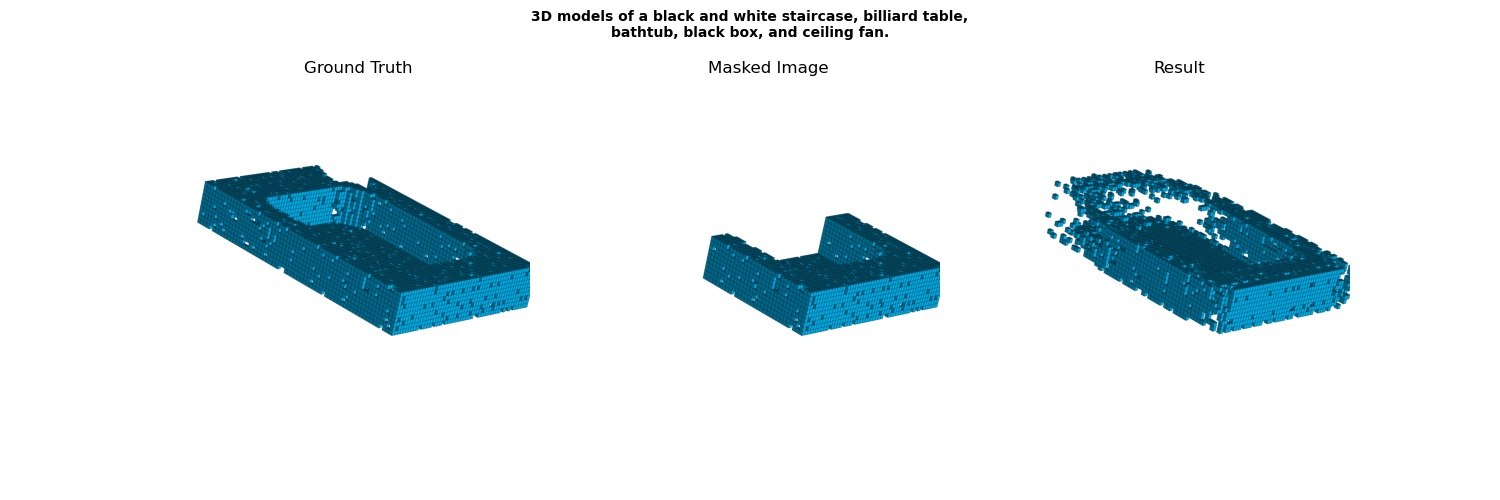}
\lrightfig{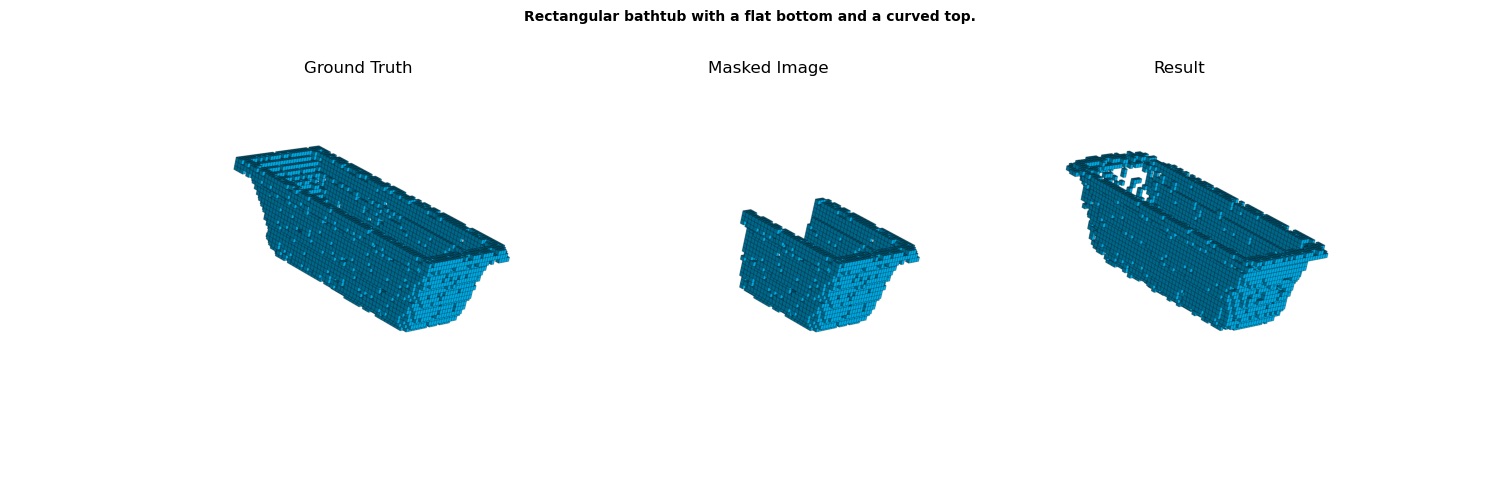}
\leftfig{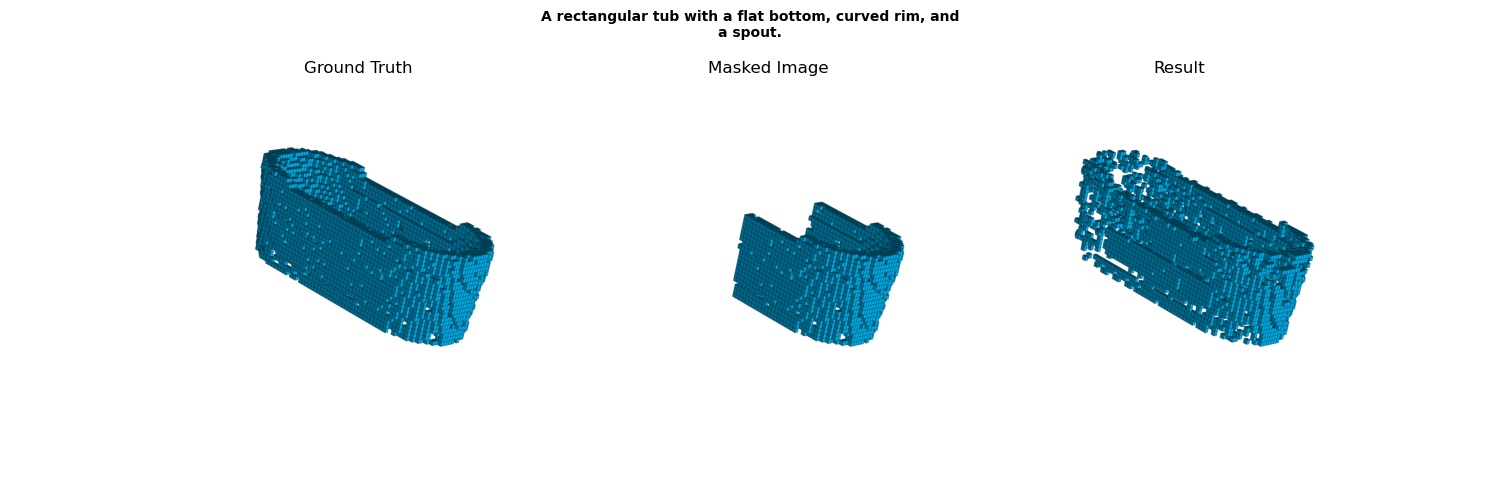}
\lrightfig{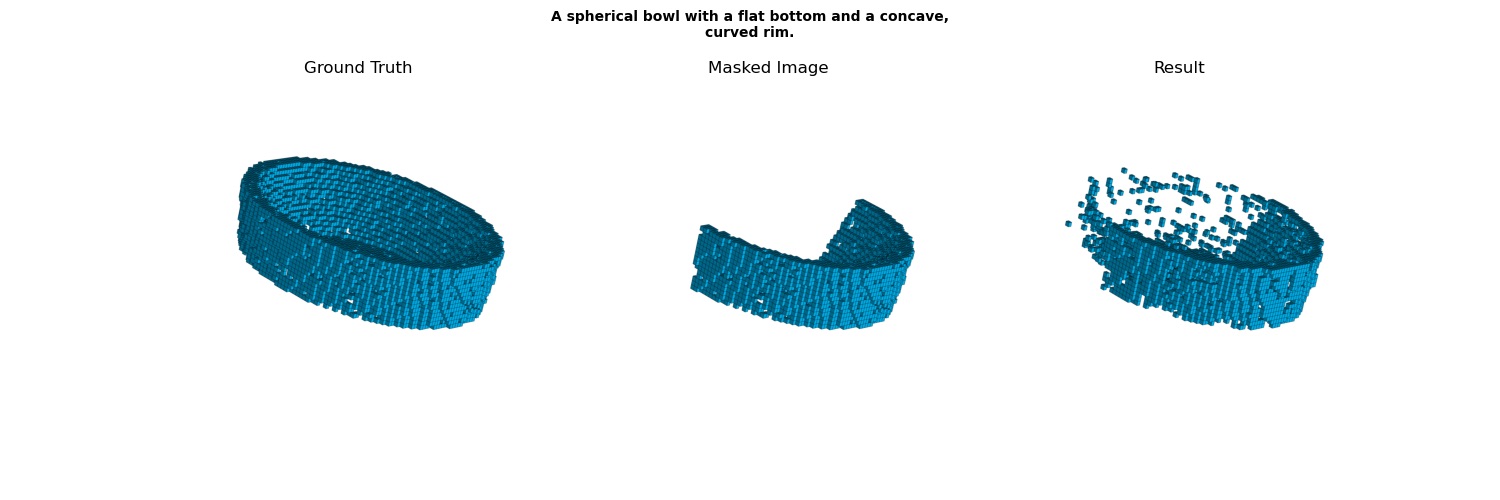}
\leftfig{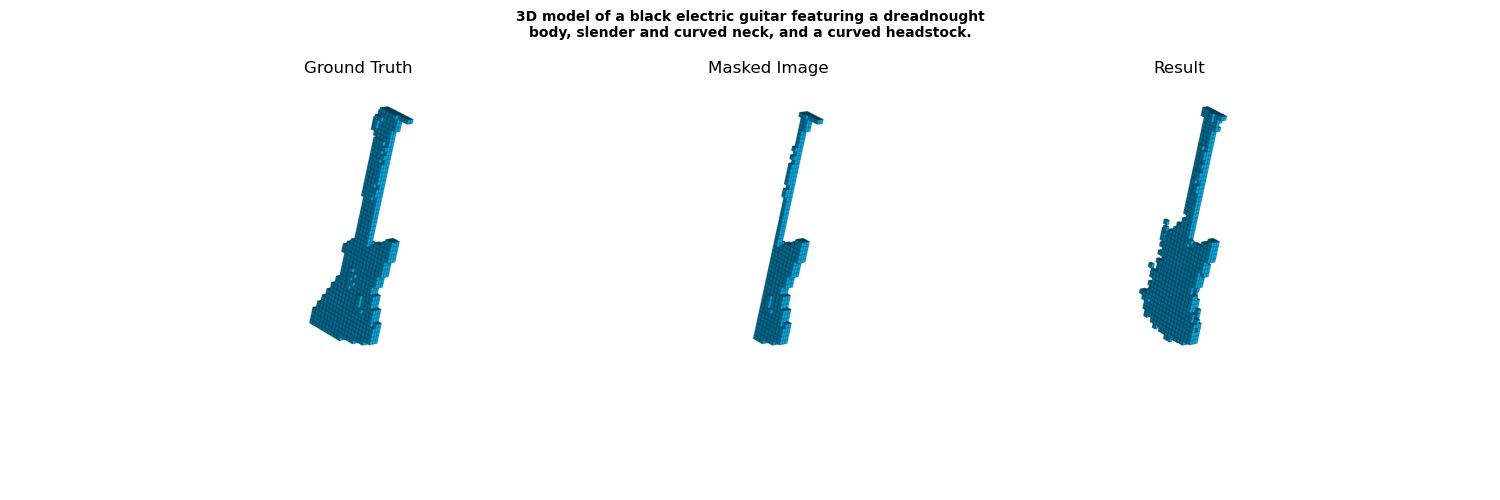}
\lrightfig{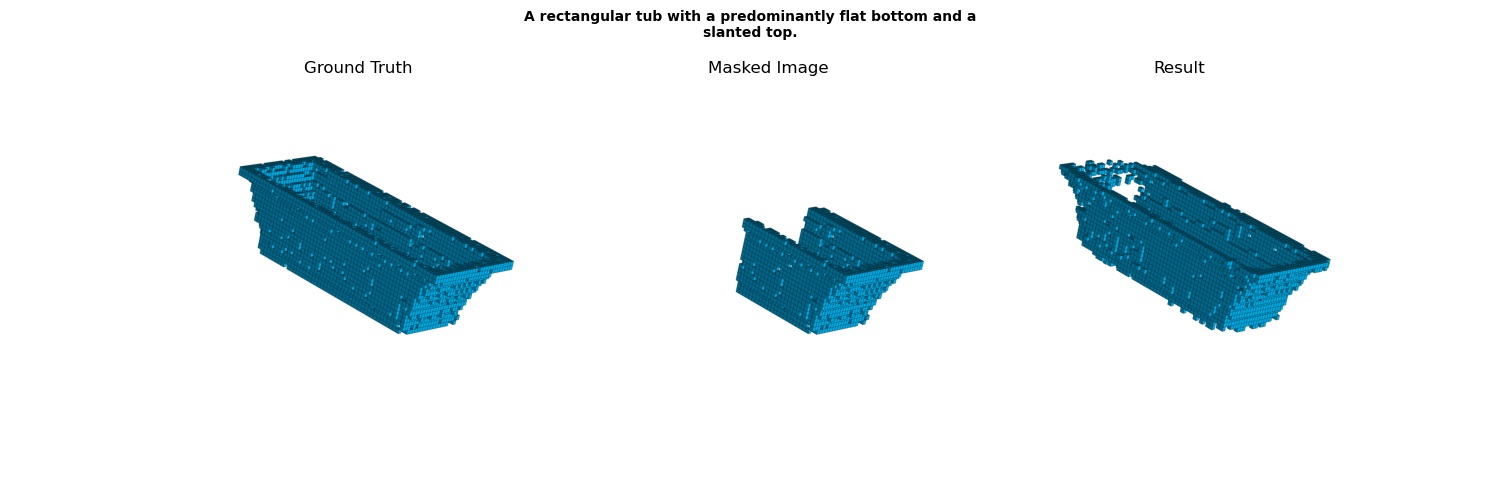}
\leftfig{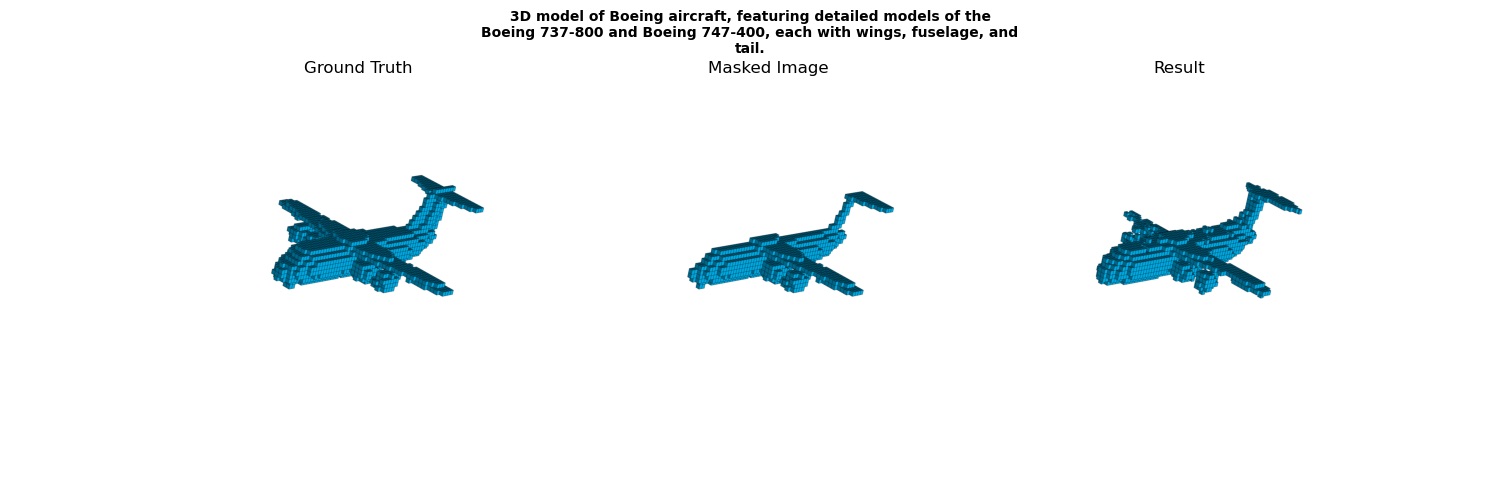}
\lrightfig{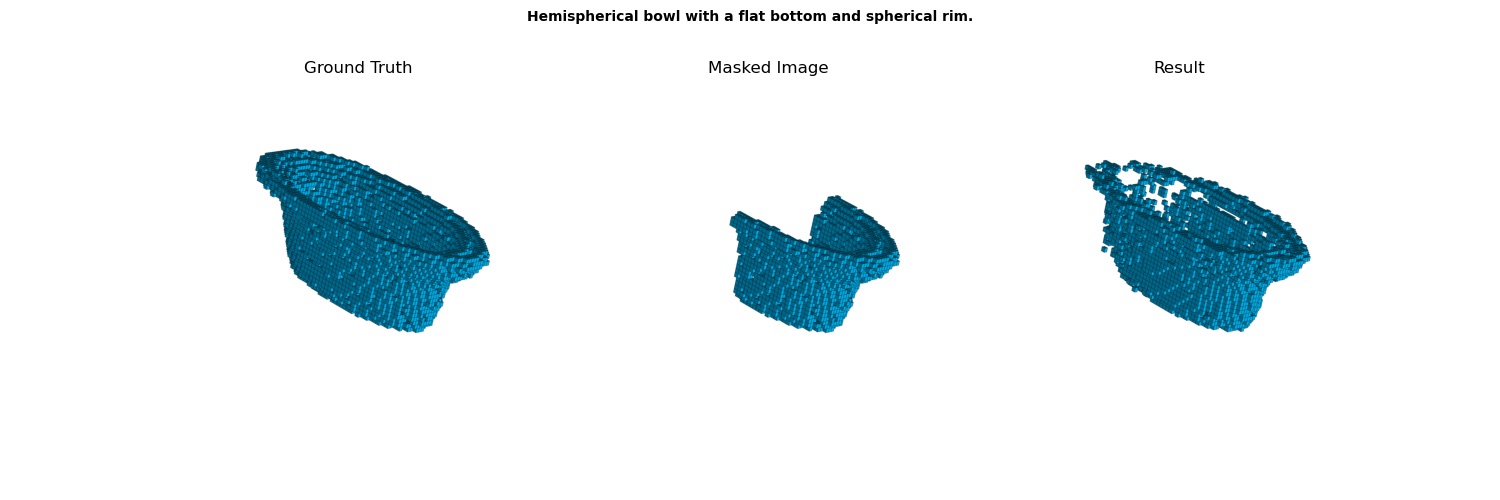}
\leftfig{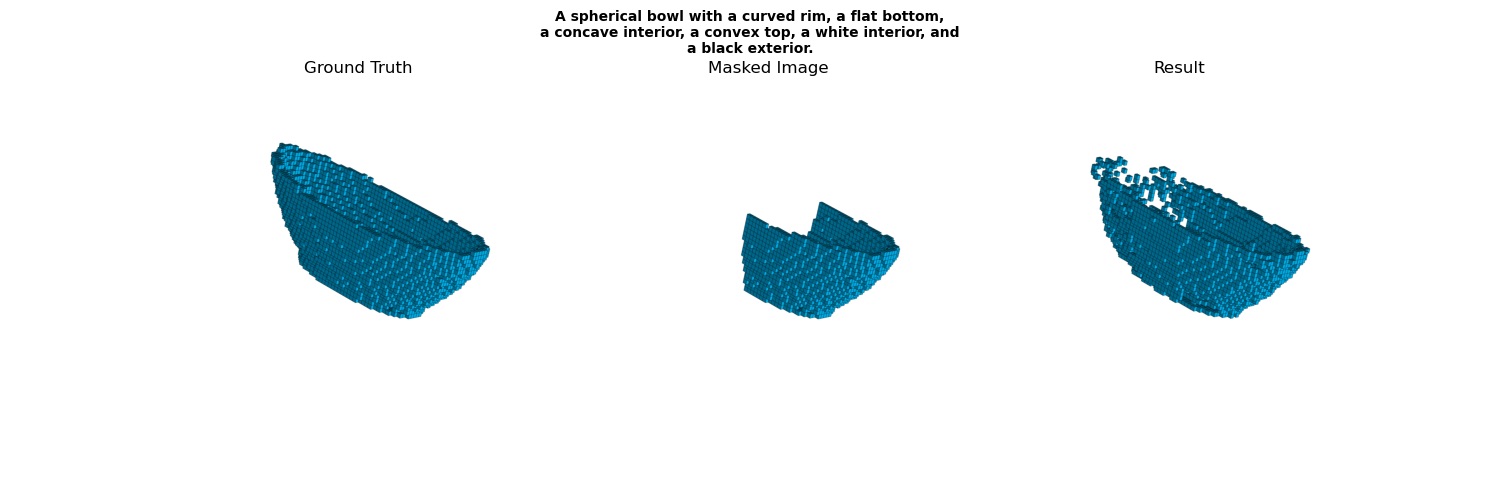}
\lrightfig{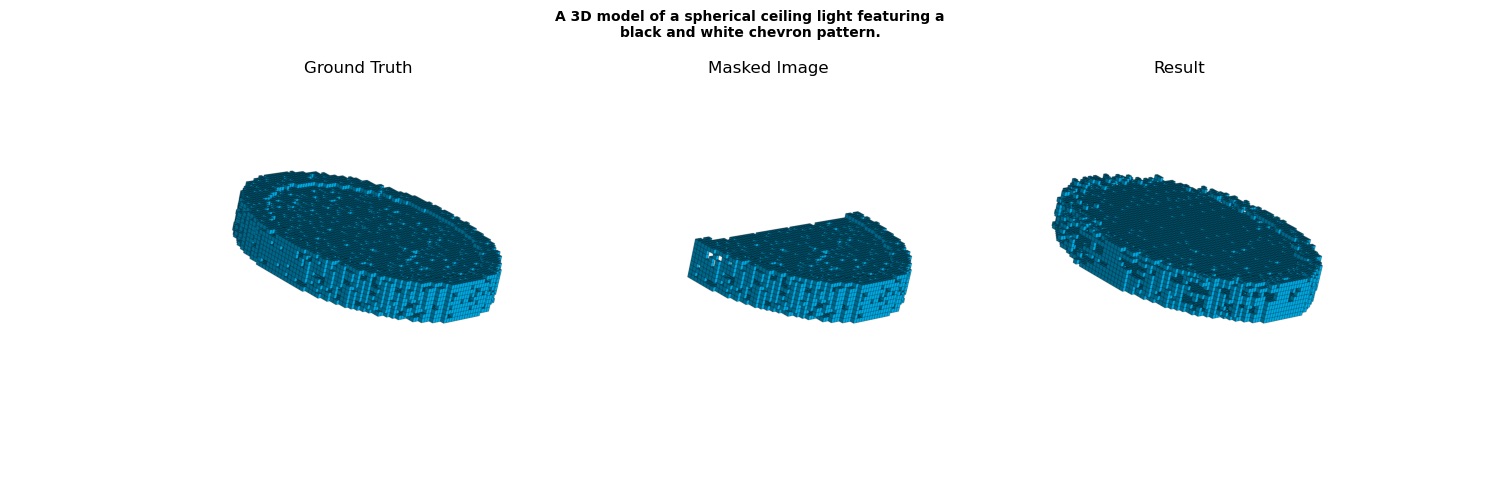}
\leftfig{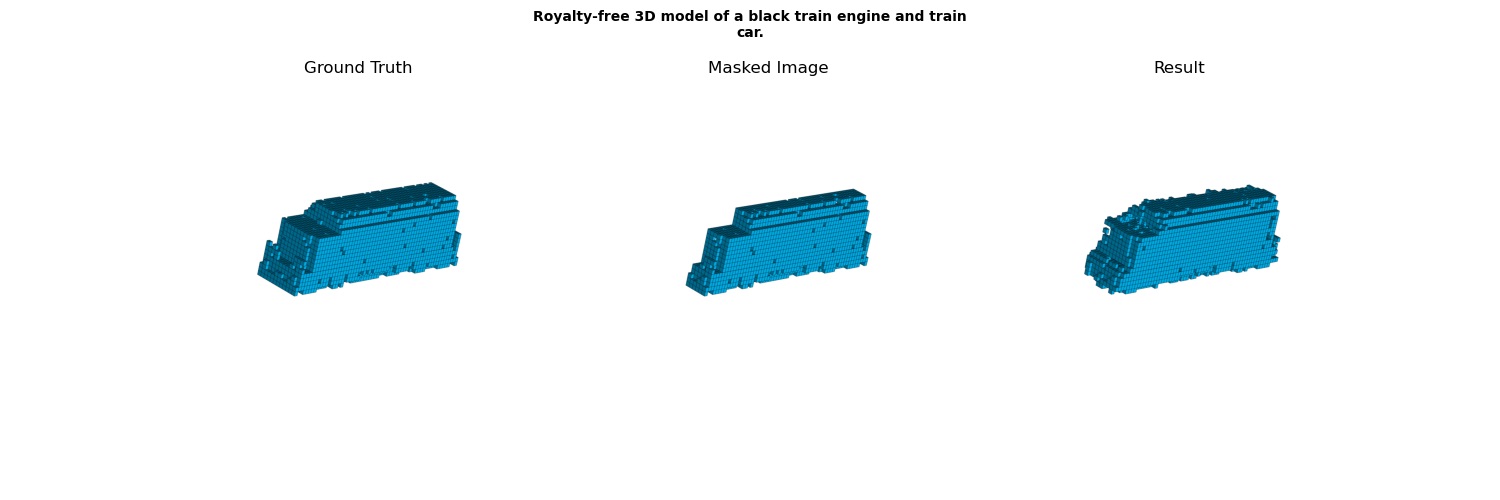}
\lrightfig{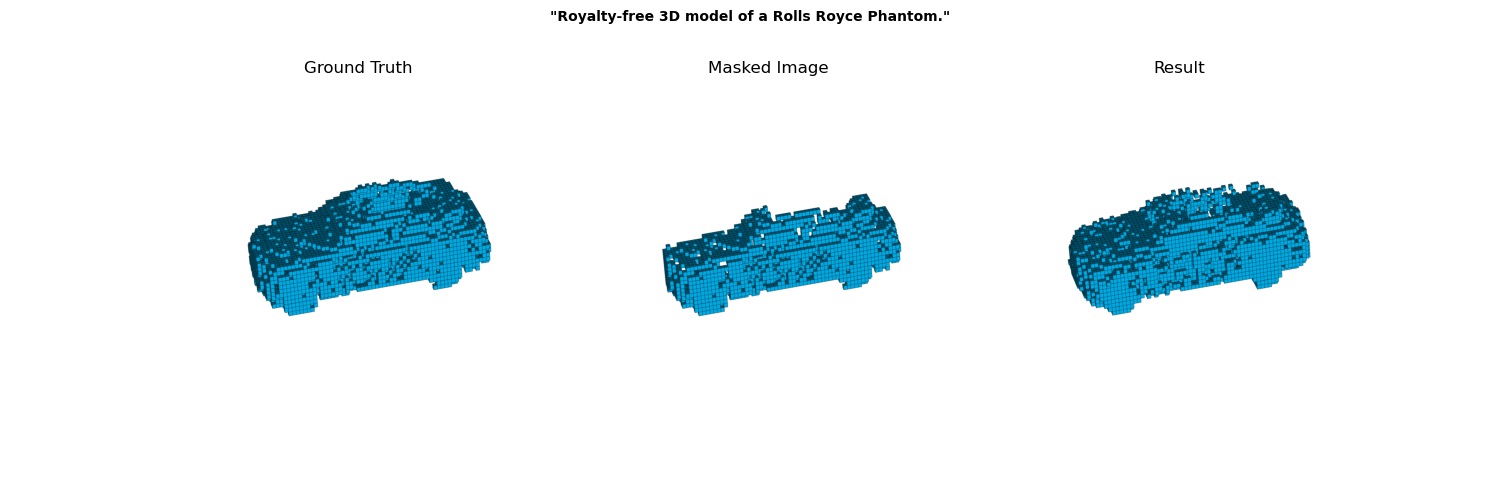}
\caption{Results Seg$50\%$}
\label{fig:Plane05-2}
\end{figure}

\begin{figure}[H] 
 \centering 
\leftfig{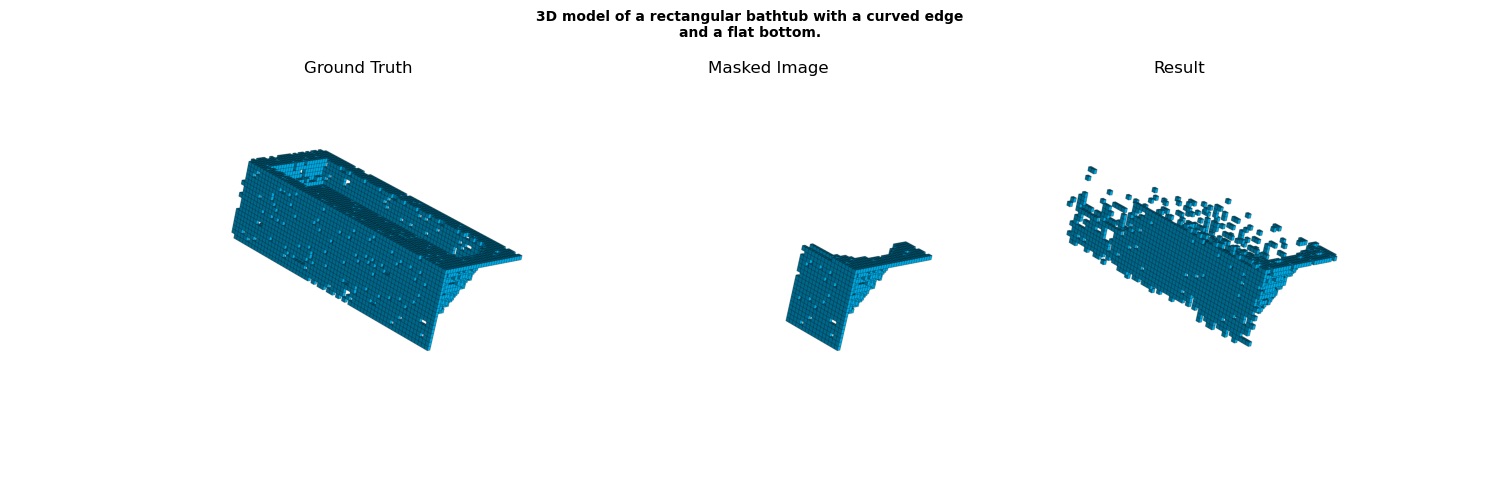}
\lrightfig{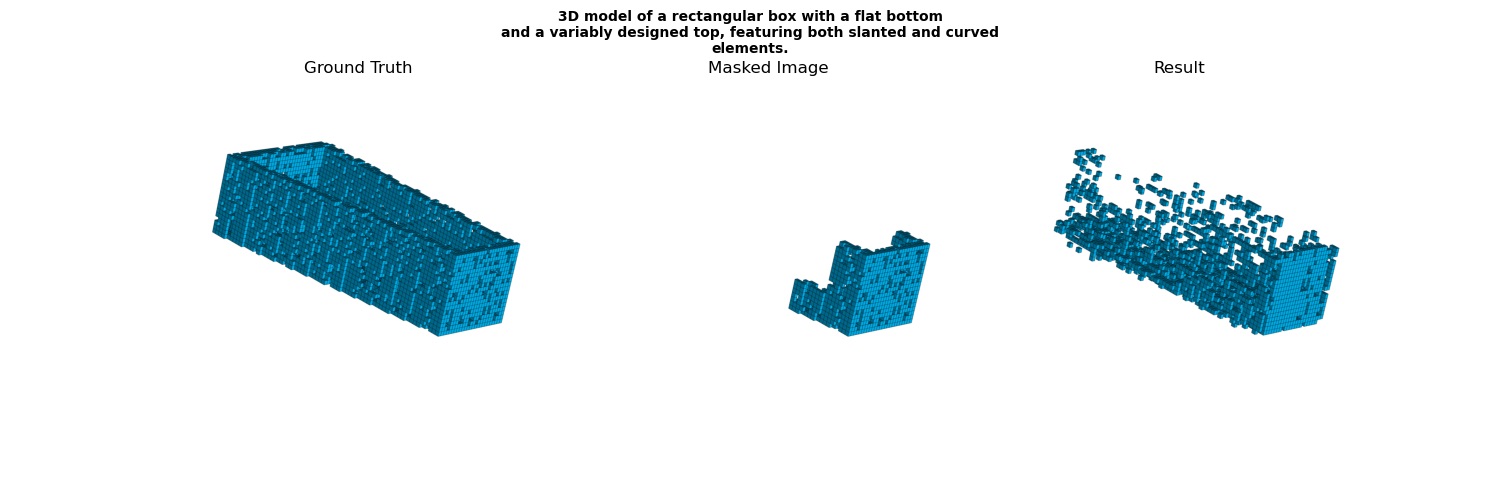}
\leftfig{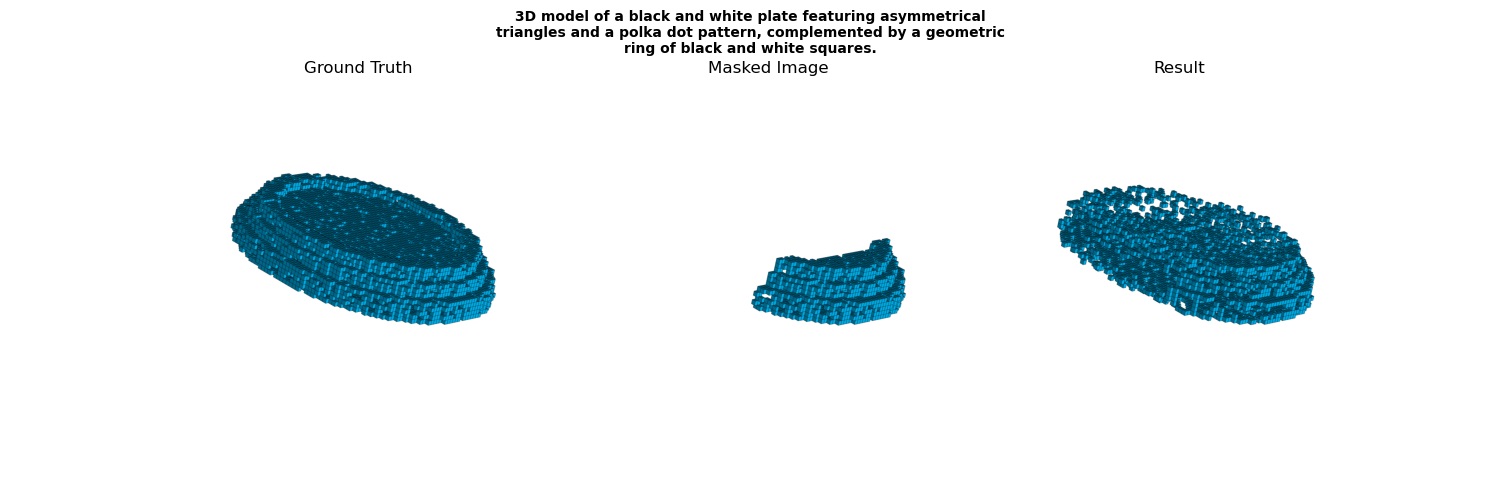}
\lrightfig{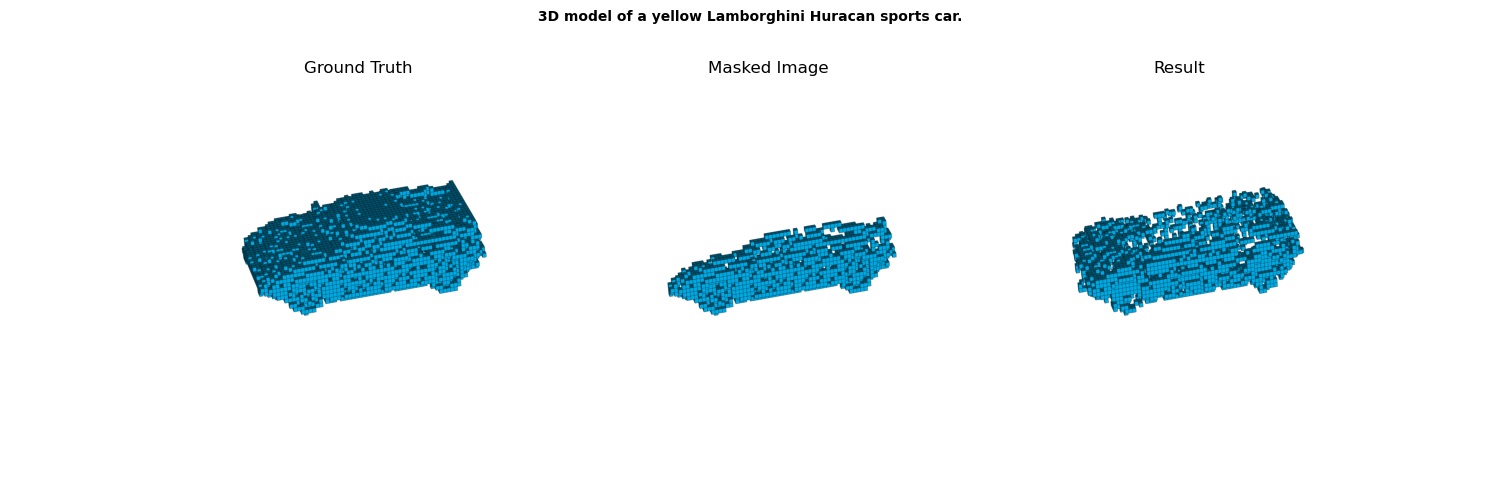}
\leftfig{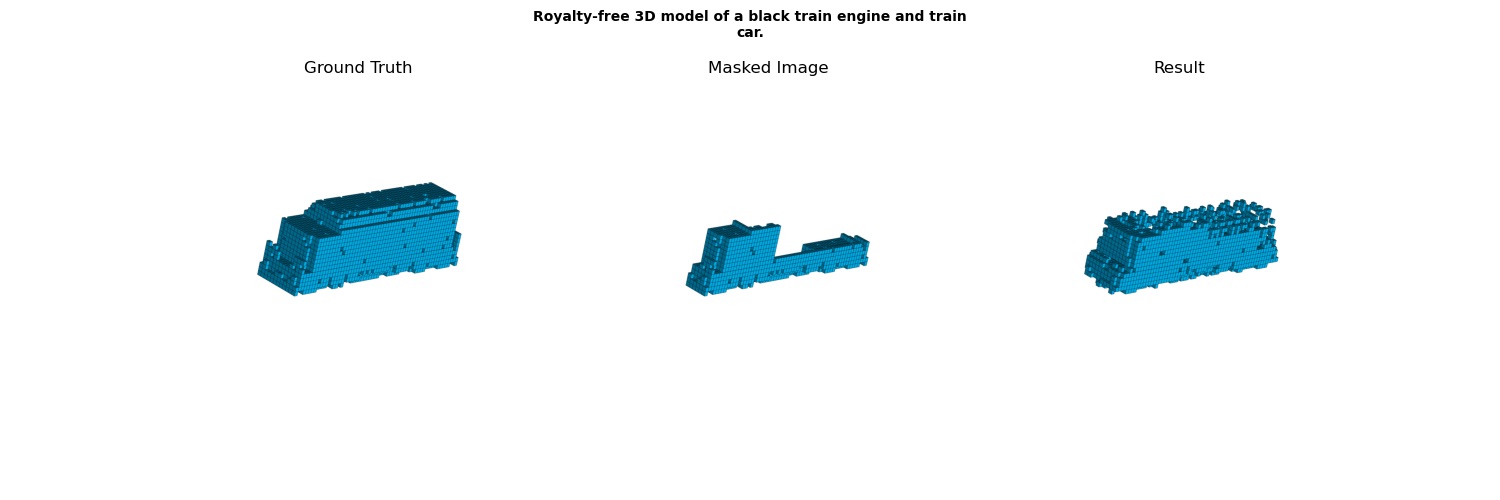}
\lrightfig{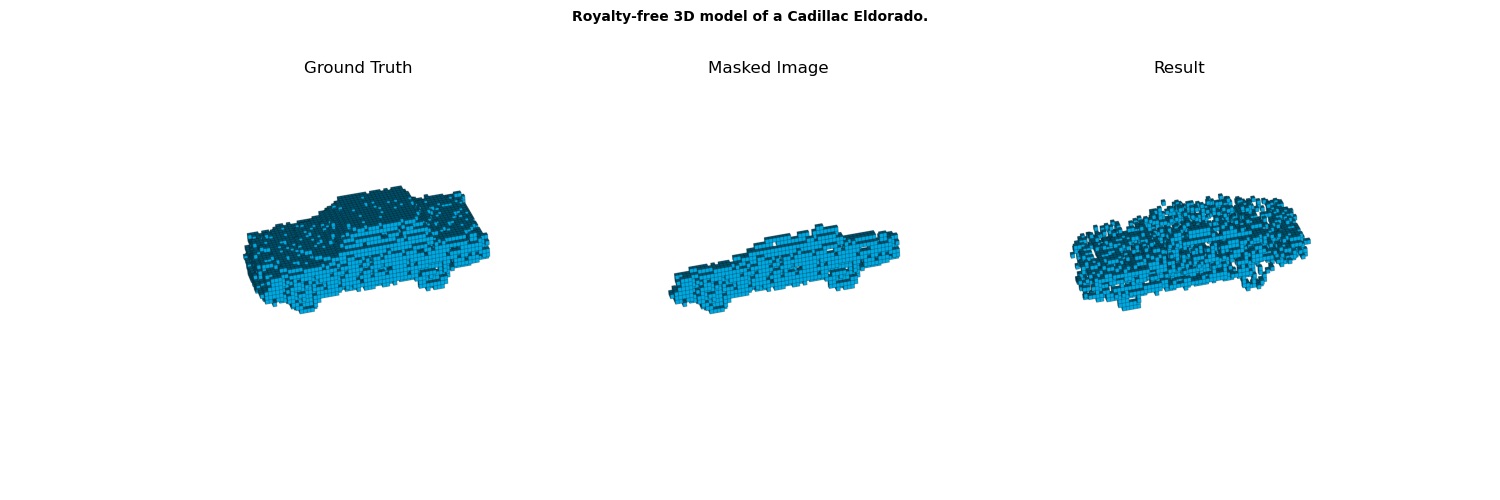}
\leftfig{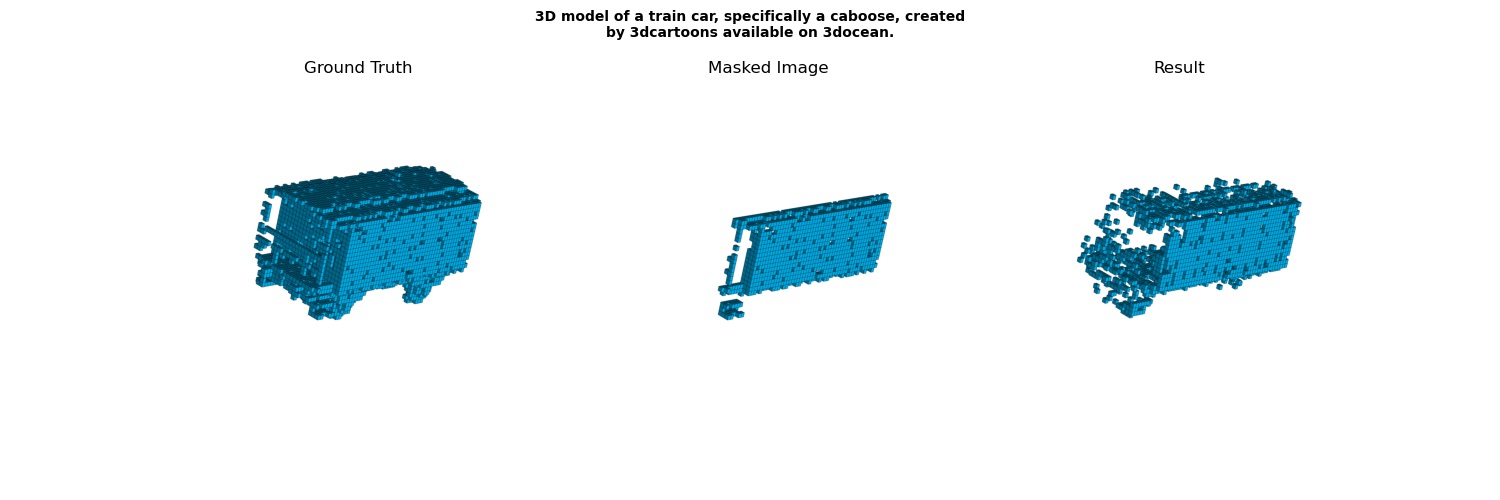}
\lrightfig{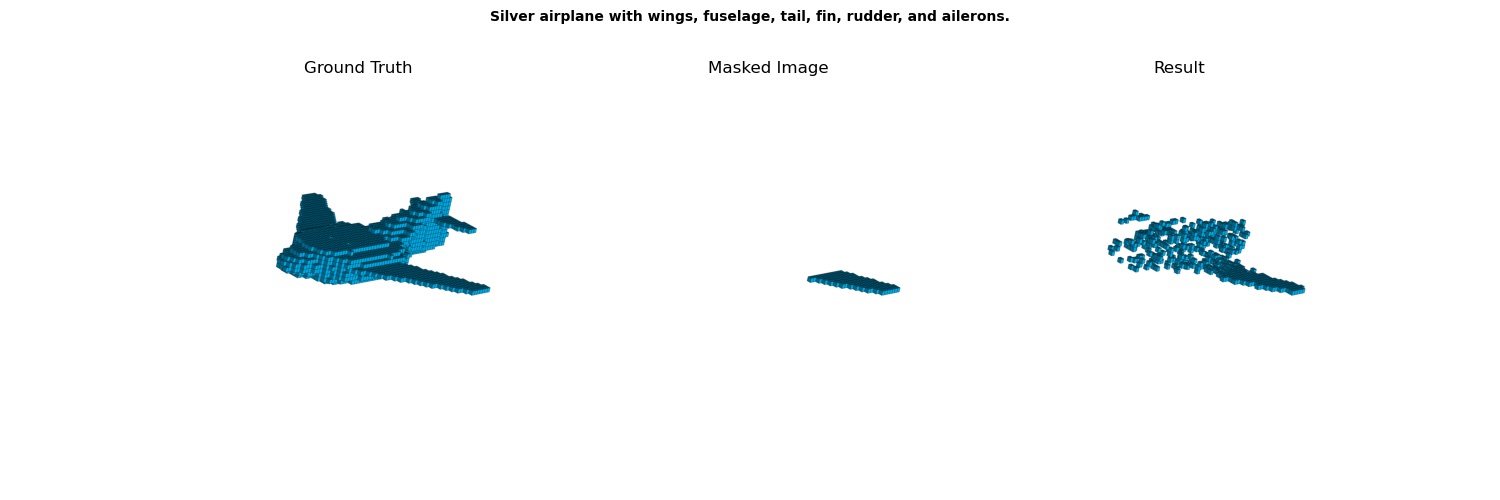}
\leftfig{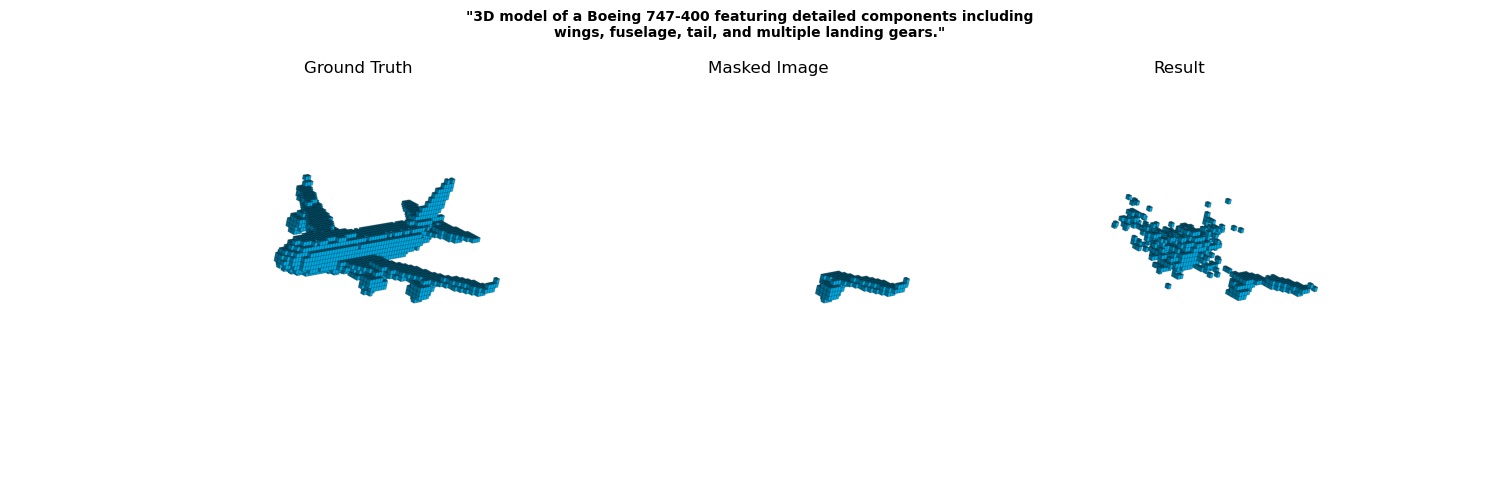}
\lrightfig{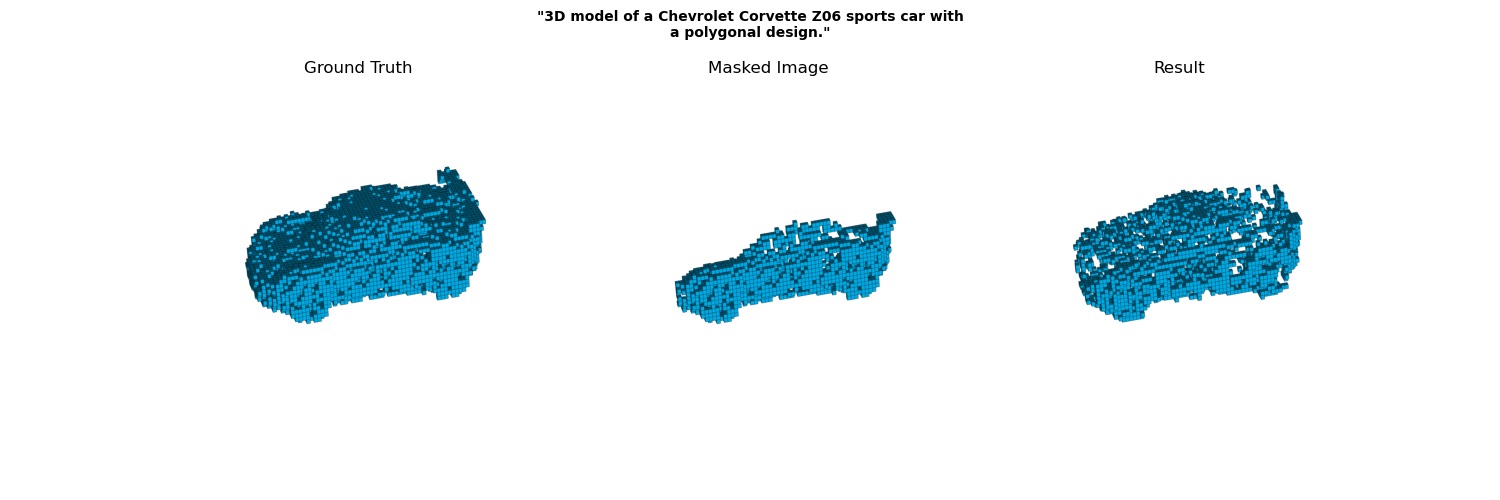}
\leftfig{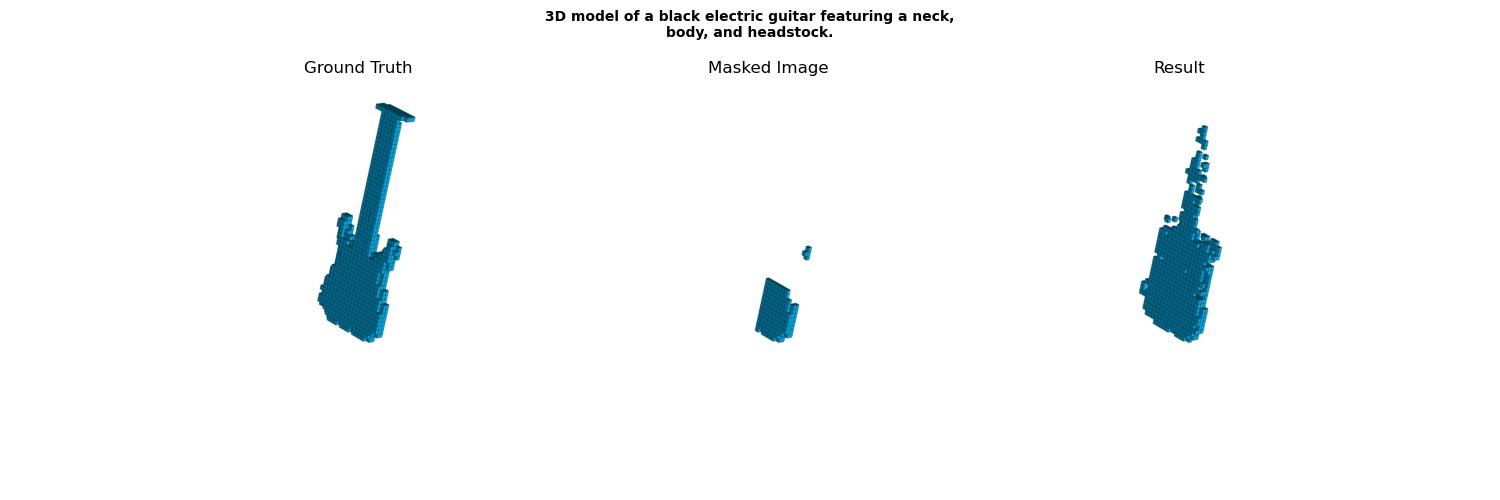}
\lrightfig{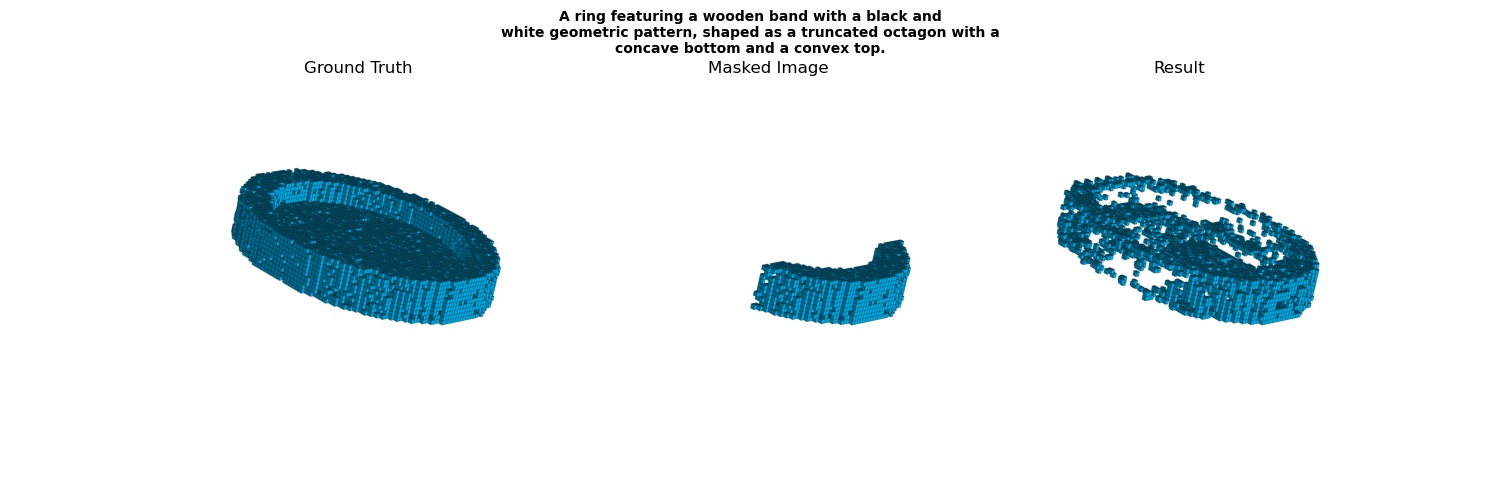}
\leftfig{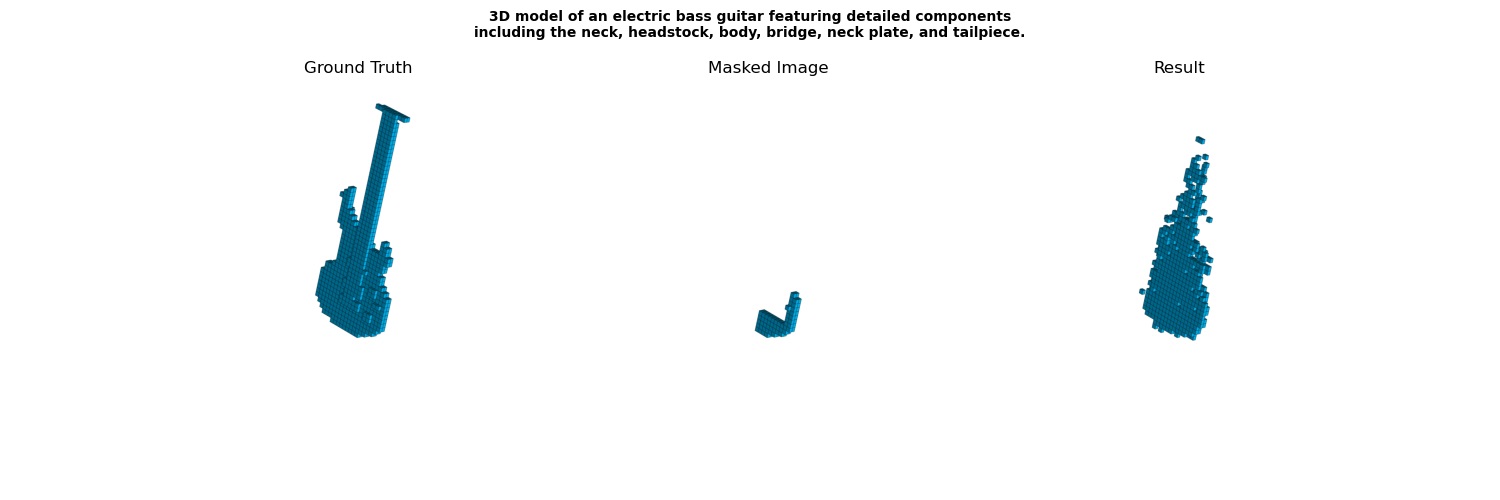}
\lrightfig{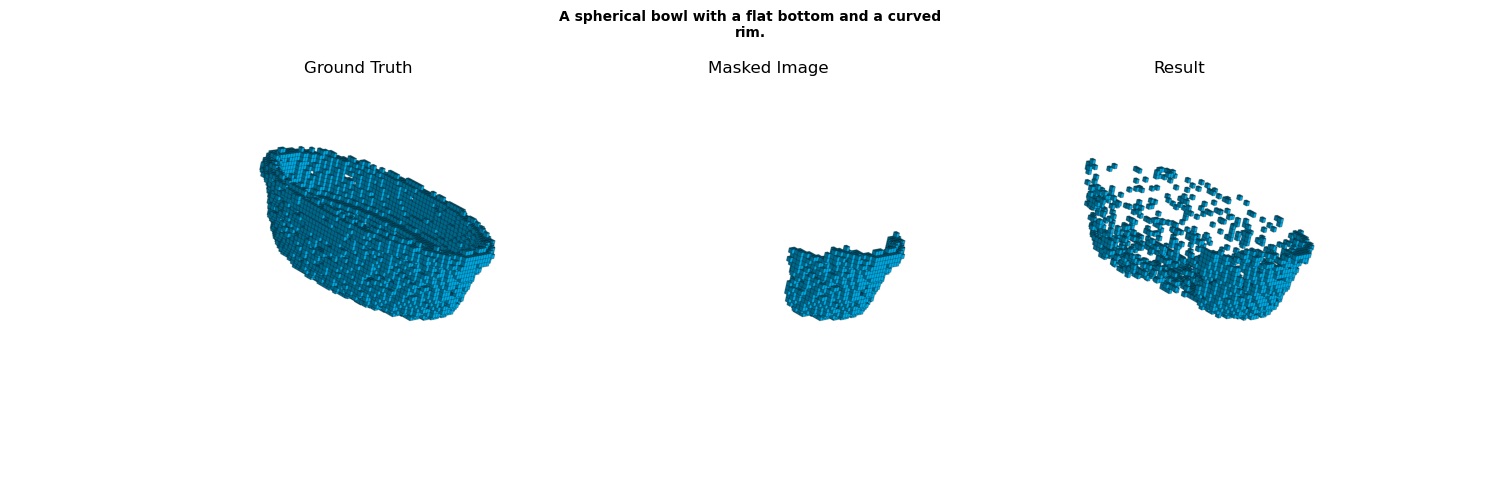}
\leftfig{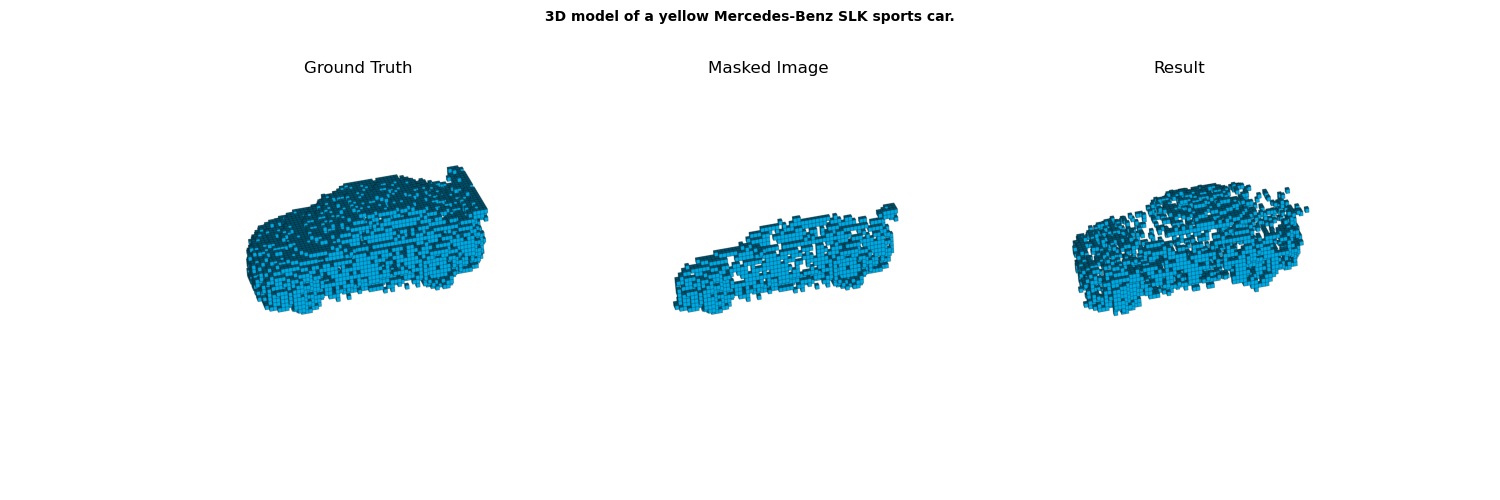}
\lrightfig{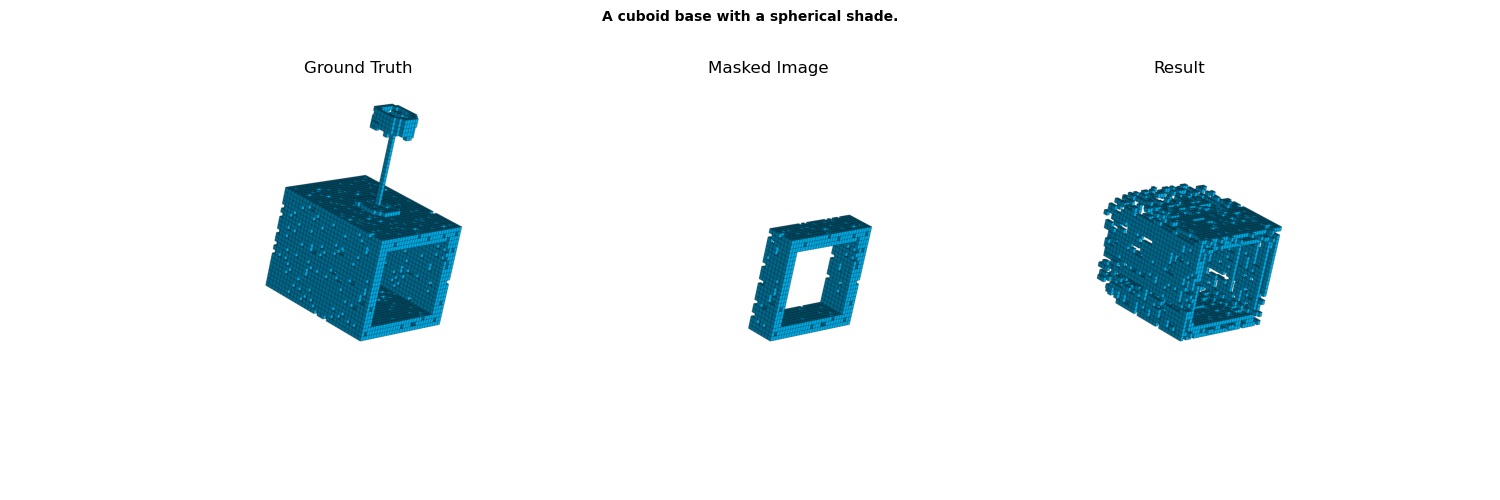}
\caption{Results Seg$80\%$}
\label{fig:plane08-1}
\end{figure}

\begin{figure}[H] 
 \centering 
\leftfig{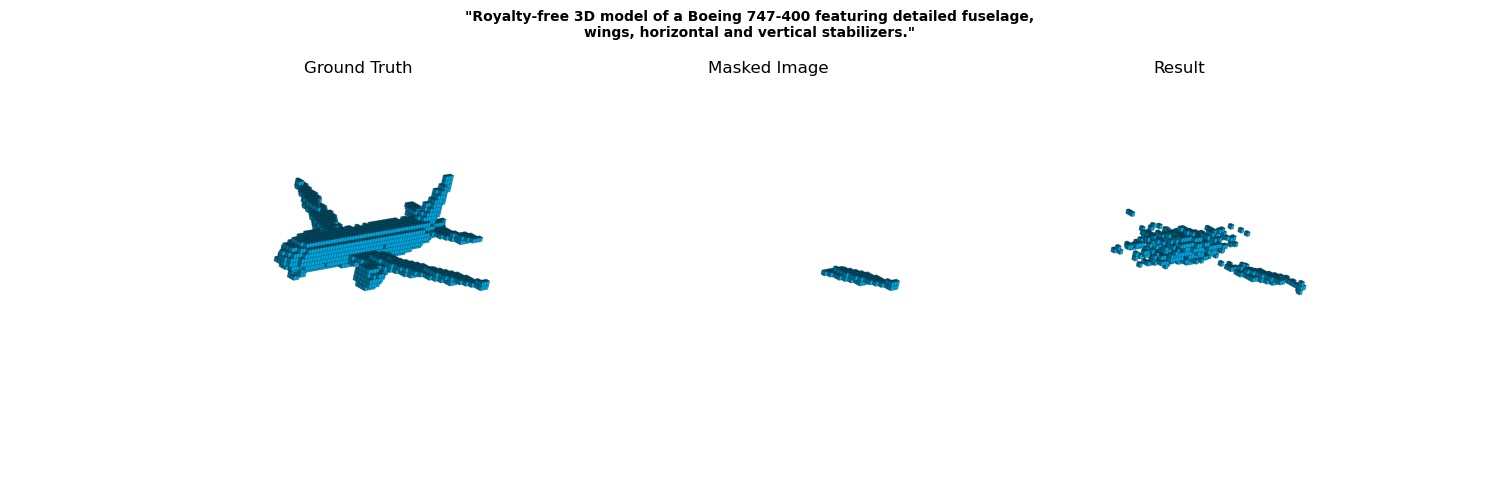}
\lrightfig{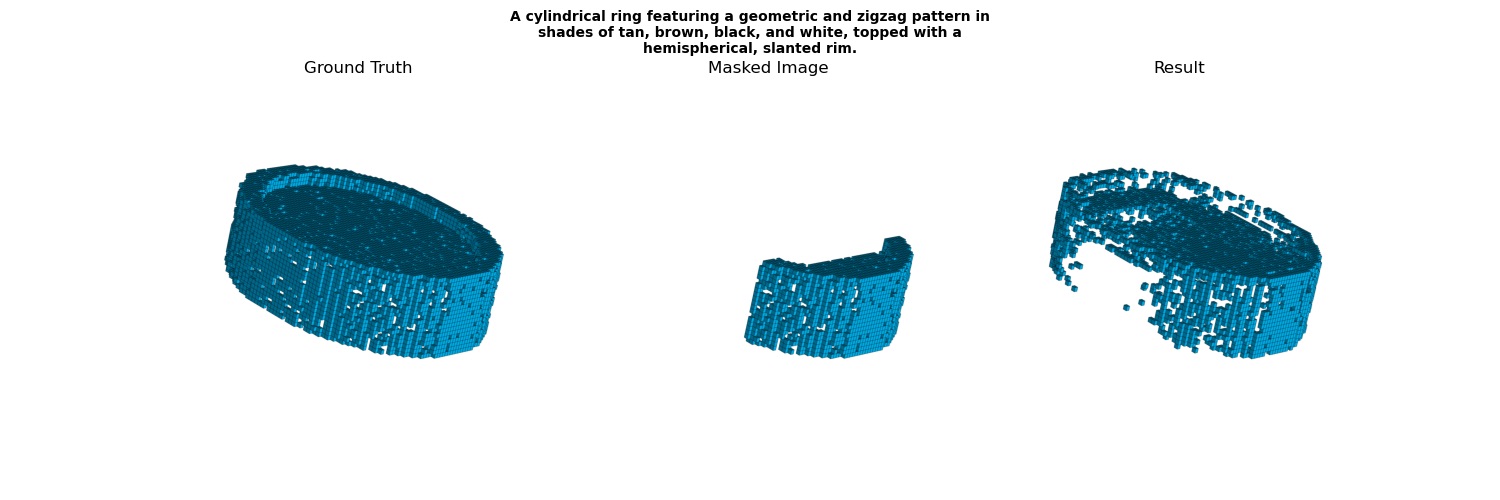}
\leftfig{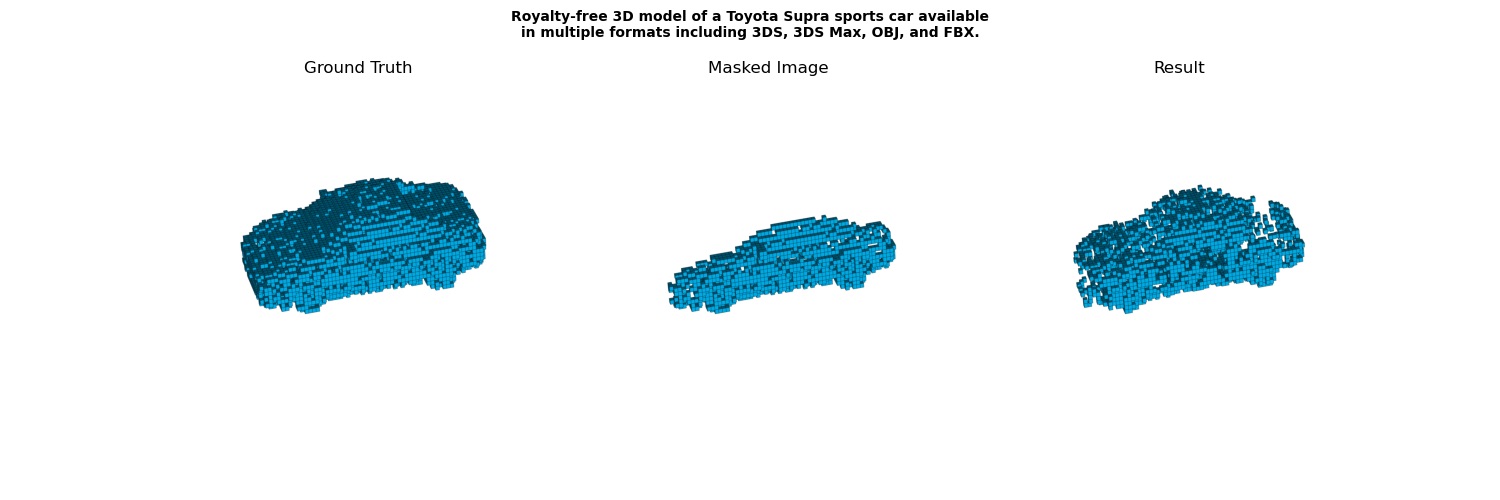}
\lrightfig{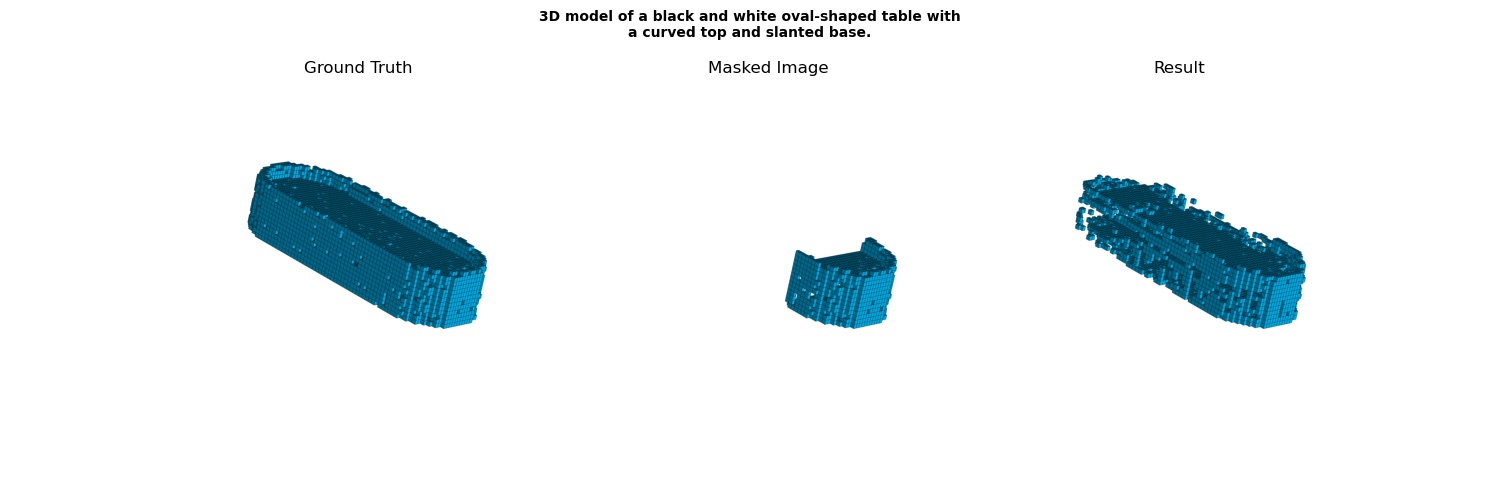}
\leftfig{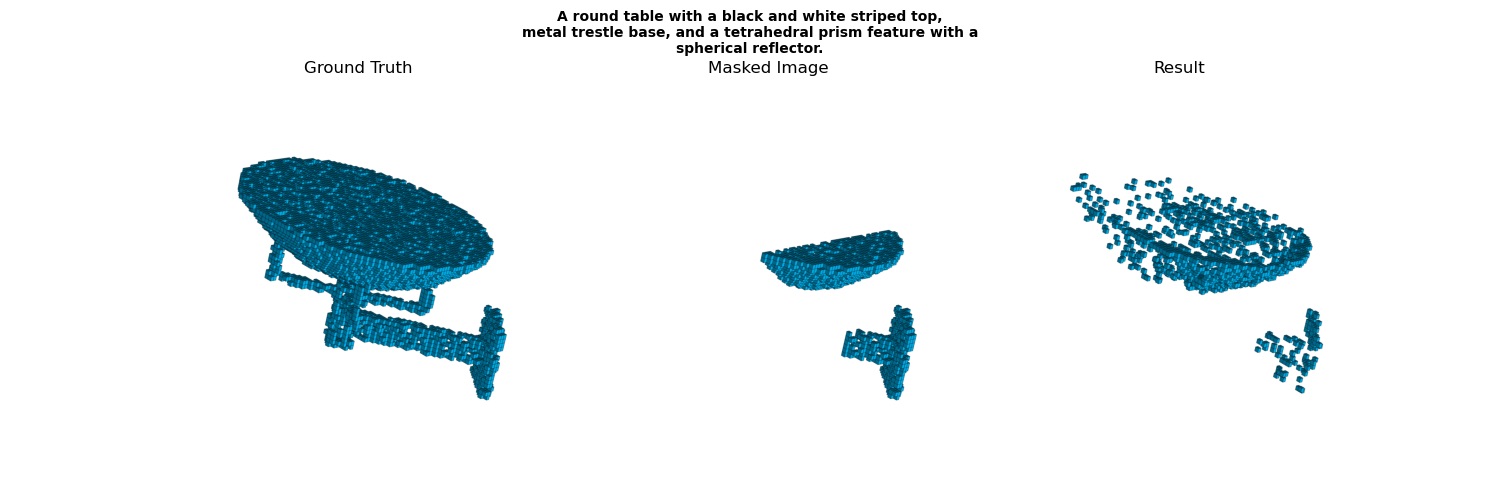}
\lrightfig{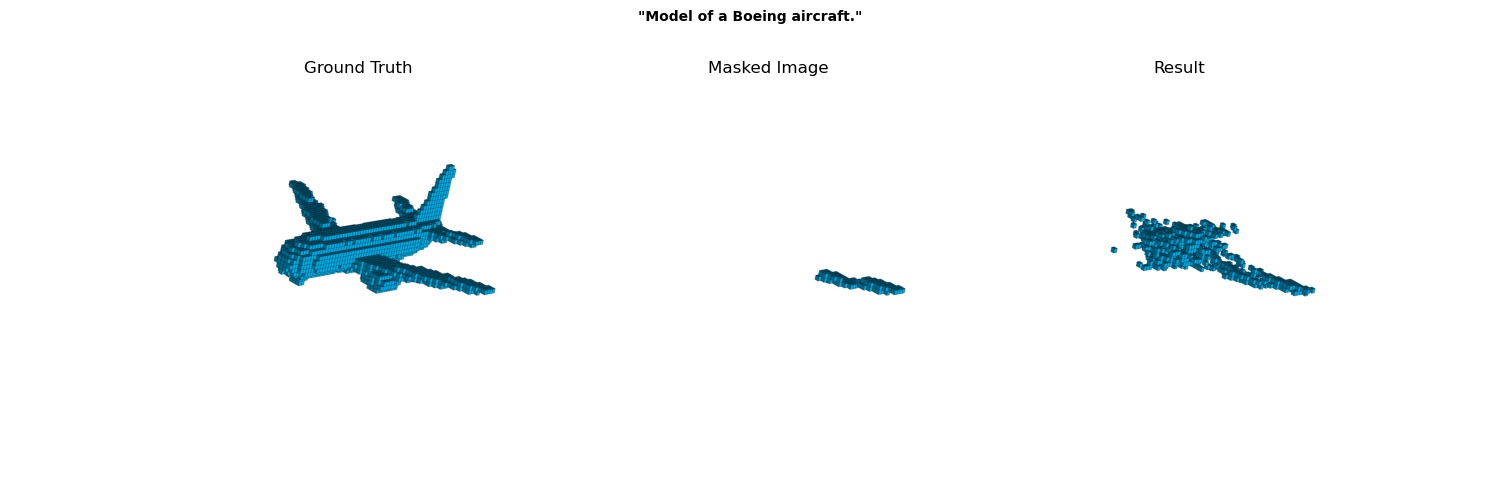}
\leftfig{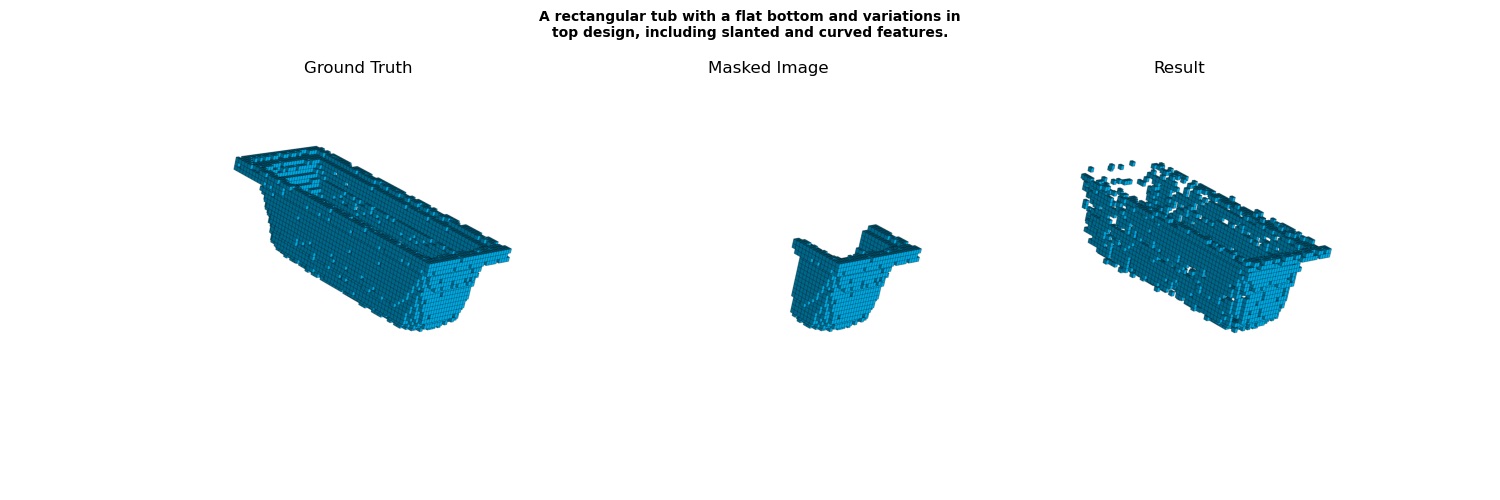}
\lrightfig{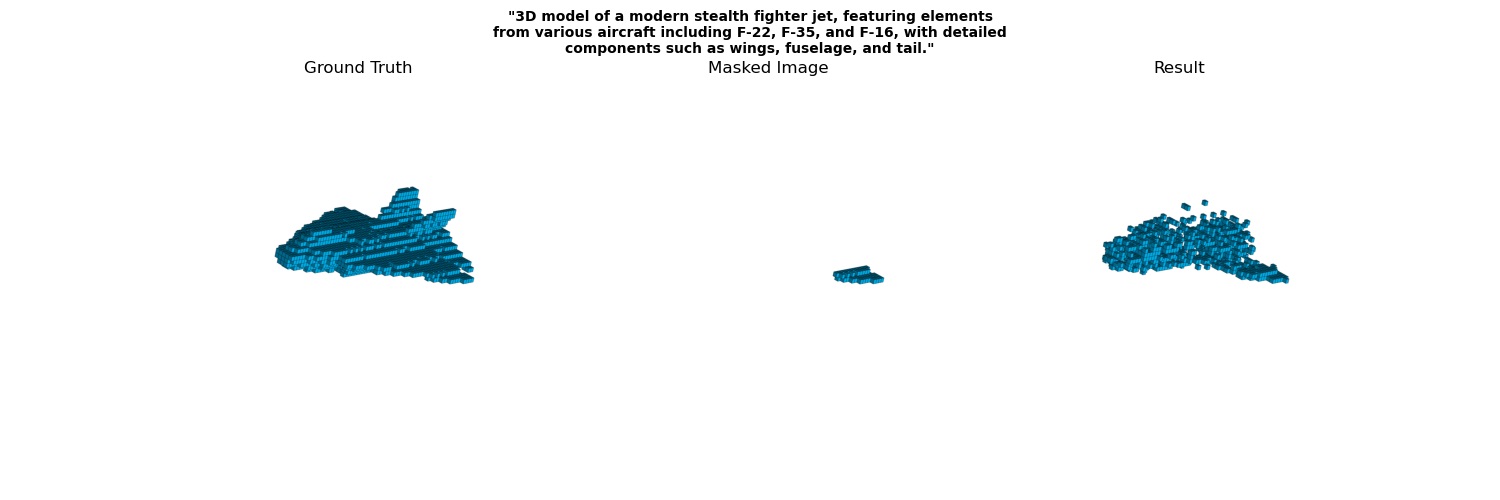}
\leftfig{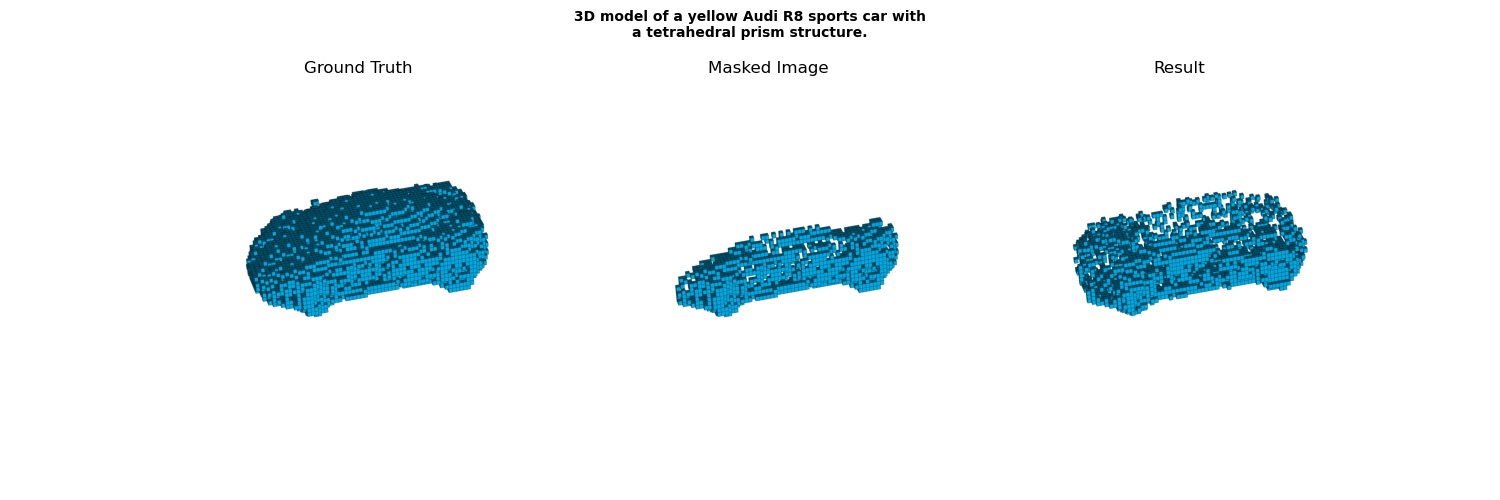}
\lrightfig{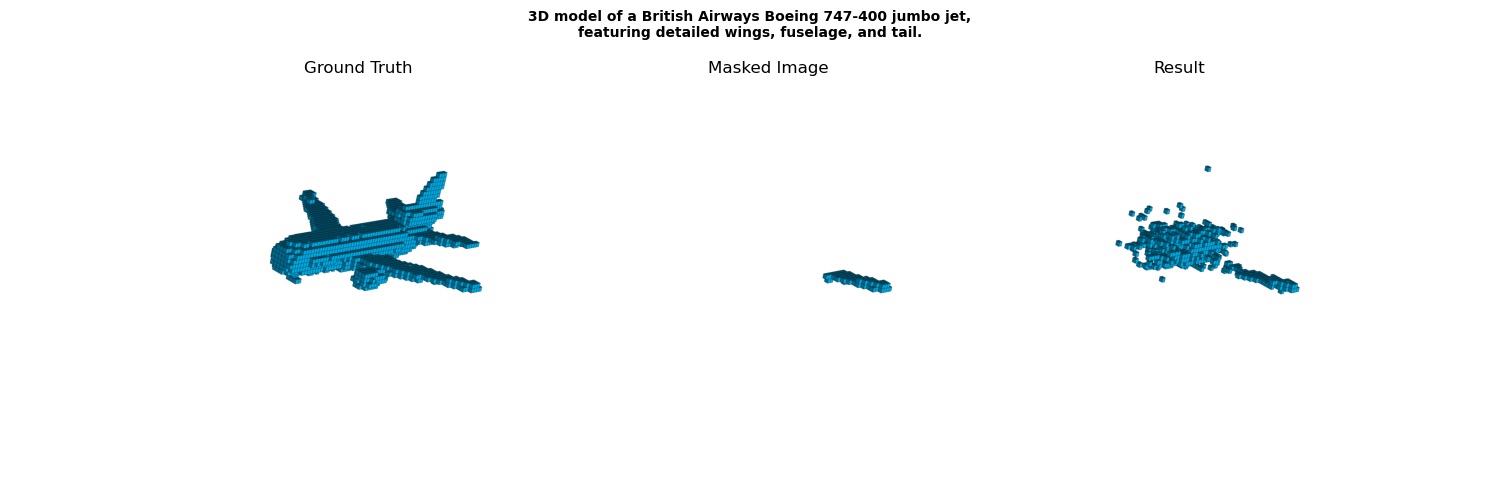}
\leftfig{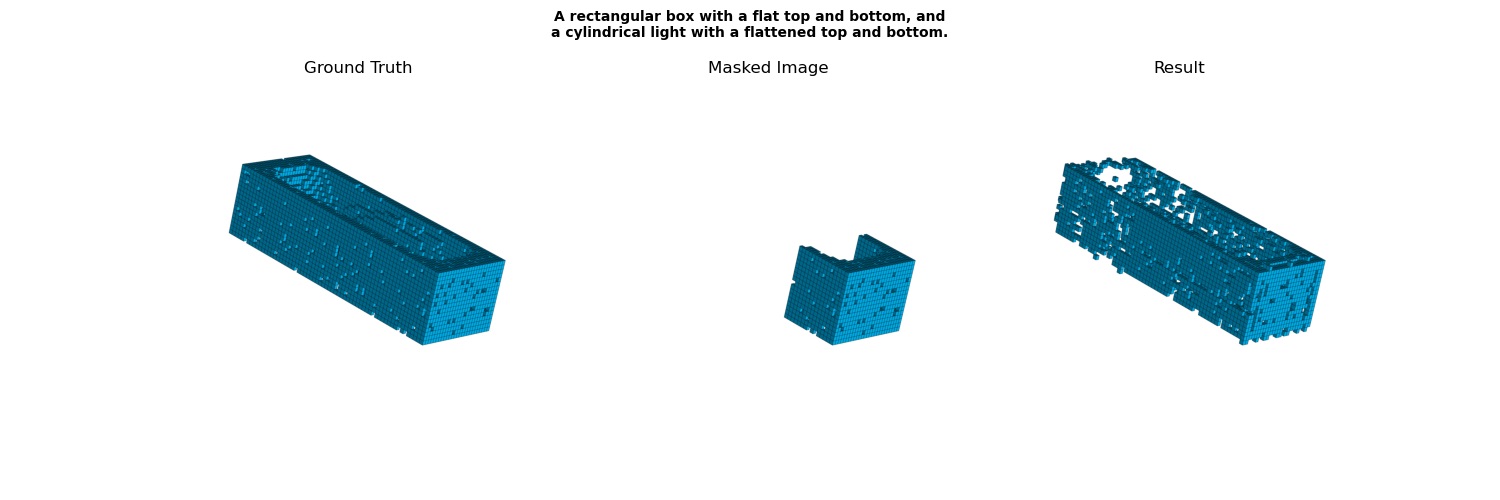}
\lrightfig{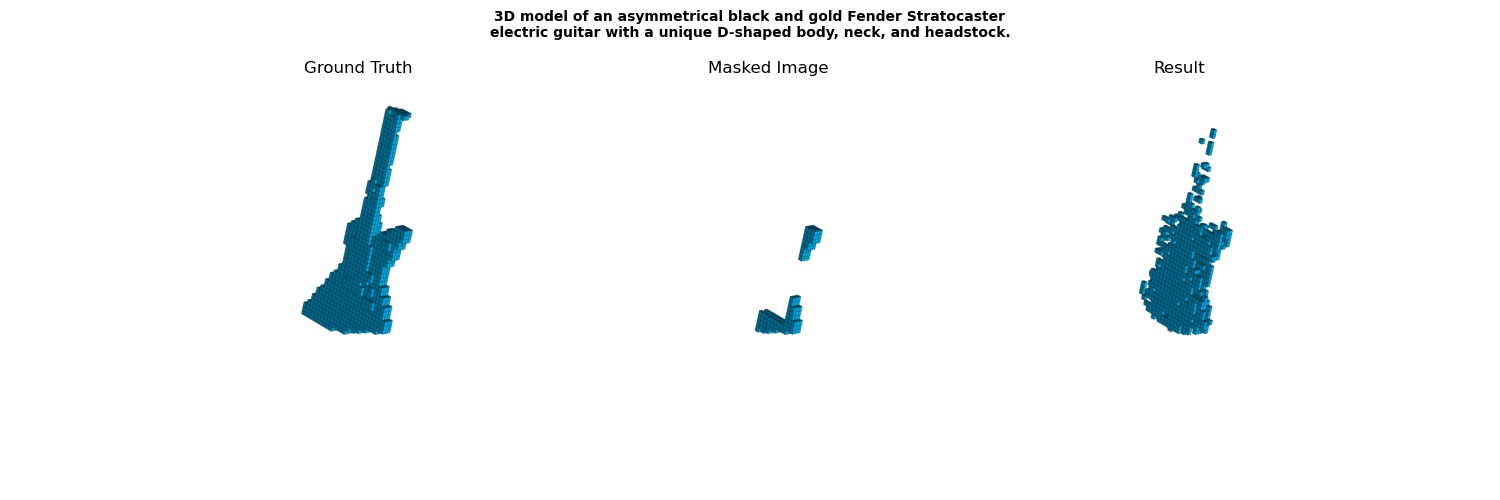}
\leftfig{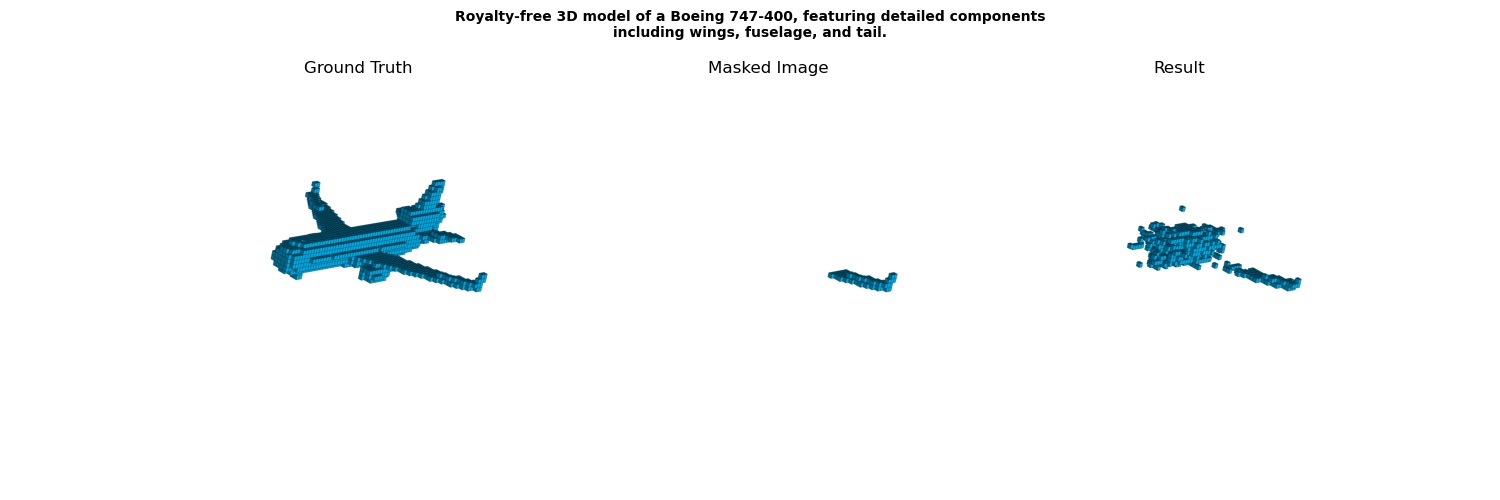}
\lrightfig{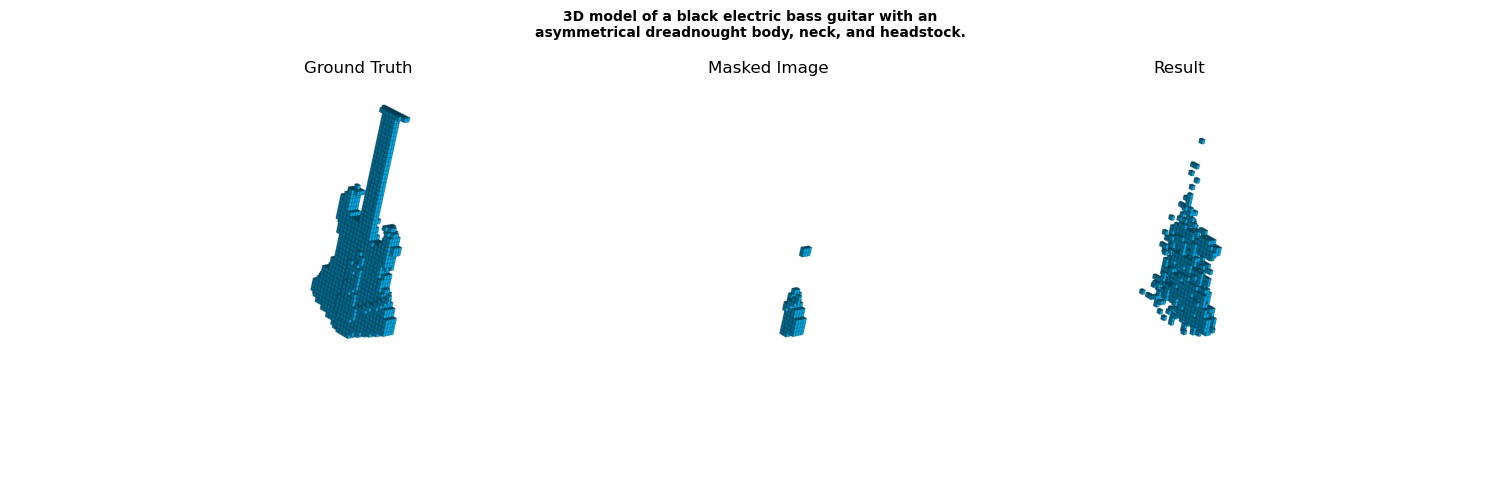}
\leftfig{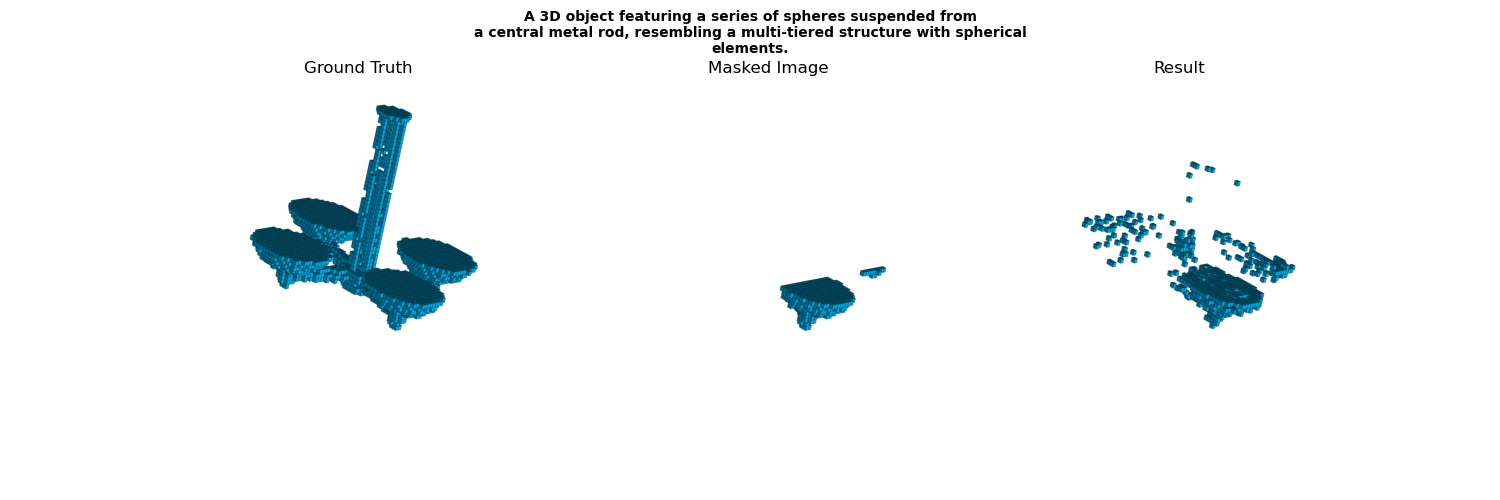}
\lrightfig{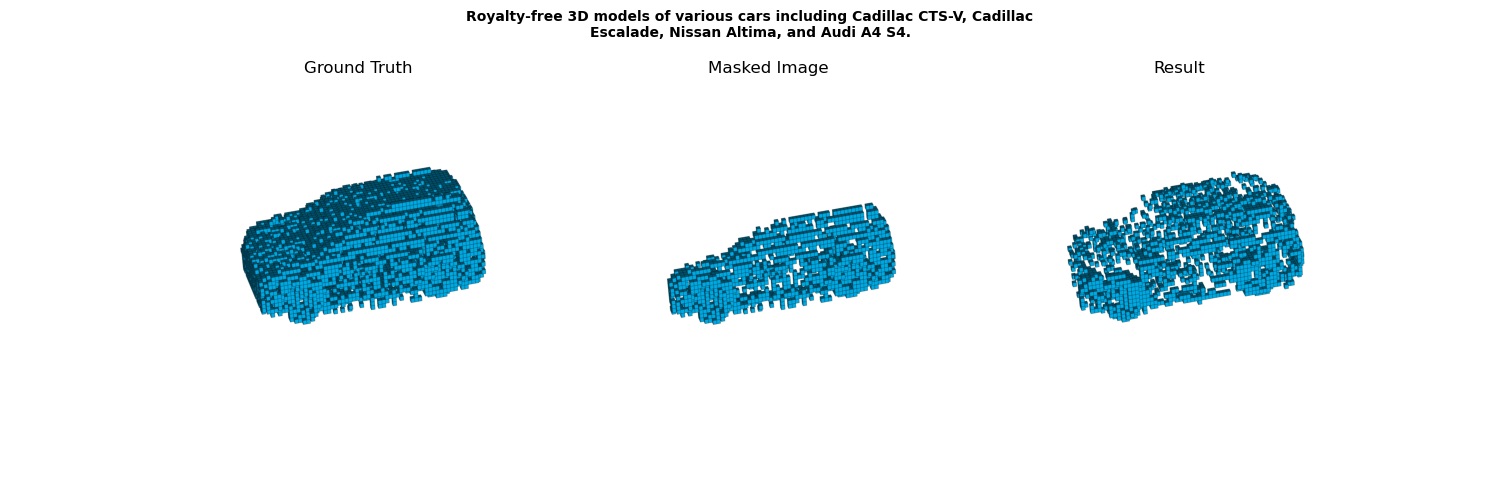}
\caption{Results Seg$80\%$}
\label{fig:Plane08-2}
\end{figure}

\begin{figure}[H]
    \centering
    \leftfig{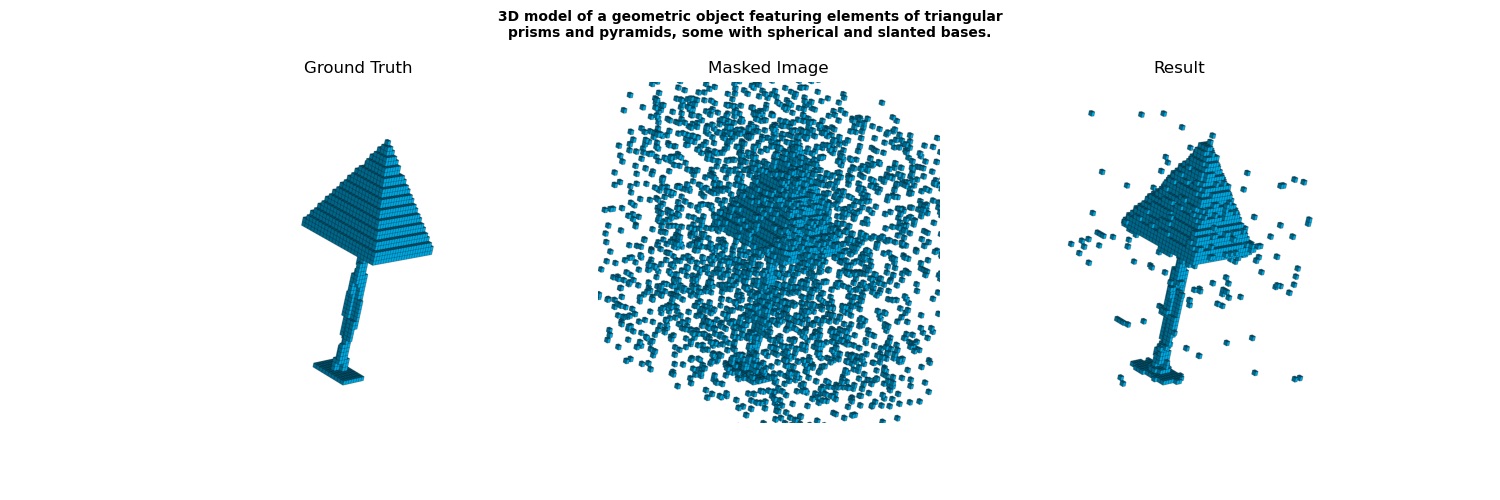}
    \lrightfig{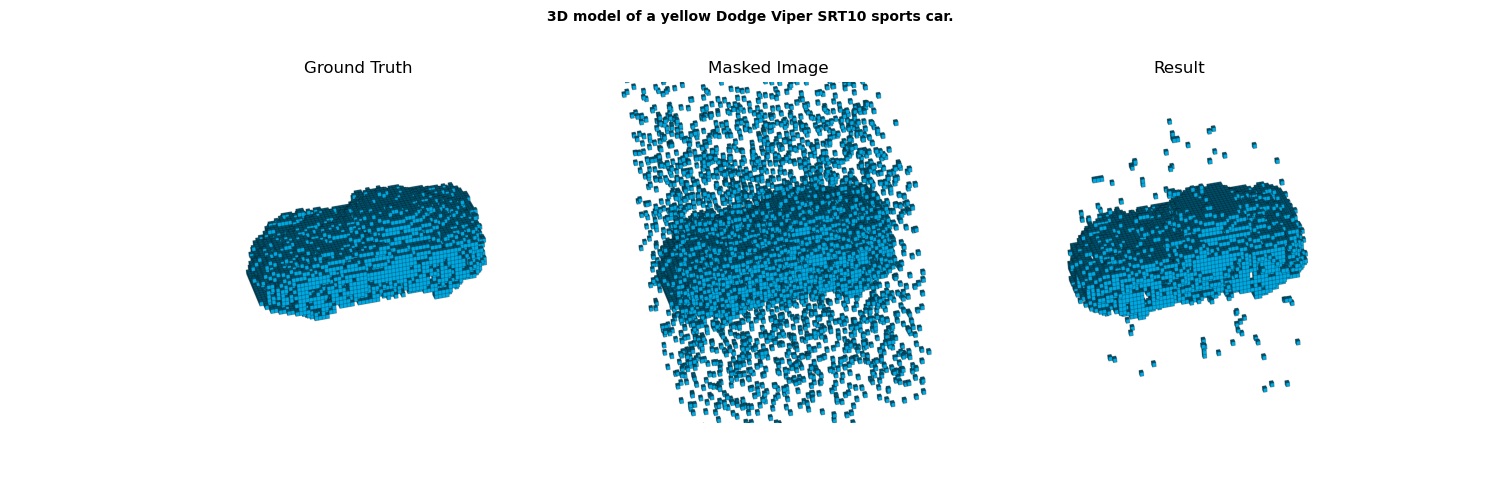}
    \leftfig{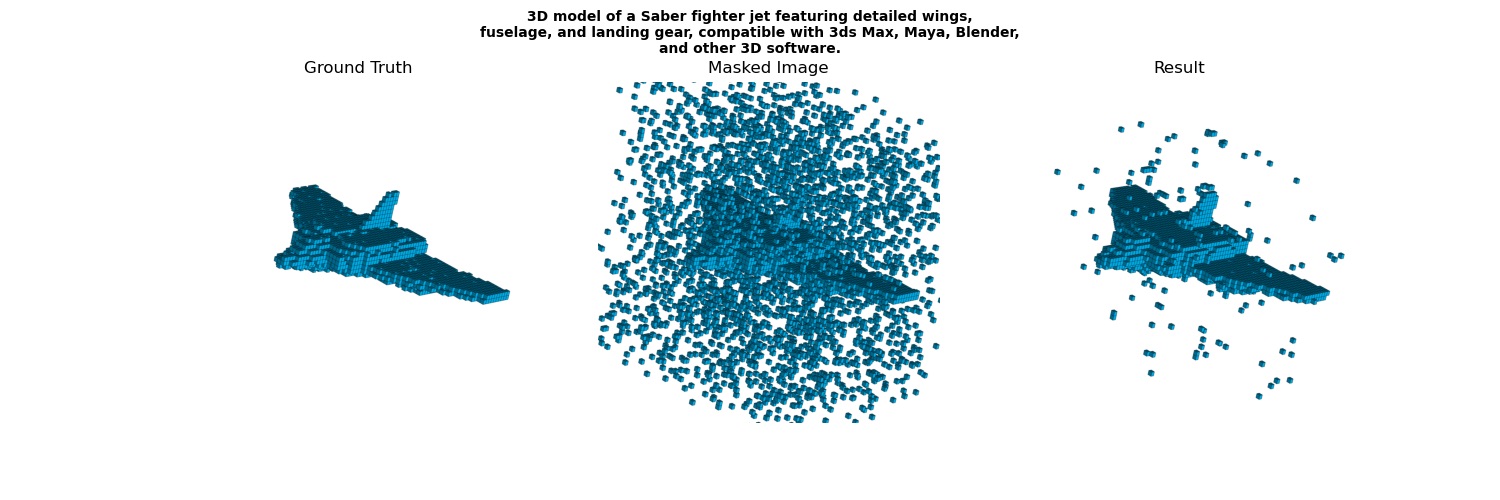}
    \lrightfig{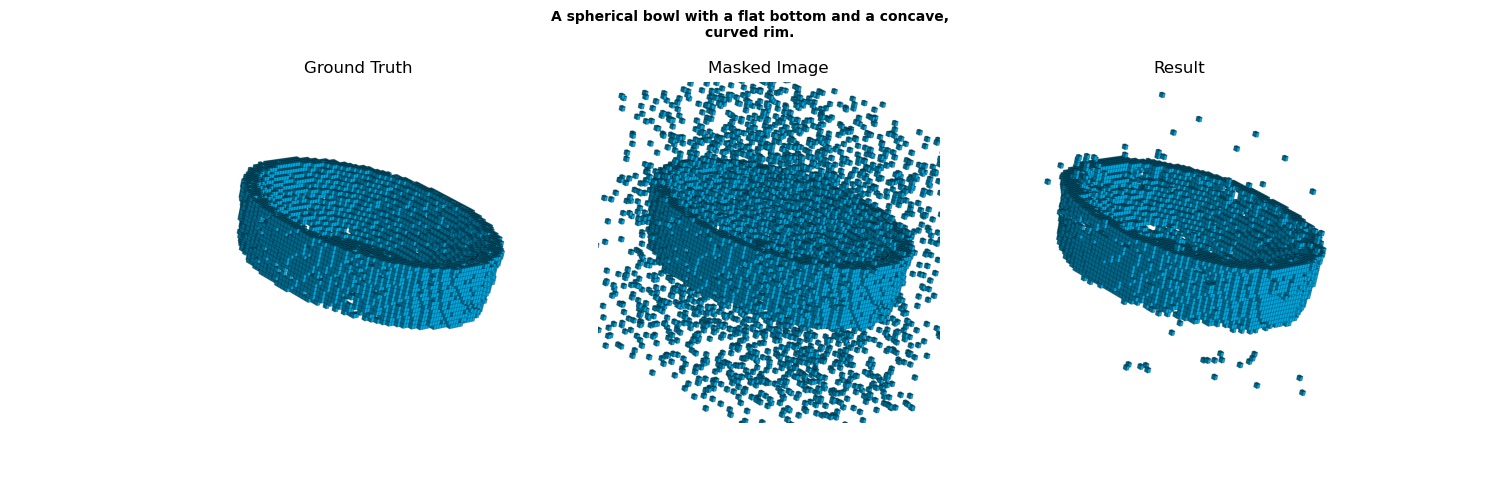}
    \leftfig{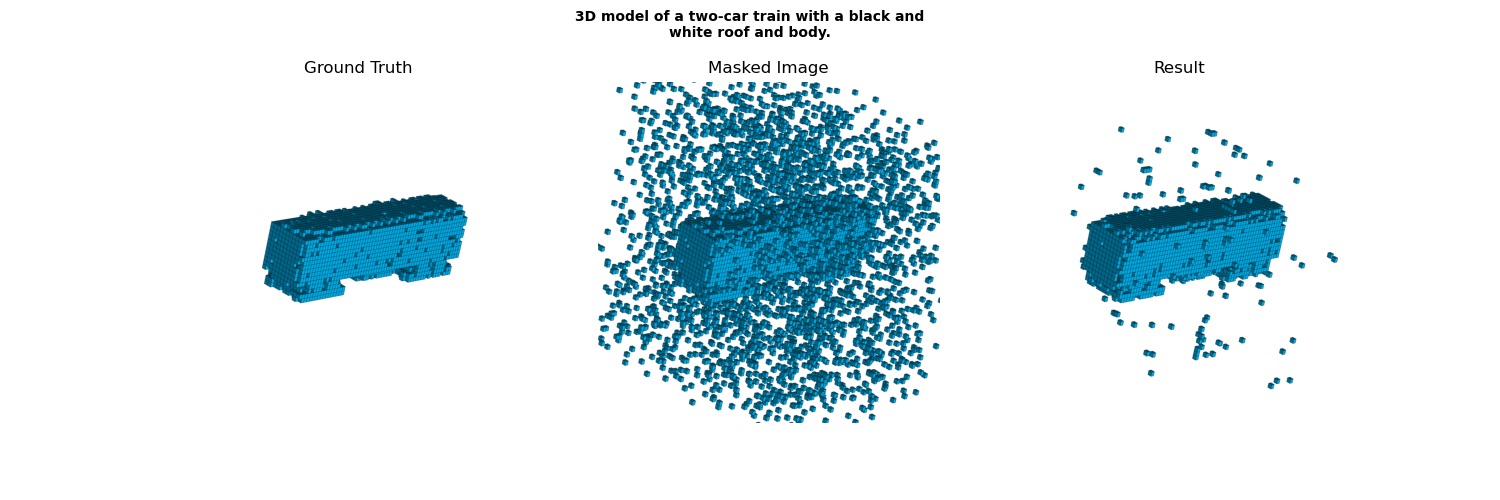}
    \lrightfig{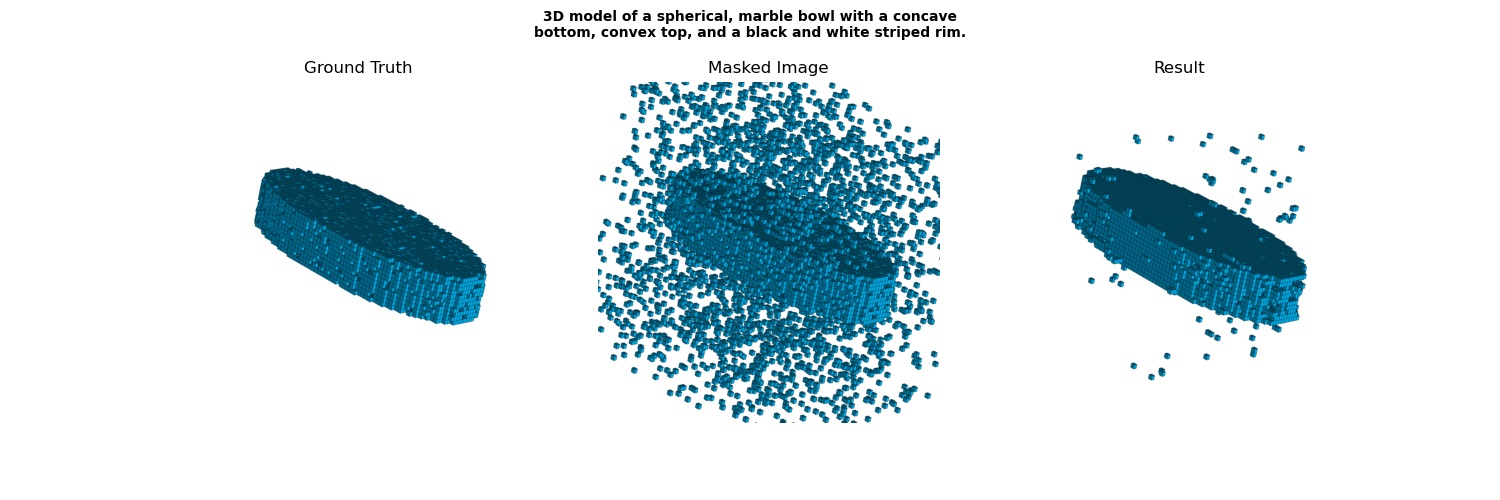}
    \leftfig{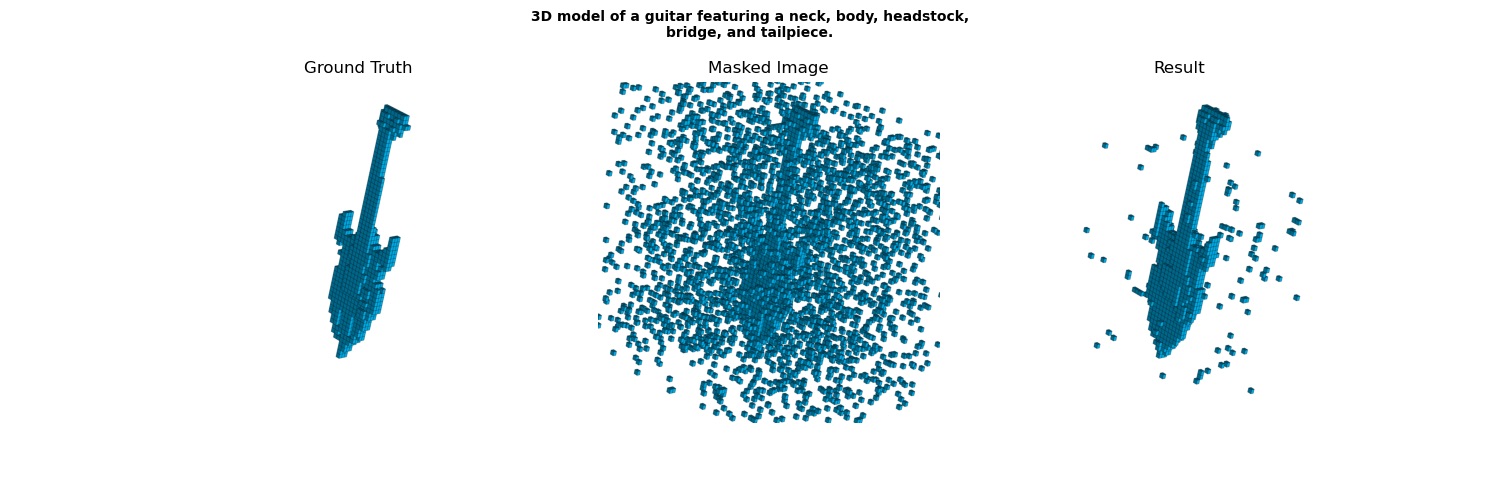}
    \lrightfig{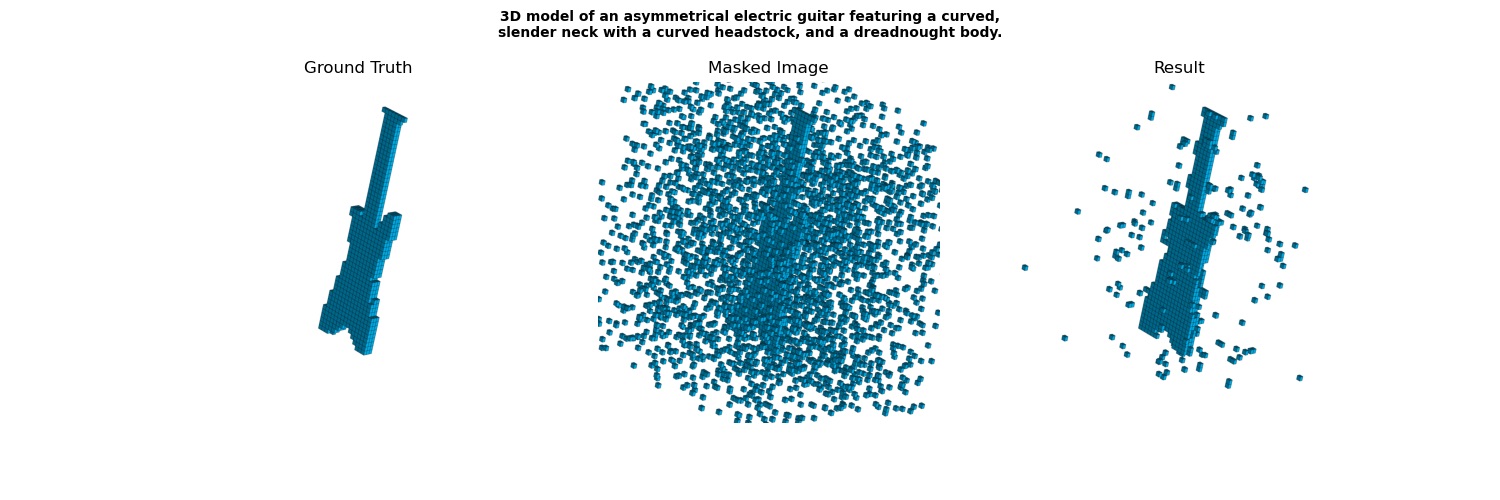}
    \leftfig{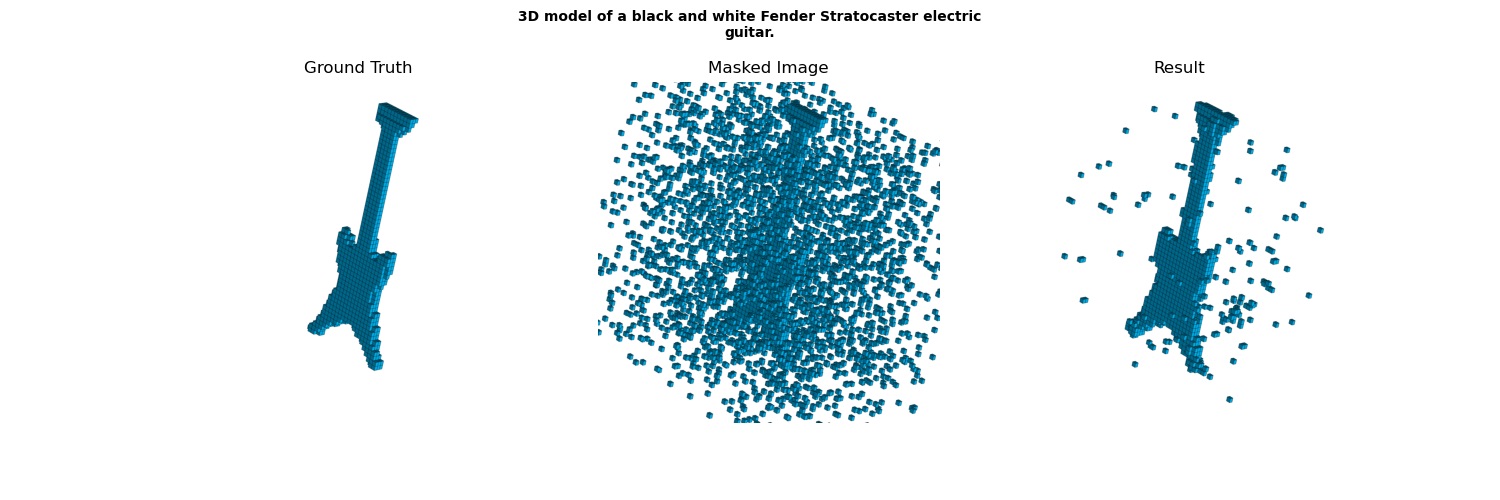}
    \lrightfig{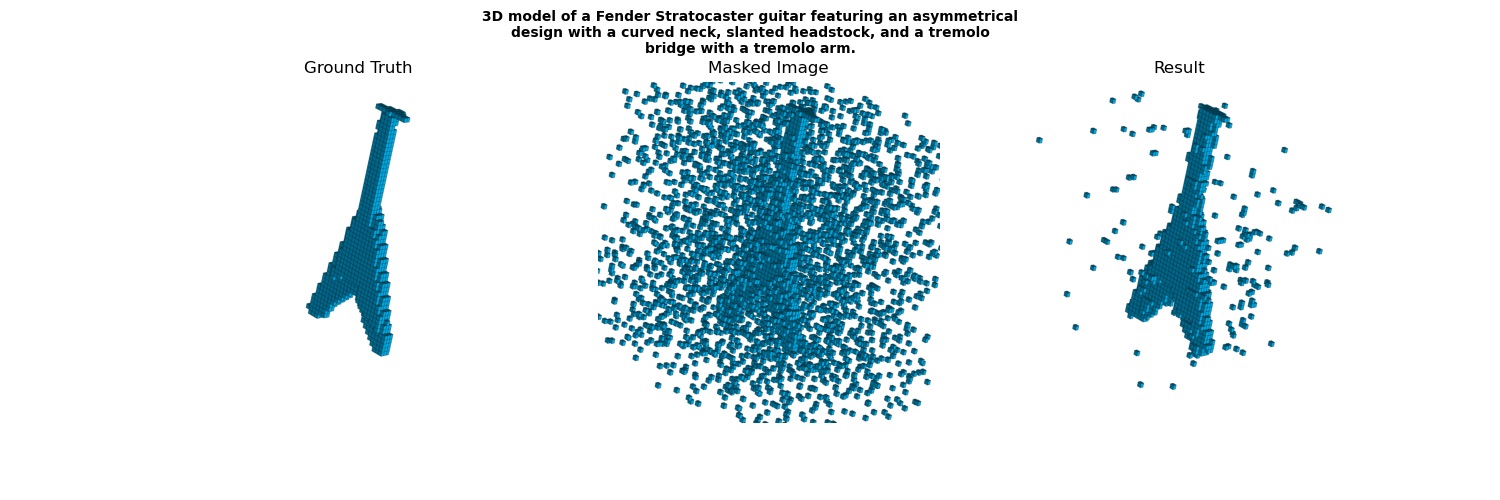}
    \leftfig{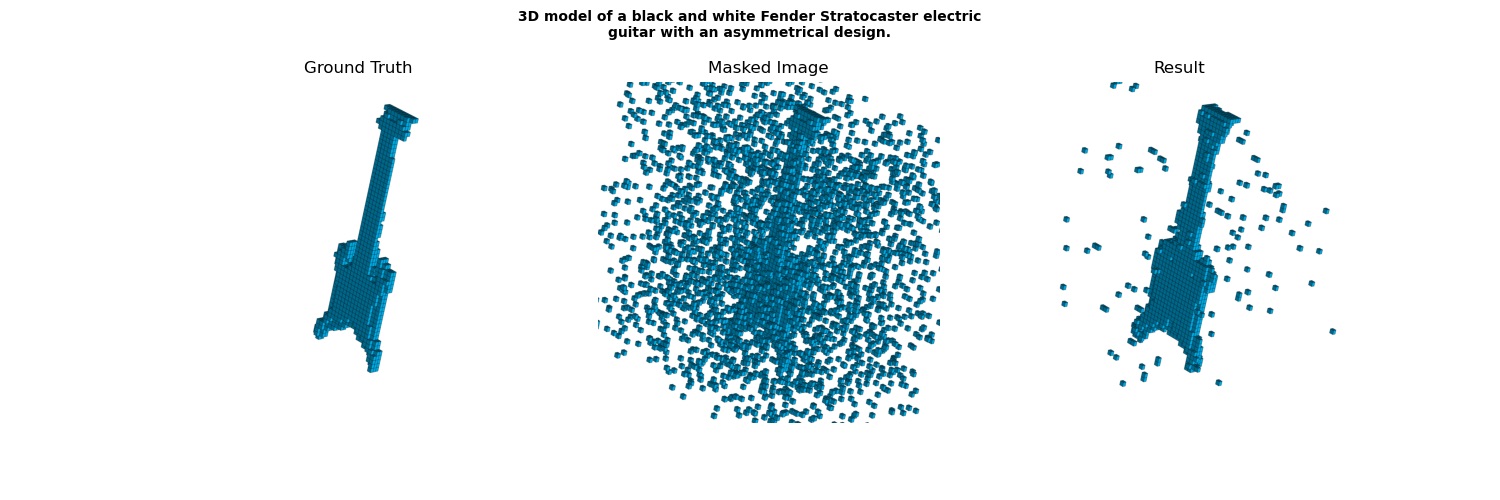}
    \lrightfig{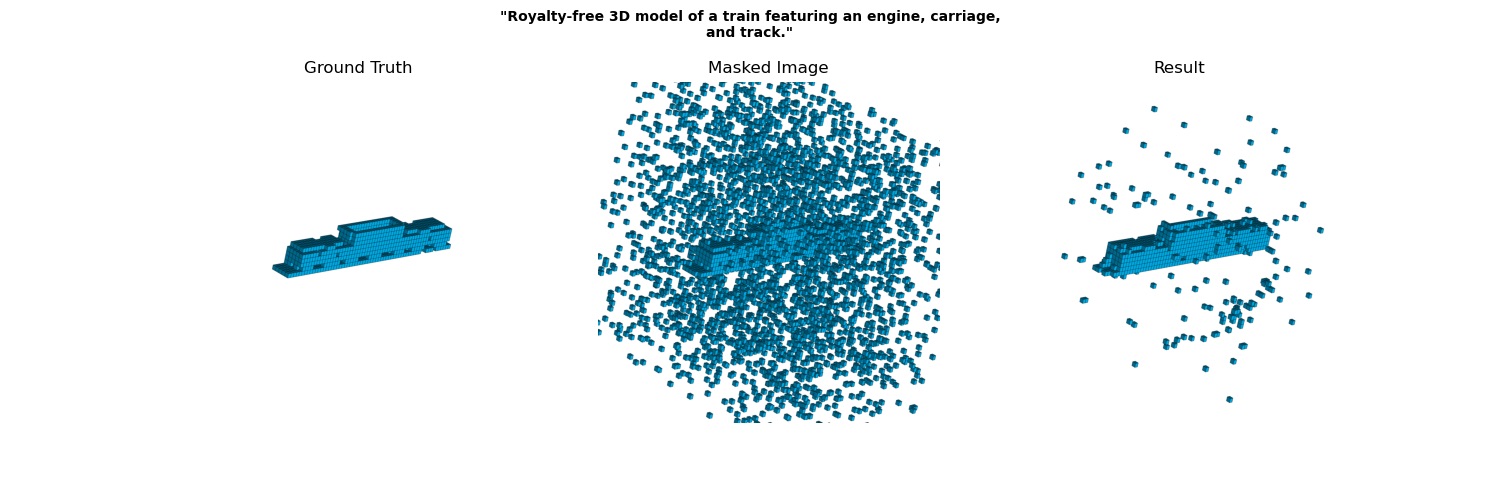}
    \leftfig{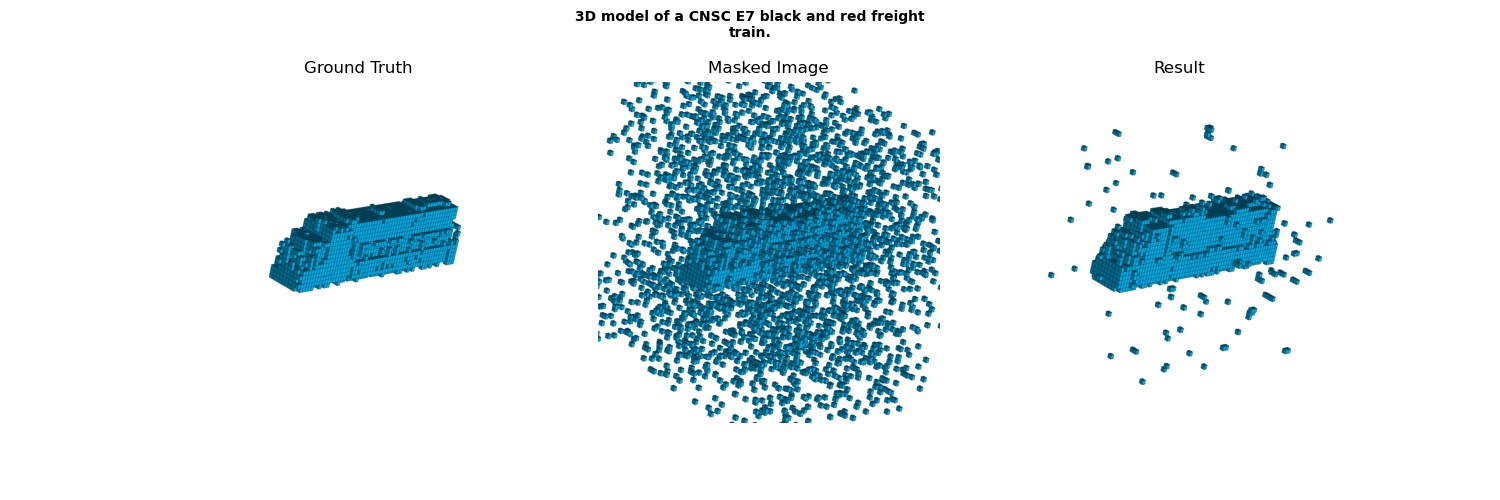}
    \lrightfig{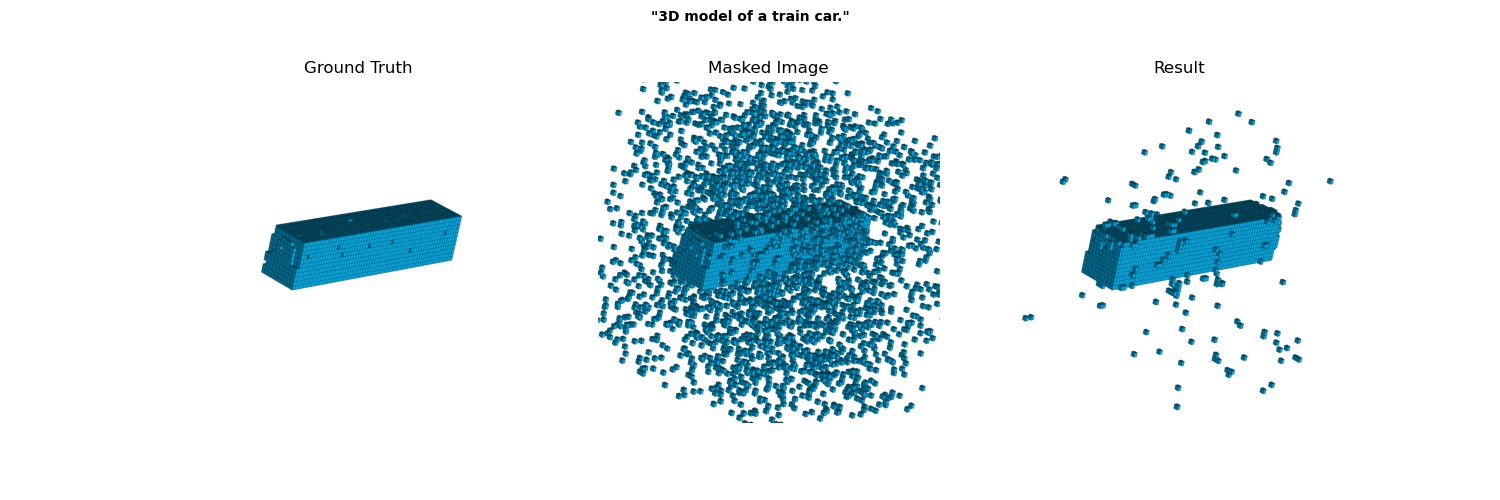}
    \caption{Results Noise$1\%$}
    \label{fig:Noise001-1}
\end{figure}

\begin{figure}[H]
    \centering
    \leftfig{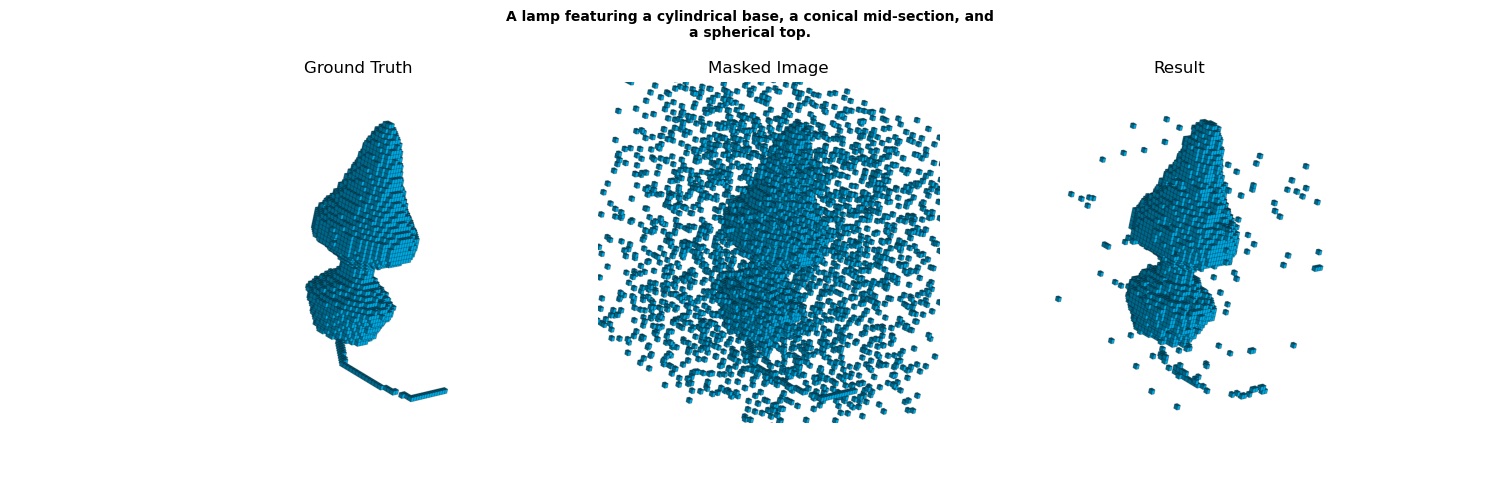}
    \lrightfig{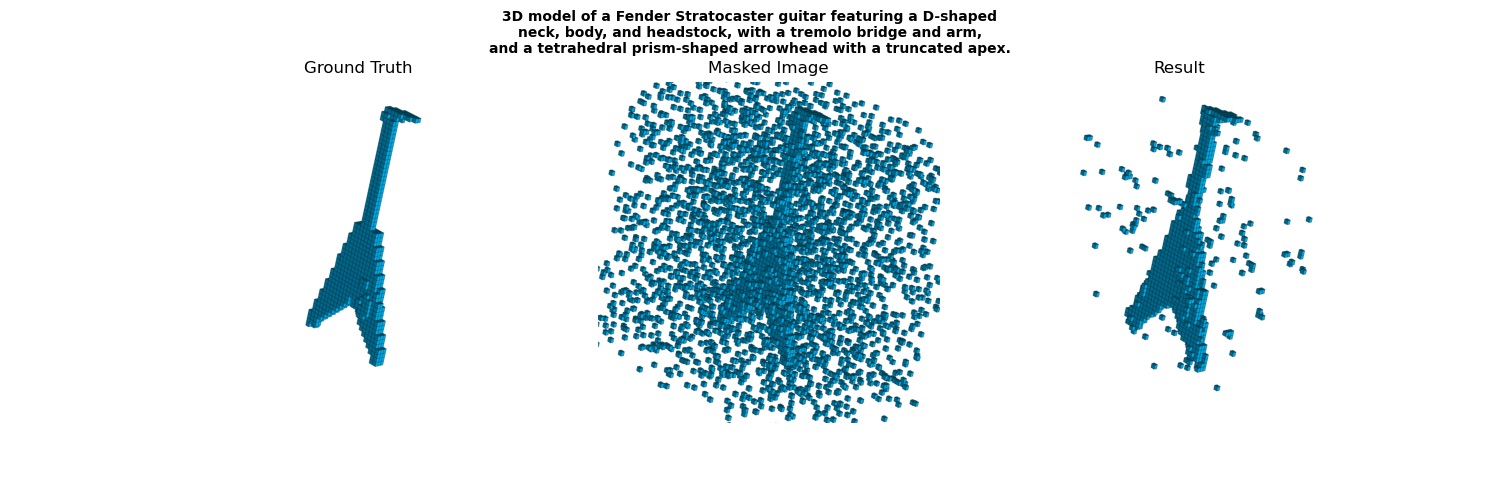}
    \leftfig{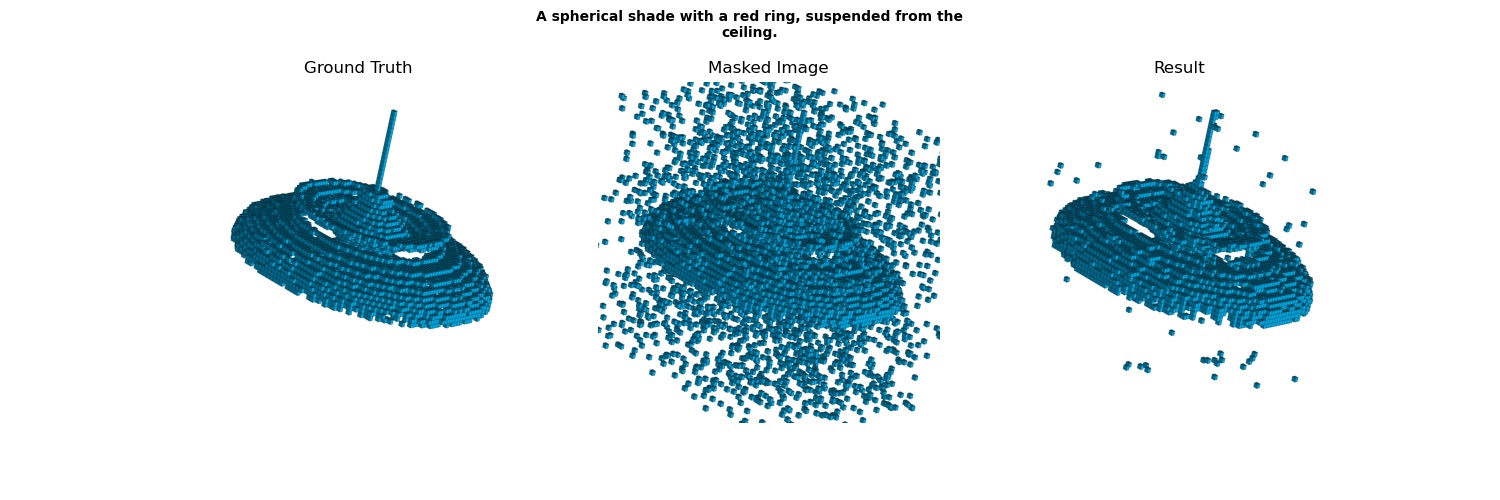}
    \lrightfig{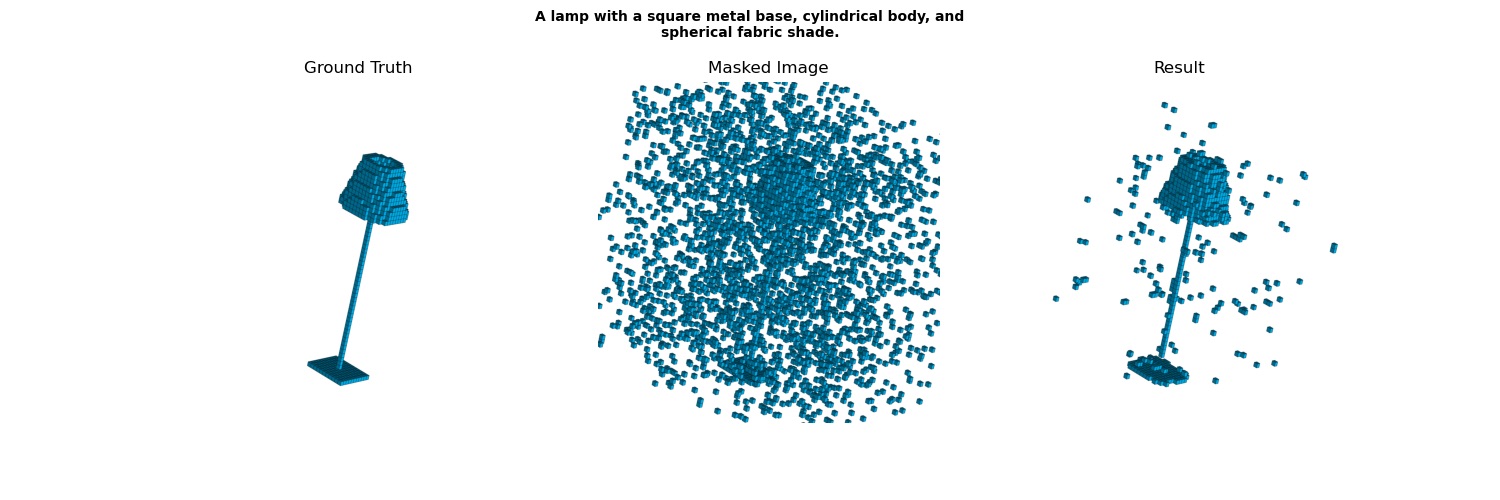}
    \leftfig{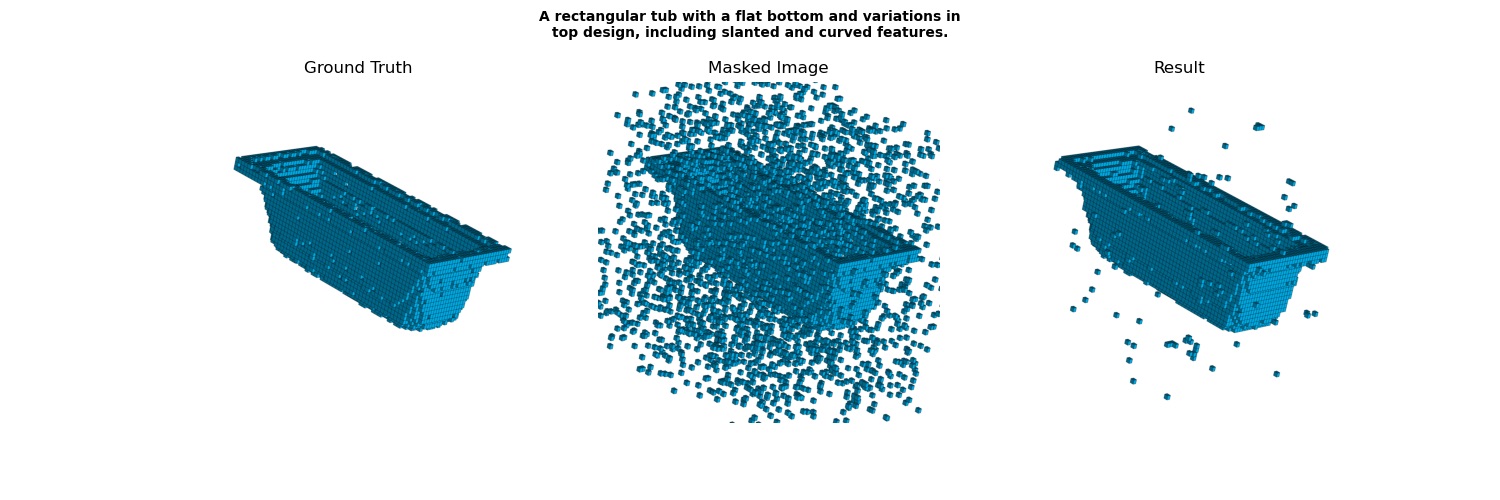}
    \lrightfig{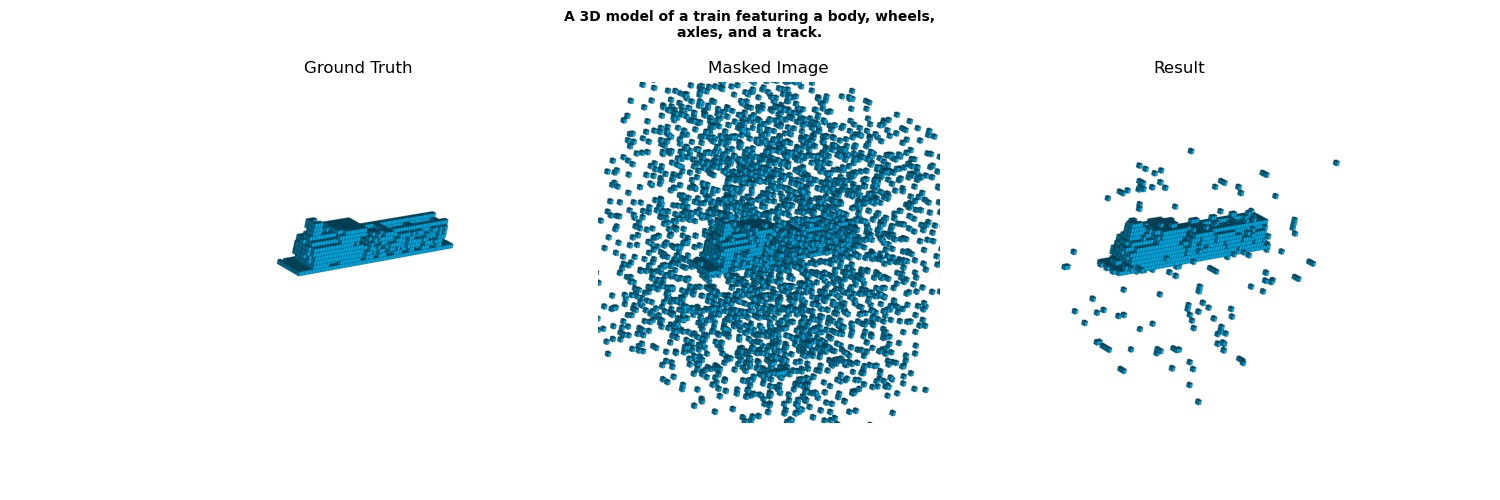}
    \leftfig{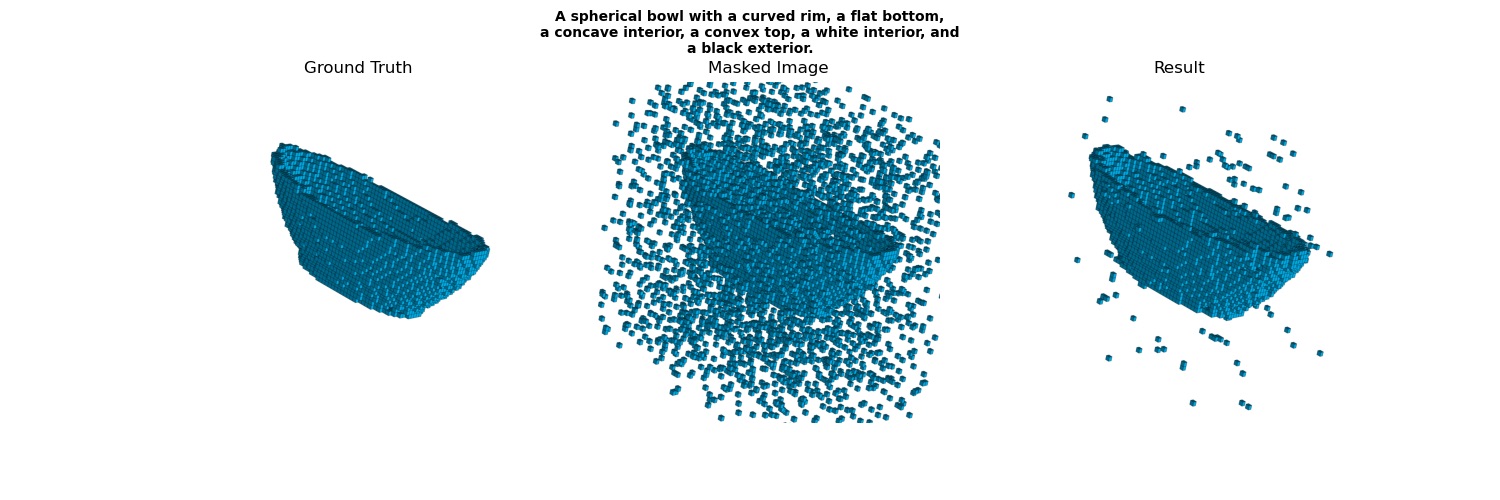}
    \lrightfig{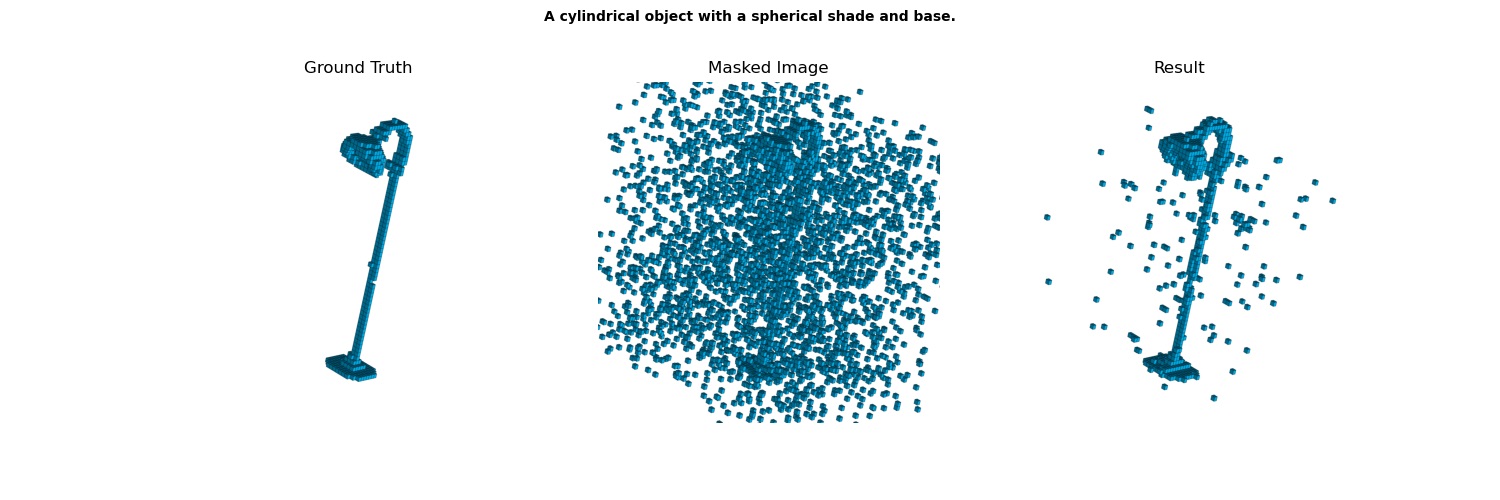}
    \leftfig{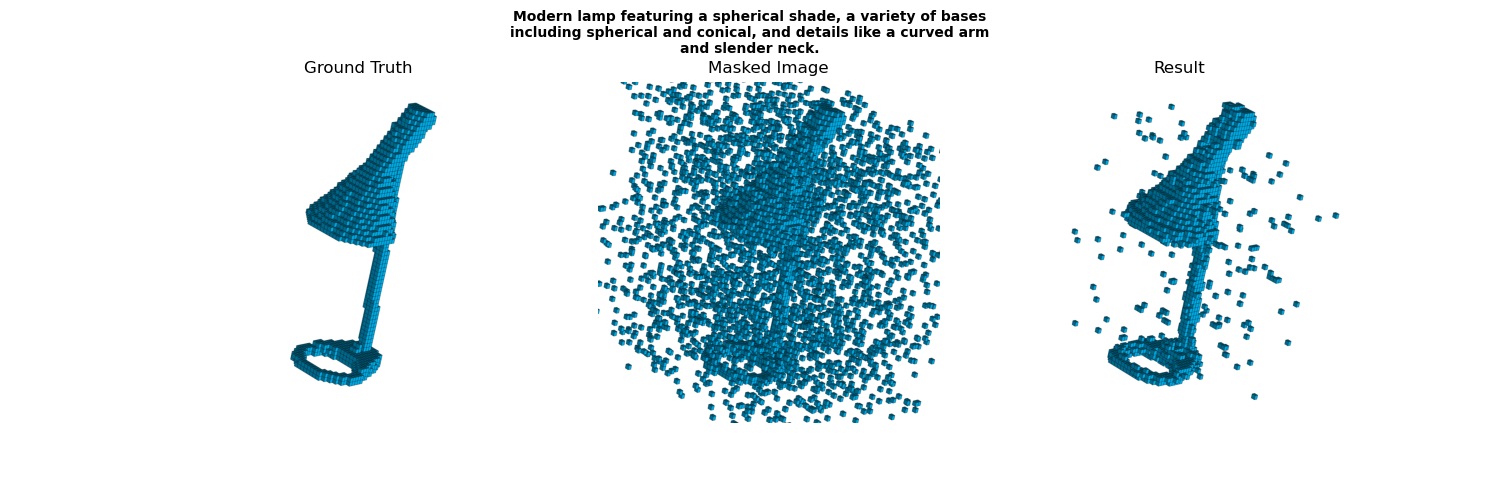}
    \lrightfig{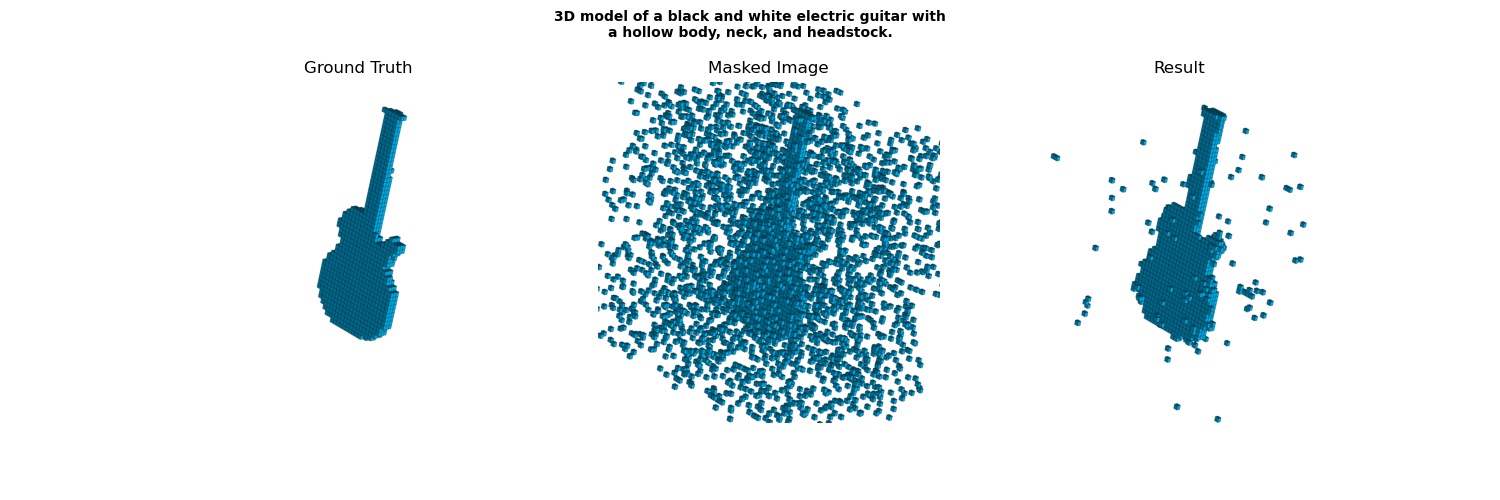}
    \leftfig{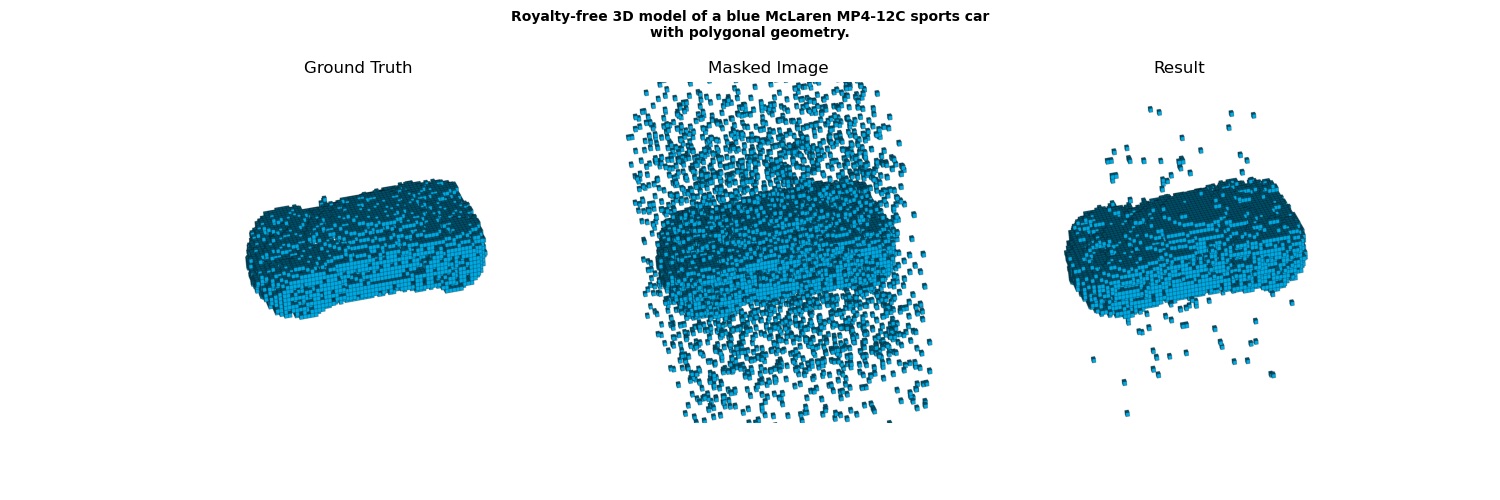}
    \lrightfig{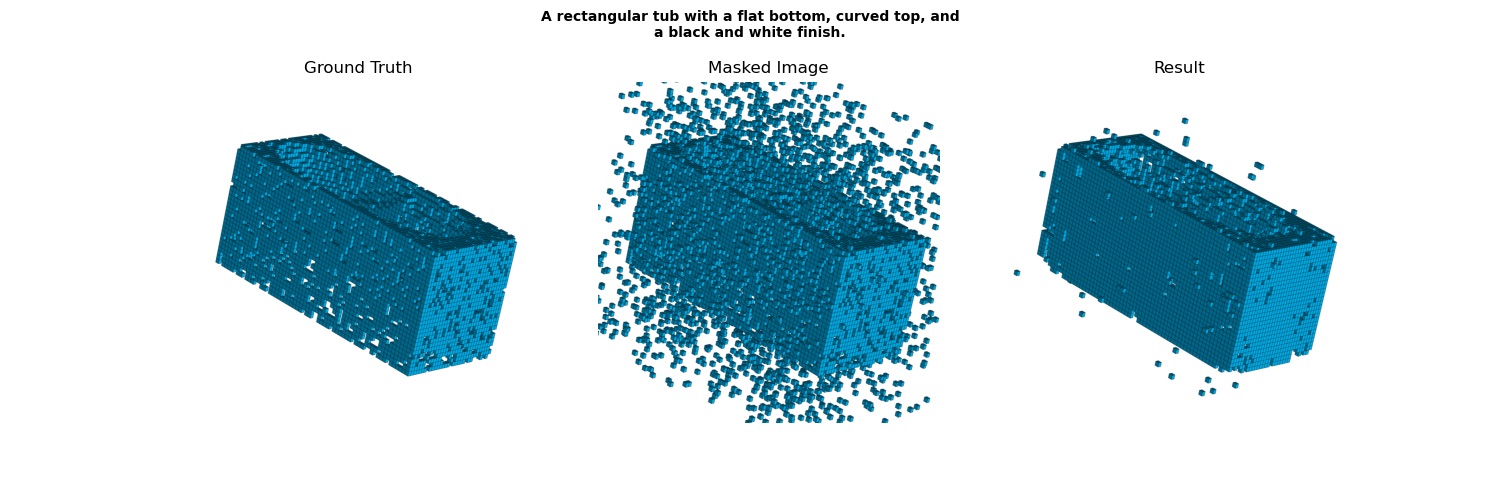}
    \leftfig{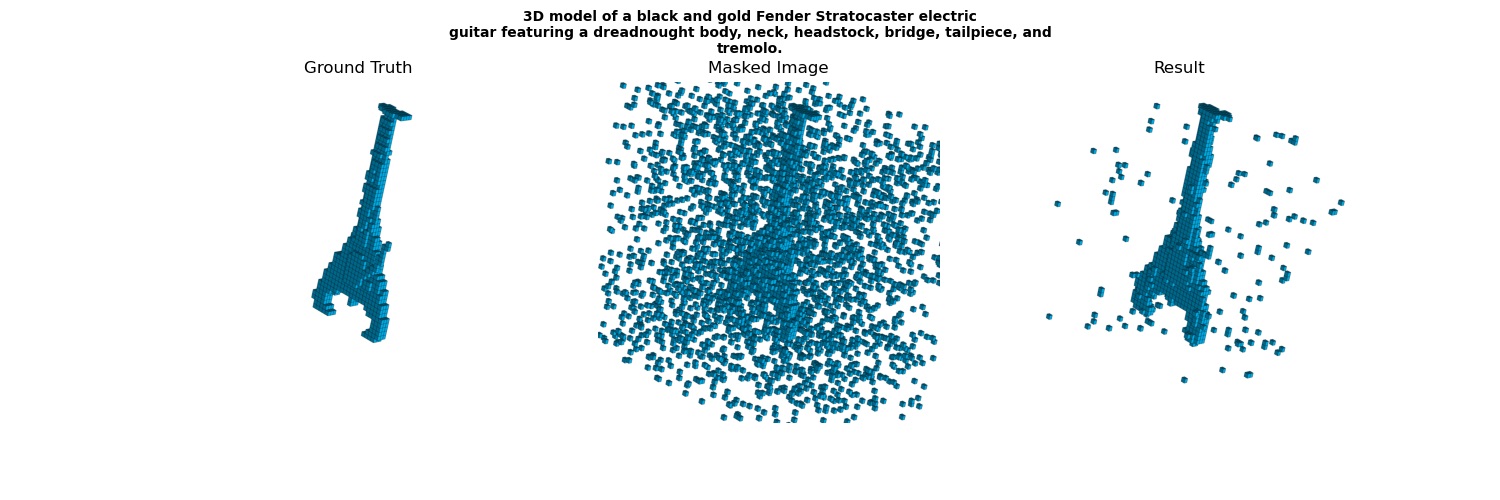}
    \lrightfig{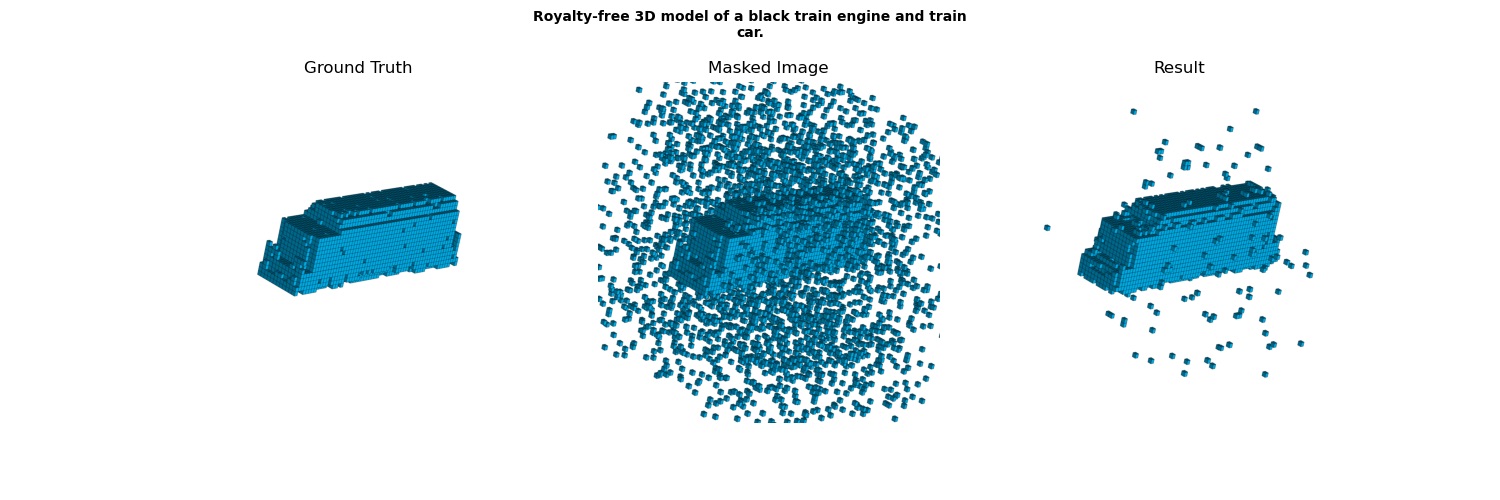}
    \leftfig{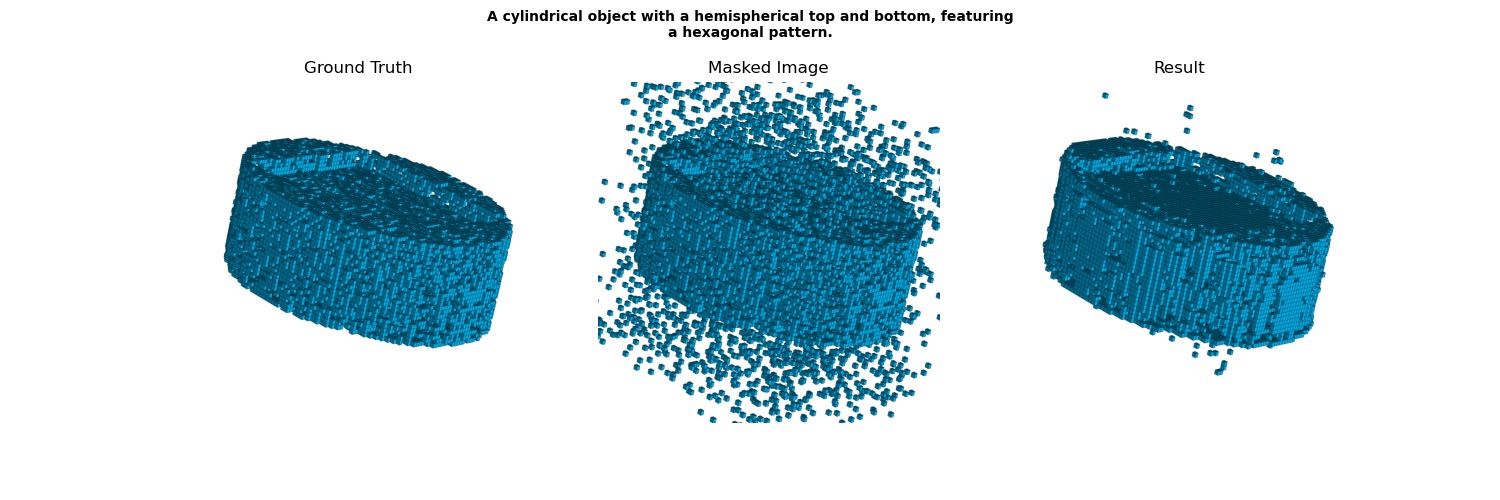}
    \lrightfig{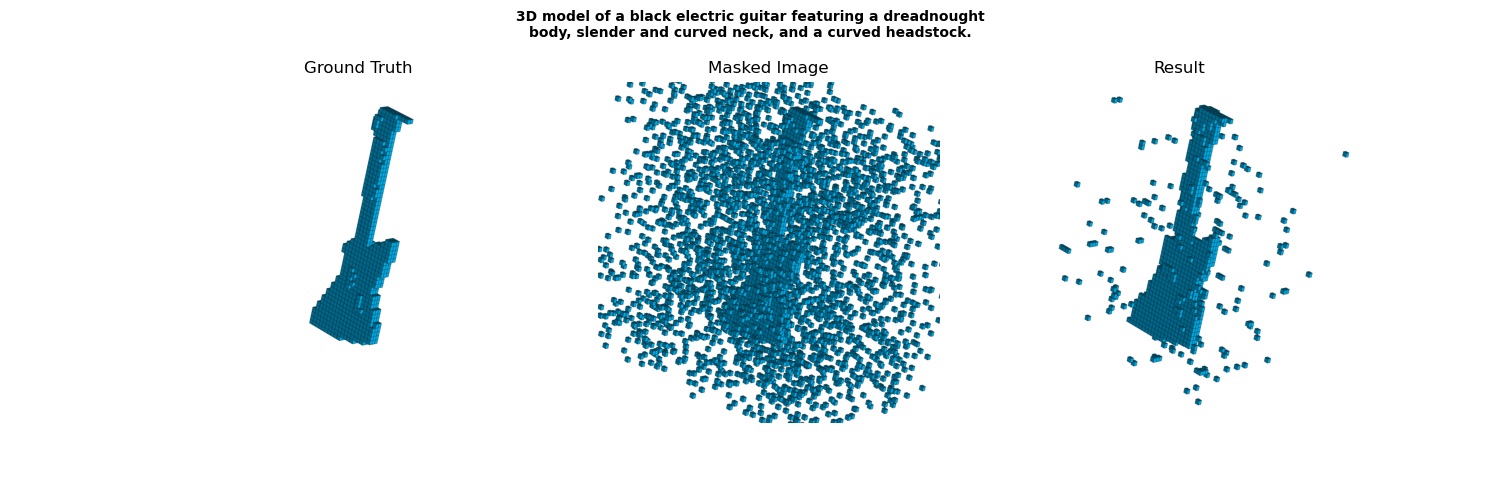}
    \caption{Results Noise$1\%$}
    \label{fig:Noise001-2}
\end{figure}

\begin{figure}[H] 
 \centering 
\leftfig{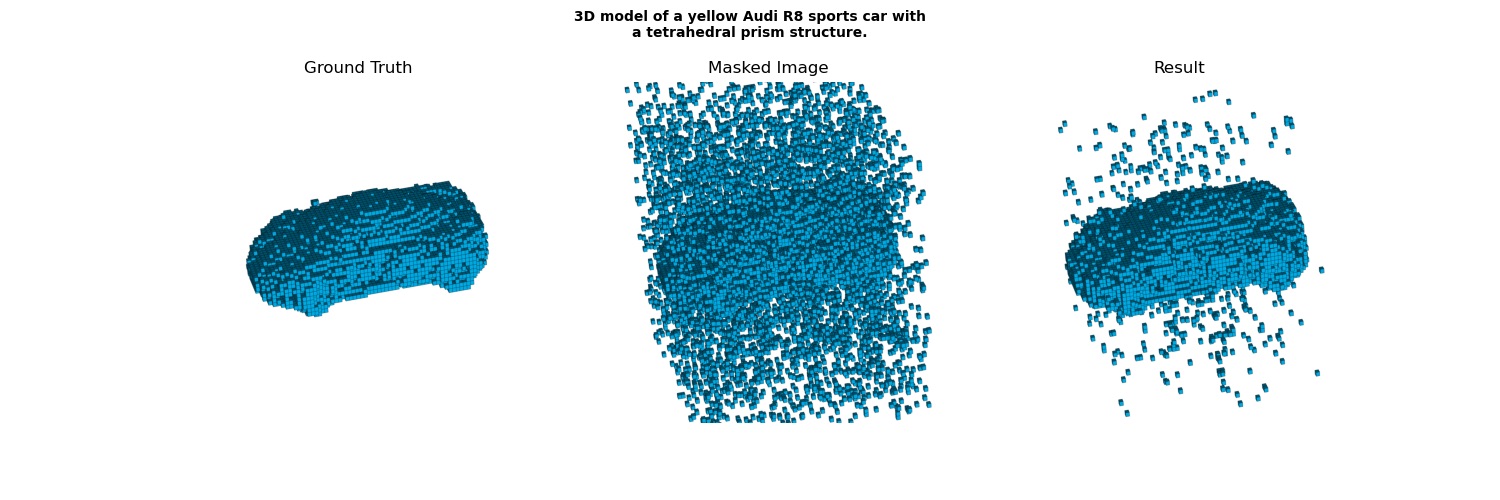}
\lrightfig{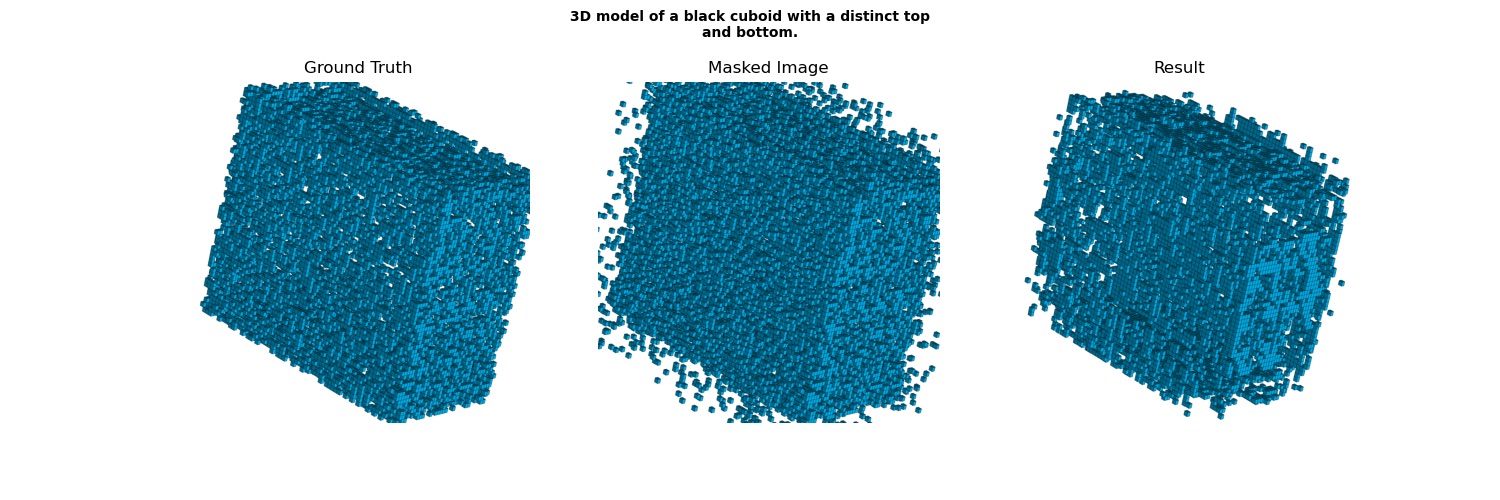}
\leftfig{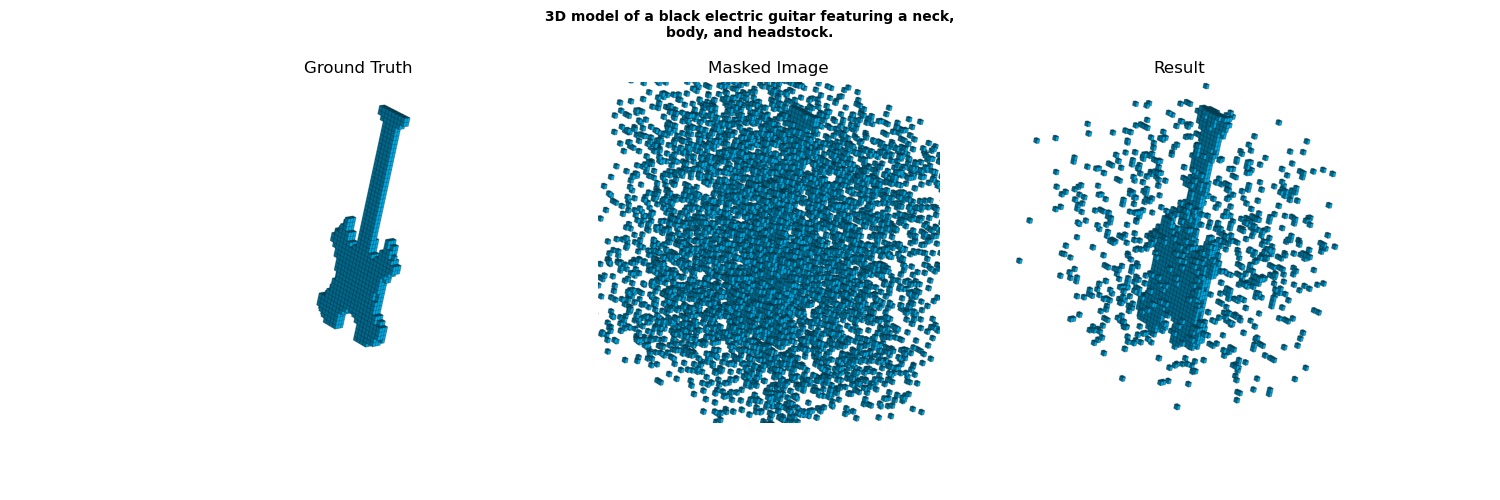}
\lrightfig{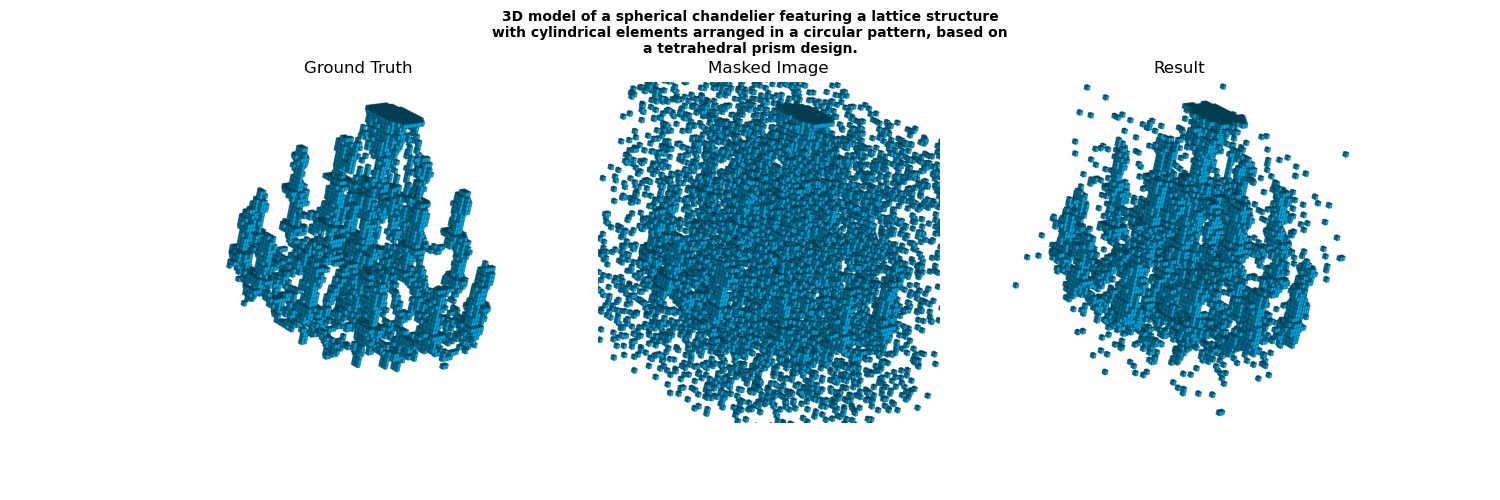}
\leftfig{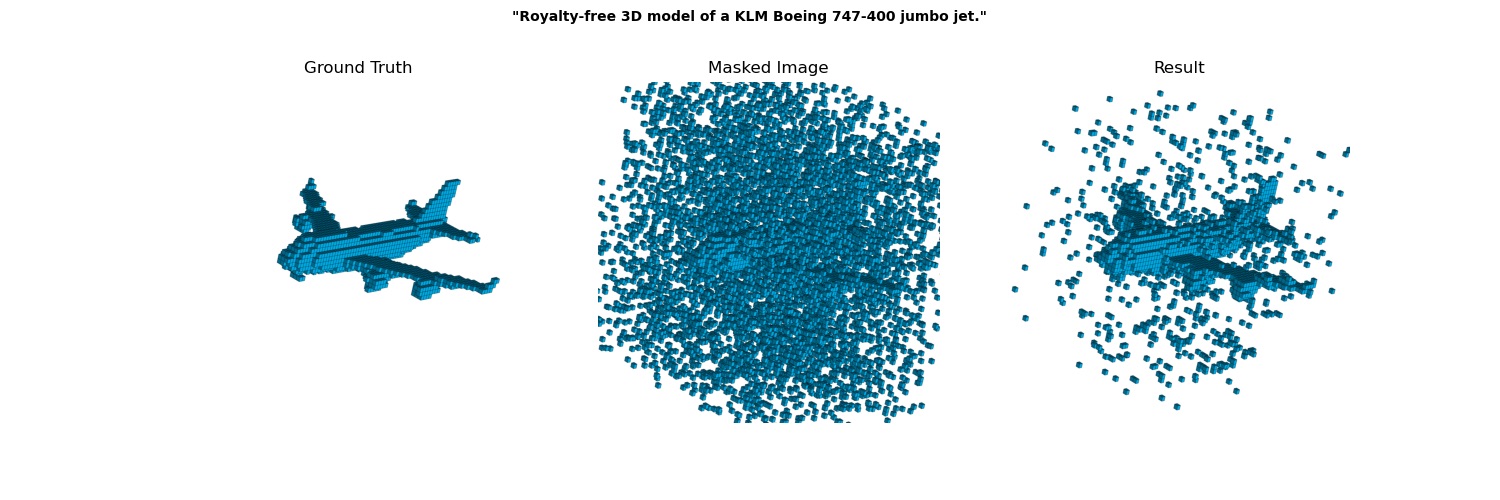}
\lrightfig{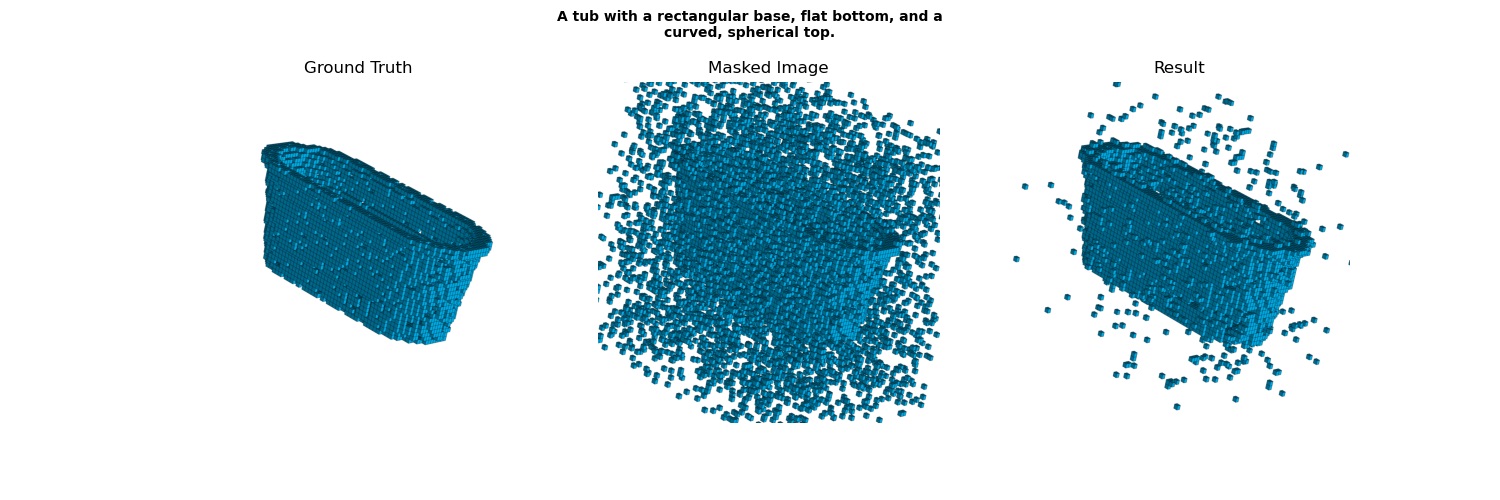}
\leftfig{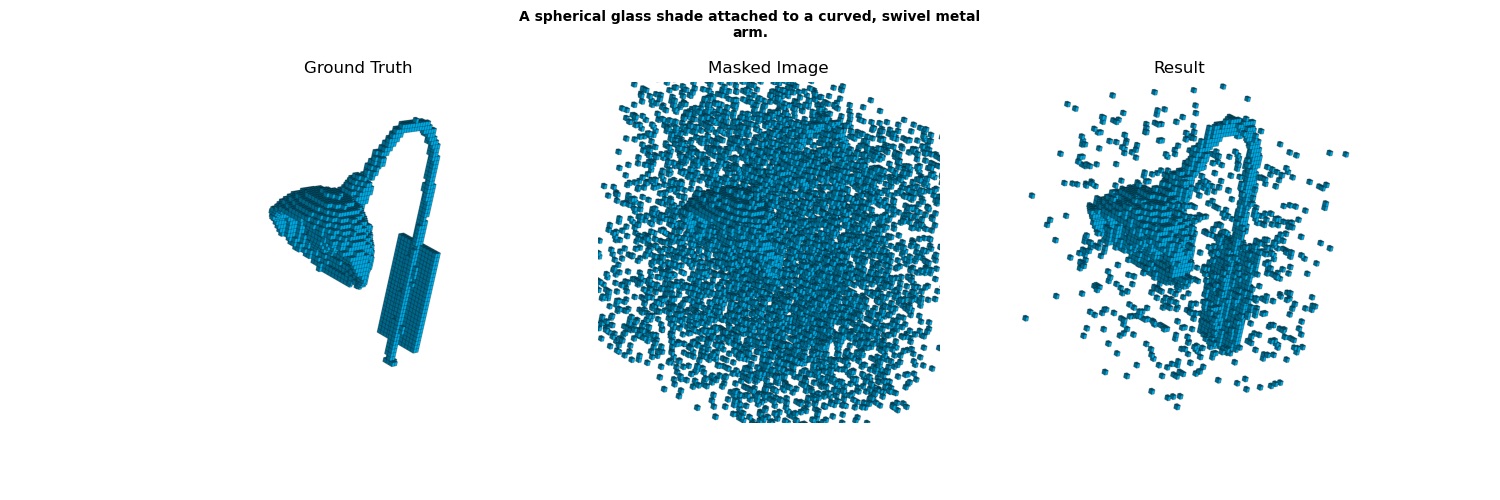}
\lrightfig{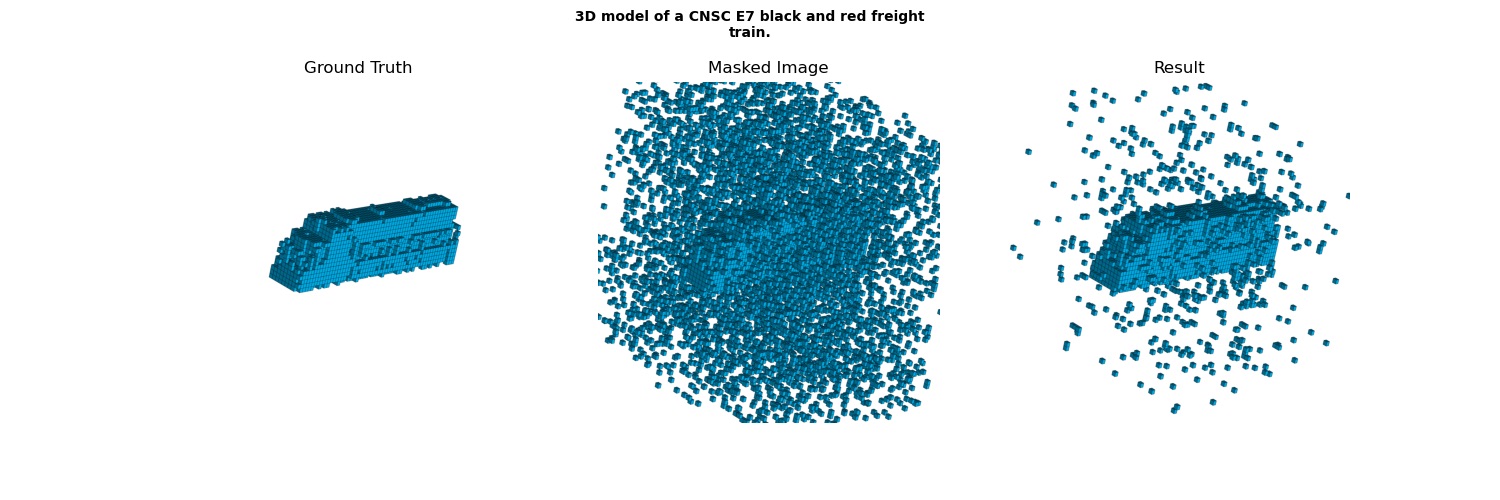}
\leftfig{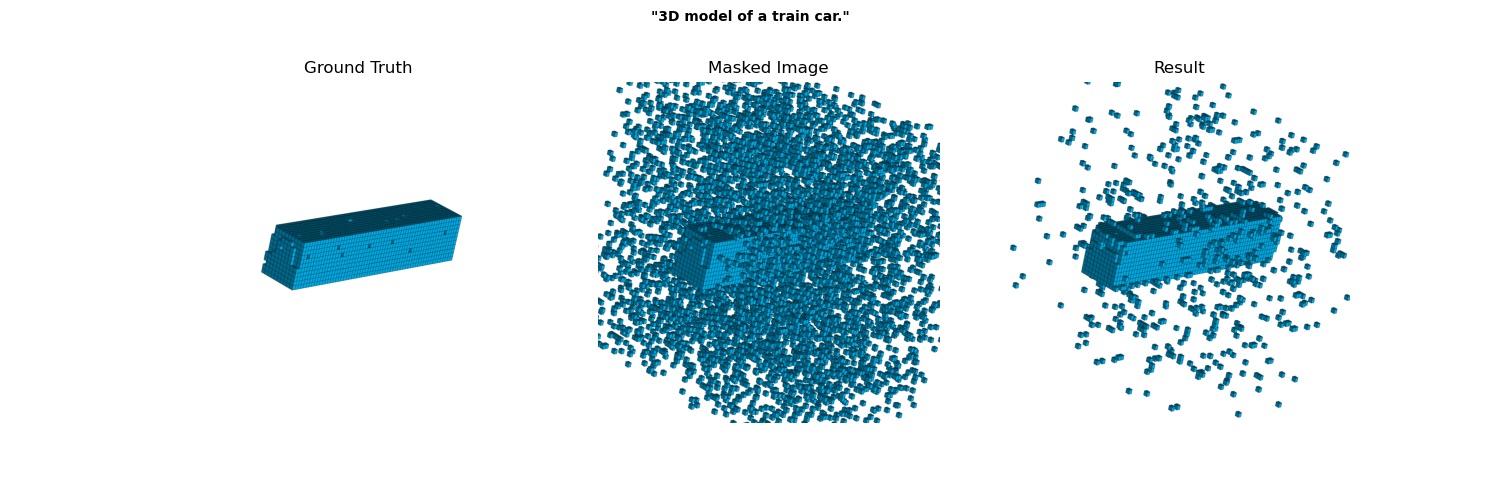}
\lrightfig{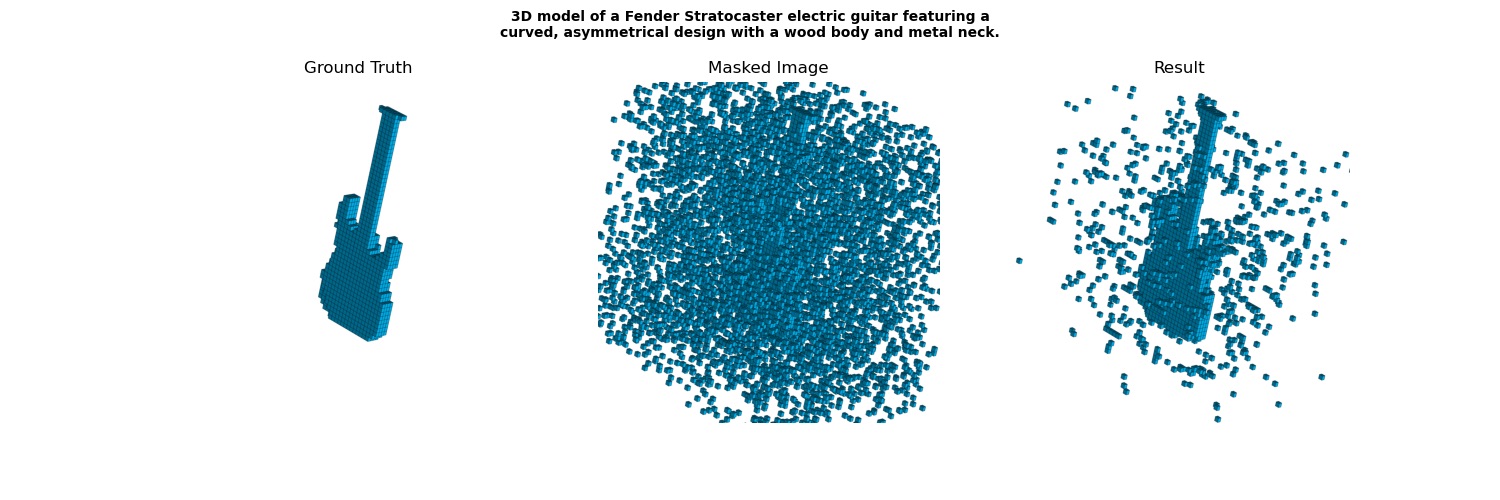}
\leftfig{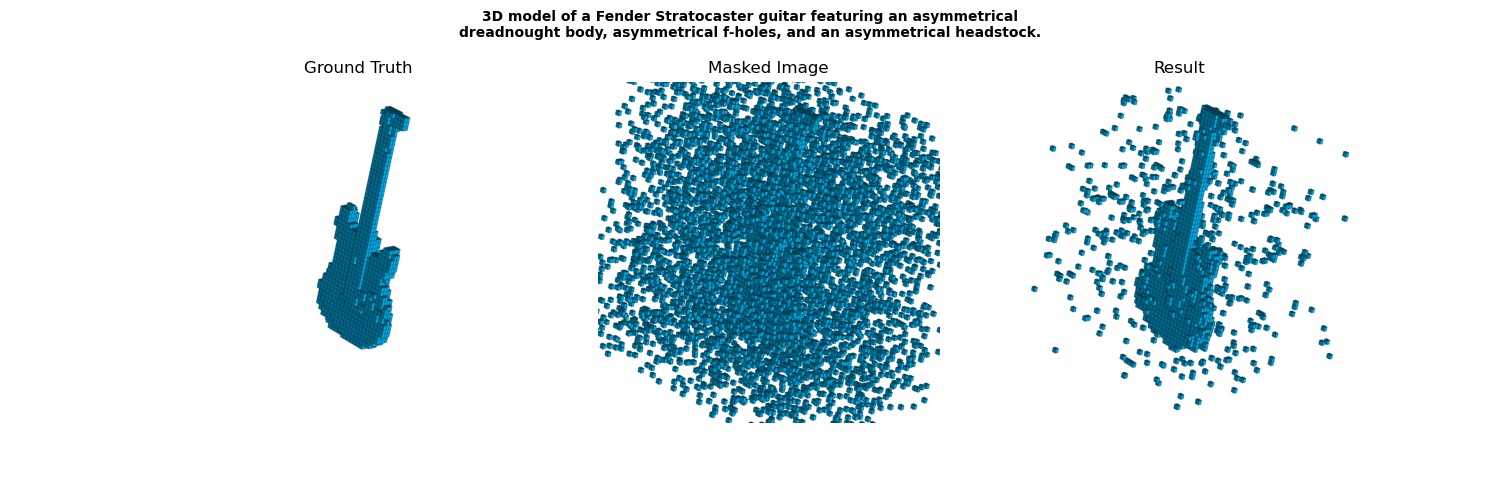}
\lrightfig{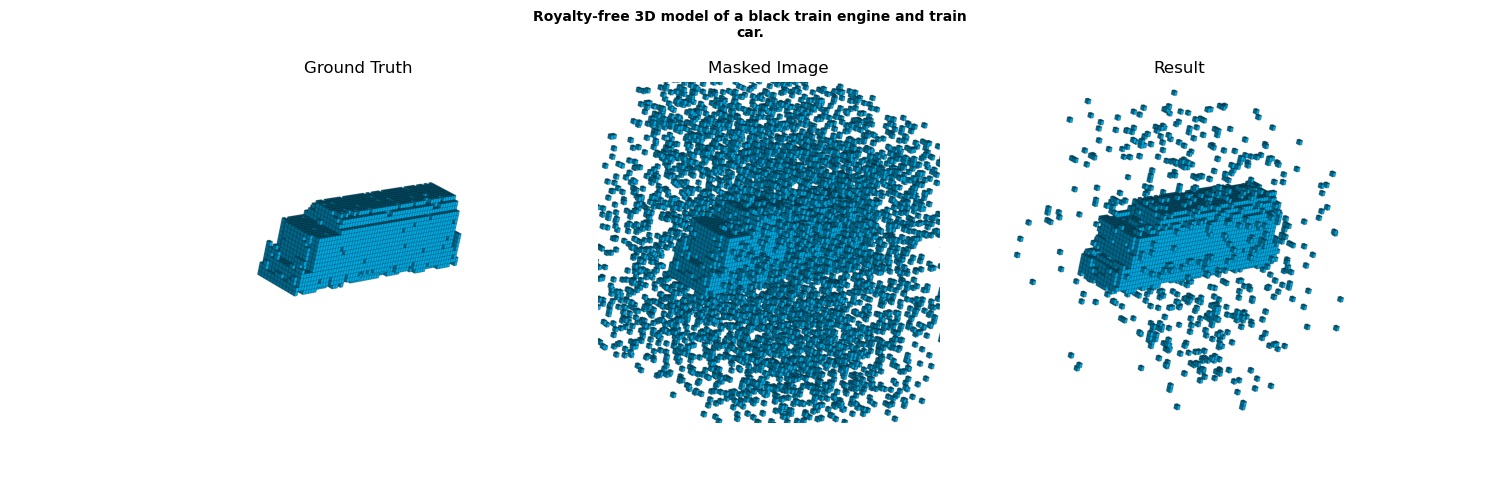}
\leftfig{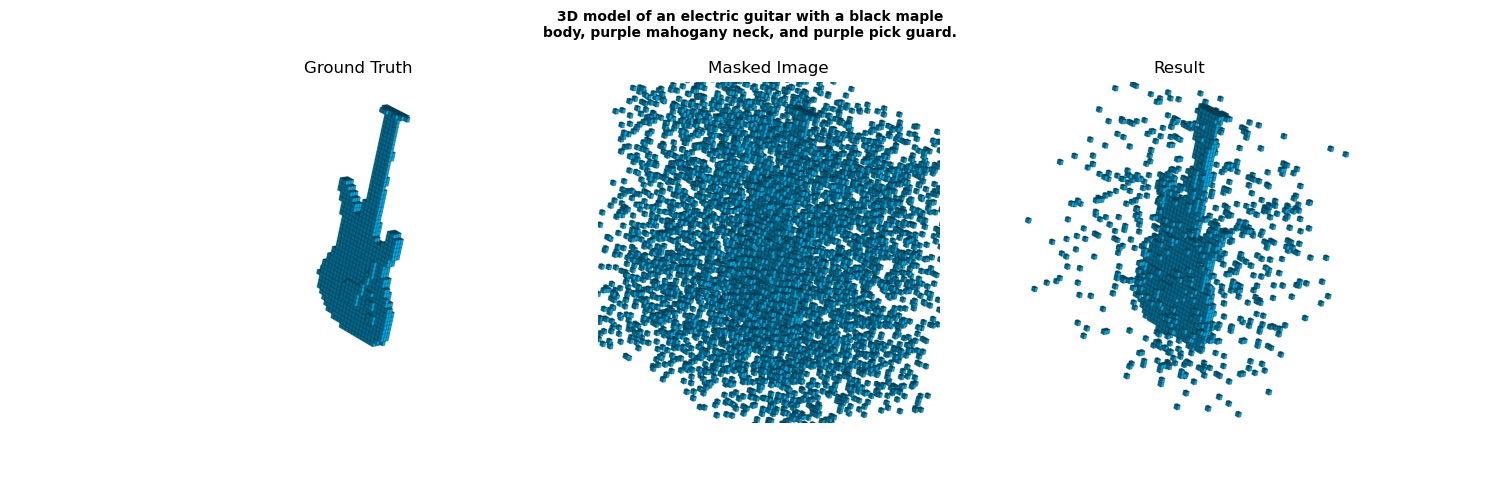}
\lrightfig{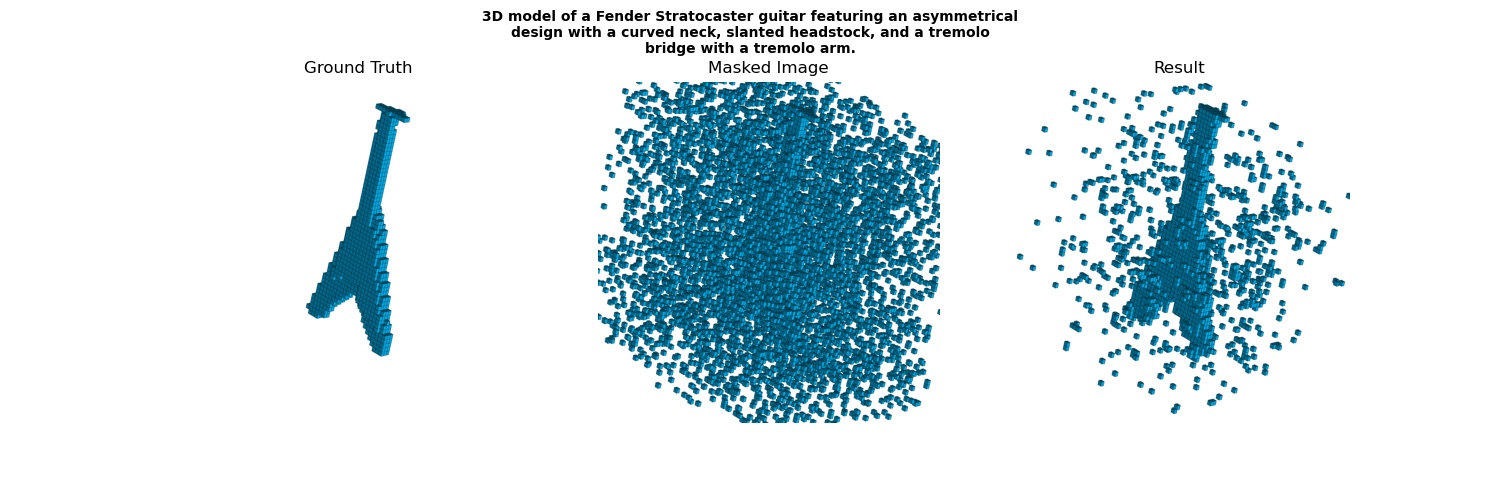}
\leftfig{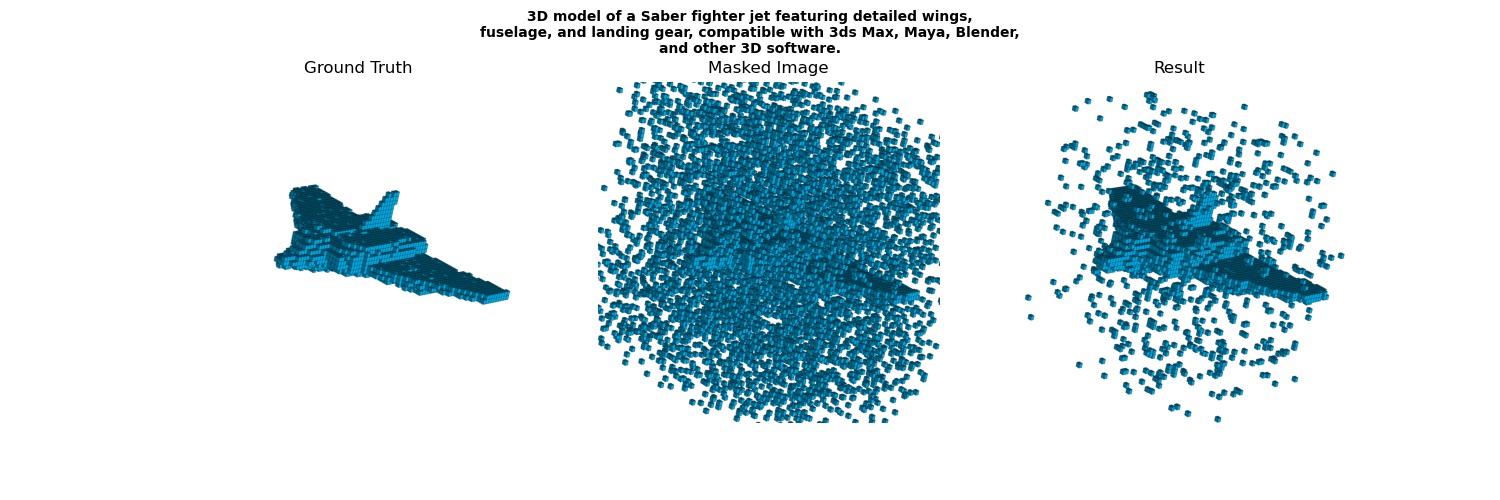}
\lrightfig{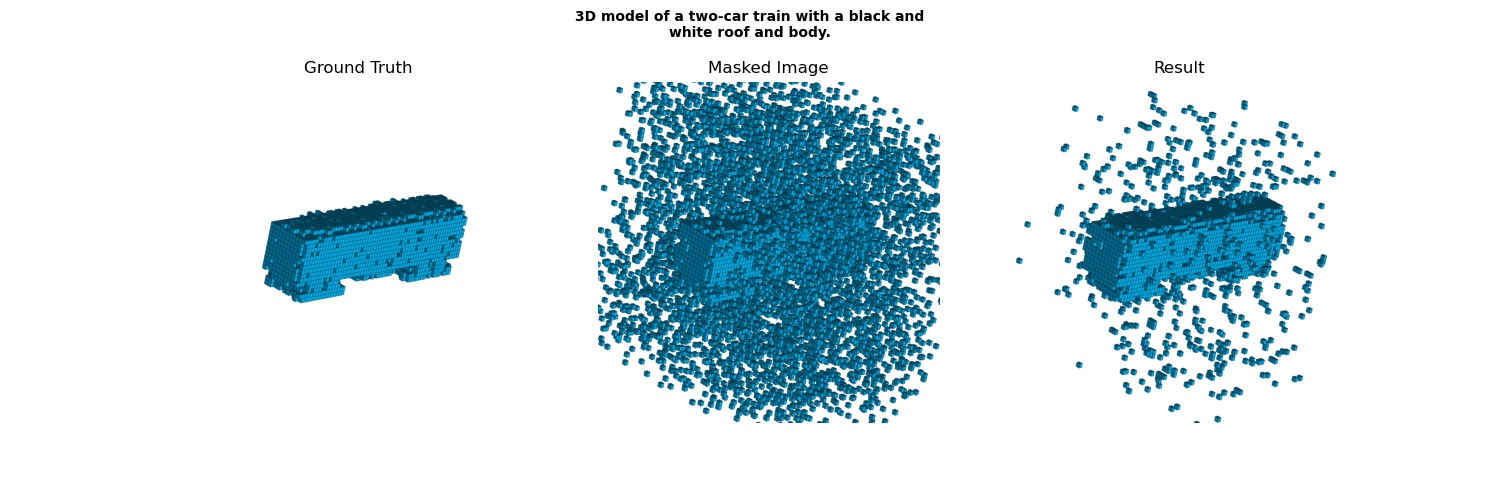}
\caption{Results Noise$2\%$}
\label{fig:Noise002-1}
\end{figure}

\begin{figure}[H] 
 \centering 
\leftfig{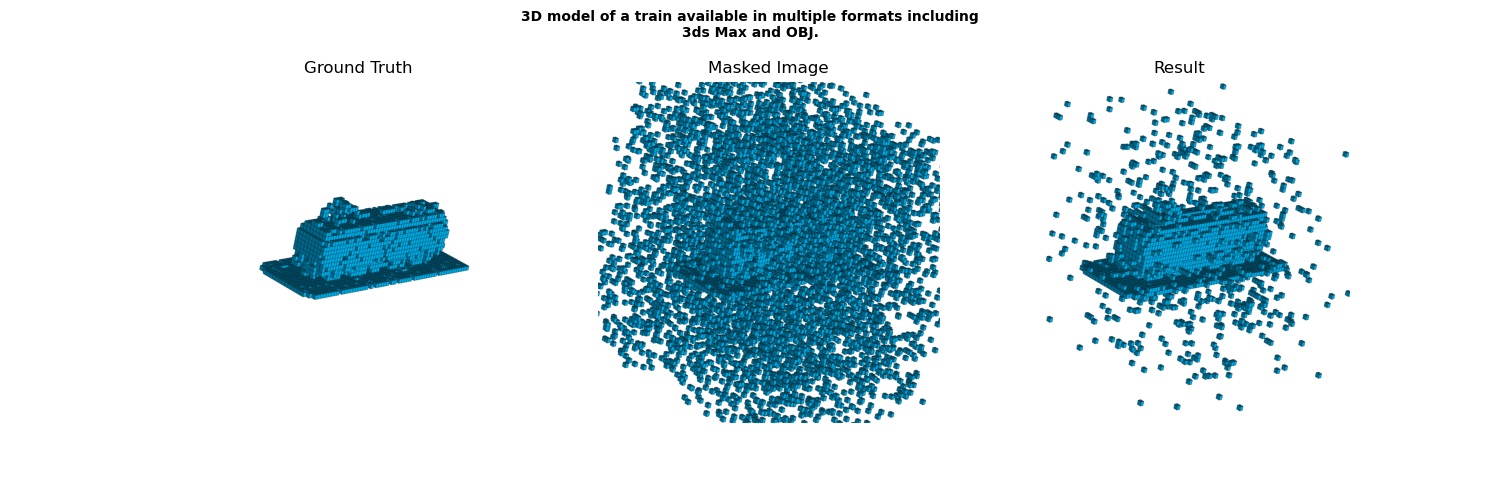}
\lrightfig{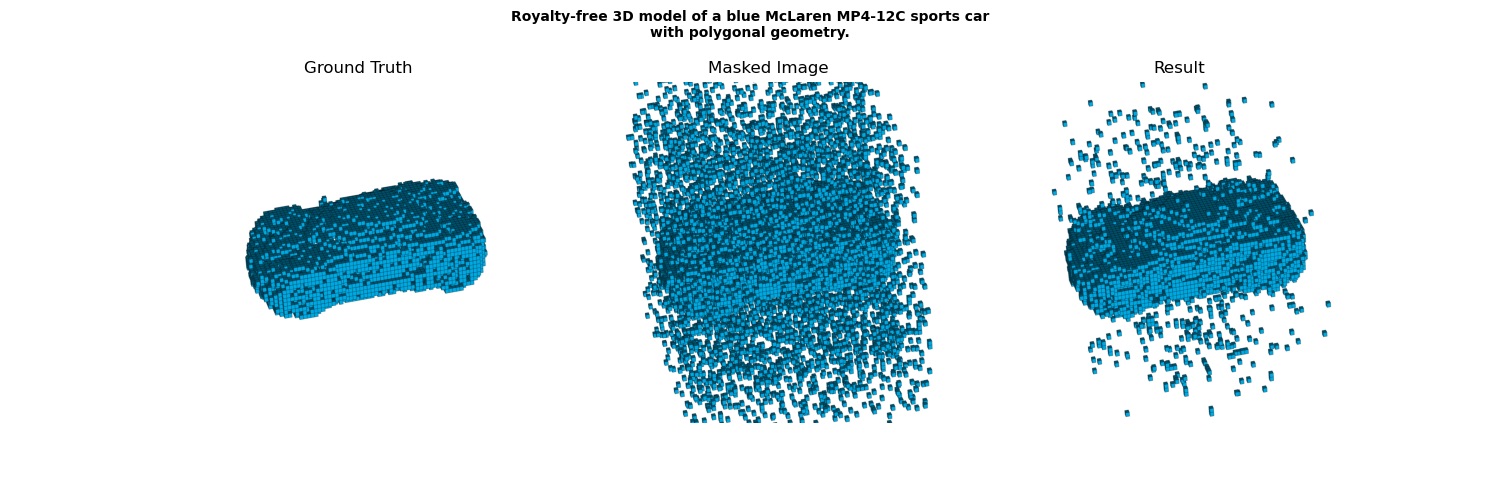}
\leftfig{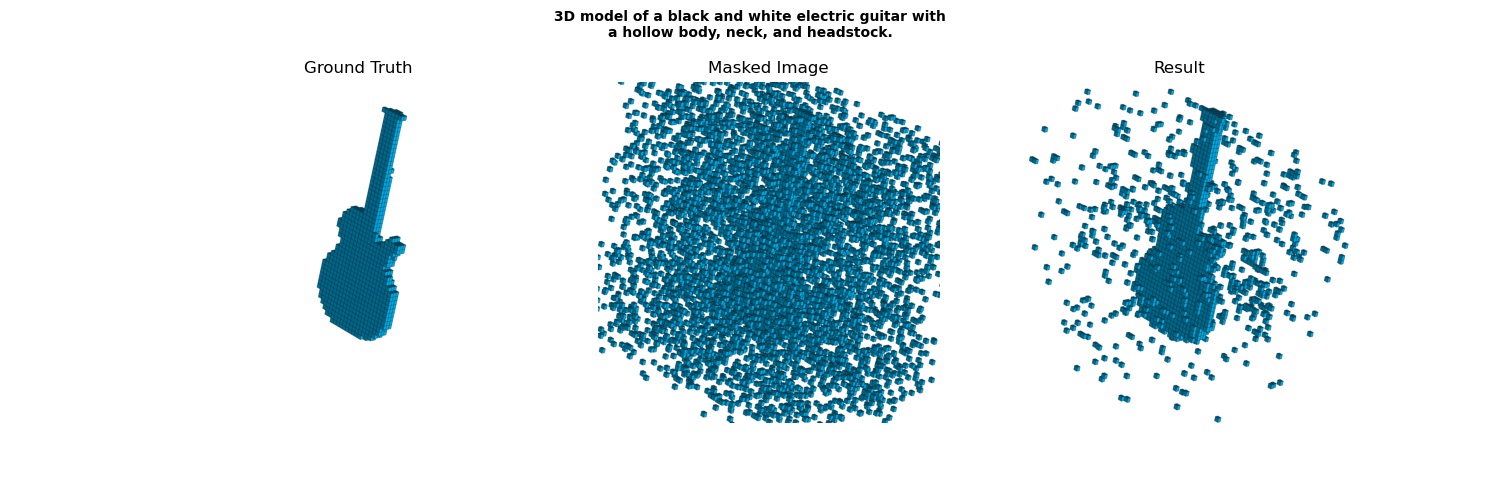}
\lrightfig{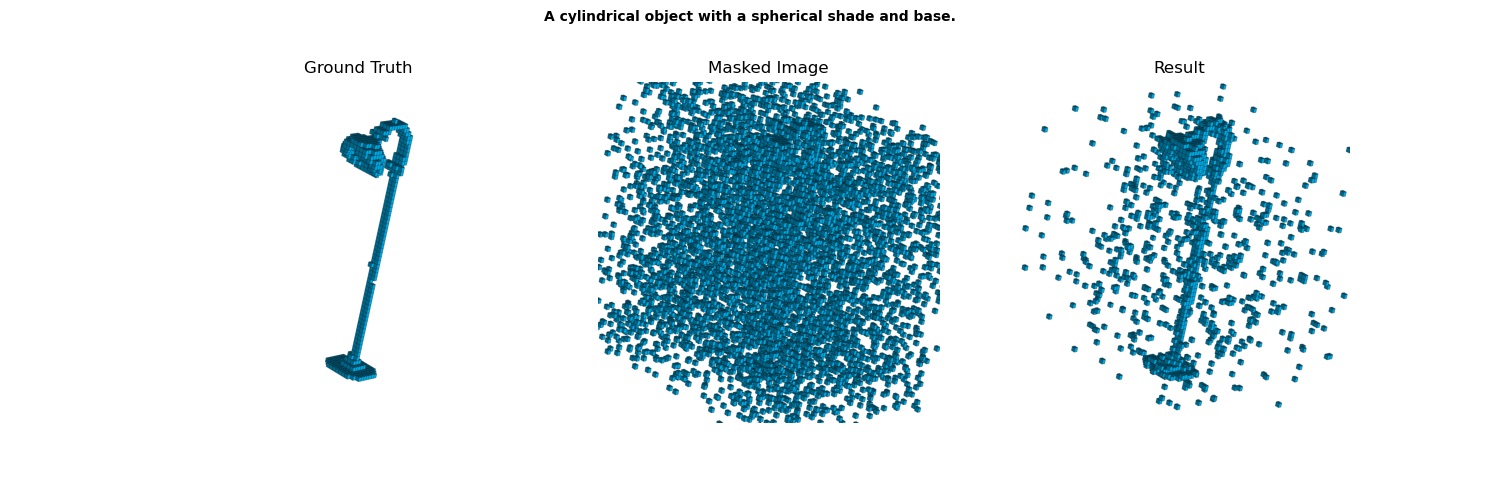}
\leftfig{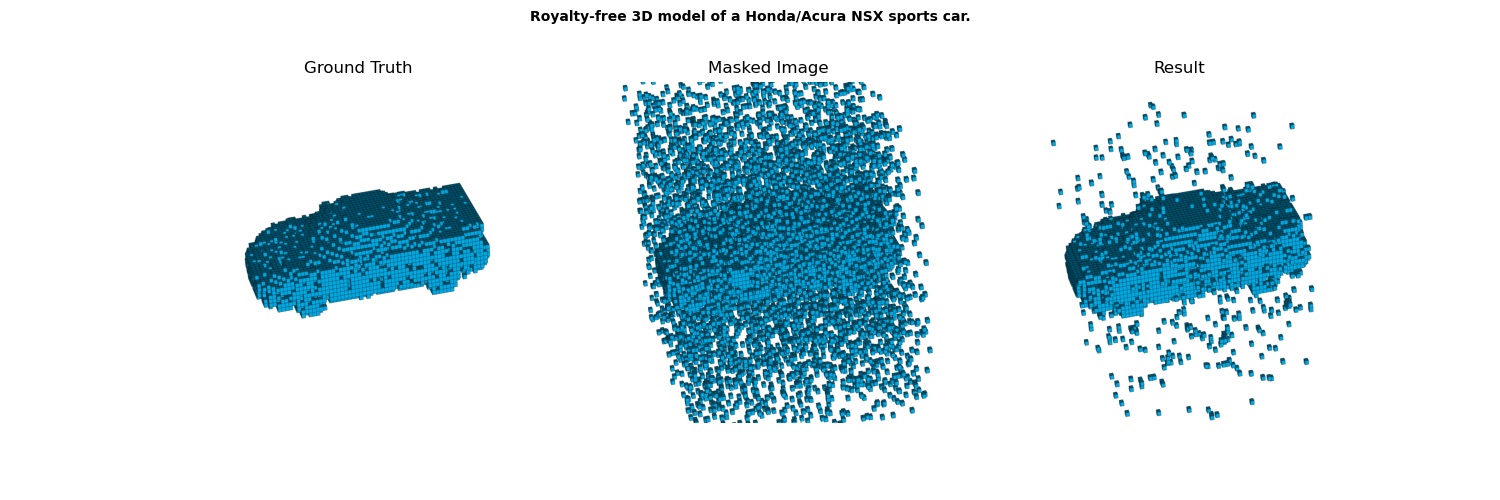}
\lrightfig{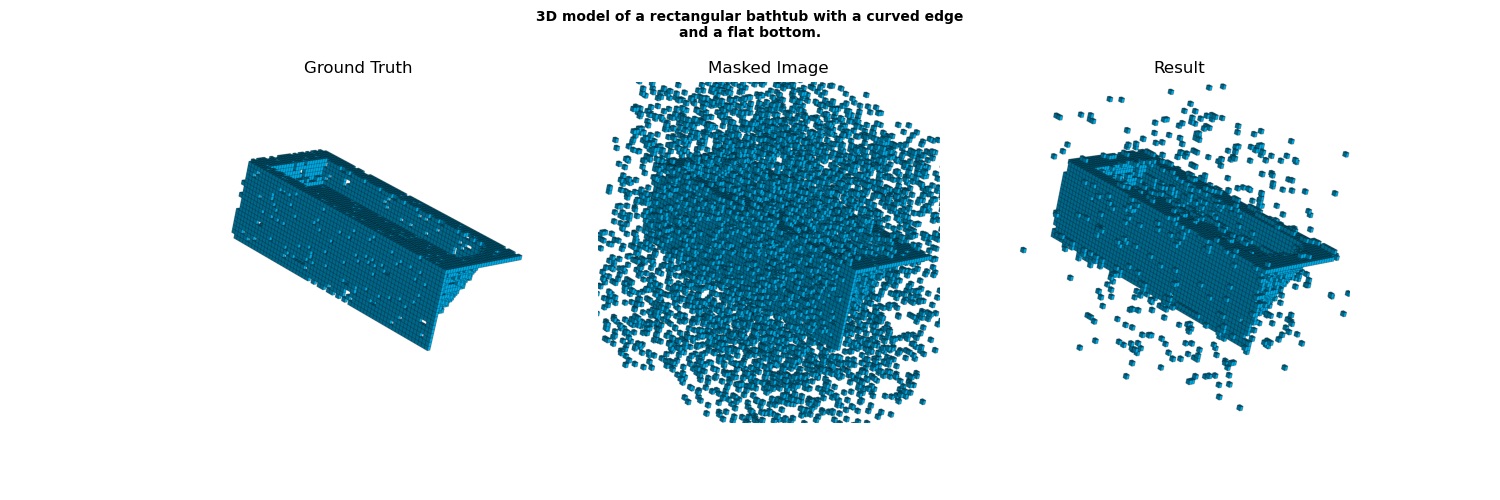}
\leftfig{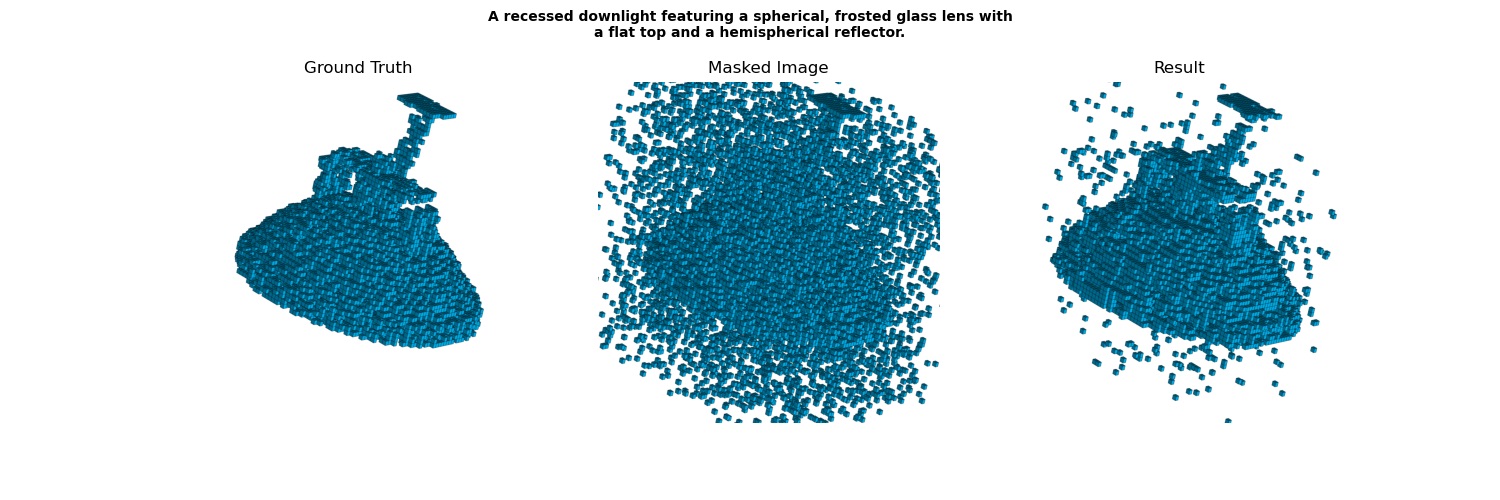}
\lrightfig{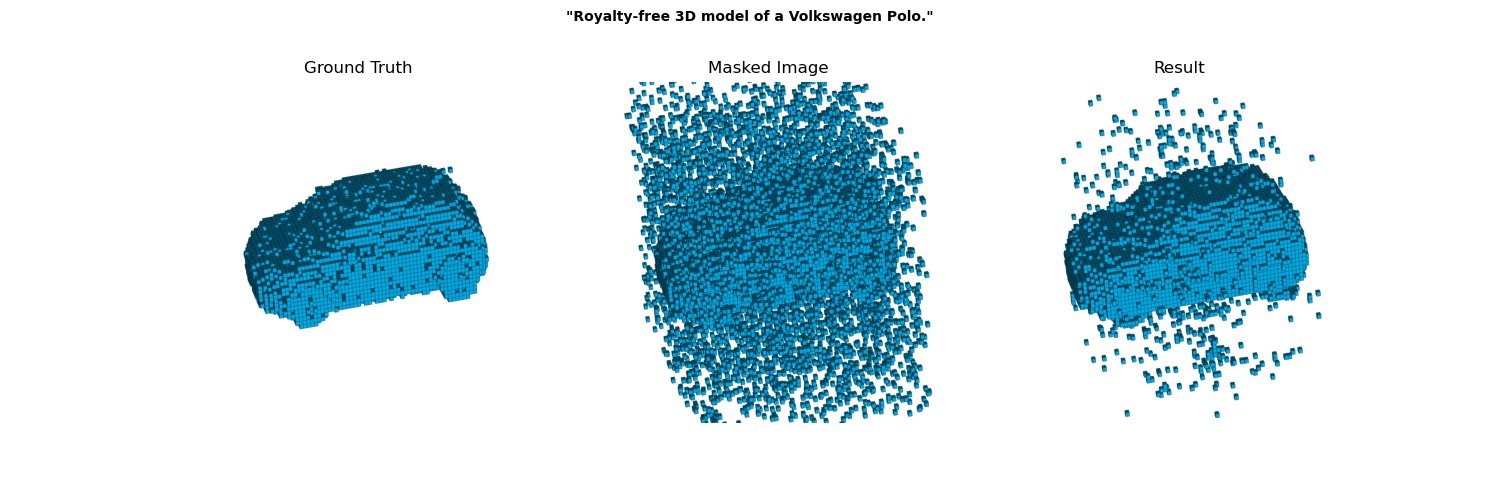}
\leftfig{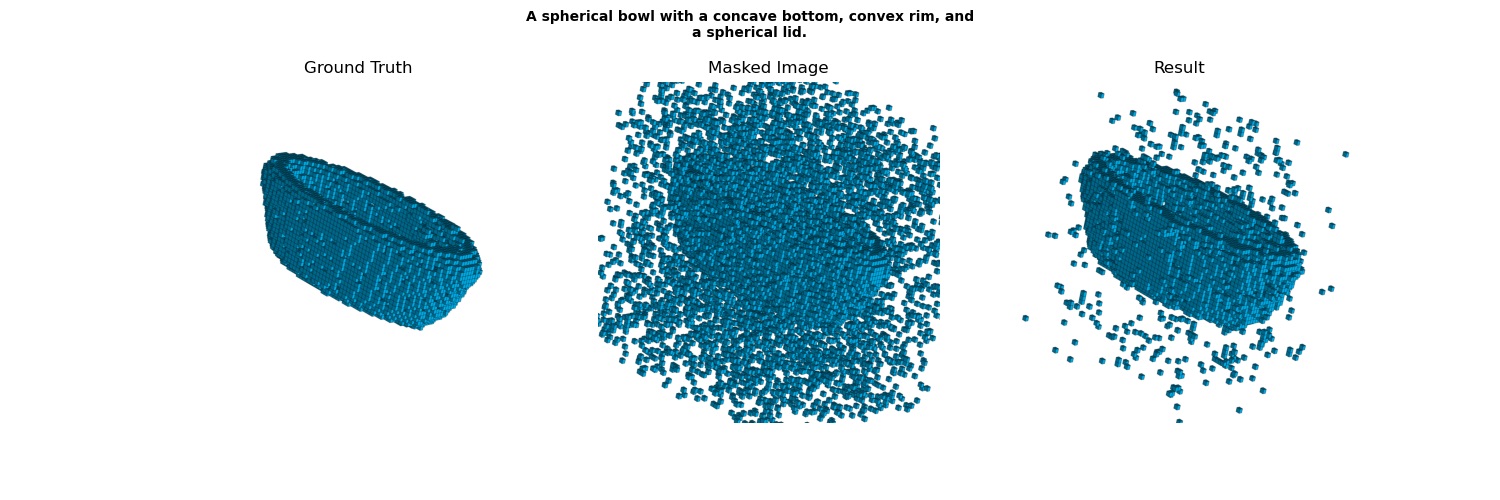}
\lrightfig{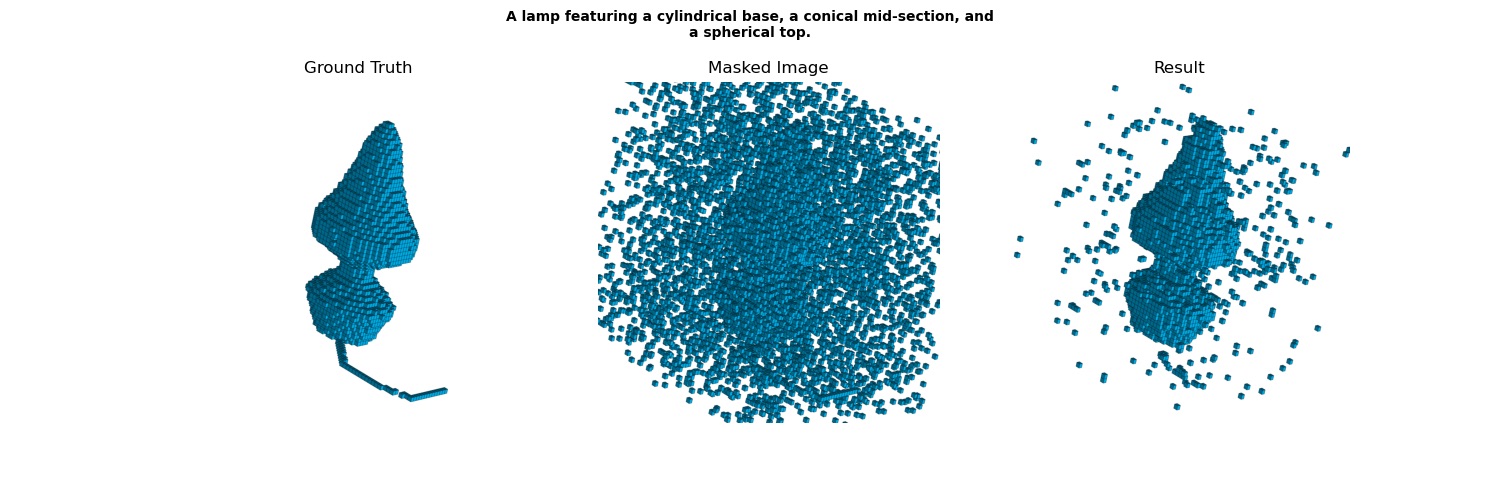}
\leftfig{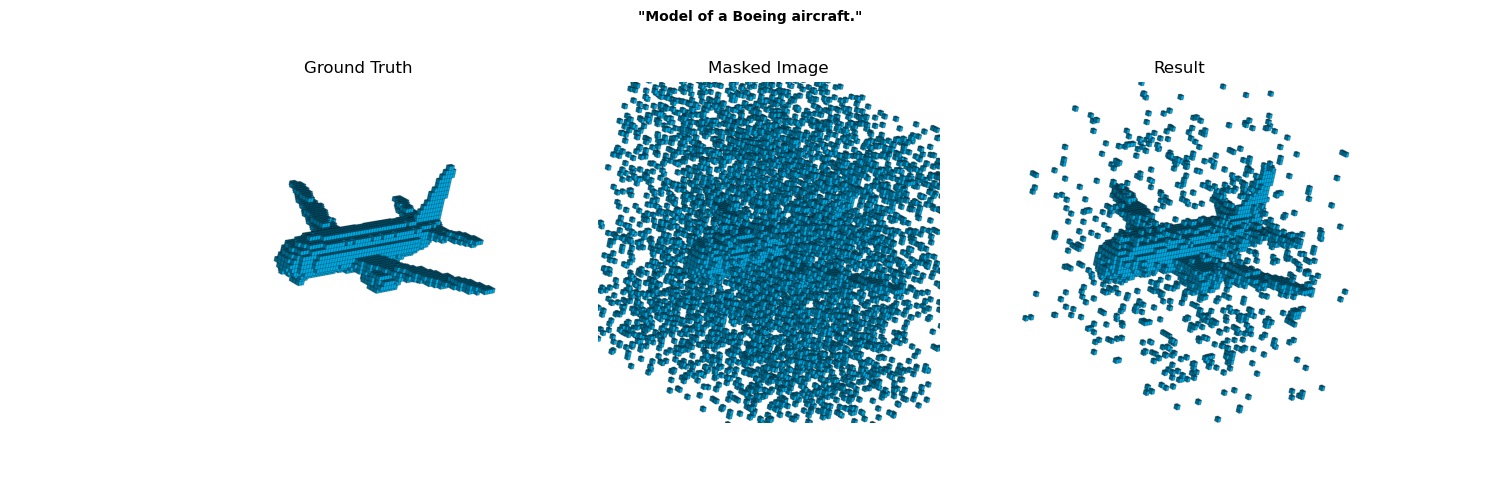}
\lrightfig{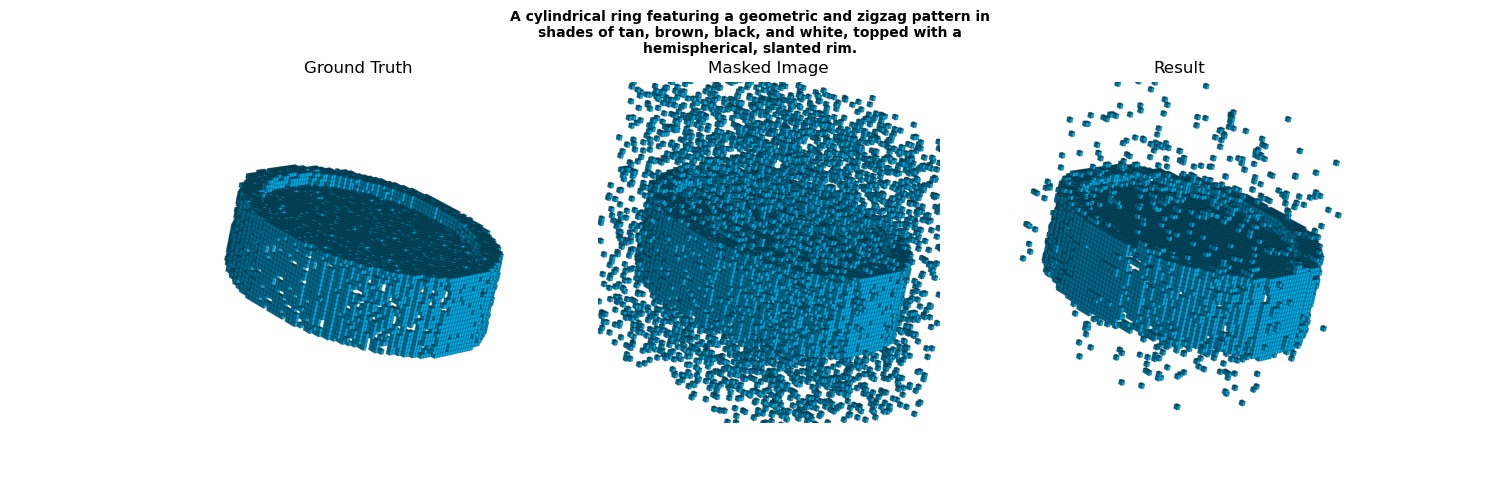}
\leftfig{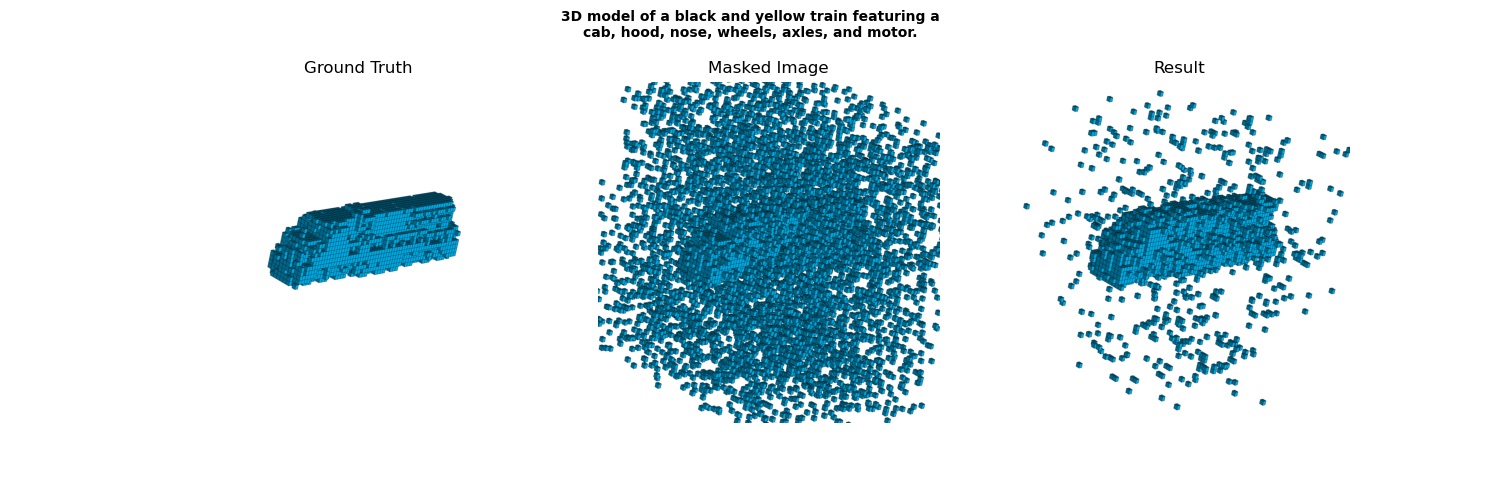}
\lrightfig{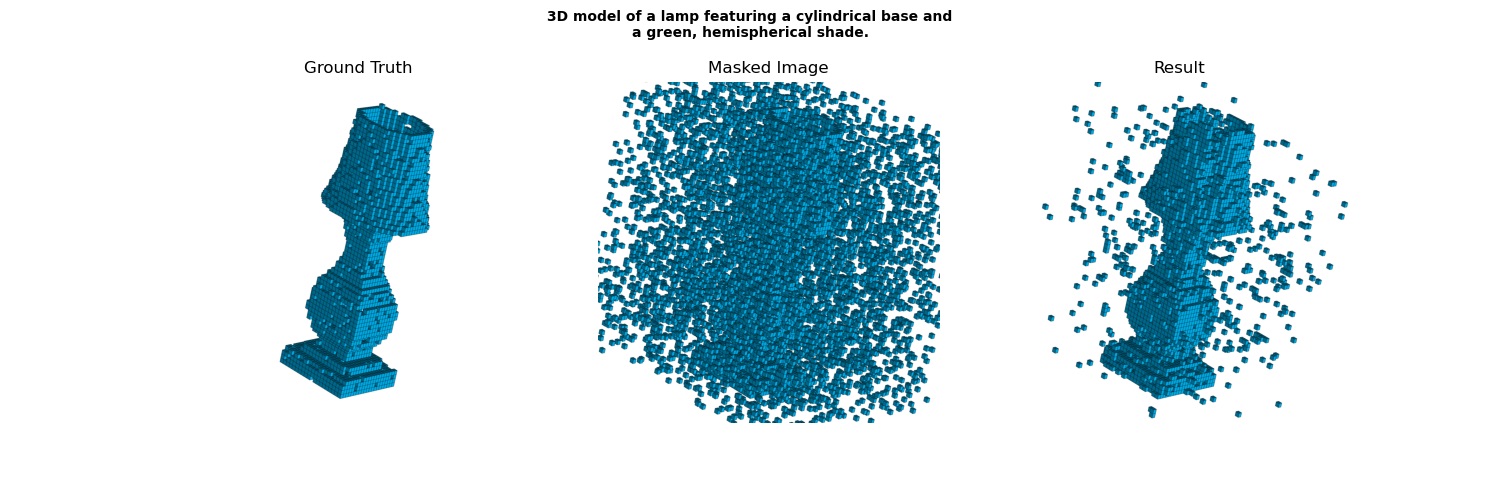}
\leftfig{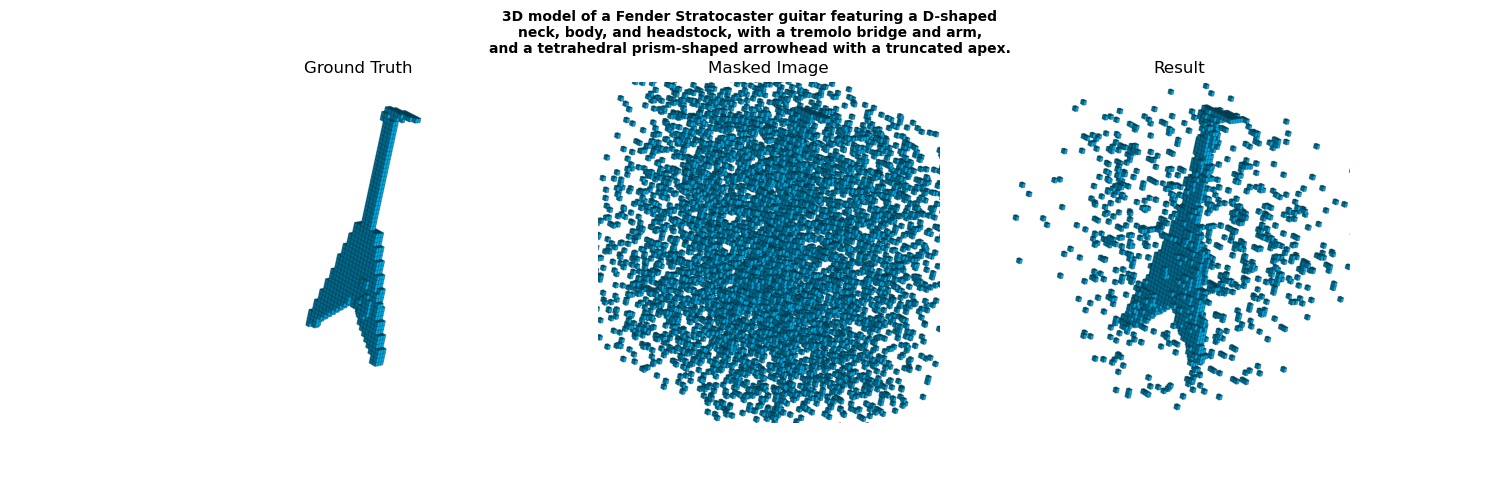}
\lrightfig{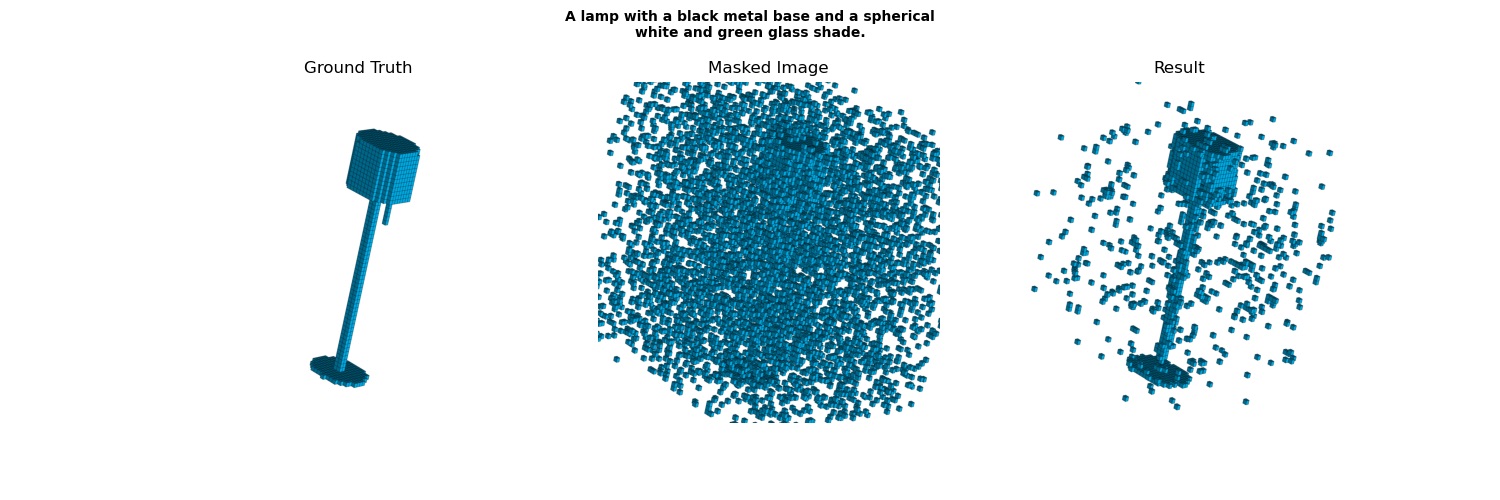}
\caption{Results Noise$2\%$}
\label{fig:Noise002-2}
\end{figure}

Figure~\ref{fig:plane02-1} to~\ref{fig:Noise002-2}
% Fig.~\ref{fig:Noise001-1}, Fig.~\ref{fig:Noise001-2}, Fig.~\ref{fig:Noise002-1}, Fig.~\ref{fig:Noise002-2}, Fig.~\ref{fig:plane02-1}, Fig.~\ref{fig:Plane02-2}, Fig.~\ref{fig:plane05-1}, Fig.~\ref{fig:Plane05-2}, Fig.~\ref{fig:plane08-1}, and Fig.~\ref{fig:Plane08-2} 
present more results of more categories, notice that all these results are inferenced from the same checkpoint as used in the experiment session.

\newpage
%\noindent\textbf{Masked by Plane Ratio=0.2}

%\noindent\textbf{Masked by Plane Ratio=0.5}

%\noindent\textbf{Masked by Plane Ratio=0.8}

%\noindent\textbf{Masked by Noise Ratio=0.01}
%\noindent\textbf{Masked by Noise Ratio=0.02}

\end{document}